%% file: valada18ijcv.tex
\RequirePackage{fix-cm}
\documentclass[twocolumn]{svjour3}          
\smartqed  

\input{packageincludes.tex}
\input{sections/colors.tex}

\newcommand{\secref}[1]{Section~\ref{#1}}

\renewcommand{\eqref}[1]{Equation~(\ref{#1})}
\newcommand{\figref}[1]{Figure~\ref{#1}}
\newcommand{\tabref}[1]{Table~\ref{#1}}

\newcommand\crule[3][black]{\textcolor{#1}{\rule{#2}{#3}}}
\newcommand{\rot}[1]{\rotatebox[origin=c]{90}{#1}}

\DeclareSIUnit{\million}{\text{M}}
\DeclareSIUnit{\billion}{\text{B}}
\DeclareMathAlphabet\mathbfcal{OMS}{cmsy}{b}{n}

\newcolumntype{d}[1]{D{.}{.}{#1}}
\newcolumntype{Y}{D{.}{.}{1.2}}
\newcolumntype{P}[1]{>{\centering\arraybackslash}p{#1}}
 \journalname{International Journal of Computer Vision}
\begin{document}

\title{Self-Supervised Model Adaptation for Multimodal\\Semantic Segmentation
}


\author{Abhinav Valada\protect\affmark[1] \and
        Rohit Mohan\affmark[1] \and 
        Wolfram Burgard\affmark[1]$^\textbf{,}$\affmark[2]
}

\authorrunning{Valada et al.} 

\institute{Abhinav Valada \at
            {valada@cs.uni-freiburg.de}
            \and Rohit Mohan \at
            {mohan@cs.uni-freiburg.de}
            \and Wolfram Burgard \at
            {burgard@cs.uni-freiburg.de}\and
            \affaddr{\affmark[1]~University of Freiburg, Germany}\\
            \affaddr{\affmark[2]~Toyota Research Institute, Los Altos, USA}
}

\date{ }

    \maketitle
    \begin{abstract}
Learning to reliably perceive and understand the scene is an integral enabler for robots to operate in the real-world. This problem is inherently challenging due to the multitude of object types as well as appearance changes caused by varying illumination and weather conditions. Leveraging complementary modalities can enable learning of semantically richer representations that are resilient to such perturbations. Despite the tremendous progress in recent years, most multimodal convolutional neural network approaches directly concatenate feature maps from individual modality streams rendering the model incapable of focusing only on the relevant complementary information for fusion. To address this limitation, we propose a mutimodal semantic segmentation framework that dynamically adapts the fusion of modality-specific features while being sensitive to the object category, spatial location and scene context in a self-supervised manner. Specifically, we propose an architecture consisting of two modality-specific encoder streams that fuse intermediate encoder representations into a single decoder using our proposed self-supervised model adaptation fusion mechanism which optimally combines complementary features. As intermediate representations are not aligned across modalities, we introduce an attention scheme for better correlation. In addition, we propose a computationally efficient unimodal segmentation architecture termed AdapNet++ that incorporates a new encoder with multiscale residual units and an efficient atrous spatial pyramid pooling that has a larger effective receptive field with more than $10\times$ fewer parameters, complemented with a strong decoder with a multi-resolution supervision scheme that recovers high-resolution details. Comprehensive empirical evaluations on Cityscapes, Synthia, \mbox{SUN RGB-D}, ScanNet and Freiburg Forest benchmarks demonstrate that both our unimodal and multimodal architectures achieve state-of-the-art performance while simultaneously being efficient in terms of parameters and inference time as well as demonstrating substantial robustness in adverse perceptual conditions.

    \end{abstract}

\input{sections/introduction}

\input{sections/related_works}

\input{sections/adapnet++}

\input{sections/multimodal_fusion}

\input{sections/experimental_results}

\input{sections/conclusion}

\bibliographystyle{spbasic}      

\begin{small}
\bibliography{sections/references}
\end{small}

\end{document}

%% file: packageincludes.tex
\usepackage{times}

\usepackage{textcomp}
\usepackage{lipsum}
\usepackage{flushend}

\usepackage{url}
\usepackage{ifpdf}

\ifpdf
\usepackage{xcolor}
\usepackage{graphics}
\usepackage[pdftex]{graphicx}
\DeclareGraphicsExtensions{.pdf,.PDF,.png,.PNG,.jpg,.JPG}
\usepackage{hyperref}
\hypersetup{
  colorlinks,%
  citecolor=blue,%
  filecolor=blue,%
  linkcolor=blue,%
  urlcolor=blue
}
\else
\usepackage{graphicx}
\DeclareGraphicsExtensions{.eps,.EPS,.ps,.PS,}
\fi

\usepackage{amsmath}
\usepackage{amssymb} 
\usepackage{stmaryrd}           
\usepackage{mathptmx}
\usepackage[mathscr]{euscript}  

\DeclareMathAlphabet{\mathcal}{OMS}{cmsy}{m}{n} 

\newcommand\imCMsym[4][\mathord]{%
  \DeclareFontFamily{U} {#2}{}
  \DeclareFontShape{U}{#2}{m}{n}{
    <-6> #25
    <6-7> #26
    <7-8> #27
    <8-9> #28
    <9-10> #29
    <10-12> #210
    <12-> #212}{}
  \DeclareSymbolFont{CM#2} {U} {#2}{m}{n}
  \DeclareMathSymbol{#4}{#1}{CM#2}{#3}
}

\imCMsym{cmmi}{124}{\CMjmath}

\usepackage{algorithm}
\usepackage[noend]{algpseudocode}

\usepackage{tabularx}
\usepackage{balance}
\usepackage{booktabs, multirow}
\usepackage[british]{babel}
\usepackage{dcolumn}
\usepackage{colortbl}

\usepackage[binary-units=true,load=prefixed]{siunitx}

\usepackage{natbib}

\makeatletter
\if@twocolumn
  \renewcommand\normalsize{%
   \@setfontsize\normalsize\@xpt{12.5pt}%
   \abovedisplayskip=3 mm plus6pt minus 4pt
   \belowdisplayskip=3 mm plus6pt minus 4pt
   \abovedisplayshortskip=0.0 mm plus6pt
   \belowdisplayshortskip=2 mm plus4pt minus 4pt
   \let\@listi\@listI}%

  \renewcommand\small{%
   \@setfontsize\small{8.5pt}\@xpt
   \abovedisplayskip 8.5\p@ \@plus3\p@ \@minus4\p@
   \abovedisplayshortskip \z@ \@plus2\p@
   \belowdisplayshortskip 4\p@ \@plus2\p@ \@minus2\p@
   \def\@listi{\leftmargin\leftmargini
               \parsep 0\p@ \@plus1\p@ \@minus\p@
               \topsep 4\p@ \@plus2\p@ \@minus4\p@
               \itemsep0\p@}%
   \belowdisplayskip \abovedisplayskip}

\else
  \if@smallext
   \renewcommand\normalsize{%
   \@setfontsize\normalsize\@xpt\@xiipt
   \abovedisplayskip=3 mm plus6pt minus 4pt
   \belowdisplayskip=3 mm plus6pt minus 4pt
   \abovedisplayshortskip=0.0 mm plus6pt
   \belowdisplayshortskip=2 mm plus4pt minus 4pt
   \let\@listi\@listI}%

  \renewcommand\small{%
   \@setfontsize\small\@viiipt{9.5pt}%
   \abovedisplayskip 8.5\p@ \@plus3\p@ \@minus4\p@
   \abovedisplayshortskip \z@ \@plus2\p@
   \belowdisplayshortskip 4\p@ \@plus2\p@ \@minus2\p@
   \def\@listi{\leftmargin\leftmargini
               \parsep 0\p@ \@plus1\p@ \@minus\p@
               \topsep 4\p@ \@plus2\p@ \@minus4\p@
               \itemsep0\p@}%
   \belowdisplayskip \abovedisplayskip}
 \else
  \renewcommand\normalsize{%
   \@setfontsize\normalsize{9.5pt}{11.5pt}%
   \abovedisplayskip=3 mm plus6pt minus 4pt
   \belowdisplayskip=3 mm plus6pt minus 4pt
   \abovedisplayshortskip=0.0 mm plus6pt
   \belowdisplayshortskip=2 mm plus4pt minus 4pt
   \let\@listi\@listI}%

  \renewcommand\small{%
   \@setfontsize\small\@viiipt{9.25pt}%
   \abovedisplayskip 8.5\p@ \@plus3\p@ \@minus4\p@
   \abovedisplayshortskip \z@ \@plus2\p@
   \belowdisplayshortskip 4\p@ \@plus2\p@ \@minus2\p@
   \def\@listi{\leftmargin\leftmargini
               \parsep 0\p@ \@plus1\p@ \@minus\p@
               \topsep 4\p@ \@plus2\p@ \@minus4\p@
               \itemsep0\p@}%
   \belowdisplayskip \abovedisplayskip}
  \fi
\fi
\let\footnotesize\small
\makeatother

\usepackage{microtype}

\usepackage{scrtime}
\usepackage[dont-mess-around]{fnpct}

\usepackage{etoolbox}
\newcommand*{\affaddr}[1]{#1} 
\newcommand*{\affmark}[1][*]{\textsuperscript{#1}}

\makeatletter
\patchcmd\@combinedblfloats{\box\@outputbox}{\unvbox\@outputbox}{}{\errmessage{\noexpand patch failed}}
\makeatother

%% file: sections/colors.tex
\definecolor{white}{RGB}{255, 255, 255}
\definecolor{black}{RGB}{0, 0, 0}

\definecolor{ais-blue}{RGB}{51, 102, 153}
\definecolor{ais-red}{RGB}{155, 0, 0}
\definecolor{ais-gray}{RGB}{150, 150, 150}

\definecolor{c_sky}{RGB}{70, 130, 180}
\definecolor{c_building}{RGB}{70, 70, 70}
\definecolor{c_road}{RGB}{128, 64, 128}
\definecolor{c_sidewalk}{RGB}{244, 35, 232}
\definecolor{c_fence}{RGB}{190, 153, 153}
\definecolor{c_vegetation}{RGB}{107, 142, 35}
\definecolor{c_pole}{RGB}{153, 153, 153}
\definecolor{c_car}{RGB}{0, 0, 142}
\definecolor{c_sign}{RGB}{220, 220, 0}
\definecolor{c_person}{RGB}{220, 20, 60}
\definecolor{c_cyclist}{RGB}{107, 104, 130}

\definecolor{s_sky}{RGB}{128,128,128}
\definecolor{s_building}{RGB}{128,0,0}
\definecolor{s_road}{RGB}{128, 64, 128}
\definecolor{s_sidewalk}{RGB}{0, 0, 192}
\definecolor{s_fence}{RGB}{64, 64, 128}
\definecolor{s_vegetation}{RGB}{128, 128, 0}
\definecolor{s_pole}{RGB}{192, 192, 128}
\definecolor{s_car}{RGB}{64, 0, 128}
\definecolor{s_sign}{RGB}{192, 128, 128}
\definecolor{s_person}{RGB}{64, 64, 0}
\definecolor{s_cyclist}{RGB}{0, 128, 192}

\definecolor{u_Wall}{RGB}{31,119,180}
\definecolor{u_Floor}{RGB}{174,199,232}
\definecolor{u_Cabinet}{RGB}{255,127,14}
\definecolor{u_Bed}{RGB}{255,187,120}
\definecolor{u_Chair}{RGB}{44,160,44}
\definecolor{u_Sofa}{RGB}{152,223,138}
\definecolor{u_Table}{RGB}{214,39,40}
\definecolor{u_Door}{RGB}{255,152,150}
\definecolor{u_Window}{RGB}{148,103,189}
\definecolor{u_Bookshelf}{RGB}{197,176,213}
\definecolor{u_Picture}{RGB}{140,86,75}
\definecolor{u_Counter}{RGB}{80,143,183}
\definecolor{u_Blinds}{RGB}{227,119,194}
\definecolor{u_Desk}{RGB}{247,182,210}
\definecolor{u_Shelves}{RGB}{127,127,127}
\definecolor{u_Curtain}{RGB}{199,199,199}
\definecolor{u_Dresser}{RGB}{188,189,34}
\definecolor{u_Pillow}{RGB}{219,219,141}
\definecolor{u_Mirror}{RGB}{23,190,207}
\definecolor{u_FloorMat}{RGB}{158,218,229}
\definecolor{u_Clothes}{RGB}{141,211,199}
\definecolor{u_Ceiling}{RGB}{255,255,179}
\definecolor{u_Books}{RGB}{190,186,218}
\definecolor{u_Fridge}{RGB}{251,128,114}
\definecolor{u_Tv}{RGB}{128,177,211}
\definecolor{u_Paper}{RGB}{253,180,98}
\definecolor{u_Towel}{RGB}{179,222,105}
\definecolor{u_ShowerCurtain}{RGB}{252,205,229}
\definecolor{u_Box}{RGB}{217,217,217}
\definecolor{u_Whiteboard}{RGB}{188,128,189}
\definecolor{u_Person}{RGB}{204,235,197}
\definecolor{u_NightStand}{RGB}{255,237,111}
\definecolor{u_Toilet}{RGB}{228,26,28}
\definecolor{u_Sink}{RGB}{55,126,184}
\definecolor{u_Lamp}{RGB}{77,175,74}
\definecolor{u_Bathtub}{RGB}{152,78,163}
\definecolor{u_Bag}{RGB}{255,127,0}

\definecolor{a_Wall}{RGB}{190,153,112}
\definecolor{a_Floor}{RGB}{189,198,255}
\definecolor{a_Cabinet}{RGB}{213,255,0}
\definecolor{a_Bed}{RGB}{158,0,148}
\definecolor{a_Chair}{RGB}{152,255,82}
\definecolor{a_Sofa}{RGB}{119,77,0}
\definecolor{a_Table}{RGB}{122,71,130}
\definecolor{a_Door}{RGB}{0,174,126}
\definecolor{a_Window}{RGB}{0,125,181}
\definecolor{a_Bookshelf}{RGB}{0,143,156}
\definecolor{a_Picture}{RGB}{107,104,130}
\definecolor{a_Counter}{RGB}{255,229,2}
\definecolor{a_Desk}{RGB}{1,255,254}
\definecolor{a_Curtain}{RGB}{255,166,254}
\definecolor{a_Fridge}{RGB}{232,94,190}
\definecolor{a_ShowerCurtain}{RGB}{2,100,1}
\definecolor{a_Toilet}{RGB}{133,169,0}
\definecolor{a_Sink}{RGB}{149,0,58}
\definecolor{a_Bathtub}{RGB}{187,136,0}
\definecolor{a_Other}{RGB}{0,0,255}

\definecolor{f_trail}{RGB}{170,170,170}
\definecolor{f_grass}{RGB}{0,255,0}
\definecolor{f_vegetation}{RGB}{102,102,51}
\definecolor{f_sky}{RGB}{0,120,255}
\definecolor{f_obstacle}{RGB}{255,255,0}

%% file: sections/introduction.tex
\section{Introduction}
\label{sec:intro}

Humans have the remarkable ability to instantaneously recognize and understand a complex visual scene which has piqued the interest of computer vision researches to model this ability since the 1960s~\citep{fei2004we}. There are numerous ever-expanding applications to this capability ranging from robotics~\citep{xiang2017rnn} and remote sensing~\citep{audebert2018beyond} to medical diagnostics~\citep{ronneberger2015u} and content-based image retrieval~\citep{noh2017largescale}. However, there are several challenges imposed by the multifaceted nature of this problem including the large variation in types and scales of objects, clutter and occlusions in the scene as well as outdoor appearance changes that take place throughout the day and across seasons.

\begin{figure}
\centering
\footnotesize
\setlength{\tabcolsep}{0.2cm}
\begin{tabular}{p{3.5cm} p{3.5cm}}
\includegraphics[width=\linewidth]{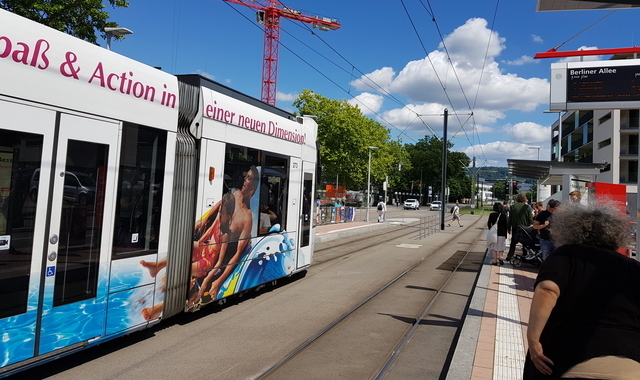} & 
\includegraphics[width=\linewidth]{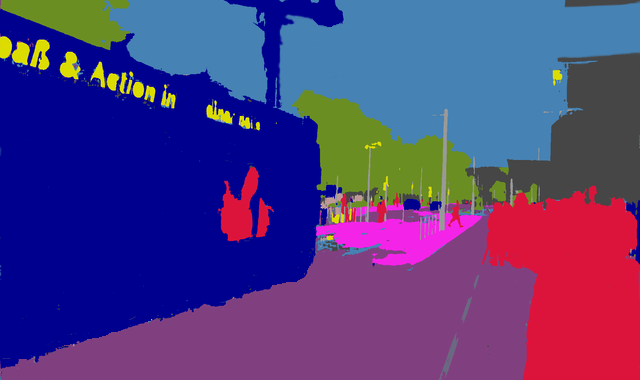} \\
\includegraphics[width=\linewidth]{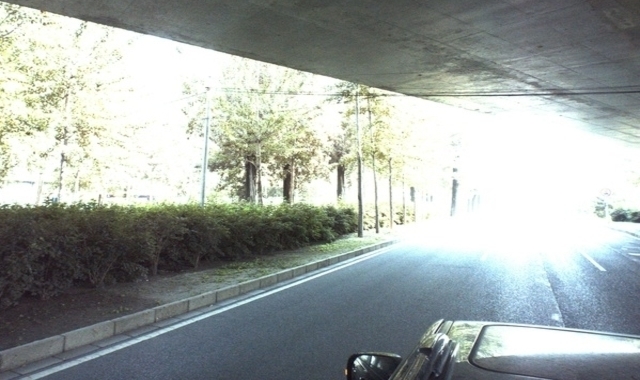} &
\includegraphics[width=\linewidth]{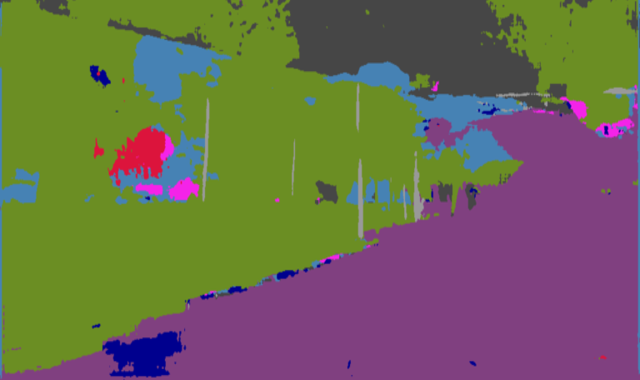} \\
(a) Input Image & (b) Segmented Output \\ 
\end{tabular}
\caption{Example real-world scenarios where current state-of-the-art approaches demonstrate misclassifications. The first row shows an issue of mismatched relationship as well as inconspicuous classes where a decal on the train is falsely predicted as a person and the decal text is falsely predicted as a sign. The second row shows misclassifications caused by overexposure of the camera due to car exiting a tunnel.}
\label{fig:cover}
\end{figure}

Deep Convolutional Neural Network (DCNN) based methods~\citep{long2015,ChenPK0Y16,YuKoltun2016} modelled as a Fully Convolutional Neural Network (FCN) have dramatically increased the performance on several semantic segmentation benchmarks. Nevertheless, they still face challenges due to the diversity of scenes in the real-world that cause mismatched relationship and inconspicuous object classes. \figref{fig:cover} shows two example scenes from real-world scenarios in which
misclassifications are produced due to the decal on the train which is falsely predicted as
a person and a traffic sign (first row), and overexposure of the camera caused by the vehicle exiting a tunnel (second row). In order to accurately predict the elements of the scene in such situations, features from complementary modalities such as depth and infrared can be leveraged to exploit properties such as geometry and reflectance, respectively. Moreover, the network can exploit complex intra-modal dependencies more effectively by directly learning to fuse visual appearance information from RGB images with learned features from complementary modalities in an end-to-end fashion. This not only enables the network to resolve inherent ambiguities and improve reliability but also obtain a more holistic scene segmentation.

\sloppy{While most existing work focuses on where to fuse modality-specific streams topologically \citep{fusenet2016accv,schneider2017multimodal,valada2016towards} and what transformations can be applied on the depth modality to enable better fusion with visual RGB features~\citep{gupta2014learning,eitel15iros}, it still remains an open question as to how to enable the network to dynamically adapt its fusion strategy based on the nature of the scene such as the types of objects, their spatial location in the world and the present scene context. This is a crucial requirement in applications such as robotics and autonomous driving where these systems run in continually changing environmental contexts. For example, an autonomous car navigating in ideal weather conditions can primarily rely on visual information but when it enters a dark tunnel or exits an underpassage, the cameras might experience under/over exposure, whereas the depth modality will be more informative. Furthermore, the strategy to be employed for fusion also varies with the types of objects in the scene, for instance, infrared might be more useful to detect categories such as people, vehicles, vegetation and boundaries of structures but it does not provide much information on object categories such as the sky. Additionally, the spatial location of objects in the scene also has an influence, for example, the depth modality provides rich information on objects that are at nearby distances but degrades very quickly for objects that are several meters away. More importantly, the approach employed should be robust to sensor failure and noise as constraining the network to always depend on both modalities and use noisy information can worsen the actual performance and lead to disastrous situations.}

Due to these complex interdependencies, naively treating modalities as multi-channel
input data or concatenating independently learned modality-specific features does not
allow the network to adapt to the aforementioned situations dynamically. Moreover, due to the nature of this dynamicity, the fusion mechanism has to be trained in a self-supervised manner in order to make the adaptivity emergent and to generalize effectively to different real-world scenarios. As a solution to this problem, we present the Self-Supervised Model Adaptation (SSMA) fusion mechanism that adaptively recalibrates and fuses modality-specific feature maps based on the object class, its spatial location and the scene context. The SSMA module takes intermediate encoder representations of modality-specific streams as input and fuses them probabilistically based on the activations of individual modality streams. As we model the SSMA block in a fully convolutional fashion, it yields a probability for each activation in the feature maps which represents the optimal combination to exploit complementary properties. These probabilities are then used to amplify or suppress the representations of the individual modality streams, followed by the fusion. As we base the fusion on modality-specific activations, the fusion is intrinsically tolerant to sensor failure and noise such as missing depth values. 

Our proposed architecture for multimodal segmentation consists of individual modality-specific encoder streams which are fused both at mid-level stages and at the end of the encoder streams using our SSMA blocks. The fused representations are input to the decoder at different stages for upsampling and refining the predictions. Note that only the multimodal SSMA fusion mechanism is self-supervised, the semantic segmentation is trained in a supervised manner. We employ a combination of mid-level and late-fusion as several experiments have demonstrated that fusing semantically meaningful representations yields better performance in comparison to early fusion~\citep{eitel15iros,valada16iser,fusenet2016accv,xiang2017rnn}. Moreover, studies of the neural dynamics of the human brain has also shown evidence of late-fusion of modalities for recognition tasks~\citep{cichy2016similarity}. However, intermediate network representations are not aligned across modality-specific streams. Hence, integrating fused multimodal mid-level features into high-level features requires explicit prior alignment. Therefore, we propose an attention mechanism that weighs the fused multimodal mid-level skip features with spatially aggregated statistics of the high-level decoder features for better correlation, followed by channel-wise concatenation.

As our fusion framework necessitates individual modality-specific encoders, the architecture that we employ for the encoder and decoder should be efficient in terms of the number of parameters and computational operations, as well as be able to learn highly discriminative deep features. State-of-the-art semantic segmentation architectures such as \mbox{DeepLab~v3} \citep{chen2017rethinking} and PSPnet~\citep{zhao2017pspnet} employ the ResNet-101~\citep{He2015} architecture which consumes $42.39\million$ parameters and $113.96\billion$ FLOPS, as the encoder backbone. Training such architectures requires a large amount of memory and synchronized training across multiple GPUs. Moreover, they have slow run-times rendering them impractical for resource constrained applications such as robotics and augmented reality. More importantly, it is infeasible to employ them in multimodal frameworks that require multiple modality-specific streams as we do in this work.\looseness=-1

With the goal of achieving the right trade-off between performance and computational complexity, we propose the \mbox{AdapNet++} architecture for unimodal segmentation. We build the encoder of our model based on the full pre-activation ResNet-50~\citep{he2016identity} architecture and incorporate our previously proposed multiscale residual units~\citep{valada17icra} to aggregate multiscale features throughout the network without increasing the number of parameters. The proposed units are more effective in learning multiscale features than the commonly employed multigrid approach introduced in DeepLab~v3~\citep{chen2017rethinking}. In addition, we propose an efficient variant of the Atrous Spatial Pyramid Pooling (ASPP)~\citep{chen2017rethinking} called eASPP that employs cascaded and parallel atrous convolutions to capture long range context with a larger effective receptive field, while simultaneously reducing the number of parameters by 87\% in comparison to the originally proposed ASPP. We also propose a new decoder that integrates mid-level features from the encoder using multiple skip refinement stages for high resolution segmentation along the object boundaries. In order to aid the optimization and to accelerate training, we propose a multiresolution supervision strategy that introduces weighted auxiliary losses after each upsampling stage in the decoder. This enables faster convergence, in addition to improving the performance of the model along the object boundaries. Our proposed architecture is compact and trainable with a large mini-batch size on a single consumer grade GPU.

Motivated by the recent success of compressing DCNNs by pruning unimportant neurons~\citep{molchanov2016pruning,liu2017learning,anwar2017structured}, we explore pruning entire convolutional feature maps of our model to further reduce the number of parameters. Network pruning approaches utilize a cost function to first rank the importance of neurons, followed by removing the least important neurons and fine-tuning the network to recover any loss in accuracy. Thus far, these approaches have only been employed for pruning convolutional layers that do not have an identity or a projection shortcut connection. Pruning residual feature maps (third convolutional layer of a residual unit) also necessitates pruning the projected feature maps in the same configuration in order to maintain the shortcut connection. This leads to a significant drop in accuracy, therefore current approaches omit pruning convolutional filters with shortcut connections. As a solution to this problem, we propose a network-wide holistic pruning approach that employs a simple and yet effective strategy for pruning convolutional filters invariant to the presence of shortcut connections. This enables our network to further reduce the number of parameters and computing operations, making our model efficiently deployable even in resource constrained applications.

Finally, we present extensive experimental evaluations of our proposed unimodal and multimodal architectures on benchmark scene understanding datasets including Cityscapes \citep{Cordts2016cityscapes}, Synthia \citep{roscvpr16}, SUN RGB-D \citep{song2015sun}, ScanNet \citep{dai2017scannet} and Freiburg Forest \citep{valada16iser}. The results demonstrate that our model sets the new state-of-the-art on all these benchmarks considering the computational efficiency and the fast inference time of $72\milli\second$ on a consumer grade GPU. More importantly, our dynamically adapting multimodal architecture demonstrates exceptional robustness in adverse perceptual conditions such as fog, snow, rain and night-time, thus enabling it to be employed in critical resource constrained applications such as robotics where not only accuracy but robustness, computational efficiency and run-time are equally important. To the best of our knowledge, this is the first multimodal segmentation work to benchmark on these wide range of datasets containing several modalities and diverse environments ranging from urban city driving scenes to indoor environments and unstructured forested scenes.

In summary, the following are the main contributions of this work:
\begin{enumerate}
\item A multimodal fusion framework incorporating our proposed SSMA fusion blocks that adapts the fusion of modality-specific features dynamically according to the object category, its spatial location as well as the scene context and learns in a self-supervised manner.
\item The novel AdapNet++ semantic segmentation architecture that incorporates our multiscale residual units, a new efficient ASPP, a new decoder with skip refinement stages and a multiresolution supervision strategy.
\item The eASPP for efficiently aggregating multiscale features and capturing long range context, while having a larger effective receptive field and over $10\times$ reduction in parameters compared to the standard ASPP.
\item An attention mechanism for effectively correlating fused multimodal mid-level and high-level features for better object boundary refinement.
\item A holistic network-wide pruning approach that enables pruning of convolutional filters invariant to the presence of identity or projection shortcuts.
\item Extensive benchmarking of existing approaches with the same input image size and evaluation setting along with quantitative and qualitative evaluations of our unimodal and multimodal architectures on five different benchmark datasets consisting of multiple modalities.
\item Implementations of our proposed architectures are made publicly available at \url{https://github.com/DeepSceneSeg} and a live demo on all the five datasets can be viewed at \url{http://deepscene.cs.uni-freiburg.de}.
\end{enumerate}


%% file: sections/related_works.tex
\section{Related Works}
\label{sec:relatedworks}

In the last decade, there has been a sharp transition in semantic segmentation approaches from employing hand engineered features with flat classifiers such as Support Vector Machines~\citep{fulkerson20009}, Boosting~\citep{sturgess2009} or Random Forests~\citep{shotton2008, brostow2008}, to end-to-end DCNN-based approaches~\citep{long2015, kendall2015}. We first briefly review some of the classical methods before delving into the state-of-the-art techniques.

{\parskip=5pt
\noindent\textbf{Semantic Segmentation:} Semantic segmentation is one of the fundamental problems in computer vision. Some of the earlier approaches for semantic segmentation use small patches to classify the center pixel using flat classifiers~\citep{shotton2008, sturgess2009} followed by smoothing the predictions using Conditional Random Fields (CRFs)~\citep{sturgess2009}. Rather than only relying on appearance based features, structure from motion features have also been used with randomized decision forests~\citep{brostow2008, sturgess2009}. View independent 3D features from dense depth maps have been shown to outperform appearance based features, that also enabled classification of all the pixels in an image, as opposed to only the center pixel of a patch ~\citep{zhang2010}. \cite{Plath2009} propose an approach to combine local and global features using a CRF and an image classification method. However, the performance of these approaches is largely bounded by the expressiveness of handcrafted features which is highly scenario-specific.}

The remarkable performance achieved by CNNs in classification tasks led to their application for dense prediction problems such as semantic segmentation, depth estimation and optical flow prediction. Initial approaches that employed neural networks for semantic  segmentation still relied on patch-wise training~\citep{grangier2009,clement2012,Pinheiro2014}. \cite{Pinheiro2014} use a recurrent CNN to aggregate several low resolution predictions for scene labeling. \cite{clement2012} transforms the input image through a Laplacian pyramid followed by feeding each scale to a CNN for hierarchical feature extraction and classification. Although these approaches demonstrated improved performance over handcrafted features, they often yield a grid-like output that does not capture the true object boundaries. One of the first end-to-end approaches that learns to directly map the low resolution representations from a classification network to a dense prediction output was the Fully Convolutional Network (FCN) model~\citep{long2015}. FCN proposed an encoder-decoder architecture in which the encoder is built upon the VGG-16~\citep{Simonyan14c} architecture with the inner-product layers replaced with convolutional layers. While, the decoder consists of successive deconvolution and convolution layers that upsample and refine the low resolution feature maps by combining them with the encoder feature maps. The last decoder then yields a segmented output with the same resolution as the input image.

DeconvNet~\citep{noh2015learning} propose an improved architecture containing stacked deconvolution and unpooling layers that perform non-linear upsampling and outperforms FCNs but at the cost of a more complex training procedure. The SegNet~\citep{kendall2015} architecture eliminates the need for learning to upsample by reusing pooling indices from the encoder layers to perform upsampling. \cite{oliveira2016} propose an architecture that builds upon FCNs and introduces more refinement stages and incorporates spatial dropout to prevent over fitting. The ParseNet~\citep{liu2015} architecture models global context directly instead of only relying on the largest receptive field of the network. Recently, there has been more focus on learning multiscale features, which was initially achieved by providing the network with multiple rescaled versions of the image~\citep{clement2012} or by fusing features from multiple parallel branches that take different image resolutions~\citep{long2015}. However, these networks still use pooling layers to increase the receptive field, thereby decreasing the spatial resolution, which is not ideal for a segmentation network.

In order to alleviate this problem, \cite{YuKoltun2016} propose dilated convolutions that allows for exponential increase in the receptive field without decrease in resolution or increase in parameters. DeepLab~\citep{ChenPK0Y16} and PSPNet~\citep{zhao2017pspnet} build upon the aforementioned idea and propose pyramid pooling modules that utilize dilated convolutions of different rates to aggregate multiscale global context. DeepLab in addition uses fully connected CRFs in a post processing step for structured prediction. However, a drawback in employing these approaches is the computational complexity and substantially large inference time even using modern GPUs that hinder them from being deployed in robots that often have limited resources. In our previous work~\citep{valada17icra}, we proposed an architecture that introduces dilated convolutions parallel to the conventional convolution layers and multiscale residual blocks that incorporates them, which enables the model to achieve competitive performance at interactive frame rates. Our proposed multiscale residual blocks are more effective at learning multiscale features compared to the widely employed multigrid approach from DeepLab~v3~\citep{chen2017rethinking}. While in this work, we propose several new improvements for learning multiscale features, capturing long range context and improving the upsampling in the decoder, while simultaneously reducing the number of parameters and maintaining a fast inference time.

\begin{figure*}
\centering
\includegraphics[width=\linewidth]{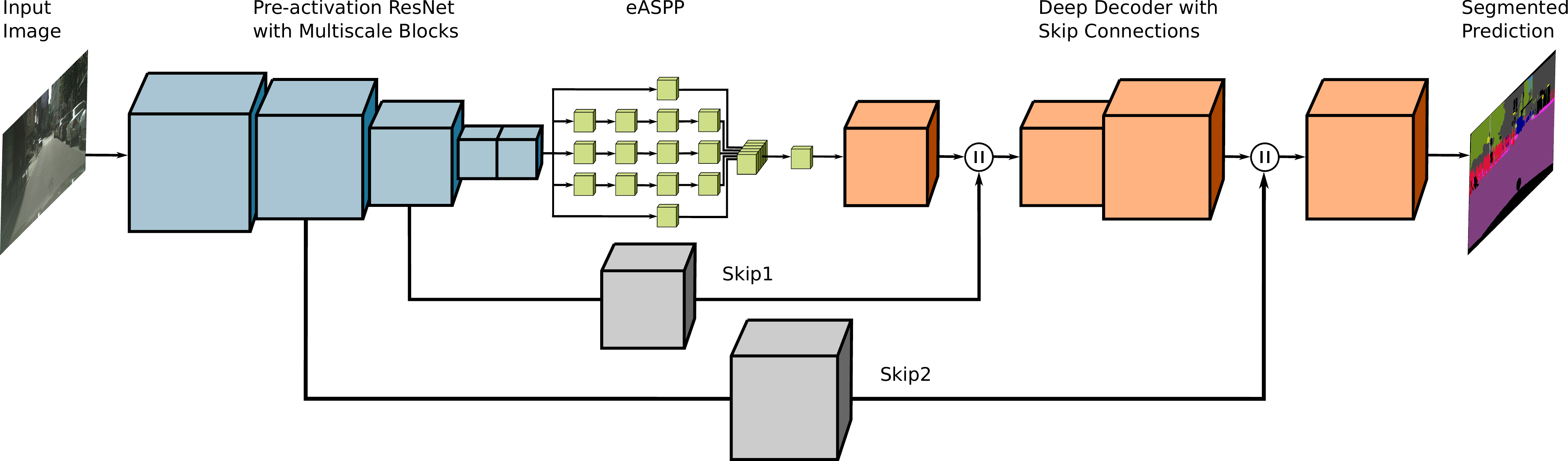}
\caption{Overview of our proposed Adapnet++ architecture. Given an input image, we use the full pre-activation ResNet-50 architecture augmented with our proposed multiscale residual blocks to yield a feature map 16-times downsampled with respect to the input image resolution, then our proposed efficient atrous spatial pyramid (eASPP) module is employed to further learn multiscale features and to capture long range context. Finally, the output of the eASPP is fed into our proposed deep decoder with skip connections for upsampling and refining the semantic pixel-level prediction.}
\label{fig:adapnet++}
\end{figure*}

{\parskip=5pt
\noindent\textbf{Multimodal Fusion:} The availability of low-cost sensors has encouraged novel approaches to exploit features from alternate modalities in an effort to improve robustness as well as the granularity of segmentation. \cite{SilbermanECCV12} propose an approach based on SIFT features and MRFs for indoor scene segmentation using RGB-D images. Subsequently, \cite{ren2012} propose improvements to the feature set by using kernel descriptors and by combining MRF with segmentation trees. \cite{munoz2012co} employ modality-specific classifier cascades that hierarchically propagate information and do not require one-to-one correspondence between data across modalities. In addition to incorporating features based on depth images, \cite{hermans2014dense} propose an approach that performs joint 3D mapping and semantic segmentation using Randomized Decision Forests. There has also been work on extracting combined RGB and depth features using CNNs~\citep{couprie2013,gupta2014learning} for object detection and semantic segmentation. In most of these approaches, hand engineered or learned features are extracted from individual modalities and combined together in a joint feature set which is then used for classification.}

More recently, there has been a series of DCNN-based fusion techniques~\citep{eitel15iros,Scherer2017,li2016lstm} that have been proposed for end-to-end learning of fused representations from multiple modalities. These fusion approaches can be categorized into early, hierarchical and late fusion methods. An intuitive early fusion technique is to stack data from multiple modalities channel-wise and feed it to the network as a four or six channel input. However, experiments have shown that this often does not enable the network to learn complementary features and cross-modal interdependencies~\citep{valada16iser, fusenet2016accv}. Hierarchical fusion approaches combine feature maps from multiple modality-specific encoders at various levels (often at each downsampling stage) and upsample the fused features using a single decoder~\citep{fusenet2016accv,Scherer2017}. Alternatively, \cite{schneider2017multimodal} propose a mid-level fusion approach in which NiN layers~\citep{lin2013network} with depth as input are used to fuse feature maps into the RGB encoder in the middle of the network. \cite{li2016lstm} propose a Long-Short Term Memory (LSTM) context fusion model that captures and fuses contextual information from multiple modalities accounting for the complex interdependencies between them. \citep{qi20173d} propose an interesting approach that employs 3D graph neural networks for RGB-D semantic segmentation that accounts for both 2D appearance and 3D geometric relations, while capturing long range dependencies within images.

In the late fusion approach, identical network streams are first trained individually on a specific modality and the feature maps are fused towards the end of network using concatenation~\citep{eitel15iros} or element-wise summation~\citep{valada16iser}, followed by learning deeper fused representations. However, this does not enable the network to adapt the fusion to changing scene context. In our previous work~\citep{valada16irosws}, we proposed a mixture-of-experts CMoDE fusion scheme for combining feature maps from late fusion based architectures. Subsequently, in~\citep{valada17icra} we extended the CMoDE framework for probabilistic fusion accounting for the types of object categories in the dataset which enables more flexibility in learning the optimal combination. Nevertheless, there are several real-world scenarios in which class-wise fusion is not sufficient, especially in outdoor scenes where different modalities perform well in different conditions. Moreover, the CMoDE module employs multiple softmax loss layers for each class to compute the probabilities for fusion which does not scale for datasets such as SUN RGB-D which has 37 object categories. Motivated by this observation, in this work, we propose a multimodal semantic segmentation architecture incorporating our SSMA fusion module that dynamically adapts the fusion of intermediate network representations from multiple modality-specific streams according to the object class, its spatial location and the scene context while learning the fusion in a self-supervised fashion.

%% file: sections/adapnet++.tex
\section{AdapNet++ Architecture}
\label{sec:technicalApproach}

In this section, we first briefly describe the overall topology of the proposed AdapNet++ architecture and our main contributions motivated by our design criteria. We then detail each of the constituting architectural components and our model compression technique.

Our network follows the general fully convolutional en-coder-decoder design principle as shown in \figref{fig:adapnet++}. The encoder (depicted in blue) is based on the full pre-activation ResNet-50~\citep{he2016identity} model as it offers a good trade-off between learning highly discriminative deep features and the computational complexity required. In order to effectively compute high resolution feature responses at different spatial densities, we incorporate our recently proposed multiscale residual units~\citep{valada17icra} at varying dilation rates in the last two blocks of the encoder. In addition, to enable our model to capture long-range context and to further learn multiscale representations, we propose an efficient variant of the atrous spatial pyramid pooling module known as eASPP which has a larger effective receptive field and reduces the number of parameters required by over $87\%$ compared to the originally proposed ASPP in DeepLab~v3~\citep{chen2017rethinking}. We append the proposed eASPP after the last residual block of the encoder, shown as green blocks in \figref{fig:adapnet++}. In order to recover the segmentation details from the low spatial resolution output of the encoder section, we propose a new deep decoder consisting of multiple deconvolution and convolution layers. Additionally, we employ skip refinement stages that fuse mid-level features from the encoder with the upsampled decoder feature maps for object boundary refinement. Furthermore, we add two auxiliary supervision branches after each upsampling stage to accelerate training and improve the gradient propagation in the network. We depict the decoder as orange blocks and the skip refinement stages as gray blocks in the network architecture shown in \figref{fig:adapnet++}. In the following sections, we discuss each of the aforementioned network components in detail and elaborate on the design choices.

\begin{figure*}
\centering
\includegraphics[width=\linewidth]{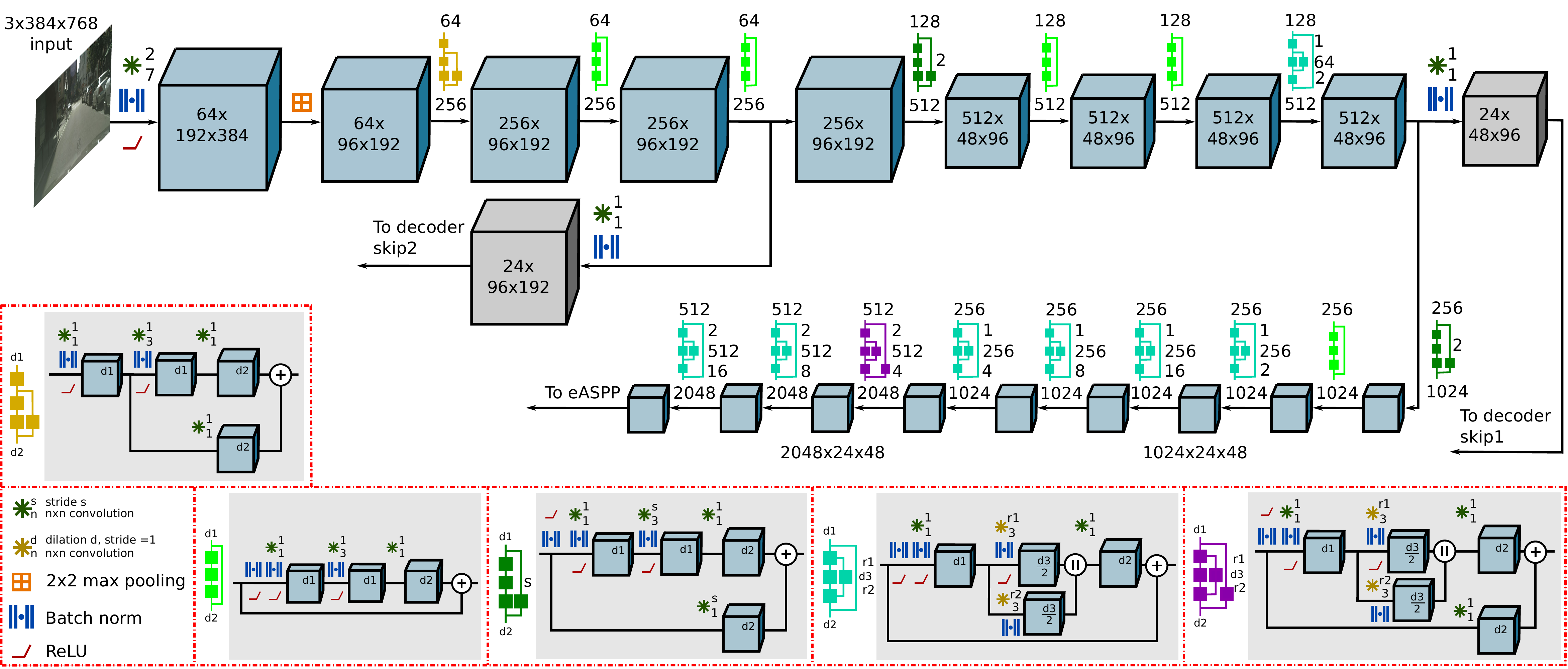}
\caption{The proposed encoder is built upon the full pre-activation ResNet-50 architecture. Specifically, we remove the last downsampling stage in ResNet-50 by setting the stride from two to one, therefore the final output of the encoder is 16-times downsampled with respect to the input. We then replace the residual units that follow the last downsampling stage with our proposed multiscale residual units. The legend enclosed in red lines show the original pre-activation residual units in the bottom left (yellow, light green and dark green), while our proposed multiscale residual units are shown in the bottom right (cyan and purple).}
\label{fig:encoder}
\end{figure*}

\subsection{Encoder}
\label{sec:encoder}

Encoders are the foundation of fully convolutional neural network architectures. Therefore, it is essential to build upon a good baseline that has a high representational ability conforming with the computational budget. Our critical requirement is to achieve the right trade-off between the accuracy of segmentation and inference time on a consumer grade GPU, while keeping the number of parameters low. As we also employ the proposed architecture for multimodal fusion, our objective is to design a topology that has a reasonable model size so that two individual modality-specific networks can be trained in a fusion framework and deployed on a single GPU. Therefore, we build upon the ResNet-50 architecture with the full preactivation residual units~\citep{he2016identity} instead of the originally proposed residual units~\citep{He2015} as they have been shown to reduce overfitting, improve the convergence and also yield better performance. The ResNet-50 architecture has four computational blocks with varying number of residual units. We use the bottleneck residual units in our encoder as they are computationally more efficient than the baseline residual units and they enable us to build more complex models that are easily trainable. The output of the last block of the ResNet-50 architecture is 32-times downsampled with respect to the input image resolution. In order to increase the spatial density of the feature responses and to prevent signal decimation, we set the stride of the convolution layer in the last block (\textit{res4a}) from two to one which makes the resolution of the output feature maps 1/16-times the input image resolution. We then replace the residual blocks that follow this last downsampling stage with our proposed multiscale residual units that incorporate parallel atrous convolutions~\citep{YuKoltun2016} at varying dilation rates. 

A naive approach to compute the feature responses at the full image resolution would be to remove the downsampling and replace all the convolutions to atrous convolutions having a dilation rate $r\geq2$ but this would be both computation and memory intensive. Therefore, we propose a novel multiscale residual unit~\citep{valada17icra} to efficiently enlarge the receptive field and aggregate multiscale features without increasing the number of parameters and the computational burden. Specifically, we replace the $3\times3$ convolution in the full pre-activation residual unit with two parallel $3\times3$ atrous convolutions with different dilation rates and half the number of feature maps each. We then concatenate their outputs before the following $1\times1$ convolution.

By concatenating their outputs, the network additionally learns to combine the feature maps of different scales. Now, by setting the dilation rate in one of the $3\times3$ convolutional layers to one and another to a rate $r\geq2$, we can preserve the original scale of the features within the block and simultaneously add a larger context. While, by varying the dilation rates in each of the parallel $3\times3$ convolutions, we can enable the network to effectively learn multiscale representations at different stages of the network. The topology of the proposed multiscale residual units and the corresponding original residual units are shown in the legend in \figref{fig:encoder}. The lower left two units show the original configuration, while the lower right two units show the proposed configuration. \figref{fig:encoder} shows our entire encoder structure with the full pre-activation residual units and the multiscale residual units.

We incorporate the first multiscale residual unit with $r_1=1, r_2=2$ before the third block at \textit{res3d} (unit before the block where we remove the downsampling as mentioned earlier). Subsequently, we replace the units \textit{res4c, res4d, res4e, res4f} with our proposed multiscale units with rates $r_1=1$ in all the units and $r_2=2, 4, 8, 16$   correspondingly. In addition, we replace the last three units of block four \textit{res5a, res5b, res5c} with the multiscale units with increasing rates in both $3\times3$ convolutions, as $(r_1=2, r_2=4)$, $(r_1=2, r_2=8)$, $(r_1=2, r_2=16)$ correspondingly. We evaluate our proposed configuration in comparison to the multigrid method of DeepLab~v3~\citep{chen2017rethinking} in \secref{sec:ablation}.

\subsection{Efficient Atrous Spatial Pyramid Pooling}

\begin{figure*}
\centering
\setlength{\tabcolsep}{0.45cm}
\begin{tabular}{p{5.5cm} p{10.5cm}}
\includegraphics[width=\linewidth]{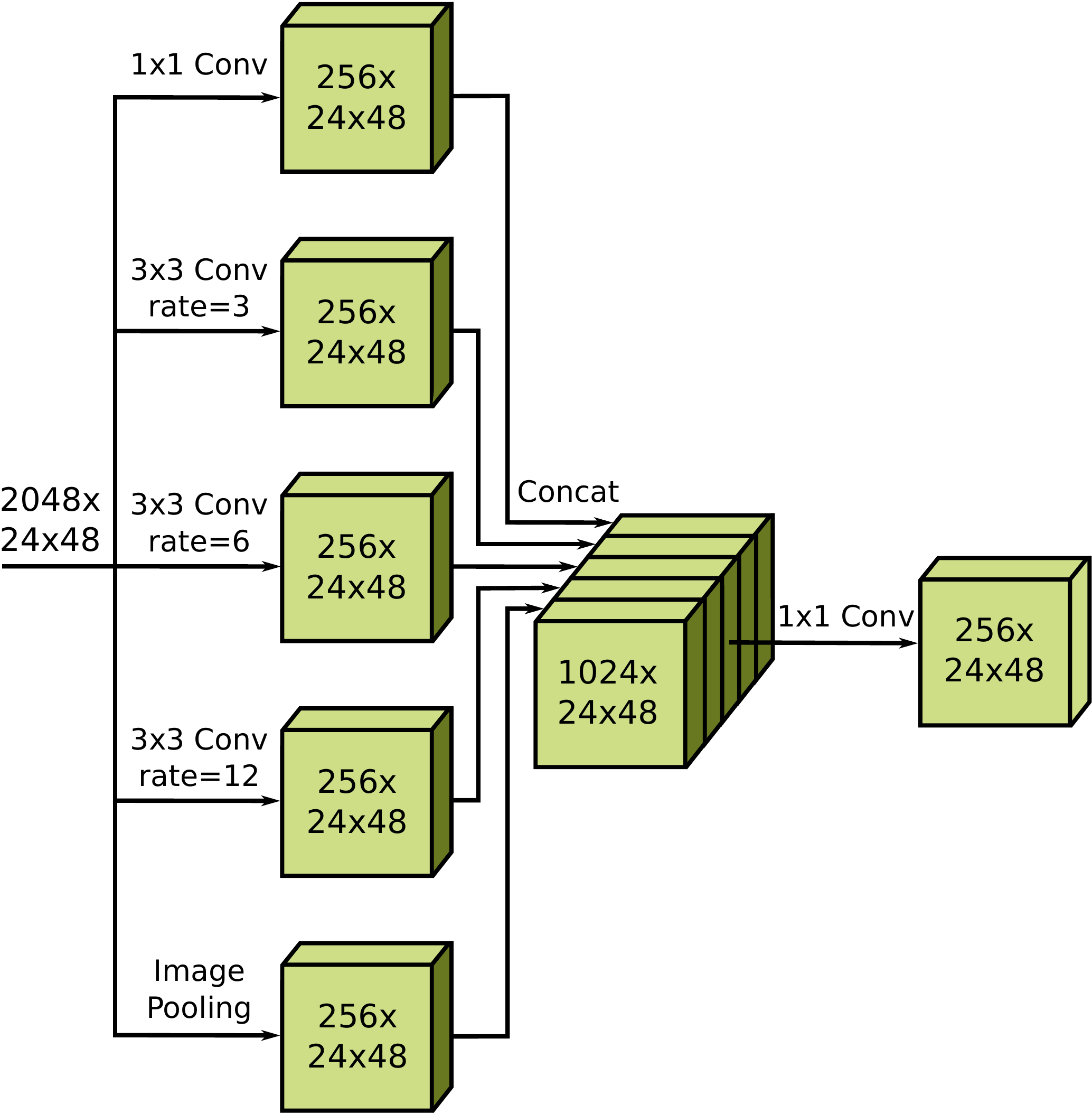} & 
\includegraphics[width=\linewidth]{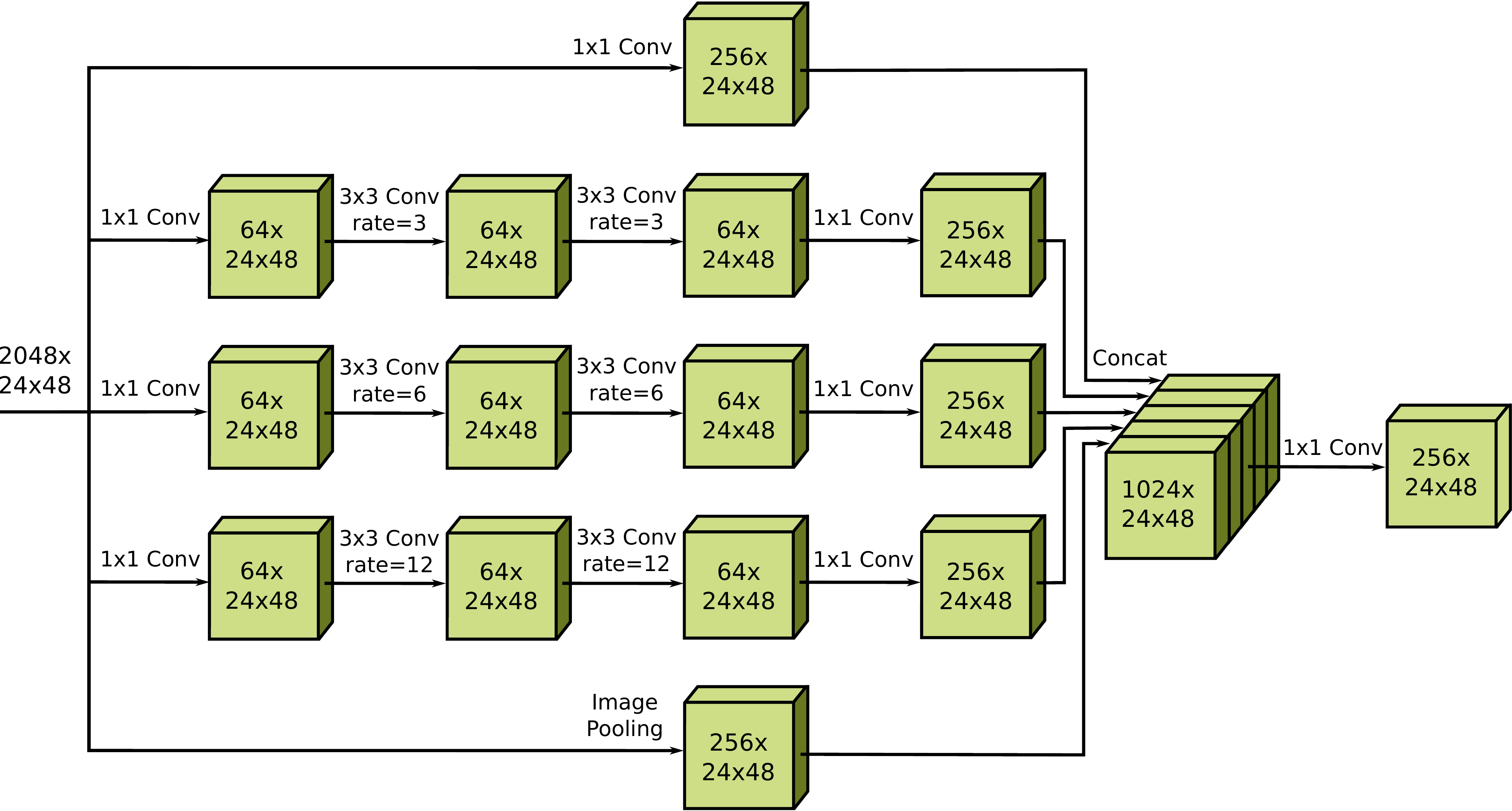} \\
\small{(a) Atrous Spatial Pyramid Pooling (ASPP)} & \small{(b) Efficient Atrous Spatial Pyramid Pooling (eASPP)} \\ 
\end{tabular}
\caption{Depiction of the ASPP module from DeepLab~v3 and our proposed efficient eASPP module. eASPP reduces the number of parameters by $87.87\%$ and the number of FLOPS by $89.88\%$, while simultaneously achieving improved performance. Note that all the convolution layers have batch normalization and we change the corresponding dilation rates in the $3\times3$ convolutions in ASPP to 3,6,12 as the input feature map to the ASPP is of dimensions $48\times23$ in our network architecture.}
\label{fig:aspp}
\end{figure*}

In this section, we first describe the topology of the Atrous Spatial Pyramid Pooling (ASPP) module, followed by the structure of our proposed efficient Atrous Spatial Pyramid Pooling (eASPP). ASPP has become prevalent in most state-of-the-art architectures due to its ability to capture long range context and multiscale information. Inspired by spatial pyramid pooling~\citep{he2015spatial}, the initially proposed ASPP in DeepLab~v2~\citep{liang2015semantic} employs four parallel atrous convolutions with different dilation rates. Concatenating the outputs of multiple parallel atrous convolutions aggregates multi-scale context with different receptive field resolutions. However, as illustrated in the subsequent DeepLab~v3~\citep{chen2017rethinking}, applying extremely large dilation rates inhibits capturing long range context due to image boundary effects. Therefore, an improved version of ASPP was proposed~\citep{chen2017rethinking} to add global context information by incorporating image-level features.

The resulting ASPP shown in \figref{fig:aspp}(a) consists of five parallel branches: one $1\times1$ convolution and three $3\times3$ convolutions with different dilation rates. Additionally, image-level features are introduced by applying global average pooling on the input feature map, followed by a $1\times1$ convolution and bilinear upsampling to yield an output with the same dimensions as the input feature map. All the convolutions have 256 filters and batch normalization layers to improve training. Finally, the resulting feature maps from each of the parallel branches are concatenated and passed through another $1\times1$ convolution with batch normalization to yield 256 output filters. The ASPP module is appended after the last residual block of the encoder where the feature maps are of dimensions $65\times65$ in the DeepLab~v3 architecture~\citep{chen2017rethinking}, therefore dilation rates of 6, 12 and 18 were used in the parallel $3\times3$ atrous convolution layers. However, as we use a smaller input image, the dimensions of the input feature map to the ASPP is $24\times48$, therefore, we reduce the dilation rates to 3, 6 and 12 in the $3\times3$ atrous convolution layers respectively. 

\begin{figure*}
\centering
\includegraphics[width=\linewidth]{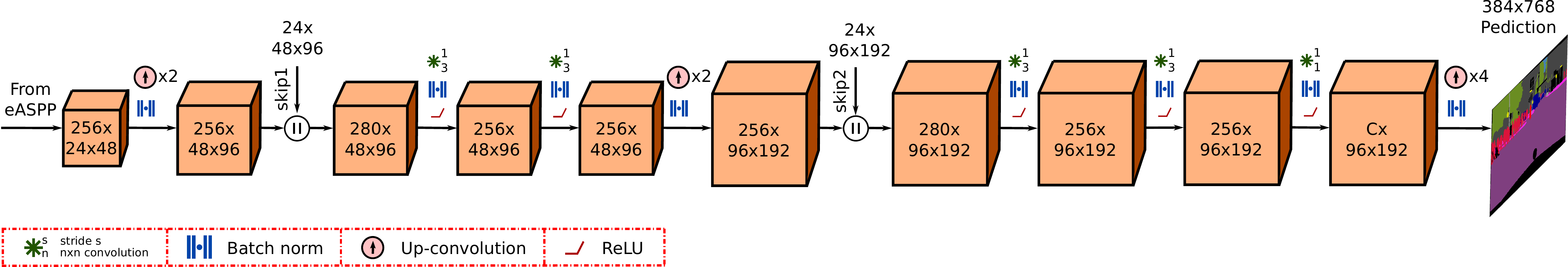} 
\caption{Our decoder consists of three upsampling stages that recover segmentation details using deconvolution layers and two skip refinement stages that fuse mid-level features from the encoder to improve the segmentation along object boundaries. Each skip refinement stage consists of concatenation of mid-level features with the upsampled decoder feature maps, followed by two $3\times3$ convolutions to improve the discriminability of the high-level features and the resolution of the refinement.}
\label{fig:decoder}
\end{figure*}

The biggest caveat of employing the ASPP is the extremely large amount of parameters and floating point operations per second (FLOPS) that it consumes. Each of the $3\times3$ convolutions have 256 filters, which in total for the entire ASPP amounts to $15.53\,\million$ parameters and $34.58\,\billion$ FLOPS which is prohibitively expensive. To address this problem, we propose an equivalent structure called eASPP that substantially reduces the computational complexity. Our proposed topology is based on two principles: cascading atrous convolutions and the bottleneck structure. Cascading atrous convolutions effectively enlarges the receptive field as the latter atrous convolution takes the output of the former atrous convolution. The receptive field size $F$ of an atrous convolution is be computed as
\begin{equation}
F = (r-1) \cdot (N-1)+N,
\end{equation}
where r is the dilation rate of the atrous convolution and $N$ is the filter size. When two atrous convolutions with the receptive field sizes as $F_1$ and $F_2$ are cascaded, the effective receptive field size is computed as
\begin{equation}
F_{eff} = F_1 + F_2 - 1.
\end{equation}

For example, if two atrous convolutions with filter size $F=3$ and dilation $r=3$ are cascaded, then each of the convolutions individually has a receptive field size of $7$, while the effective receptive field size of the second atrous convolution is $13$. Moreover, cascading atrous convolutions enables denser sampling of pixels in comparison to parallel atrous convolution with a larger receptive field. Therefore, by using both parallel and cascaded atrous convolutions in the ASPP, we can efficiently aggregate dense multiscale features with very large receptive fields. 

In order to reduce the number of parameters in the ASPP topology, we employ a bottleneck structure in the cascaded atrous convolution branches. The topology of our proposed eASPP shown in \figref{fig:aspp}(b) consists of five parallel branches similar to ASPP but the branches with the $3\times3$ atrous convolutions are replaced with our cascaded bottleneck branches. If $c$ is the number of channels in the $3 \times 3$ atrous convolution, we add a $1\times 1$ convolution with $c/4$ filters before the atrous convolution to squeeze only the most relevant information through the bottleneck. We then replace the $3 \times 3$ atrous convolution with two cascaded $3 \times 3$ atrous convolutions with $c/4$ filters, followed by another $1\times1$ convolution to restore the number of filters to $c$. The proposed eASPP only has $2.04\,\million$ parameters and consumes $3.62\,\billion$ FLOPS which accounts to a reduction of $87.87\%$ of parameters and $89.53\%$ of FLOPS in comparison to the ASPP. We evaluate our proposed eASPP in comparison to ASPP in the ablation study presented in \secref{sec:studyeASPP} and show that it achieves improved performance while being more than 10 times efficient in the number of parameters.

\subsection{Decoder}
\label{sec:decoder}

The output of the eASPP in our network is 16-times downsampled with respect to the input image and therefore it has to be upsampled back to the full input resolution. In our previous work~\citep{valada17icra}, we employed a simple decoder with two deconvolution layers and one skip refinement connection. Although the decoder was more effective in recovering the segmentation details in comparison to direct bilinear upsampling, it often produced disconnected segments while recovering the structure of thin objects such as poles and fences. In order to overcome this impediment, we propose a more effective decoder in this work. 

\begin{figure*}
\centering
\includegraphics[width=\linewidth]{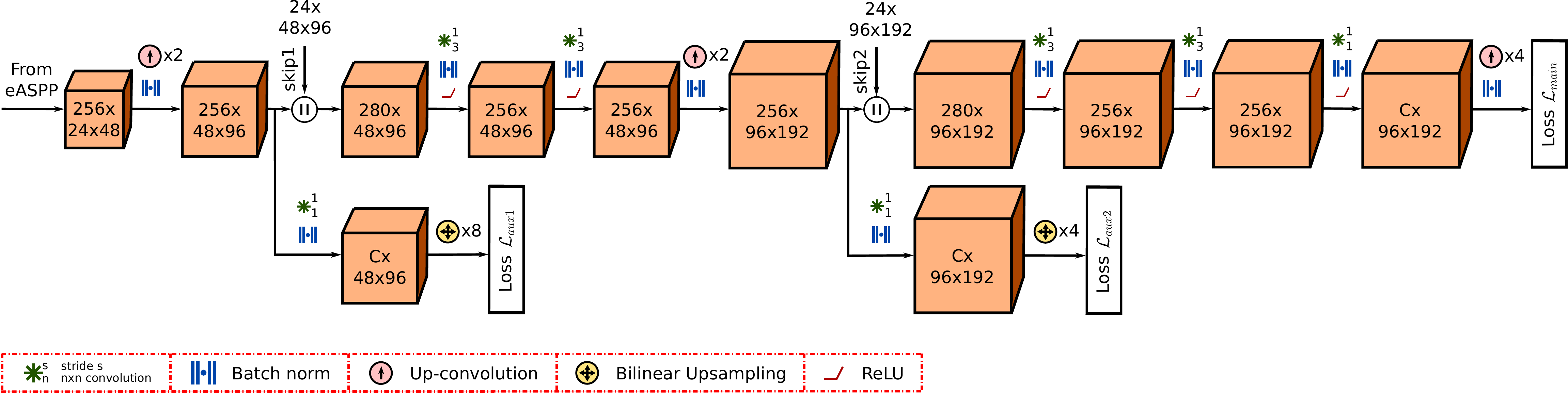}
\caption{Depiction of the two auxiliary softmax losses that we add before each skip refinement stage in the decoder in addition to the main softmax loss in the end of the decoder. The two auxiliary losses are weighed for balancing the gradient flow through all the previous layers. While testing the auxiliary branches are removed and only the main stream as shown in \figref{fig:decoder} is used.}
\label{fig:deepSupervision}
\end{figure*}

Our decoder shown in \figref{fig:decoder} consists of three stages. In the first stage, the output of the eASPP is upsampled by a factor of two using a deconvolution layer to obtain a coarse segmentation mask. The upsampled coarse mask is then passed through the second stage, where the feature maps are concatenated with the first skip refinement from \textit{Res3d}. The skip refinement consists of a $1\times1$ convolution layer to reduce the feature depth in order to not outweigh the encoder features. We experiment with varying number of feature channels in the skip refinement in the ablation study presented in \secref{sec:skipAblation}. The concatenated feature maps are then passed through two $3\times3$ convolutions to improve the resolution of the refinement, followed by a deconvolution layer that again upsamples the feature maps by a factor of two. This upsampled output is fed to the last decoder stage which resembles the previous stage consisting of concatenation with the feature maps from the second skip refinement from \textit{Res2c}, followed by two $3\times3$ convolution layers. All the convolutional and deconvolutional layers until this stage have 256 feature channels, therefore the output from the two $3\times3$ convolutions in the last stage is fed to a $1\times1$ convolution layer to reduce the number of feature channels to the number of object categories $C$. This output is finally fed to the last deconvolution layer which upsamples the feature maps by a factor of four to recover the original input resolution.

\subsection{Multiresolution Supervision}
\label{sec:deepSupervision}

Deep networks often have difficulty in training due to the intrinsic instability associated with learning using gradient descent which leads to exploding or vanishing gradient problems. As our encoder is based on the residual learning framework, shortcut connections in each unit help propagating the gradient more effectively. Another technique that can be used to mitigate this problem to a certain extent is by initializing the layers with pretrained weights, however our proposed eASPP and decoder layers still have to be trained from scratch which could lead to optimization difficulties. Recent deep architectures have proposed employing an auxiliary loss in the middle of encoder network~\citep{lee2015deeply,zhao2017pspnet}, in addition to the main loss towards the end of the network. However, as shown in the ablation study presented in \secref{sec:dissecting} this does not improve the performance of our network although it helps the optimization to converge faster.

Unlike previous approaches, in this work, we propose a multiresolution supervision strategy to both accelerate the training and improve the resolution of the segmentation. As described in the previous section, our decoder consists of three upsampling stages. We add two auxiliary loss branches at the end of the first and second stage after the deconvolution layer in addition to the main $\mathsf{softmax}$ loss $\mathcal{L}_{main}$ at the end of the decoder as shown in \figref{fig:deepSupervision}. Each auxiliary loss branch decreases the feature channels to the number of category labels $C$ using a $1\times1$ convolution with batch normalization and upsamples the feature maps to the input resolution using bilinear upsampling. We only use simple bilinear upsampling which does not contain any weights instead of a deconvolution layer in the auxiliary loss branches as our aim is to force the main decoder stream to improve its discriminativeness at each upsampling resolution so that it embeds multiresolution information while learning to upsample. We weigh the two auxiliary losses $\mathcal{L}_{aux1}$ and $\mathcal{L}_{aux2}$ to balance the gradient flow through all the previous layers. While testing, the auxiliary loss branches are discarded and only the main decoder stream is used. We experiment with different loss weightings in the ablation study presented in \secref{sec:skipAblation} and in \secref{sec:dissecting} we show that each of the auxiliary loss branches improves the segmentation performance in addition to speeding-up the training.

\subsection{Network Compression}
\label{sec:netCompress}

As we strive to design an efficient and compact semantic segmentation architecture that can be employed in resource constrained applications, we must ensure that the utilization of convolutional filters in our network is thoroughly optimized. Often, even the most compact networks have abundant neurons in deeper layers that do not significantly contribute to the overall performance of the model. Excessive convolutional filters not only increase the model size but also the inference time and the number of computing operations. These factors critically hinder the deployment of models in resource constrained real-world applications. Pruning of neural networks can be traced back to the 80s when \cite{lecun1990optimal} introduced a technique called Optimal Brain Damage for selectively pruning weights with a theoretically justified measure. Recently, several new techniques have been proposed for pruning weight matrices~\citep{wen2016learning, anwar2017structured, liu2017learning, li2016pruning} of convolutional layers as most of the computation during inference is consumed by them.

These approaches rank neurons based on their contribution and remove the low ranking neurons from the network, followed by fine-tuning of the pruned network. While the simplest neuron ranking method computes the $\ell^1$-norm of each convolutional filter~\citep{li2016pruning}, more sophisticated techniques have recently been proposed \citep{anwar2017structured, liu2017learning, molchanov2016pruning}. Some of these approaches are based on sparsity based regularization of network parameters which additionally increases the computational overhead during training~\citep{liu2017learning, wen2016learning}. Techniques have also been proposed for structured pruning of entire kernels with strided sparsity~\citep{anwar2017structured} that demonstrate impressive results for pruning small networks. However, their applicability to complex networks that are to be evaluated on large validation sets has not been explored due its heavy computational processing. Moreover, until a year ago these techniques were only applied to simpler architectures such as VGG~\citep{Simonyan14c} and AlexNet~\citep{krizhevsky2012imagenet}, as pruning complex deep architectures such as ResNets requires a holistic approach. Thus far, pruning of residual units has only been performed on convolutional layers that do not have an identity or shortcut connection as pruning them additionally requires pruning the added residual maps in the exact same configuration. Attempts to prune them in the same configuration have resulted in a significant drop in performance~\citep{li2016pruning}. Therefore, often only the
first and the second convolutional layers of a residual unit are pruned.

Our proposed AdapNet++ architecture has shortcut and skip connections both in the encoder as well the decoder. Therefore, in order to efficiently maximize the pruning of our network, we propose a holistic network-wide pruning technique that is invariant to the presence of skip or shortcut connections. Our proposed technique first involves pruning all the convolutional layers of a residual unit, followed by masking out the pruned indices of the last convolutional layer of a residual unit with zeros before the addition of the residual maps from the shortcut connection. As masking is performed after the pruning, we efficiently reduce the parameters and computing operations in a holistic fashion, while optimally pruning all the convolutional layers and preserving the shortcut or skip connections. After each pruning iteration, we fine-tune the network to recover any loss in accuracy. We illustrate this strategy adopting a recently proposed greedy criteria-based oracle pruning technique that incorporates a novel ranking method based on a first order Taylor expansion of the network cost function~\citep{molchanov2016pruning}. The pruning problem is framed as a combinatorial optimization problem such that when the weights $B$ of the network are pruned, the change in cost value will be minimal.
\begin{equation}
\min_{\mathcal{W'}} \lvert \mathcal{C}(\mathcal{T}|\mathcal{W}') - \mathcal{C}(\mathcal{T}|\mathcal{W}) \rvert \quad \text{s.t.} \quad \|{\mathcal{W}'}\|_0 \leq B,
\end{equation}
where $\mathcal{T}$ is the training set, $\mathscr{W}$ is the network parameters and $\mathcal{C}(\cdot)$ is the negative log-likelihood function. Based on Taylor expansion, the change in the loss function from removing a specific parameter can be approximated. Let $h_i$ be the output feature maps produced by parameter $i$ and $h_i=\{z_{0}^{1},z_{0}^{2},\cdots,z_{L}^{C_l}\}$. The output $h_i$ can be pruned by setting it to zero and the ranking can be given by
\begin{equation}
|\Delta\mathcal{C}(h_i)| = |\mathcal{C}(\mathcal{T},h_i=0) - \mathcal{C}(\mathcal{T},h_i)|,
\end{equation}

Approximating with Taylor expansion, we can write 
\begin{align}
\Theta_{TE}(h_i) &= |\Delta\mathcal{C}(h_i)| = |\mathcal{C}(\mathcal{T},h_i) - \frac{\delta \mathcal{C}}{\delta h_i} h_i - \mathcal{C}(\mathcal{T},h_i)| \nonumber\\
&= \left|\frac{\delta \mathcal{C}}{\delta h_i} h_i \right|. \\
\Theta_{TE} (z_{l}^{(k)}) &= \left| \frac{1}{M} \sum_m \frac{\delta \mathcal{C}}{\delta z_{l,m}^{(k)}} z_{l,m}^{(k)}\right|,
\end{align}
where $M$ is the length of the vectorized feature map. This ranking can be easily computed using the standard back-propagation computation as it requires the gradient of the cost function with respect to the activation and the product of the activation. Furthermore, in order to achieve adequate rescaling across layers, a layer-wise $\ell^2$-norm of the rankings is computed as
\begin{equation}
\hat{\Theta}(z_{l}^{(k)}) = \frac{\Theta (z_{l}^{(k)})}{\sqrt{\sum_j \Theta^2 (z_{l}^{(j)})}}.
\end{equation}

The entire pruning procedure can be summarized as follows: first the AdapNet++ network is trained until convergence using the training protocol described in \secref{sec:training}. Then the importance of the feature maps is evaluated using the aforementioned ranking method and subsequently the unimportant feature maps are removed. The pruned convolution layers that have shortcut connections are then masked at the indices where the unimportant feature maps are removed to maintain the shortcut connections. The network is then fine-tuned and the pruning process is reiterated until the desired trade-off between accuracy and the number of parameters has been achieved. We present results from pruning our AdapNet++ architecture in \secref{sec:resultsCompression}, where we perform pruning of both the convolutional and deconvolutional layers of our network in five stages by varying the threshold for the rankings. For each of these stages, we quantitatively evaluate the performance versus number of parameters trade-off obtained using our proposed pruning strategy in comparison to the standard approach.

%% file: sections/multimodal_fusion.tex
\section{Self-Supervised Model Adaptation}
\label{sec:multimodalFusion}

In this section, we describe our approach to multimodal fusion using our proposed self-supervised model adaptation (SSMA) framework. Our framework consists of three components: a modality-specific encoder as described in \secref{sec:encoder}, a decoder built upon the topology described in \secref{sec:decoder} and our proposed SSMA block for adaptively recalibrating and fusing modality-specific feature maps. In the following, we first formulate the problem of semantic segmentation from multimodal data, followed by a detailed description of our proposed SSMA units and finally we describe the overall topology of our fusion architecture.

We represent the training set for multimodal semantic segmentation as $\mathbfcal{T} = \{(I_n,K_n,M_n) \mid n = 1,\dots,N\}$, where $I_n=\{u_r \mid r=1,\dots,\rho\}$ denotes the input frame from modality $a$, $K_n=\{k_r \mid r=1,\dots,\rho\}$ denotes the corresponding input frame from modality $b$ and the groundtruth label is given by $M_n=\{m_{r} \mid r=1,\dots,\rho\}$, where $m_{r} \in \{1,...,C\}$ is the set of semantic classes. The image $I_n$ is only shown to the modality-specific encoder $E_a$ and similarly, the corresponding image $K_n$ from a complementary modality is only shown to the modality-specific encoder $E_b$. This enables each modality-specific encoder to specialize in a particular sub-space learning their own hierarchical representations individually. We assume that the input images $I_n$ and $K_n$, as well as the label $M_n$ have the same dimensions $\rho = H\times W$ and that the pixels are drawn as $i.i.d.$ samples following a categorical distribution. Let $\theta$ be the network parameters consisting of weights and biases. Using the classification scores $s_j$ at each pixel $u_r$, we obtain probabilities $\mathbf{P} = (p_1, \dots, p_C)$ with the $\mathsf{softmax}$ function such that
\begin{equation}
p_j(u_r,\theta \mid I_n,K_n) = \sigma\left(s_j\left(u_r, \theta\right)\right) = \frac{exp\left(s_j\left(u_r, \theta\right)\right)}{\sum^{C}_{k} exp\left(s_k\left(u_r,\theta\right)\right)}
\end{equation}
denotes the probability of pixel $u_r$ being classified with label $j$. The optimal $\theta$ is estimated by minimizing
\begin{equation}\label{eq:entropyloss}
\mathcal{L}_{seg}(\mathbfcal{T}, \theta) = - \sum^{N}_{n=1} \sum^{\rho}_{r=1} \sum^{C}_{j=1} \delta_{m_r, j}  \log p_j(u_r,\theta \mid I_n, K_n),
\end{equation}
for $(I_n, K_n, M_n) \in \mathbfcal{T}$, where $\delta_{m_r, j}$ is the Kronecker delta.

\subsection{SSMA Block}
\label{sec:ssmaBlock}

\begin{figure}
\centering
\includegraphics[width=\linewidth]{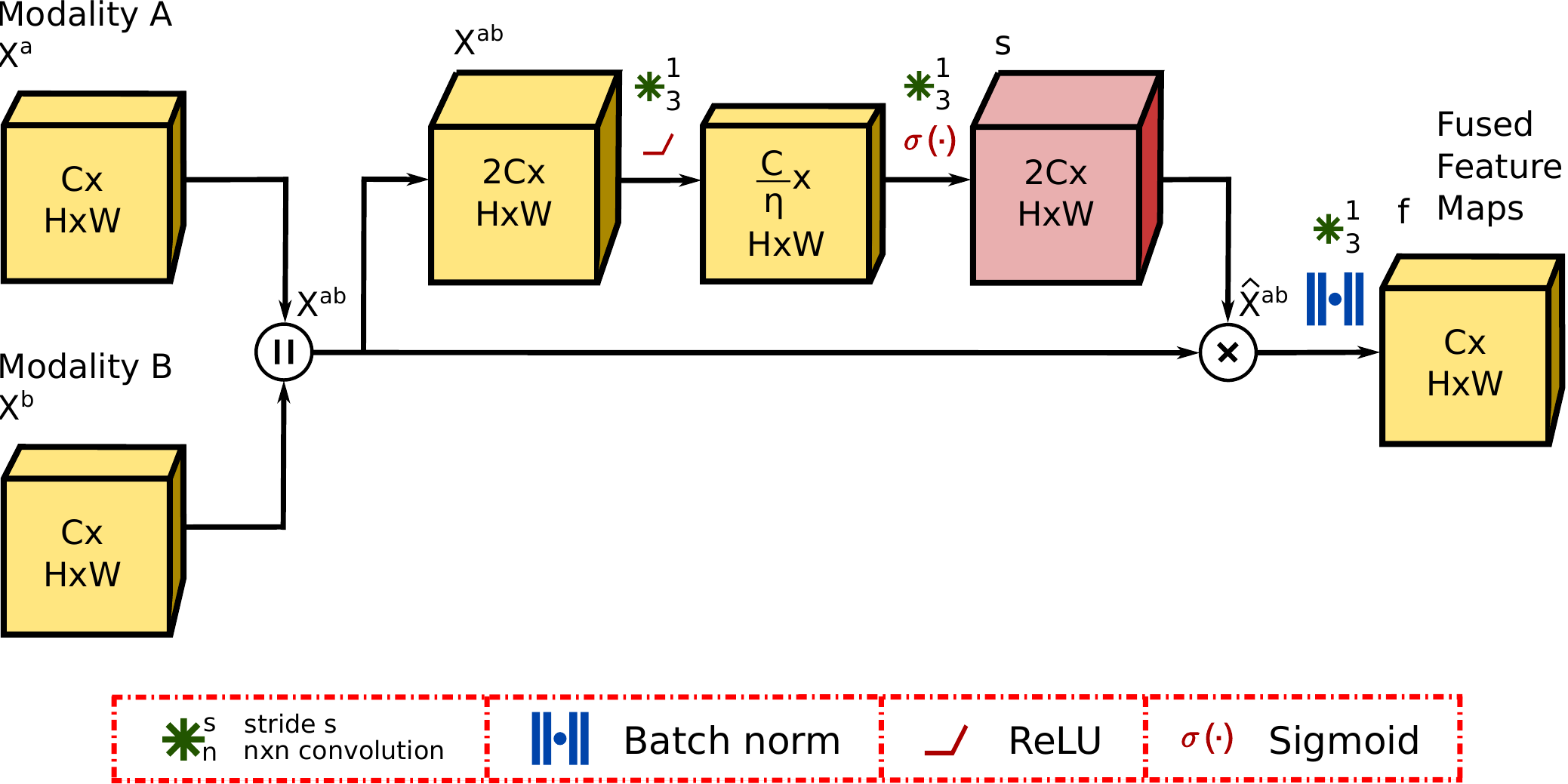}
\caption{The topology of our proposed SSMA unit that adaptively recalibrates and fuses modality-specific feature maps based on the inputs in order to exploit the more informative features from the modality-specific streams. $\eta$ denotes the bottleneck compression rate.}
\label{fig:ssmaUnit}
\end{figure}

In order to adaptively recalibrate and fuse feature maps from modality-specific networks, we propose a novel architectural unit called the SSMA block. The goal of the SSMA block is to explicitly model the correlation between the two modality-specific feature maps before fusion so that the network can exploit the complementary features by learning to selectively emphasize more informative features from one modality, while suppressing the less informative features from the other. We construct the topology of the SSMA block in a fully-convolutional fashion which empowers the network with the ability to emphasize features from a modality-specific network for only certain spatial locations or object categories, while emphasizing features from the complementary modality for other locations or object categories. Moreover, the SSMA block dynamically recalibrates the feature maps based on the input scene context.

The structure of the SSMA block is shown in \figref{fig:ssmaUnit}. Let $\mathbf{X}^{a} \in \mathbb{R}^{C \times H\times W}$ and $\mathbf{X}^{b} \in \mathbb{R}^{C \times H\times W}$ denote the modality-specific feature maps from modality $A$ and modality $B$ respectively, where $C$ is the number of feature channels and $H \times W$ is the spatial dimension. First, we concatenate the modality-specific feature maps $\mathbf{X}^{a}$ and $\mathbf{X}^{b}$ to yield $\mathbf{X}^{ab} \in \mathbb{R}^{2\cdot C \times H\times W}$. We then employ a recalibration technique to adapt the concatenated feature maps before fusion. In order to achieve this, we first pass the concatenated feature map $\mathbf{X}^{ab}$ through a bottleneck consisting of two $3\times 3$ convolutional layers for dimensionality reduction and to improve the representational capacity of the concatenated features. The first convolution has weights $\mathcal{W}_1 \in \mathbb{R}^{\frac{1}{\eta} \cdot C\times H\times W}$ with a channel reduction ratio $\eta$ and a non-linearity function $\delta(\cdot)$. We use ReLU for the non-linearity, similar to the other activations in the encoders and experiment with different reductions ratios in \secref{sec:fusionAblationDetails}. Note that we omit the bias term to simplify the notation. The subsequent convolutional layer with weights $\mathcal{W}_2 \in \mathbb{R}^{2 \cdot C\times H\times W}$ increases the dimensionality of the feature channels back to concatenation dimension $2C$ and a sigmoid function $\sigma(\cdot)$ scales the dynamic range of the activations to the $[0,1]$ interval. This can be represented as
\begin{align}
\mathbf{s} &= F_{ssma}(\mathbf{X}^{ab};\mathcal{W}) = \sigma \left(g\left(\mathbf{X}^{ab}; \mathcal{W}\right)\right) \nonumber\\
&= \sigma \left(\mathcal{W}_2 \delta \left(\mathcal{W}_1 \mathbf{X}^{ab}\right)\right).
\end{align}

The resulting output $\mathbf{s}$ is used to recalibrate or emphasize/de-emphasize regions in $\mathbf{X}^{ab}$ as
\begin{equation}
\hat{\mathbf{X}}^{ab} = F_{scale} (\mathbf{X}^{ab};\mathbf{s}) = \mathbf{s} \circ \mathbf{X}^{ab},
\end{equation}
where $F_{scale} (\mathbf{X}^{ab},\mathbf{s})$ denotes Hadamard product of the feature maps $\mathbf{X}^{ab}$ and the matrix of scalars $\mathbf{s}$ such that each element $x_{c,i,j}$ in $\mathbf{X}^{ab}$ is multiplied with a corresponding activation $s_{c,i,j}$ in $\mathbf{s}$ with $c \in \{1,2,\dots,2C\}$, $i \in \{1,2,\dots,H\}$ and $j \in \{1,2,\dots,W\}$. The activations $\mathbf{s}$ adapt to the concatenated input feature map $\mathbf{X}^{ab}$, enabling the network to weigh features element-wise spatially and across the channel depth based on the multimodal inputs $I_n$ and $K_n$. With new multimodal inputs, the network dynamically weighs and reweighs the feature maps in order to optimally combine complementary features. Finally, the recalibrated feature maps $\hat{\mathbf{X}}^{ab}$ are passed through a $3\times 3$ convolution with weights $\mathcal{W}_3 \in \mathbb{R}^{C \times H\times W}$ and a batch normalization layer to reduce the feature channel depth and yield the fused output $\mathbf{f}$ as
\begin{equation} \label{eq:fusion}
\mathbf{f} = F_{fused}(\hat{\mathbf{X}}^{ab};\mathcal{W}) = g(\hat{\mathbf{X}}^{ab};\mathcal{W}) = \mathcal{W}_{3}\hat{\mathbf{X}}^{ab}.
\end{equation}

\begin{figure}
\centering
\includegraphics[width=\linewidth]{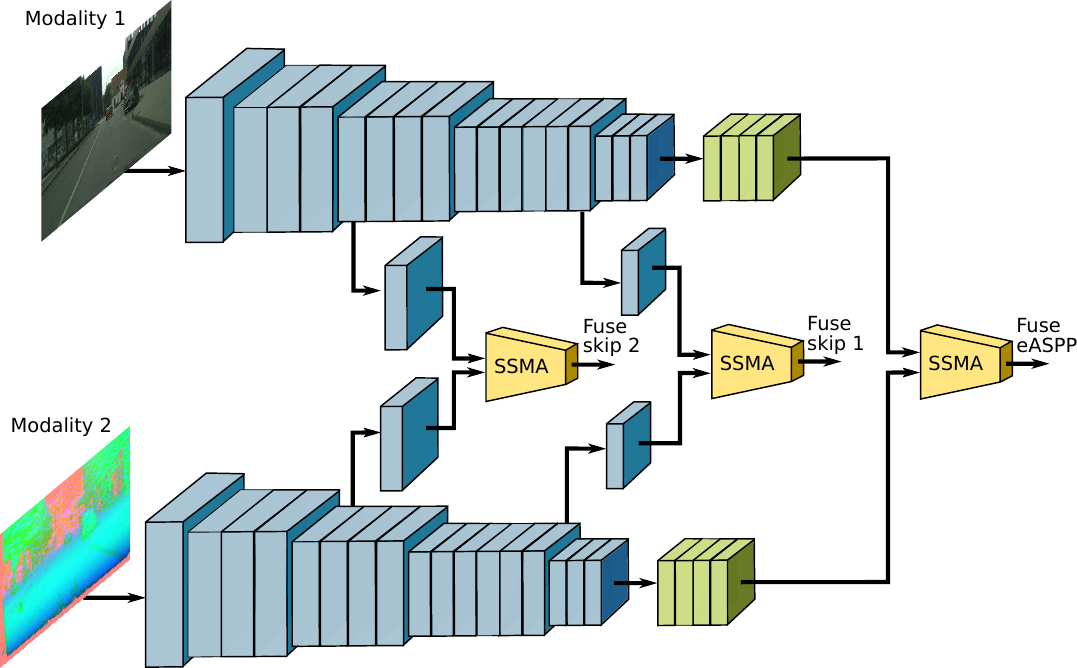}
\caption{Topology of our Adapnet++ encoder for multimodal fusion. The encoder employs a late fusion technique to fuse feature maps from modality-specific streams using our proposed SSMA block. The SSMA block is employed to fuse the latent features from the eASPP as well as the feature maps from the skip refinements.}
\label{fig:fusionEncoder}
\end{figure}

\begin{figure*}
\centering
\includegraphics[width=\linewidth]{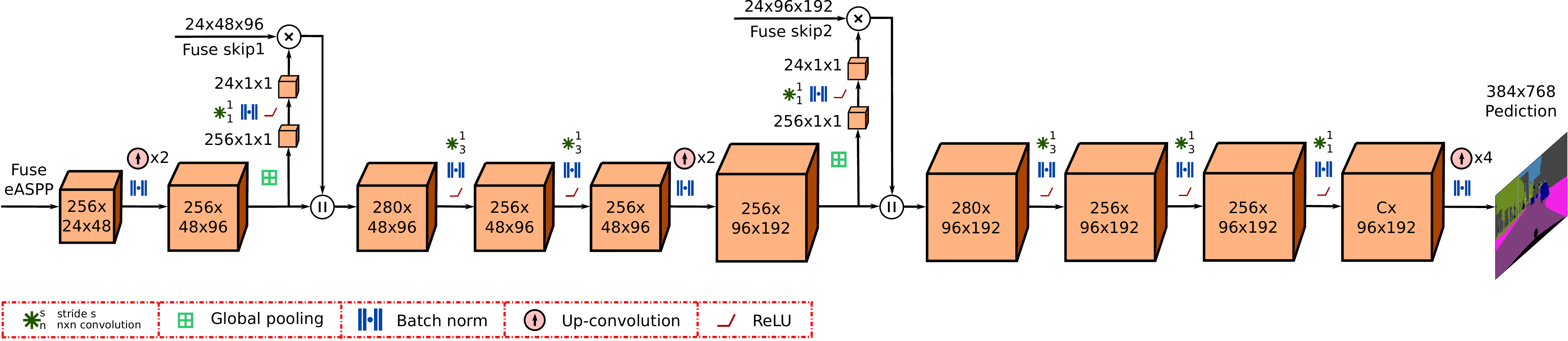}
\caption{Topology of the modified AdapNet++ decoder used for multimodal fusion. We propose a mechanism to better correlate the fused mid-level skip refinement features with the high-level decoder feature before integrating into the decoder. The correlation mechanism is depicted following the fuse skip connections.}
\label{fig:fusionDecoder}
\end{figure*}

As described in the following section, we employ our proposed SSMA block to fuse modality-specific feature maps both at intermediate stages of the network and towards the end of the encoder. Although we utilize a bottleneck structure to conserve the number of parameters consumed, further reduction in the parameters can be achieved by replacing the $3\times 3$ convolution layers with $1\times 1$ convolutions, which yields comparable performance. We also remark that the SSMA blocks can be used for multimodal fusion in other tasks such as scene classification as shown in \secref{sec:multimodalGeneralization}.

\subsection{Fusion Architecture}
\label{sec:approachFusion}

We propose a framework for multimodal semantic segmentation using a modified version of our AdapNet++ architecture and the proposed SSMA blocks. For simplicity, we consider the fusion of two modalities, but the framework can be easily extended to arbitrary number of modalities. The encoder of our framework shown in \figref{fig:fusionEncoder} contains two streams, where each stream is based on the encoder topology described in \secref{sec:encoder}. Each encoder stream is modality-specific and specializes in a particular sub-space. In order to fuse the feature maps from both streams, we adopt a combination of mid-level and late fusion strategy in which we fuse the latent representations of both encoders using the SSMA block and pass the fused feature map to the first decoder stage. We denote this as latent SSMA fusion as it takes the output of the eASPP from each modality-specific encoder as input. We set the reduction ratio $\eta = 16$ in the latent SSMA. As the AdapNet++ architecture contains skip connections for high-resolution refinement, we employ an SSMA block at each skip refinement stage after the $1\times1$ convolution as shown in \figref{fig:fusionEncoder}. As the $1\times 1$ convolutions reduce the feature channel depth to 24, we only use a reduction ratio $\eta = 6$ in the two skip SSMAs as identified from the ablation experiments presented in \secref{sec:fusionAblationDetails}.

In order to upsample the fused predictions, we build upon our decoder described in \secref{sec:decoder}. The main stream of our decoder resembles the topology of the decoder in our AdapNet++ architecture consisting of three upsampling stages. The output of the latent SSMA block is fed to the first upsampling stage of the decoder. Following the AdapNet++ topology, the outputs of the skip SSMA blocks would be concatenated into the decoder at the second and third upsampling stages (\textit{skip1} after the first deconvolution and \textit{skip2} after the second deconvolution). However, we find that concatenating the fused mid-level features into the decoder does not substantially improve the resolution of the segmentation, as much as in the unimodal AdapNet++ architecture. We hypothesise that directly concatenating the fused mid-level features and fused high-level features causes a feature localization mismatch as each SSMA block adaptively recalibrates at different stages of the network where the resolution of the feature maps and channel depth differ by one half of their dimensions. Moreover, training the fusion network end-to-end from scratch also contributes to this problem as without initializing the encoders with modality-specific pre-trained weights, concatenating the uninitialized mid-level fused encoder feature maps into the decoder does not yield any performance gains, rather it hampers the convergence.

With the goal of mitigating this problem, we propose two strategies. In order to facilitate better fusion, we adopt a multi-stage training protocol where we first initialize each encoder in the fusion architecture with pre-trained weights from the unimodal AdapNet++ model. We describe this procedure in \secref{sec:SSMAtraining}. Secondly, we propose a mechanism to better correlate the mid and high-level fused features before concatenation in  the decoder. We propose to weigh the fused mid-level skip features with the spatially aggregated statistics of the high-level decoder features before the concatenation. Following the notation convention, we define $\mathbf{D} \in \mathbb{R}^{C \times H\times W}$ as the high-level decoder feature map before the skip concatenation stage. A feature statistic $\mathbf{s} \in \mathbb{R}^C$ is produced by projecting $\mathbf{D}$ along the spatial dimensions $H\times W$ using a global average pooling layer as
\begin{equation}
s_c = F_{shrink}(d_c) = \frac{1}{H\times W} \sum^{H}_{i=1}\sum^{W}_{j=1} d_c(i,j),
\end{equation}
where $s_c$ represents a statistic or a local descriptor of the $c^{th}$ element of $\mathbf{D}$. We then reduce the number of feature channels in $\mathbf{s}$ using a $1\times 1$ convolution layer with weights $\mathcal{W}_{4} \in \mathbb{R}^{C \times H\times W}$, batch normalization and an ReLU activation function $\delta$ to match the channels of the fused mid-level feature map $\mathbf{f}$, where $\mathbf{f}$ is computed as shown in \eqref{eq:fusion}. We can represent resulting output as
\begin{equation}
z = F_{reduce}(\mathbf{s};\mathcal{W}) = \delta(\mathcal{W}_{4}\mathbf{s}).
\end{equation}

Finally, we weigh the fused mid-level feature map $\mathbf{f}$ with the reduced aggregated descriptors $\mathbf{z}$ using channel-wise multiplication as
\begin{equation}
\hat{\mathbf{f}} = F_{loc}(\mathbf{f}_{c};z_c) = \left(z_1 \mathbf{f}_1, z_2 \mathbf{f}_2,\dots,z_c \mathbf{f}_c \right).
\end{equation}

As shown in \figref{fig:fusionDecoder}, we employ the aforementioned mechanism to the fused feature maps from skip1 SSMA as well as skip2 SSMA and concatenate their outputs with the decoder feature maps at the second and third upsampling stages respectively. We find that this mechanism guides the fusion of mid-level skip refinement features with the high-level decoder feature more effectively than direct concatenation and yields a notable improvement in the resolution of the segmentation output.

%% file: sections/experimental_results.tex
\section{Experimental Results}
\label{sec:experimentalResults}

In this section, we first describe the datasets that we benchmark on, followed by comprehensive quantitative results for unimodal segmentation using our proposed AdapNet++ architecture in \secref{sec:benchm} and the results for model compression in \secref{sec:resultsCompression}. We then present detailed ablation studies that describe our architectural decisions in \secref{sec:ablation}, followed by the qualitative unimodal segmentation results in \secref{sec:unimodalQual}. We present the multimodal fusion benchmarking experiments with the various modalities contained in the datasets in \secref{sec:multimodalFusionEx} and the ablation study on our multimodal fusion architecture in \secref{sec:multimodalFusionAblation}. We finally present the qualitative multimodal segmentation results in \secref{sec:qualitative} and in challenging perceptual conditions in \secref{sec:seasonsQual}. 

All our models were implemented using the TensorFlow \citep{tensorflow2015-whitepaper} deep learning library and the experiments were carried out on a system with an Intel Xeon E5 with $2.4\giga\hertz$ and an NVIDIA TITAN X GPU. We primarily use the standard Jaccard Index, also known as the intersection-over-union (IoU) metric to quantify the performance. The IoU for each object class is computed as IoU = TP/(TP + FP + FN), where TP, FP and FN correspond to true positives, false positives and false negatives respectively. We report the mean intersection-over-union (mIoU) metric for all the models and also the pixel-wise accuracy (Acc), average precision (AP), global intersection-over-union (gIoU) metric, false positive rate (FPR), false negative rate (FNR) in the detailed analysis. All the metrics are computed as defined in the PASCAL VOC challenge~\citep{everingham2015pascal} and additionally, the gIoU metric is computed as $\text{gIoU} = \sum_{\text{C}} \text{TP}_{\text{C}}/\sum_{\text{C}} (\text{TP}_{\text{C}} + \text{FP}_{\text{C}} + \text{FN}_{\text{C}})$, where C is the number of object categories. The implementations of our proposed architectures are publicly available at \url{https://github.com/DeepSceneSeg} and a live demo can be viewed at \url{http://deepscene.cs.uni-freiburg.de}.

\subsection{Training Protocol}
\label{sec:training}

In this section, we first describe the procedure that we employ for training our proposed AdapNet++ architecture, followed by the protocol for training the SSMA fusion scheme. We then detail the various data augmentations that we perform on the training set.

\subsubsection{AdapNet++ Training}
\label{sec:adapnetTraining}

We train our network with an input image of resolution $768\times384$ pixels, therefore we employ bilinear interpolation for resizing the RGB images and the nearest-neighbor interpolation for the other modalities as well as the groundtruth labels. We initialize the encoder section of the network with weights pre-trained on the ImageNet dataset~\citep{deng2009imagenet}, while we use the He initialization~\citep{he2015delving} for the other convolutional and deconvolutional layers. We use the Adam solver for optimization with $\beta_1=0.9, \beta_2=0.999$ and $\epsilon=10^{-10}$. We train our model for 150K iterations using an initial learning rate of $\lambda_0 = 10^{-3}$ with a mini-batch size of $8$ and a dropout probability of $0.5$. We use the cross-entropy loss function and set the weights $\lambda_1 =0.6$ and $\lambda_2 =0.5$ to balance the auxiliary losses. The final loss function can be given as $\mathcal{L} = \mathcal{L}_{main} + \lambda_1 \mathcal{L}_{aux1} + \lambda_2 \mathcal{L}_{aux2}$.

\subsubsection{SSMA Training}
\label{sec:SSMAtraining}

We employ a multi-stage procedure for training the multimodal models using our proposed SSMA fusion scheme. We first train each modality-specific Adapnet++ model individually using the training procedure described in \secref{sec:adapnetTraining}. In the second stage, we leverage transfer learning to train the joint fusion model in the SSMA framework by initializing only the encoders with the weights from the individual modality-specific encoders trained in the previous stage. We then set the learning rate of the encoder layers to $\lambda_0 = 10^{-4}$ and the decoder layers to $\lambda_0 = 10^{-3}$, and train the fusion model with a mini-batch of 7 for a maximum of 100K iterations. This enables the SSMA blocks to learn the optimal combination of multimodal feature maps from the well trained encoders, while slowly adapting the encoder weights to improve the fusion. In the final stage, we fix the learning rate of the encoder layers to $\lambda_0 = 0$ while only training the decoder and the SSMA blocks with a learning rate of $\lambda_0 = 10^{-5}$ and a mini-batch size of 12 for 50K iterations. This enables us to train the network with a larger batch size, while focusing more on the upsampling stages to yield the high-resolution segmentation output.

\subsubsection{Data Augmentation}

The training of deep networks can be significantly improved by expanding the dataset to introduce more variability. In order to achieve this, we apply a series of augmentation strategies randomly on the input data while training. The augmentations that we apply include rotation ($-13\degree$ to $13\degree$), skewing ($0.05$ to $0.10$), scaling ($0.5$ to $2.0$), vignetting ($210$ to $300$), cropping ($0.8$ to $0.9$), brightness modulation ($-40$ to $40$), contrast modulation ($0.5$ to $1.5$) and flipping.

\subsection{Datasets}

We evaluate our proposed AdapNet++ architecture on five publicly available diverse scene understanding benchmarks ranging from urban driving scenarios to unstructured forested scenes and cluttered indoor environments. The datasets were particularly chosen based on the criteria of containing scenes with challenging perceptual conditions including rain, snow, fog, night-time, glare, motion blur and other seasonal appearance changes. Each of the datasets contain multiple modalities that we utilize for benchmarking our fusion approach. We briefly describe the datasets and their constituting semantic categories in this section.

{\parskip=5pt
\noindent\textbf{Cityscapes:} The Cityscapes dataset~\citep{Cordts2016cityscapes} is one of the largest labeled RGB-D dataset for urban scene understanding. Being one of the standard benchmarks, it is highly challenging as it contains images of complex urban scenes, collected from over 50 cities during varying seasons, lighting and weather conditions. The images were captured using a automotive-grade $22\centi\meter$ baseline stereo camera at a resolution of $2048\times1024$ pixels. The dataset contains 5000 finely annotated images, of which 2875 are provided for training, 500 are provided for validation and 1525 are used for testing. As a supplementary training set, 20000 coarse annotations are also provided. The testing images are not publicly released, they are used by the evaluation server for benchmarking on 19 semantic object categories. We report results on the full 19 class label set for both the validation and test sets. Additionally, in order to facilitate comparison with previous fusion approaches we also report results on the reduced 11 class label set consisting of: \textit{sky, building, road, sidewalk, fence, vegetation, pole, car/truck/bus, traffic sign, person, rider/bicycle/motorbike} and \textit{background}.}

\begin{figure}
\centering
\footnotesize
\setlength{\tabcolsep}{0.2cm}
\begin{tabular}{p{3.68cm} p{3.68cm}}
\includegraphics[width=\linewidth]{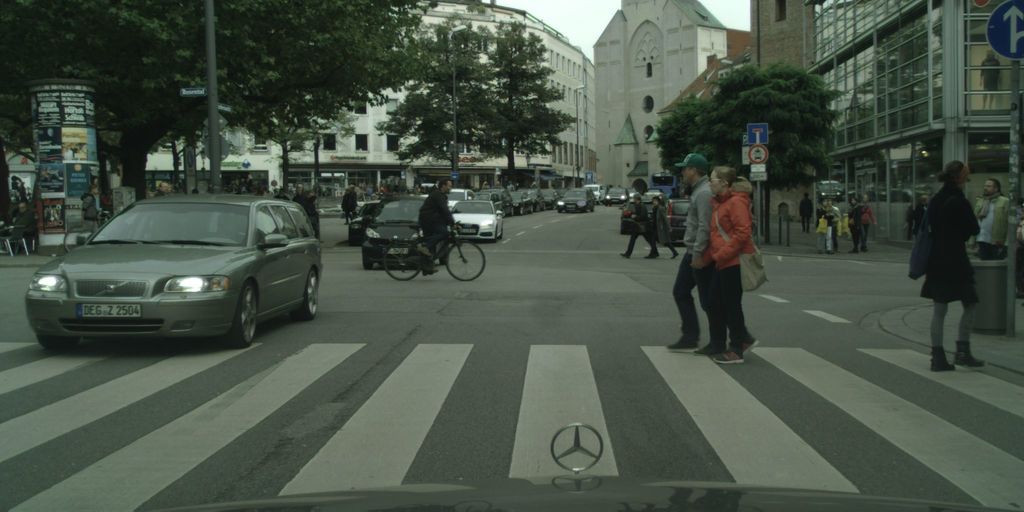} & 
\includegraphics[width=\linewidth]{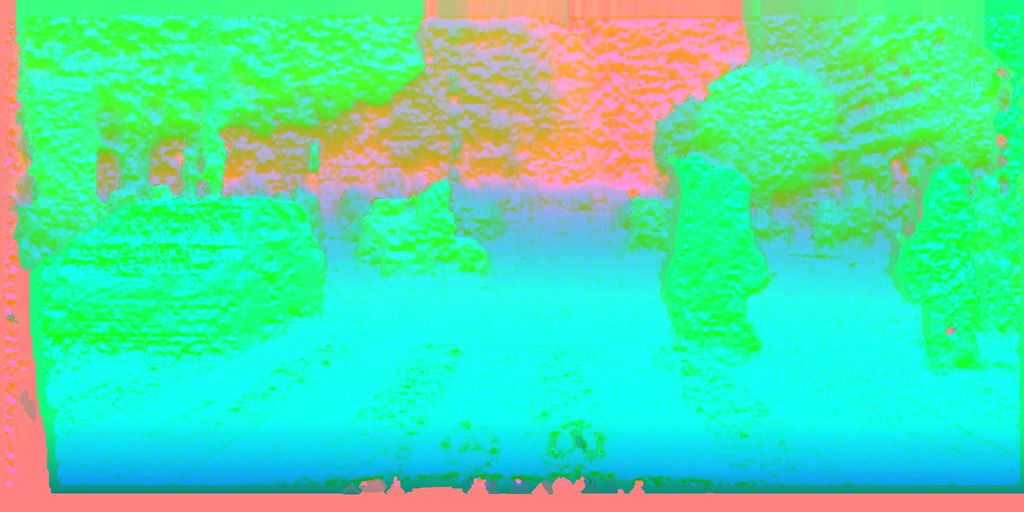} \\
(a) RGB & (b) HHA \\ 
\includegraphics[width=\linewidth]{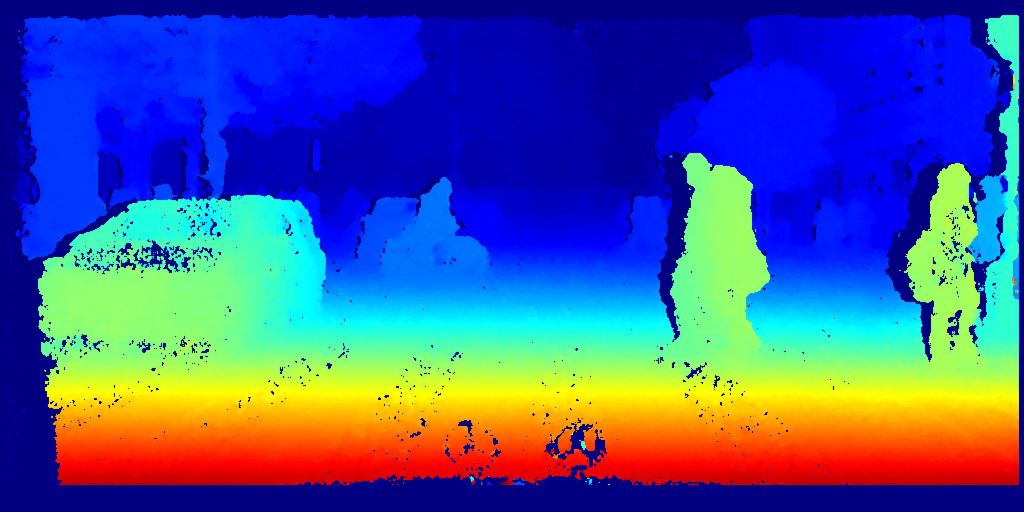} &
\includegraphics[width=\linewidth]{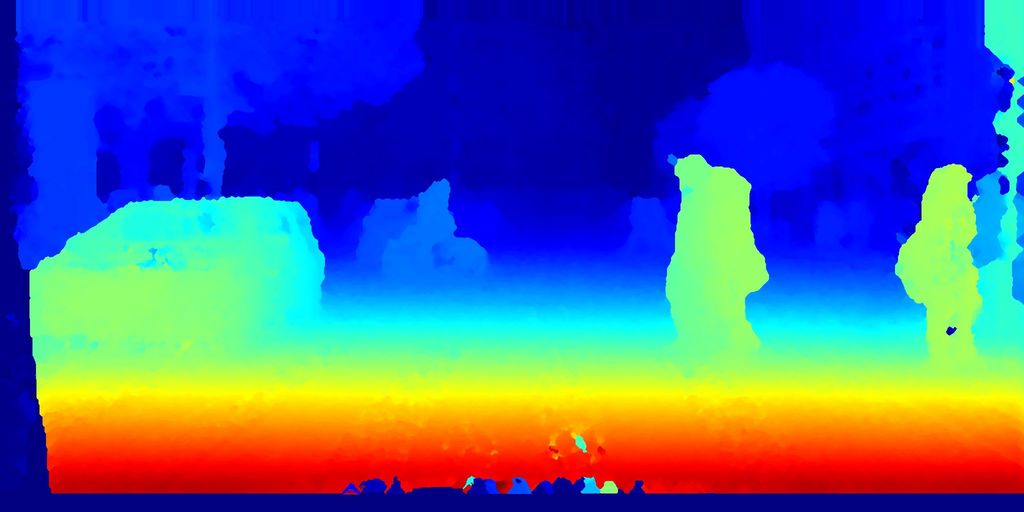} \\
(c) Depth & (d) Depth Filled \\ 
\end{tabular}
\caption{Example image from the Cityscapes dataset showing a complex urban scene with many dynamic objects and the corresponding depth map representations.}
\label{fig:cityscapesEx}
\end{figure}

In our previous work~\citep{valada17icra}, we directly used the colorized depth image as input to our network. We converted the stereo disparity map to a three-channel colorized depth image by normalizing and applying the standard jet color map \figref{fig:cityscapesEx}(a) and \figref{fig:cityscapesEx}(c) show an example image and the corresponding colorized depth map from the dataset. However, as seen in the figure, the depth maps have considerable amount of noise and missing depth values due to occlusion, which are undesirable especially when utilizing depth maps as an input modality for pixel-wise segmentation. Therefore, in this work, we employ a recently proposed state-of-the-art fast depth completion technique~\citep{ku2018defense} to fill any holes that may be present. The resulting filled depth map is shown in \figref{fig:cityscapesEx}(d). The depth completion algorithm can easily be incorporated into our pipeline as a preprocessing step as it only requires $11\milli\second$ while running on the CPU and it can be further parallelized using a GPU implementation. Additionally, \cite{gupta2014learning} proposed an alternate representation of the depth map known as the HHA encoding to enable DCNNs to learn more effectively. The authors demonstrate that the HHA representation encodes properties of geocentric pose that emphasizes on complementary discontinuities in the image which are extremely hard for the network to learn, especially from limited training data. This representation also yields a three-channel image consisting of: horizontal disparity, height above ground, and the angle between the local surface normal of a pixel and the inferred gravity direction. The resulting channels are then linearly scaled and mapped to the 0 to 255 range. However, it is still unclear if this representation enables the network to learn features complementary to that learned from visual RGB images as different works show contradicting results \citep{fusenet2016accv,gupta2014learning,eitel15iros}. In this paper, we perform in-depth experiments with both the jet colorized and the HHA encoded depth map on a larger and more challenging dataset than previous works to investigate the utility of these encodings.

\begin{figure}
\centering
\setlength{\tabcolsep}{0.2em}
\begin{tabular}{p{2.68cm} p{2.68cm} p{2.68cm}}
\includegraphics[width=\linewidth]{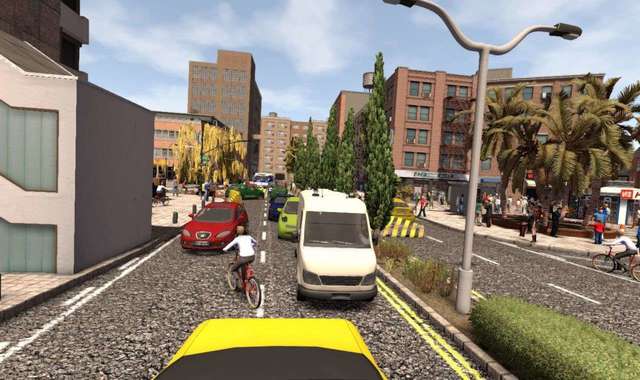} & \includegraphics[width=\linewidth]{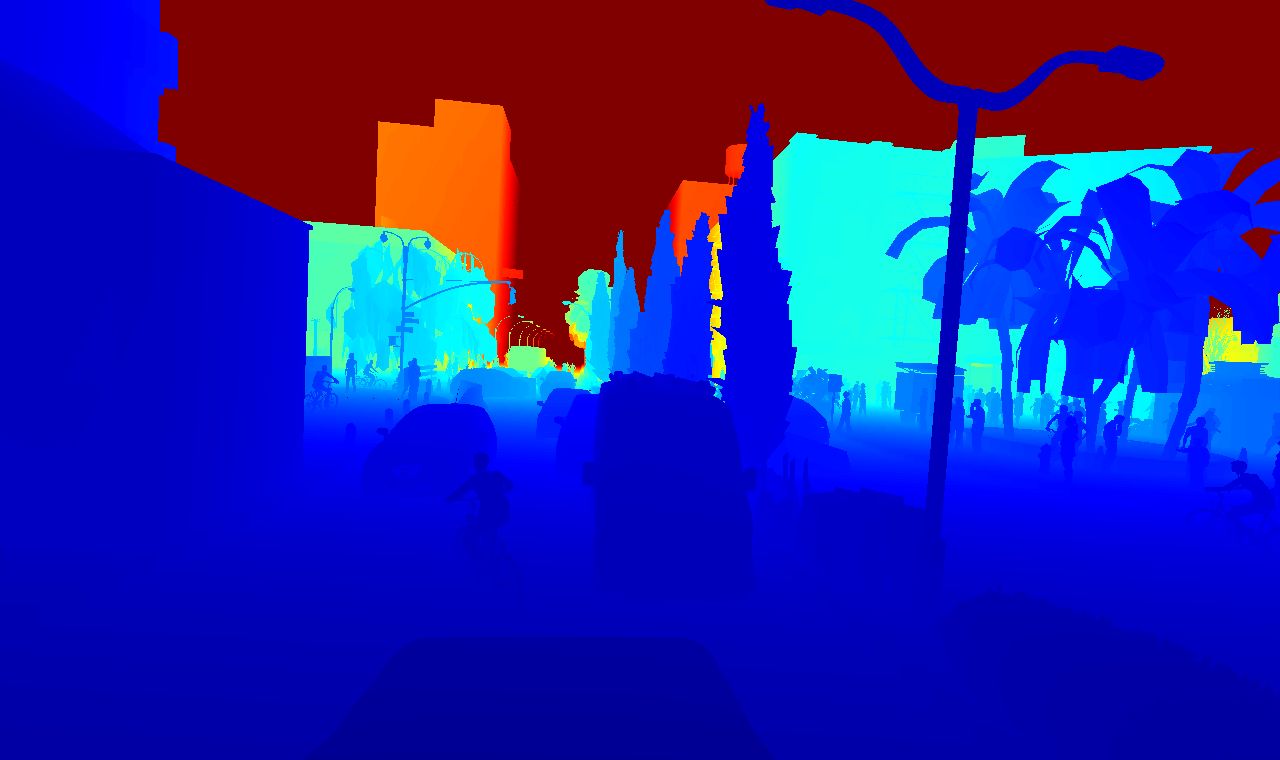} &
\includegraphics[width=\linewidth]{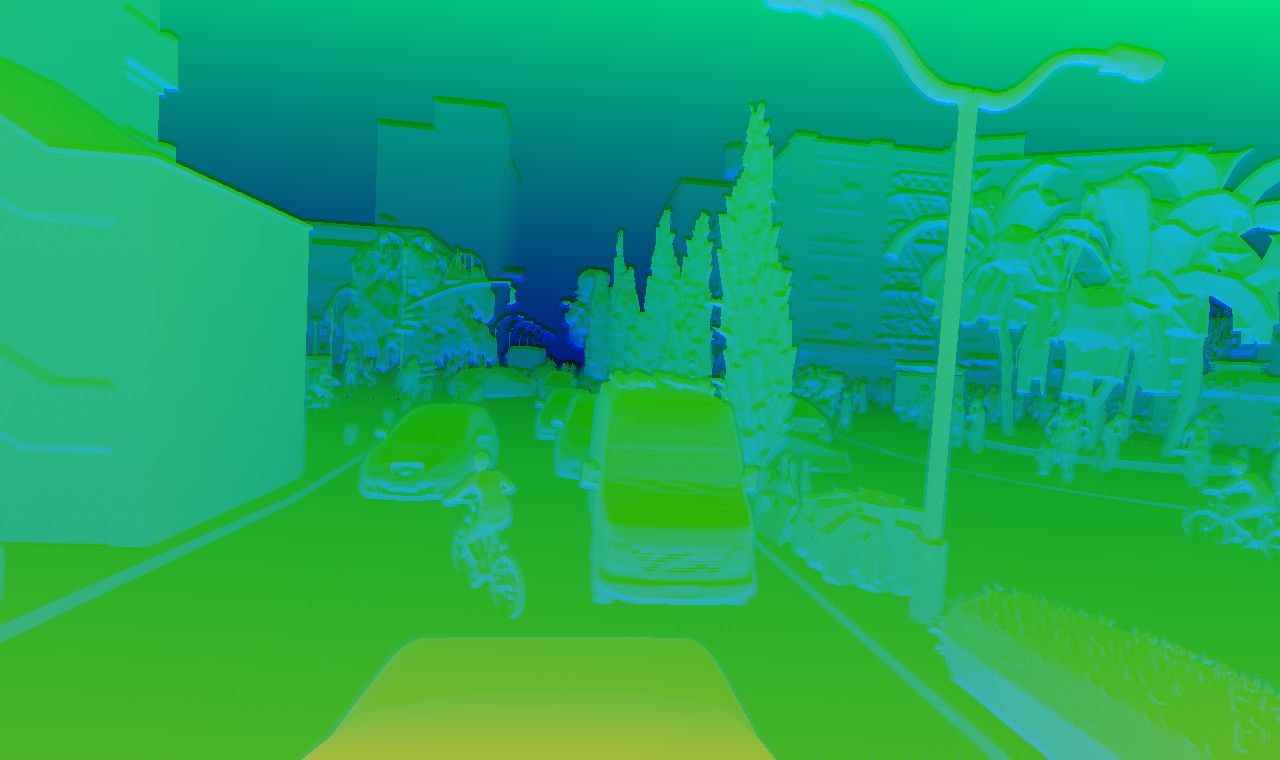} \\
\small{(a) RGB} & \small{(b) Depth} & \small{(c) HHA} \\ 
\end{tabular}
\caption{Example image from the Synthia dataset showing an outdoor urban scene and the corresponding depth map representations.}
\label{fig:synthiaEx}
\end{figure}

{\parskip=5pt
\noindent\textbf{Synthia:} The Synthia dataset~\citep{roscvpr16} is a large-scale urban outdoor dataset that contains photo realistic images and depth data rendered from a virtual city built using the Unity engine. It consists of several annotated label sets. In this work, we use the Synthia-Rand-Cityscapes and the video sequences which have images of resolution $1280\times760$ with a $100\degree$ horizontal field of view. This dataset is of particular interest for benchmarking the fusion approaches as it contains diverse traffic situations under different weather conditions. Synthia-Rand-Cityscapes consists of 9000 images and the sequences contain 8000 images with groundtruth labels for 12 classes. The categories of object labels are the same as the aforementioned Cityscapes label set.}

\begin{figure}
\centering
\footnotesize
\setlength{\tabcolsep}{0.2em}
\begin{tabular}{p{2.68cm} p{2.68cm} p{2.68cm}}
\includegraphics[width=\linewidth]{figures/train_132_rgb} & \includegraphics[width=\linewidth]{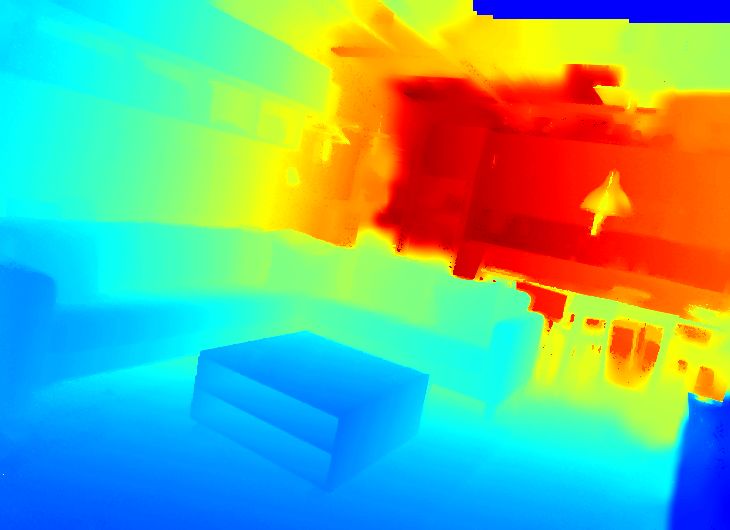} &
\includegraphics[width=\linewidth]{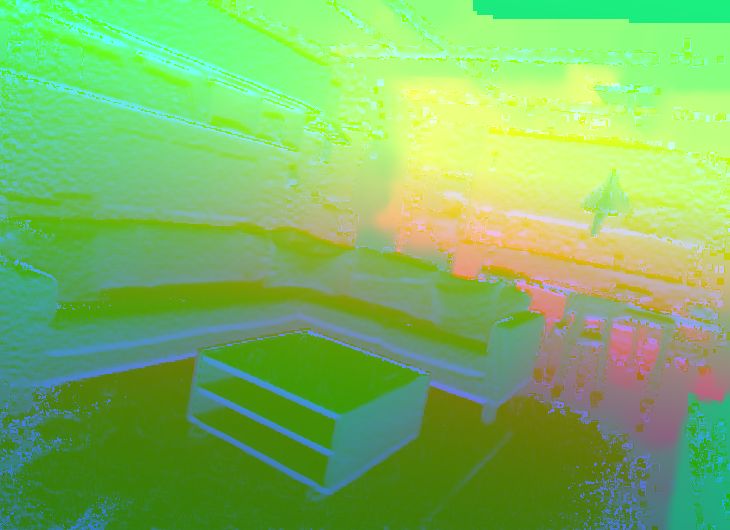} \\
(a) RGB & (b) Depth & (c) HHA \\ 
\end{tabular}
\caption{Example image from the SUN RGB-D dataset showing an indoor scene and the corresponding depth map representations.}
\label{fig:sunrgbdEx}
\end{figure}

{\parskip=5pt
\noindent\textbf{SUN RGB-D:} The SUN RGB-D dataset~\citep{song2015sun} is one of the most challenging indoor scene understanding benchmarks to date. It contains 10,335 RGB-D images that were captured with four different types of RGB-D cameras (Kinect~V1, Kinect~V2, Xtion and RealSense) with different resolutions and fields of view. This benchmark also combines several other datasets including 1449 images from the NYU~Depth~v2~\citep{SilbermanECCV12}, 554 images from the Berkeley~B3DO~\citep{janoch2013category} and 3389 images from the SUN3D~\citep{xiao2013sun3d}. We use the original train-val split consisting of 5285 images for training and 5050 images for testing. We use the refined in-painted depth images from the dataset that were processed using a multi-view fusion technique. However, some refined depth images still have missing depth values at distances larger than a few meters. Therefore, as mentioned in previous works~\citep{fusenet2016accv}, we exclude the 587 training images that were captured using the RealSense RGB-D camera as they contain a significant amount of invalid depth measurements that are further intensified due to the in-painting process.}

This dataset provides pixel-level semantic annotations for 37 categories, namely: \textit{wall, floor, cabinet, bed, chair, sofa, table, door, window, bookshelf, picture, counter, blinds, desk, shelves, curtain, dresser, pillow, mirror, floor mat, clothes, ceiling, books, fridge, tv, paper, towel, shower curtain, box, whiteboard, person, night stand, toilet, sink, lamp, bathtub} and \textit{bag}. Although we benchmark on all the object categories, 16 out of the 37 classes are rarely present in the images and about $0.25\%$ of the pixels are not assigned to any of the classes, making it extremely unbalanced. Moreover, as each scene contains many different types of objects, they are often partially occluded and may appear completely different in the test images.

\begin{figure}
\centering
\footnotesize
\setlength{\tabcolsep}{0.1cm}
\begin{tabular}{p{1.3cm} p{1.3cm} p{1.3cm}p{1.3cm} p{1.3cm} p{1.3cm}}
\multicolumn{2}{p{2.6cm}}{\includegraphics[width=\linewidth]{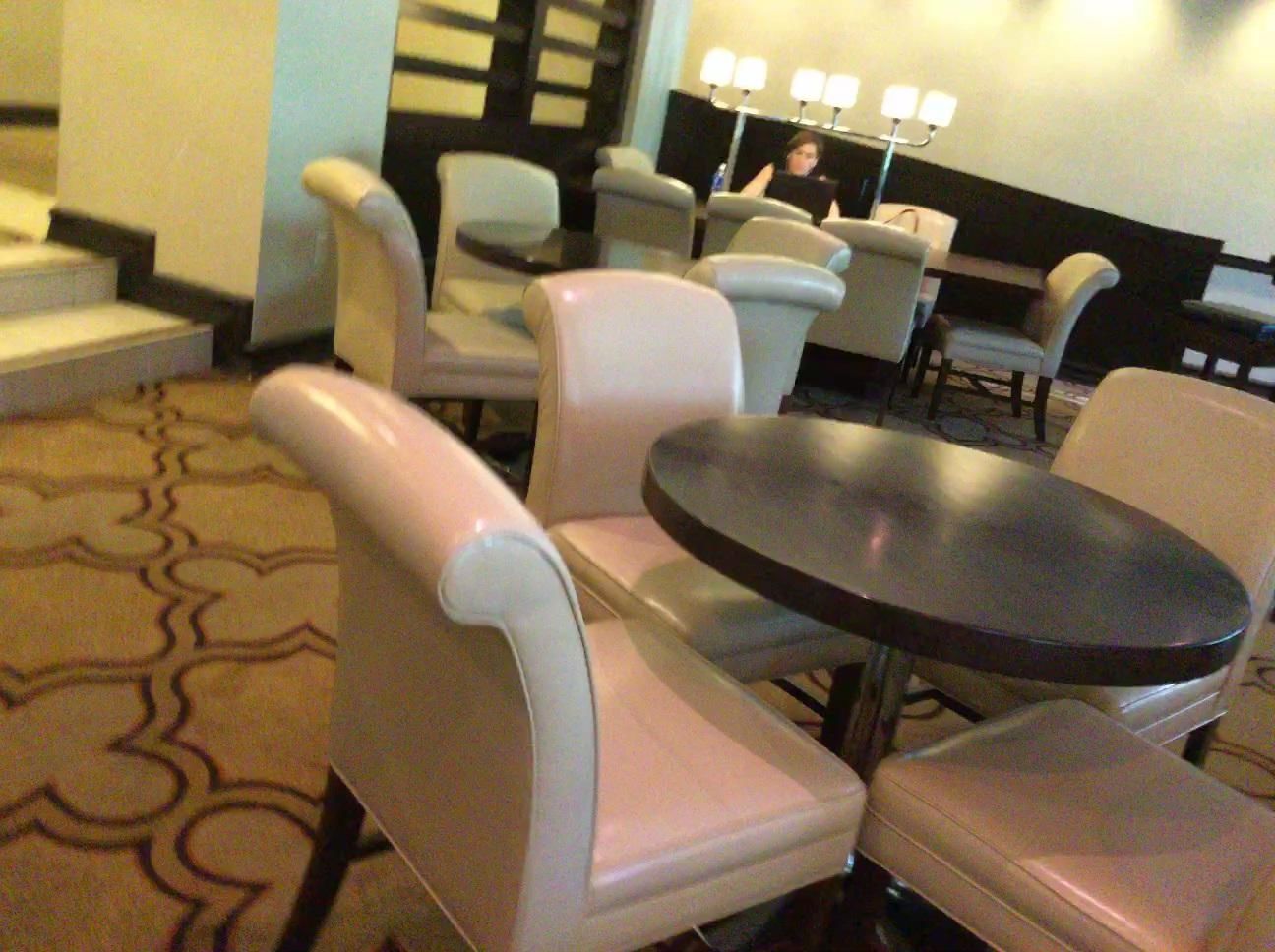}} & \multicolumn{2}{p{2.6cm}}{\includegraphics[width=\linewidth]{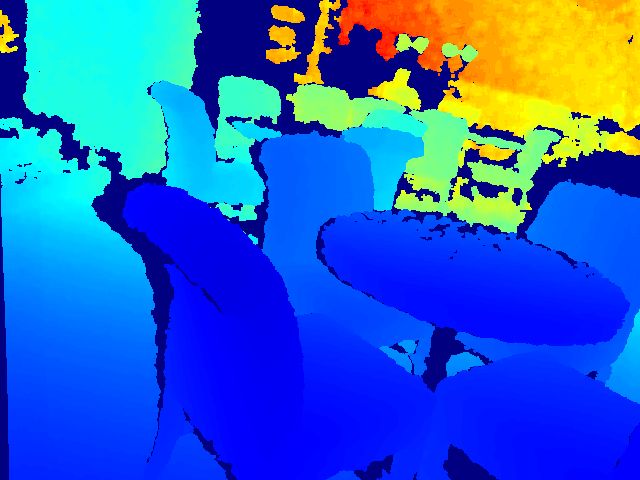}} &
\multicolumn{2}{p{2.6cm}}{\includegraphics[width=\linewidth]{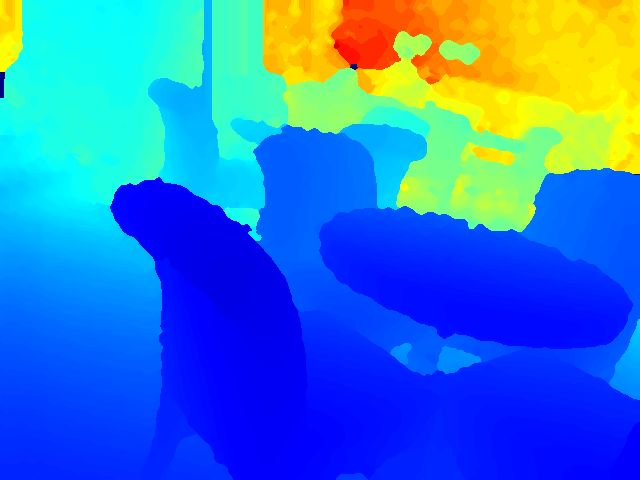}} \\
\multicolumn{2}{p{2.6cm}}{(a) RGB} & \multicolumn{2}{p{2.6cm}}{(b) Depth} & \multicolumn{2}{p{2.6cm}}{(c) Depth Filled} \\
& \multicolumn{2}{p{2.6cm}}{\includegraphics[width=\linewidth]{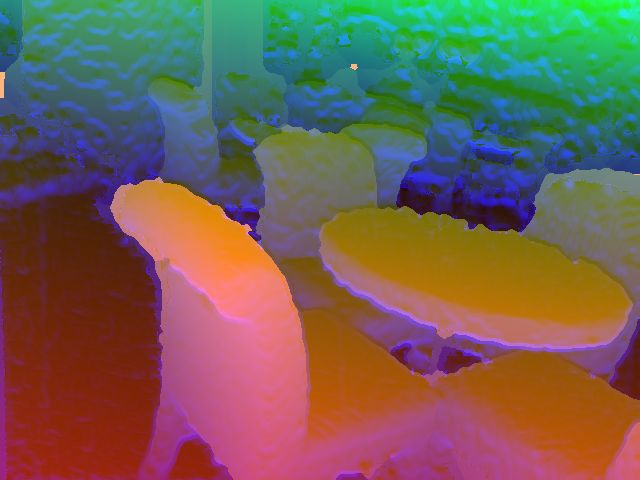}} & \multicolumn{2}{p{2.6cm}}{\includegraphics[width=\linewidth]{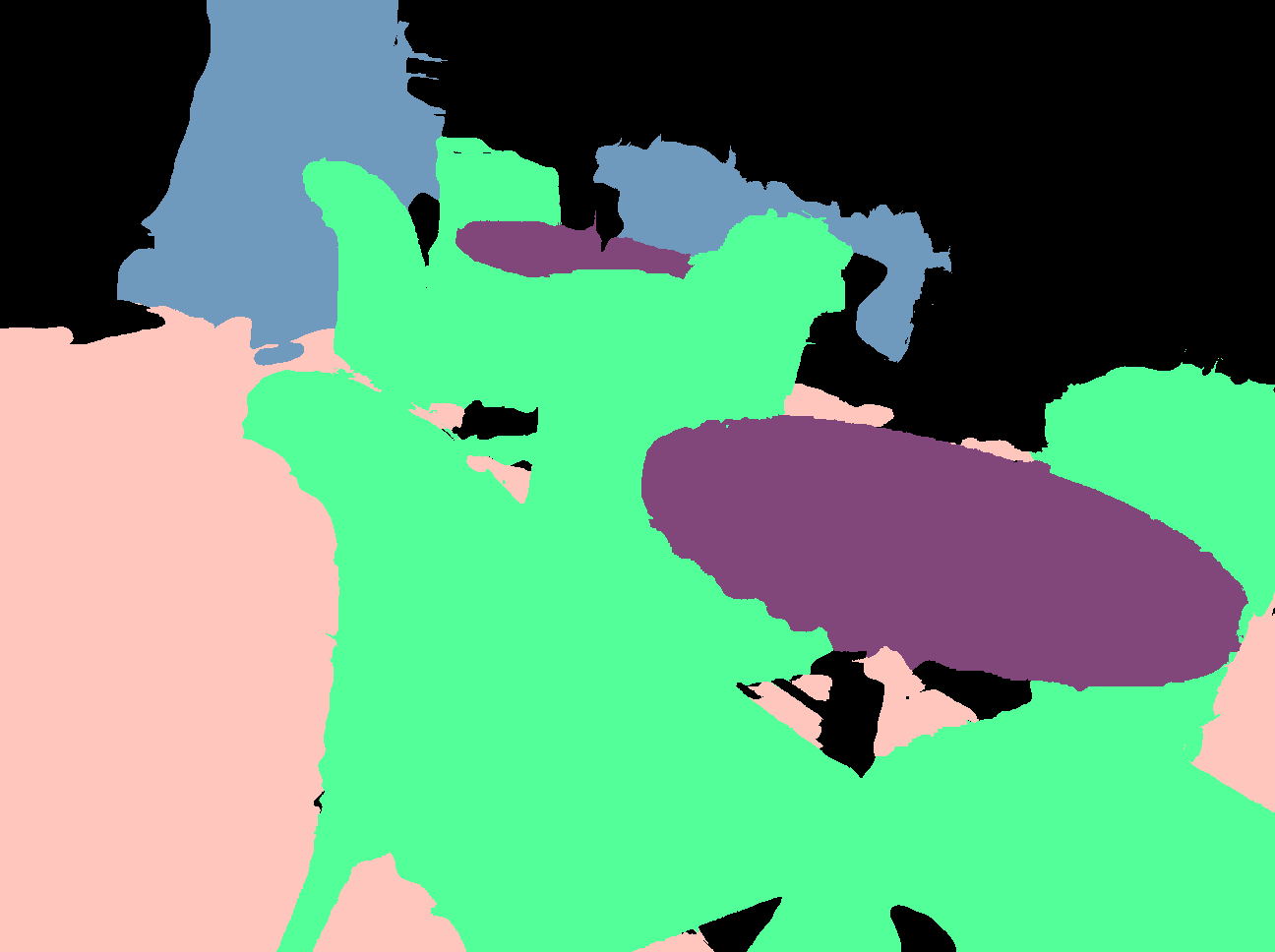}} \\
& \multicolumn{2}{p{2.6cm}}{(d) HHA} & \multicolumn{2}{p{2.6cm}}{(e) Ground truth label} \\ 
\end{tabular}
\caption{Example image from the Scannet dataset showing a complex indoor scene, the corresponding depth map representations and the ground truth semantic segmentation mask.}
\label{fig:scannetEx}
\end{figure}

{\parskip=5pt
\noindent\textbf{ScanNet:} The ScanNet RGB-D video dataset~\citep{dai2017scannet} is a recently introduced large-scale indoor scene understanding benchmark. It contains $2.5\million$ RGB-D images accounting to 1512 scans acquired in 707 distinct spaces. The data was collected using an iPad Air2 mounted with a depth camera similar to the Microsoft Kinect v1. Both the iPad camera and the depth camera were hardware synchronized and frames were captured at $30\hertz$. The RGB images were captured at a resolution of $1296\times968$ pixels and the depth frames were captured at $640\times480$ pixels. The semantic segmentation benchmark contains 16,506 labelled training images and 2537 testing images. From the example depth image shown in \figref{fig:scannetEx}(b), we can see that there are a number of missing depth values at the object boundaries and at large distances. Therefore, similar to the preprocessing that we perform on the cityscapes dataset, we use a fast depth completion technique~\citep{ku2018defense} to fill the holes. The corresponding filled depth image is shown in \figref{fig:scannetEx}(c). We also compute the HHA encoding for the depth maps and use them as an additional modality in our experiments.

The dataset provides pixel-level semantic annotations for 21 object categories, namely: \textit{wall, floor, chair, table, desk, bed, bookshelf, sofa, sink, bathtub, toilet, curtain, counter, door, window, shower curtain, refrigerator, picture, cabinet, other furniture} and \textit{void}. Similar to the SUN RGB-D dataset, many object classes are rarely present making the dataset unbalanced. Moreover, the annotations at the object boundaries are often irregular and parts of objects at large distances are unlabelled as shown in \figref{fig:scannetEx}(e). These factors make the task even more challenging on this dataset.}

\begin{figure}
\centering
\footnotesize
\setlength{\tabcolsep}{0.2em}
\begin{tabular}{p{2.68cm} p{2.68cm} p{2.68cm}}
\includegraphics[width=\linewidth]{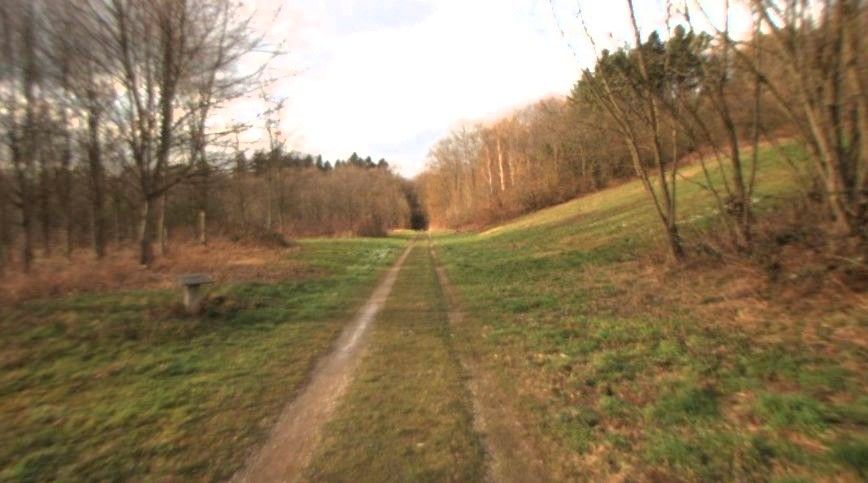} & \includegraphics[width=\linewidth]{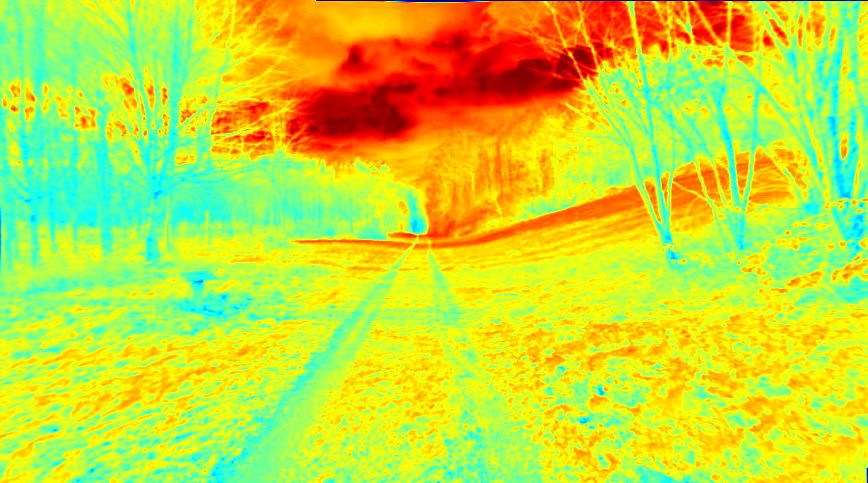} &
\includegraphics[width=\linewidth]{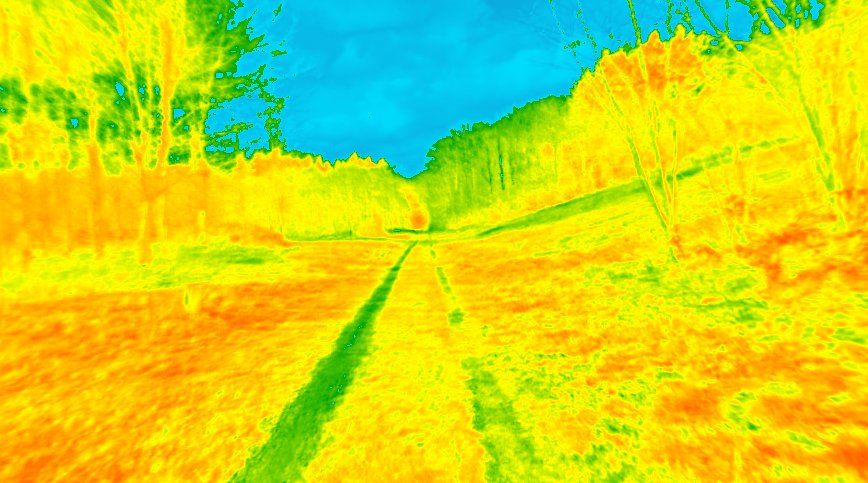} \\
(a) RGB & (b) NIR & (c) NDVI \\ 
\includegraphics[width=\linewidth]{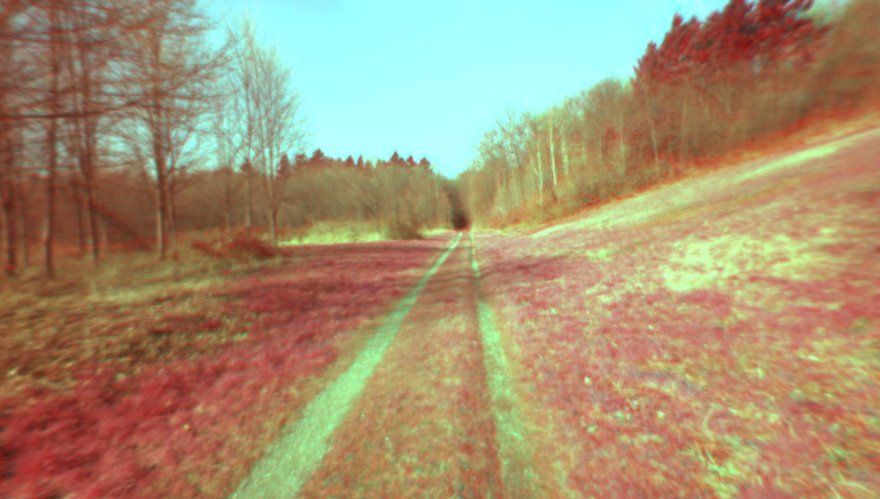} &
\includegraphics[width=\linewidth]{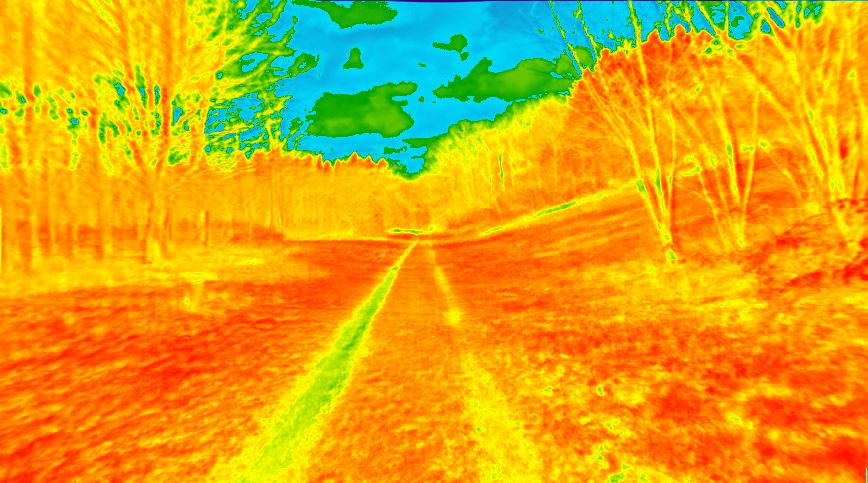} &
\includegraphics[width=\linewidth]{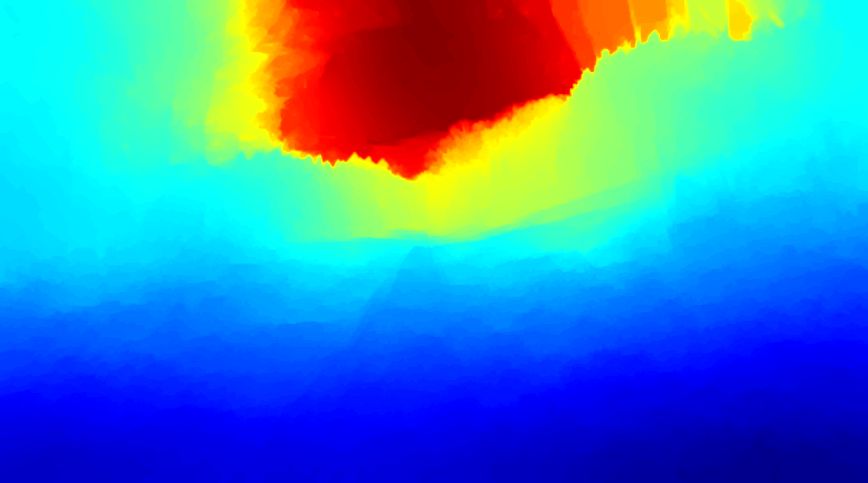} \\
(d) NRG & (e) EVI & (f) Depth \\
\end{tabular}
\caption{Example image from the Freiburg Forest dataset showing the various spectra and modalities: Near-InfraRed (NIR), Normalized Difference Vegetation Index (NDVI), Near-InfraRed-Red-Green (NRG), Enhanced Vegetation Index (EVI) and Depth.}
\label{fig:forestEx}
\end{figure}

{\parskip=5pt
\noindent\textbf{Freiburg Forest:} In our previous work~\citep{valada16iser}, we introduced the Freiburg Multispectral Segmentation benchmark, which is a first-of-a-kind dataset of unstructured forested environments. Unlike urban and indoor scenes which are highly structured with rigid objects that have distinct geometric properties, objects in unstructured forested environments are extremely diverse and moreover, their appearance completely changes from month to month due to seasonal variations. The primary motivation for the introduction of this dataset is to enable robots to discern obstacles that can be driven over such as tall grass and bushes to obstacles that should be avoided such as tall trees and boulders. Therefore, we proposed to exploit the presence of chlorophyll in these objects which can be detected in the Near-InfraRed (NIR) wavelength. NIR images provide a high fidelity description on the presence of vegetation in the scene and as demonstrated in our previous work~\citep{valada17icra}, it enhances border accuracy for segmentation.}

The dataset was collected over an extended period of time using our Viona autonomous robot equipped with a Bumblebee2 camera to capture stereo images and a modified camera with the NIR-cut filter replaced with a Wratten 25A filter for capturing the NIR wavelength in the blue and green channels. The dataset contains over 15,000 images that were sub-sampled at $1\hertz$, corresponding to traversing over $4.7\kilo\meter$ each day. In order to extract consistent spatial and global vegetation information we computed vegetation indices such as Normalized Difference Vegetation Index (NDVI) and Enhanced Vegetation Index (EVI) using the approach presented by~\cite{huete1999modis}. NDVI is resistant to noise caused due to changing sun angles, topography and shadows but is susceptible to error due to variable atmospheric and canopy background conditions~\citep{huete1999modis}. EVI was proposed to compensate for these defects with improved sensitivity to high biomass regions and improved detection though decoupling of canopy background signal and reduction in atmospheric influences. \figref{fig:forestEx} shows an example image from the dataset and the corresponding modalities. The dataset contains hand-annotated segmentation groundtruth for six classes: \textit{sky, trail, grass, vegetation, obstacle} and \textit{void}. We use the original train and test splits provided by the dataset.

\subsection{AdapNet++ Benchmarking}
\label{sec:benchm}

In this section, we report results comparing the performance of our proposed AdapNet++ architecture against several well adopted state-of-the-art models including DeepLab v3~\citep{chen2017rethinking}, ParseNet~\citep{liu2015}, FCN-8s~\citep{long2015}, SegNet~\citep{kendall2015}, FastNet~\citep{oliveira2016}, DeepLab v2~\citep{ChenPK0Y16}, DeconvNet~\citep{noh2015learning} and Adapnet~\citep{valada17icra}. We use the official implementations of these architectures that are publicly released by the authors to train on the input image resolution of $768\times 384$ pixels. For the Cityscapes and \mbox{ScanNet} benchmarking results reported in \tabref{tab:benchmarkingCityscapes} and \tabref{tab:benchmarkingScannet}, we report results directly from the official benchmark leaderboard. For each of the datasets, we report the mIoU score, as well as the per-class IoU score. In order to have a fair comparison, we also evaluate the models at the same input resolution using the same evaluation settings. We do not apply multiscale inputs or left-right flips during testing as these techniques require each crop of each image to be evaluated several times which significantly increases the computational complexity and runtime (Note: We do not use crops for testing, we evaluate on the full image in a single forward-pass). Moreover, these techniques do not improve the performance of the model in real-time applications. However, we show the potential gains that can be obtained in the evaluation metric utilizing these techniques and with a higher resolution input image in the ablation study presented in \secref{sec:highRes}. Additionally, we report results with full resolution evaluation on the test set of the datasets when available, namely for Cityscapes and ScanNet.

\begin{table*}
\footnotesize
\centering
\caption{Performance comparison of AdapNet++ with baseline models on the Cityscapes validation set with 11 semantic class labels (input image dim: $768\times384$). Note that no left-right flips or multiscale testing is performed. Note: Corresponding multimodal results are reported in \tabref{tab:multimodalCityscapes}.}
\label{tab:baselineCityscapes}
\begin{tabular}{p{2.4cm}p{0.7cm}p{0.8cm}p{0.6cm}p{0.9cm}p{0.7cm}p{1cm}p{0.6cm}p{0.6cm}p{0.6cm}p{0.7cm}p{0.9cm}|p{0.7cm}}
\noalign{\smallskip}\hline\noalign{\smallskip}
Network & Sky & Building & Road & Sidewalk & Fence & Vegetation & Pole & Car & Sign & Person & Cyclist & mIoU \\
 & \crule[c_sky]{0.2cm}{0.2cm} & \crule[c_building]{0.2cm}{0.2cm} & \crule[c_road]{0.2cm}{0.2cm} & \crule[c_sidewalk]{0.2cm}{0.2cm} & \crule[c_fence]{0.2cm}{0.2cm} & \crule[c_vegetation]{0.2cm}{0.2cm} & \crule[c_pole]{0.2cm}{0.2cm} & \crule[c_car]{0.2cm}{0.2cm} & \crule[c_sign]{0.2cm}{0.2cm} & \crule[c_person]{0.2cm}{0.2cm} & \crule[c_cyclist]{0.2cm}{0.2cm} & (\%) \\
\noalign{\smallskip}\hline\hline\noalign{\smallskip}
FCN-8s & $76.51$ & $83.97$ & $93.82$ & $67.67$ & $24.91$ & $86.38$ & $31.71$ & $84.80$ & $50.92$ & $59.89$ & $59.11$ &  $59.97$ \\
SegNet & $73.74$ & $79.29$ & $92.70$ & $59.88$ & $13.63$ & $81.89$ & $26.18$ & $78.83$ & $31.44$ & $45.03$ & $43.46$ & $52.17$ \\
FastNet & $77.69$ & $86.25$ & $94.97$ & $72.99$ & $31.02$ & $88.06$ & $38.34$ & $88.42$ & $52.34$ & $61.76$ & $61.83$ & $68.52$ \\
ParseNet & $77.57$ & $86.81$ & $95.27$ & $74.02$ & $33.31$ & $87.37$ & $38.24$ & $88.99$ & $53.34$ & $63.25$ & $63.87$ & $69.28$ \\
DeconvNet & $89.38$ & $83.08$ & $95.26$ & $68.07$ & $27.58$ & $85.80$ & $34.20$ & $85.01$ & $27.62$ & $45.11$ & $41.11$ & $62.02$ \\
DeepLab v2 & $74.28$ & $81.66$ & $90.86$ & $63.3$ & $26.29$ & $84.33$ & $27.96$ & $86.24$ & $44.79$ & $58.89$ & $60.92$ & $63.59$ \\
AdapNet & $92.45$ & $89.98$ & $97.43$ & $81.43$ & $49.93$ & $91.44$ & $53.43$ & $92.23$ & $65.32$ & $69.86$ & $69.62$ & $77.56$ \\
DeepLab v3 & $92.82$ & $89.02$ & $96.74$ & $78.13$ & $41.00$ & $90.81$ & $49.74$ & $91.02$ & $64.48$ & $66.52$ & $66.98$ & $75.21$ \\
\hline\noalign{\smallskip}
\textbf{AdapNet++ (ours)} & $\mathbf{94.18}$ & $\mathbf{91.49}$ & $\mathbf{97.93}$ & $\mathbf{84.40}$ & $\mathbf{54.98}$ & $\mathbf{92.09}$ & $\mathbf{58.85}$ & $\mathbf{93.86}$ & $\mathbf{72.61}$ & $\mathbf{75.52}$ & $\mathbf{72.90}$ & $\mathbf{80.80}$ \\
\noalign{\smallskip}\hline\noalign{\smallskip}
\end{tabular}
\end{table*}

\tabref{tab:baselineCityscapes} shows the results on the 11 class Cityscapes validation set. AdapNet++ outperforms all the baselines in each individual object category as well in the mIoU score. AdapNet++ outperforms the highest baseline by a margin of $3.24\%$. Analyzing the individual class IoU scores, we can see that AdapNet++ yields the highest improvement in object categories that contain thin structures such as \textit{poles} for which it gives a large improvement of $5.42\%$, a similar improvement of $5.05\%$ for \textit{fences} and the highest improvement for $7.29\%$ for \textit{signs}. Most architectures struggle to recover the structure of thin objects due to downsampling by pooling and striding in the network which causes such information to be lost. However, these results show that AdapNet++ efficiently recovers the structure of such objects by learning multiscale features at several stages of the encoder using the proposed multiscale residual units and the eASPP. We further show the improvement in performance due to the incorporation of the multiscale residual units and the eASPP in the ablation study presented
in \secref{sec:dissecting}. In driving scenarios, information of objects such as \textit{pedestrians} and \textit{cyclists} can also be lost when they appear at far away distances. A large improvement can also be seen in categories such as \textit{person} in which AdapNet++ achieves an improvement of $5.66\%$. The improvement in larger object categories such as \textit{cars} and \textit{vegetation} can be attributed to the new decoder which improves the segmentation performance near object boundaries. This is more evident in the qualitative results presented in \secref{sec:qualitative}. Note that the colors shown below the object category names serve as a legend for the qualitative results.

\begin{table*}
\footnotesize 
\centering
\caption{Benchmarking results on the Cityscapes dataset with full resolution evaluation on 19 semantic class labels. Only the eight top performing published models in the leaderboard are listed in this table. The inference time is reported for an input image resolution of $768 \times 384$ pixels and it was computed on an NVIDIA TITAN X (PASCAL) GPU using the official implementation of each method.}
\label{tab:benchmarkingCityscapes}
\begin{tabular}{p{2.6cm}p{2.6cm}|p{1cm}p{1cm}p{1cm}p{1cm}}
\noalign{\smallskip}\hline\noalign{\smallskip}
Network & Backbone & \multicolumn{2}{c}{mIoU (\%)} & Parms. & Time \\
\cline{3-4}
 & & val & test & ($\million$) & ($\milli\second$) \\
\noalign{\smallskip}\hline\hline\noalign{\smallskip}
PSPNet & ResNet-101 & $80.91$ & $81.19$ & $56.27$ & $172.42$ \\
DeepLab v3 & ResNet-101 & $79.30$ & $81.34$ & $58.16$ & $79.90$ \\
Mapillary & WideResNet-38 & $78.31$ & $82.03$ & $135.86$ & $214.46$ \\
DeepLab v3+ & Modified Xception & $79.55$ & $82.14$ & $43.48$ & $127.97$ \\
DPC & Modified Xception & $80.85$ & $82.66$ & $41.82$ & $144.41$ \\
DRN & WideResNet-38 & $79.69$ & $82.82$ & $129.16$ & $1259.67$ \\
\hline\noalign{\smallskip}
\textbf{AdapNet++ (ours)} & ResNet50 & $81.24$ & $81.34$ & $30.20$ & $72.92$ \\
\textbf{SSMA (ours)} & ResNet50 & $82.19$ & $82.31$ & $56.44$ & $101.95$ \\
\noalign{\smallskip}\hline\noalign{\smallskip}
\end{tabular}
\end{table*}

\begin{table*}
\footnotesize 
\centering
\caption{Performance comparison of AdapNet++ with baseline models on the Synthia validation set (input image dim: $768\times384$). Note that no left-right flips or multiscale testing is performed. Note: Corresponding multimodal results are reported in \tabref{tab:multimodalSynthia}.}
\label{tab:baselineSynthia}
\begin{tabular}{p{2.4cm}p{0.7cm}p{0.8cm}p{0.6cm}p{0.9cm}p{0.7cm}p{1cm}p{0.6cm}p{0.6cm}p{0.6cm}p{0.7cm}p{0.9cm}|p{0.7cm}}
\noalign{\smallskip}\hline\noalign{\smallskip}
Network & Sky & Building & Road & Sidewalk & Fence & Vegetation & Pole & Car & Sign & Person & Cyclist & mIoU \\
 & \crule[s_sky]{0.2cm}{0.2cm} & \crule[s_building]{0.2cm}{0.2cm} & \crule[s_road]{0.2cm}{0.2cm} & \crule[s_sidewalk]{0.2cm}{0.2cm} & \crule[s_fence]{0.2cm}{0.2cm} & \crule[s_vegetation]{0.2cm}{0.2cm} & \crule[s_pole]{0.2cm}{0.2cm} & \crule[s_car]{0.2cm}{0.2cm} & \crule[s_sign]{0.2cm}{0.2cm} & \crule[s_person]{0.2cm}{0.2cm} & \crule[s_cyclist]{0.2cm}{0.2cm} & (\%) \\
\noalign{\smallskip}\hline\hline\noalign{\smallskip}
FCN-8s & $92.36$ & $91.92$ & $88.94$ & $86.46$ & $48.22$ & $77.41$ & $36.02$ & $82.63$ & $30.37$ & $57.10$ & $46.84$ & $67.11$ \\
SegNet & $91.90$ & $87.19$ & $83.72$ & $80.94$ & $50.02$ & $71.63$ & $26.12$ & $71.31$ & $1.01$ & $52.34$ & $32.64$ & $58.98$ \\
FastNet & $92.21$ & $92.41$ & $91.85$ & $89.89$ & $56.64$ & $78.59$ & $51.17$ & $84.75$ & $32.03$ & $69.87$ & $55.65$ & $72.28$ \\
ParseNet & $93.80$ & $93.09$ & $91.05$ & $88.98$ & $53.22$ & $79.48$ & $46.15$ & $85.37$ & $36.00$ & $63.30$ & $50.82$ & $71.02$ \\
DeconvNet & $95.88$ & $93.83$ & $92.85$ & $90.79$ & $66.40$ & $81.04$ & $48.23$ & $84.65$ & $0.00$ & $69.46$ & $52.79$ & $70.54$ \\
DeepLab v2 & $94.07$ & $93.34$ & $88.07$ & $88.93$ & $55.57$ & $80.22$ & $45.97$ & $85.87$ & $38.73$ & $64.40$ & $52.54$ & $71.61$ \\
AdapNet & $96.95$ & $95.88$ & $95.60$ & $94.46$ & $76.30$ & $86.59$ & $67.14$ & $92.20$ & $58.85$ & $80.18$ & $66.89$ & $82.83$ \\
DeepLab v3 & $95.30$ & $92.75$ & $93.58$ & $91.56$ & $73.37$ & $80.71$ & $55.83$ & $88.09$ & $44.17$ & $75.65$ & $60.15$ & $77.38$ \\
\hline\noalign{\smallskip}
\textbf{AdapNet++ (ours)} & $\mathbf{97.77}$ & $\mathbf{96.98}$ & $\mathbf{96.60}$ & $\mathbf{95.70}$ & $\mathbf{79.87}$ & $\mathbf{89.63}$ & $\mathbf{74.94}$ & $\mathbf{94.22}$ & $\mathbf{71.99}$ & $\mathbf{83.64}$ & $\mathbf{72.31}$ & $\mathbf{86.70}$ \\
\noalign{\smallskip}\hline\noalign{\smallskip}
\end{tabular}
\end{table*}

We also report results on the full 19 class Cityscapes validation and test sets in \tabref{tab:benchmarkingCityscapes}. We compare against the top six published models on the leaderboard, namely,  PSPNet~\citep{zhao2017pspnet}, DeepLab~v3~\citep{chen2017rethinking}, Mapilary~\citep{bulo2018place}, DeepLab~v3+~\citep{chen2018encoder}, DPC~\citep{chen2018searching}, and DRN~\citep{zhuang2018dense}. The results of the competing methods reported in this table are directly taken from the benchmark leaderboard for the test set and from the corresponding manuscripts of the methods for the validation set. We trained our models on $768\times 768$ crops from the full image resolution for benchmarking on the leaderboard. Our AdapNet++ model with a much smaller network backbone achieves a comparable performance as other top performing models on the leaderboard. Moreover, our network is the most efficient architecture in terms of both the number of parameters that it consumes as well as the inference time compared to other networks on the entire first page of the Cityscapes leaderboard.

We benchmark on the Synthia dataset largely due to the variety of seasons and adverse perceptual conditions where the improvement due to multimodal fusion can be seen. However, even for baseline comparison shown in \tabref{tab:baselineSynthia}, it can be seen that AdapNet++ outperforms all the baselines, both in the overall mIoU score as well as in the score of the individual object categories. It achieves an overall improvement of $3.87\%$ and a similar observation can be made in the improvement of scores for thin structures, reinforcing the utility of our proposed multiscale feature learning configuration. The largest improvement of $13.14\%$ was obtained for the \textit{sign} class, followed by an improvement of $7.8\%$ for the \textit{pole} class. In addition a significant improvement of $5.42\%$ can also be seen for the \textit{cyclist} class.

\begin{table*}
\footnotesize 
\centering
\caption{Performance comparison of AdapNet++ with baseline models on the SUN RGB-D validation set (input image dim: $768\times384$). Note that no left-right flips or multiscale testing is performed. Note: Corresponding multimodal results are reported in \tabref{tab:multimodalSun}.}
\label{tab:baselineSun}
\begin{tabular}{p{2.4cm}p{0.6cm}p{0.6cm}p{0.6cm}p{0.6cm}p{0.6cm}p{0.6cm}p{0.6cm}p{0.6cm}p{0.6cm}p{0.6cm}p{0.6cm}p{0.6cm}p{0.6cm}p{0.7cm}}
\noalign{\smallskip}\hline\noalign{\smallskip}
Network & Wall & Floor & Cabinet & Bed & Chair & Sofa & Table & Door & Window & Bshelf & Picture & Counter & Blinds & Desk \\
 & \crule[u_Wall]{0.2cm}{0.2cm} & \crule[u_Floor]{0.2cm}{0.2cm} & \crule[u_Cabinet]{0.2cm}{0.2cm} & \crule[u_Bed]{0.2cm}{0.2cm} & \crule[u_Chair]{0.2cm}{0.2cm} & \crule[u_Sofa]{0.2cm}{0.2cm} & \crule[u_Table]{0.2cm}{0.2cm} & \crule[u_Door]{0.2cm}{0.2cm} & \crule[u_Window]{0.2cm}{0.2cm} & \crule[u_Bookshelf]{0.2cm}{0.2cm} & \crule[u_Picture]{0.2cm}{0.2cm} & \crule[u_Counter]{0.2cm}{0.2cm} & \crule[u_Blinds]{0.2cm}{0.2cm} & \crule[u_Desk]{0.2cm}{0.2cm} \\
\noalign{\smallskip}\hline\hline\noalign{\smallskip}
FCN-8s & $68.57$ & $79.94$ & $37.27$ & $52.85$ & $58.42$ & $42.79$ & $43.69$ & $27.58$ & $43.49$ & $28.46$ & $42.21$ & $27.08$ & $28.34$ & $11.38$ \\
SegNet & $70.18$ & $82.01$ & $37.62$ & $45.77$ & $57.22$ & $35.86$ & $40.60$ & $30.32$ & $39.37$ & $29.67$ & $40.67$ & $22.48$ & $32.13$ & $14.79$ \\
FastNet & $68.39$ & $79.73$ & $34.98$ & $44.93$ & $55.26$ & $36.68$ & $38.57$ & $26.88$ & $38.28$ & $26.02$ & $39.72$ & $20.55$ & $22.47$ & $11.71$ \\
ParseNet & $70.83$ & $81.85$ & $40.25$ & $55.64$ & $61.12$ & $45.06$ & $45.86$ & $30.41$ & $41.73$ & $\mathbf{34.45}$ & $44.11$ & $28.63$ & $32.59$ & $12.78$ \\
DeconvNet & $61.53$ & $75.52$ & $26.97$ & $38.52$ & $48.46$ & $30.94$ & $37.12$ & $21.99$ & $27.48$ & $0.00$ & $0.00$ & $0.00$ & $0.00$ & $9.40$ \\
DeepLab v2 & $63.68$ & $74.07$ & $31.15$ & $47.00$ & $55.46$ & $40.81$ & $40.60$ & $29.10$ & $37.95$ & $18.67$ & $35.67$ & $23.20$ & $25.08$ & $13.52$ \\
AdapNet & $72.82$ & $\mathbf{86.61}$ & $41.90$ & $56.63$ & $64.99$ & $46.14$ & $46.74$ & $36.14$ & $44.43$ & $28.40$ & $\mathbf{46.53}$ & $25.17$ & $31.07$ & $13.41$ \\
DeepLab v3 & $\mathbf{74.99}$ & $85.44$ & $39.64$ & $54.14$ & $64.23$ & $45.84$ & $47.35$ & $38.30$ & $45.50$ & $30.70$ & $45.30$ & $15.21$ & $24.92$ & $18.18$ \\
\hline\noalign{\smallskip}
\textbf{AdapNet++ (ours)} & $73.81$ & $84.79$ & $\mathbf{47.45}$ & $\mathbf{64.31}$ & $\mathbf{65.76}$ & $\mathbf{52.15}$ & $\mathbf{49.97}$ & $\mathbf{41.66}$ & $\mathbf{46.99}$ & $32.83$ & $45.19$ & $\mathbf{32.71}$ & $\mathbf{34.72}$ & $\mathbf{21.16}$ \\
\noalign{\smallskip}\hline\noalign{\smallskip}
Network & Shelves & Curtain & Dresser & Pillow & Mirror & FMat & Clothes & Ceiling & Books & Fridge & Tv & Paper & Towel & ShwrC \\
 & \crule[u_Shelves]{0.2cm}{0.2cm} & \crule[u_Curtain]{0.2cm}{0.2cm} & \crule[u_Dresser]{0.2cm}{0.2cm} & \crule[u_Pillow]{0.2cm}{0.2cm} & \crule[u_Mirror]{0.2cm}{0.2cm} & \crule[u_FloorMat]{0.2cm}{0.2cm} & \crule[u_Clothes]{0.2cm}{0.2cm} & \crule[u_Ceiling]{0.2cm}{0.2cm} & \crule[u_Books]{0.2cm}{0.2cm} & \crule[u_Fridge]{0.2cm}{0.2cm} & \crule[u_Tv]{0.2cm}{0.2cm} & \crule[u_Paper]{0.2cm}{0.2cm} & \crule[u_Towel]{0.2cm}{0.2cm} & \crule[u_ShowerCurtain]{0.2cm}{0.2cm} \\
\noalign{\smallskip}\hline\hline\noalign{\smallskip}
FCN-8s & $6.01$ & $46.96$ & $28.80$ & $27.73$ & $18.97$ & $0.00$ & $20.34$ & $51.98$ & $30.98$ & $9.06$ & $\mathbf{34.23}$ & $15.95$ & $12.79$ & $0.00$ \\
SegNet & $5.15$ & $39.69$ & $27.46$ & $26.05$ & $15.64$ & $0.00$ & $16.01$ & $53.89$ & $26.21$& $12.95$ & $24.23$ & $14.57$ & $10.88$ & $0.00$ \\
FastNet & $5.67$ & $41.98$ & $25.08$ & $22.76$ & $11.76$ & $0.10$ & $13.37$ & $49.14$ & $29.38$ & $17.49$ & $19.80$ & $17.95$ & $13.55$ & $2.93$ \\
ParseNet & $\mathbf{8.58}$ & $46.51$ & $34.05$ & $28.07$ & $24.82$ & $0.02$ & $19.10$ & $52.28$ & $\mathbf{34.36}$ & $27.05$ & $31.49$ & $\mathbf{21.65}$ & $21.49$ & $\mathbf{6.46}$ \\
DeconvNet & $0.00$ & $29.59$ & $0.00$ & $0.00$ & $8.32$ & $0.00$ & $0.00$ & $46.40$ & $0.00$ & $0.00$ & $0.00$ & $0.00$ & $0.00$ & $0.00$ \\
DeepLab v2 & $2.55$ & $42.91$ & $24.56$ & $29.81$ & $\mathbf{27.16}$ & $0.00$ & $15.85$ & $43.47$ & $25.39$ & $35.55$ & $23.21$ & $16.53$ & $22.00$ & $0.34$ \\
AdapNet & $3.98$ & $47.90$ & $28.58$ & $\mathbf{32.90}$ & $19.25$ & $0.00$ & $17.30$ & $52.60$ & $30.63$ & $20.32$ & $16.56$ & $18.21$ & $13.69$ & $0.00$ \\
DeepLab v3 & $66.37$ & $48.84$ & $28.20$ & $32.87$ & $31.23$ & $0.00$ & $16.98$ & $60.09$ & $31.74$ & $37.86$ & $33.80$ & $17.84$ & $20.51$ & $0.00$ \\
\hline\noalign{\smallskip}
\textbf{AdapNet++ (ours)} & $5.03$ & $\mathbf{50.32}$ & $\mathbf{37.45}$ & $32.35$ & $25.89$ & $\mathbf{0.12}$ & $\mathbf{21.68}$ & $\mathbf{60.65}$ & $32.41$ & $\mathbf{46.96}$ & $18.81$ & $20.79$ & $\mathbf{27.95}$ & $1.32$ \\
\noalign{\smallskip}\hline\noalign{\smallskip}
Network & Box & Wboard & Person & NStand & Toilet & Sink & Lamp & Bathtub & Bag & \vline~~mIoU \\
 & \crule[u_Box]{0.2cm}{0.2cm} & \crule[u_Whiteboard]{0.2cm}{0.2cm} & \crule[u_Person]{0.2cm}{0.2cm} & \crule[u_NightStand]{0.2cm}{0.2cm} & \crule[u_Toilet]{0.2cm}{0.2cm} & \crule[u_Sink]{0.2cm}{0.2cm} & \crule[u_Lamp]{0.2cm}{0.2cm} & \crule[u_Bathtub]{0.2cm}{0.2cm} & \crule[u_Bag]{0.2cm}{0.2cm} & \vline~~(\%) \\
\noalign{\smallskip}\cline{1-11}
\noalign{\smallskip}\cline{1-11}\noalign{\smallskip}
FCN-8s & $18.24$ & $41.38$ & $8.92$ & $0.97$ & $59.64$ & $44.21$ & $22.83$ & $26.20$ & $8.81$ & \vline~~$30.46$ \\
SegNet & $13.50$ & $38.02$ & $3.83$ & $0.76$ & $60.98$ &$45.97$ & $22.82$ & $31.94$ & $8.97$ & \vline~~$29.14$ \\
FastNet & $15.33$ & $38.12$ & $15.82$ & $7.93$ & $54.08$ & $39.59$ & $22.20$ & $23.71$ & $9.80$ &  \vline~~$28.15$ \\
ParseNet & $20.77$ & $44.48$ & $34.59$ & $15.37$ & $66.96$ & $47.15$ & $31.49$ & $28.14$ & $13.19$ & \vline~~$34.68$ \\
DeconvNet & $4.43$ & $32.06$ & $0.00$ & $0.00$ & $47.35$ & $32.85$ & $0.00$ & $0.00$ & $0.00$ & \vline~~$15.64$ \\
DeepLab v2 & $13.56$ & $22.32$ & $\mathbf{46.18}$ & $11.45$ & $55.94$ & $43.33$ & $29.52$ & $32.73$ & $7.63$ & \vline~~$30.06$ \\
AdapNet & $15.72$ & $43.02$ & $17.48$ & $9.44$ & $61.11$ & $49.43$ & $31.32$ & $21.56$ & $9.38$ & \vline~~$32.47$ \\
DeepLab v3 & $19.58$ & $44.10$ & $31.36$ & $18.15$ & $68.76$ & $52.01$ & $35.36$ & $\mathbf{45.97}$ & $10.37$ & \vline~~$35.74$ \\
\cline{1-11}\noalign{\smallskip}
\textbf{AdapNet++ (ours)} & $\mathbf{22.70}$ & $\mathbf{51.08}$ & $33.77$ & $\mathbf{20.59}$ & $\mathbf{71.28}$ & $\mathbf{55.31}$ & $\mathbf{36.74}$ & $38.60$ & $\mathbf{15.34}$ & \vline~~$\mathbf{38.40}$ \\
\noalign{\smallskip}\cline{1-11}\noalign{\smallskip}
\end{tabular}
\end{table*}

Compared to outdoor driving datasets, indoor benchmarks such as SUN RGB-D and ScanNet pose a different challenge. Indoor datasets have vast amounts of object categories in various different configurations and images captured from many different view points compared to driving scenarios where the camera is always parallel to the ground with similar viewpoints from the perspective of the vehicle driving on the road. Moreover, indoor scenes are often extremely cluttered which causes occlusions, in addition to the irregular frequency distribution of the object classes that make the problem even harder. Due to these factors SUN RGB-D is considered one of the hardest datasets to benchmark on. Despite these factors, as shown in \tabref{tab:baselineSun}, AdapNet++ outperforms all the baseline networks overall by a margin of $2.66\%$ compared to the highest performing DeepLab~v3 baseline which took $30,000$ iterations more to reach this score. Unlike the performance in the Cityscapes and Synthia datasets where our previously proposed AdapNet architecture yields the second highest performance, AdapNet is outperformed by DeepLab~v3 in the SUN RGB-D dataset. AdapNet++ on the other hand, outperforms the baselines in most categories by a large margin, while it is outperformed in 13 of the 37 classes by small margin. It can also be observed that the classes in which AdapNet++ get outperformed are the most infrequent classes. This can be alleviated by adding supplementary training images containing the low-frequency classes from other datasets or by employing class balancing techniques. However, our initial experiments employing techniques such as median frequency class balancing, inverse median frequency class balancing, normalized inverse frequency balancing, severely affected the performance of our model.

\begin{table*}
\footnotesize 
\centering
\caption{Performance comparison of AdapNet++ with baseline models on the ScanNet validation set (input image dim: $768\times384$). Note that no left-right flips or multiscale testing is performed. Note: Corresponding multimodal results are reported in \tabref{tab:multimodalScannet}.}
\label{tab:scannet}
\begin{tabular}{p{2.4cm}p{0.6cm}p{0.7cm}p{0.6cm}p{0.6cm}p{0.6cm}p{0.6cm}p{0.6cm}p{0.6cm}p{0.6cm}p{0.6cm}p{0.6cm}p{0.6cm}p{0.6cm}p{0.7cm}}
\noalign{\smallskip}\hline\noalign{\smallskip}
Network & Wall & Floor & Cabinet & Bed & Chair & Sofa & Table & Door & Window & Bshelf & Picture & Counter & Desk & Curtain \\
 & \crule[a_Wall]{0.2cm}{0.2cm} & \crule[a_Floor]{0.2cm}{0.2cm} & \crule[a_Cabinet]{0.2cm}{0.2cm} & \crule[a_Bed]{0.2cm}{0.2cm} & \crule[a_Chair]{0.2cm}{0.2cm} & \crule[a_Sofa]{0.2cm}{0.2cm} & \crule[a_Table]{0.2cm}{0.2cm} & \crule[a_Door]{0.2cm}{0.2cm} & \crule[a_Window]{0.2cm}{0.2cm} & \crule[a_Bookshelf]{0.2cm}{0.2cm} & \crule[a_Picture]{0.2cm}{0.2cm} & \crule[a_Counter]{0.2cm}{0.2cm} & \crule[a_Desk]{0.2cm}{0.2cm} & \crule[a_Curtain]{0.2cm}{0.2cm} \\
\noalign{\smallskip}\hline\hline\noalign{\smallskip}
FCN-8s & $70.00$ & $74.70$ & $39.31$ & $66.23$ & $51.96$ & $48.21$ & $45.19$ & $42.61$ & $42.15$ & $\mathbf{62.59}$ & $\mathbf{27.66}$ & $24.69$ & $33.25$ & $\mathbf{48.29}$ \\
SegNet & $64.03$ & $63.46$ & $28.44$ & $30.22$ & $39.84$ & $27.09$ & $35.68$ & $31.86$ & $29.36$ & $35.23$ & $21.19$ & $12.08$ & $16.45$ & $14.44$ \\
FastNet & $66.31$ & $71.35$ & $34.22$ & $56.32$ & $47.86$ & $44.87$ & $39.34$ & $33.20$ & $34.32$ & $50.11$ & $23.27$ & $24.48$ & $28.87$ & $42.56$ \\
ParseNet & $\mathbf{71.29}$ & $76.42$ & $39.92$ & $71.11$ & $52.72$ & $51.07$ & $48.74$ & $43.06$ & $45.25$ & $58.81$ & $31.33$ & $27.84$ & $38.46$ & $54.81$ \\
DeconvNet & $66.41$ & $69.83$ & $30.57$ & $51.30$ & $38.27$ & $42.98$ & $38.74$ & $32.51$ & $28.86$ & $22.31$ & $0.00$ & $0.00$ & $19.70$ & $31.62$ \\
DeepLab v2 & $59.80$ & $74.21$ & $39.23$ & $50.21$ & $48.95$ & $46.25$ & $47.39$ & $41.27$ & $34.32$ & $48.27$ & $23.36$ & $29.26$ & $35.13$ & $33.89$ \\
AdapNet & $70.62$ & $76.73$ & $43.89$ & $54.10$ & $50.21$ & $48.58$ & $51.95$ & $49.33$ & $35.75$ & $50.05$ & $26.41$ & $30.93$ & $37.23$ & $34.64$ \\
DeepLab v3 & $70.83$ & $77.65$ & $44.21$ & $55.38$ & $54.36$ & $50.23$ & $54.26$ & $51.13$ & $36.14$ & $51.48$ & $25.78$ & $31.59$ & $38.11$ & $36.20$ \\
\hline\noalign{\smallskip}
\textbf{AdapNet++ (ours)} & $71.21$ & $\mathbf{80.41}$ & $\mathbf{46.48}$ & $\mathbf{60.72}$ & $\mathbf{58.89}$ & $\mathbf{55.12}$ & $\mathbf{58.42}$ & $\mathbf{55.15}$ & $\mathbf{37.18}$ & $52.15$ & $26.73$ & $\mathbf{32.79}$ & $\mathbf{39.66}$ & $37.12$ \\
\noalign{\smallskip}\hline\noalign{\smallskip}
Network & Fridge & SCurtain & Toilet & Sink & Bathtub & OtherF & \vline~~mIoU \\
 & \crule[a_Fridge]{0.2cm}{0.2cm} & \crule[a_ShowerCurtain]{0.2cm}{0.2cm} & \crule[a_Toilet]{0.2cm}{0.2cm} & \crule[a_Sink]{0.2cm}{0.2cm} & \crule[a_Bathtub]{0.2cm}{0.2cm} & \crule[a_Other]{0.2cm}{0.2cm} & \vline~~(\%) \\
\noalign{\smallskip}\cline{1-8}
\noalign{\smallskip}\cline{1-8}\noalign{\smallskip}
FCN-8s & $51.96$ & $32.15$ & $55.67$ & $35.80$ & $38.27$ & $26.72$ & \vline~~$45.87$ \\
SegNet & $0.09$ & $24.34$ & $30.70$ & $32.50$ & $3.47$ & $9.73$ & \vline~~$27.51$ \\
FastNet & $37.77$ & $20.27$ & $49.55$ & $27.27$ & $31.99$ & $24.39$ & \vline~~$39.42$ \\
ParseNet & $52.40$ & $28.60$ & $58.53$ & $36.27$ & $41.91$ & $25.85$ & \vline~~$47.72$ \\
DeconvNet & $1.31$ & $0.12$ & $0.00$ & $0.00$ & $0.00$ & $16.66$ & \vline~~$24.56$ \\
DeepLab v2 & $43.69$ & $42.57$ & $62.26$ & $39.97$ & $55.40$ & $22.36$ & \vline~~$43.89$ \\
AdapNet & $43.27$ & $45.83$ & $68.22$ & $43.45$ & $60.37$ & $24.04$ & \vline~~$47.28$ \\
DeepLab v3 & $51.76$ & $51.46$ & $76.22$ & $51.24$ & $64.36$ & $29.32$ & \vline~~$50.09$ \\
\cline{1-8}\noalign{\smallskip}
\textbf{AdapNet++ (ours)} & $\mathbf{54.54}$ & $\mathbf{54.88}$ & $\mathbf{81.92}$ & $\mathbf{54.91}$ & $\mathbf{68.65}$ & $\mathbf{31.46}$ & \vline~~$\mathbf{52.92}$ \\
\noalign{\smallskip}\cline{1-8}\noalign{\smallskip}
\end{tabular}
\end{table*}

\begin{table}
\footnotesize 
\centering
\caption{Bechmarking results on the ScanNet test set with full resolution evaluation. Results were obtained from the ScanNet benchmark leaderboard.}
\label{tab:benchmarkingScannet}
\begin{tabular}{p{2.4cm}cp{0.2cm}|p{1.2cm}}
\noalign{\smallskip}\hline\noalign{\smallskip}
Network & Multimodal & & mIoU (\%) \\
\noalign{\smallskip}\hline\hline\noalign{\smallskip}
Enet & - && $37.6$ \\
PSPNet & - && $47.5$ \\
3DMV (2d proj) & \checkmark && $49.8$ \\
FuseNet & \checkmark && $52.1$  \\
\hline\noalign{\smallskip}
\textbf{AdapNet++ (ours)} & - && $\mathbf{50.3}$ \\
\textbf{SSMA (ours)} & \checkmark && $\mathbf{57.7}$ \\
\noalign{\smallskip}\hline\noalign{\smallskip}
\end{tabular}
\end{table}

We report results on the ScanNet validation set in \tabref{tab:scannet}. AdapNet++ outperforms the state-of-the-art overall by a margin of $2.83\%$. The large improvement can be attributed to the proposed eASPP which efficiently captures long range context. Context aggregation plays an important role in such cluttered indoor datasets as different parts of an object are occluded from different viewpoints and across scenes. As objects such as the legs of a chair have thin structures, multiscale learning contributes to recovering such structures. We see a similar trend in the performance as in the SUN RGB-D dataset, where our network outperforms the baselines in most of the object categories (16 of the 20 classes) significantly, while yielding a comparable performance for the other categories. The largest improvement of $5.70\%$ is obtained for the \textit{toilet} class, followed by an improvement of $5.34\%$ for the \textit{bed} class which appears as many different variations in the dataset. An interesting observation that can be made is that the highest parametrized network DeconvNet which has $252\million$ parameters has the lowest performance in both SUN RGB-D and ScanNet datasets, while AdapNet++ which has about 1/9th of the parameters, outperforms it by more than twice the margin. However, this is only observed in the indoor datasets, while in the outdoor datasets DeconvNet performs comparable to the other networks. This is primarily due to the fact that indoor datasets have more number of small classes and the predictions of DeconvNet do not retain them.

\tabref{tab:benchmarkingScannet} shows the results on the ScanNet test set. We compare against the top performing models on the leaderboard, namely, FuseNet~\citep{fusenet2016accv}, 3DMV (2d proj)~\citep{dai20183dmv}, PSPNet~\citep{zhao2017pspnet}, and Enet~\citep{paszke2016enet}. Note that 3DMV and FuseNet are multimodal fusion methods. Our proposed AdapNet++ model outperforms all the unimodal networks and achieves state-of-the-art performance for unimodal semantic segmentation on the ScanNet benchmark.

\begin{table*}
\footnotesize 
\centering
\caption{Performance comparison of AdapNet++ with baseline models on the Freiburg Forest validation set (input image dim: $768\times384$). Note that no left-right flips or multiscale testing is performed. Note: Corresponding multimodal results are reported in \tabref{tab:multimodalForest}.}
\label{tab:baselineForest}
\begin{tabular}{p{2.4cm}p{0.7cm}p{0.7cm}p{0.7cm}p{0.7cm}p{0.8cm}|p{0.7cm}p{0.7cm}p{0.7cm}p{1cm}d{3.2}d{3.2}}
\noalign{\smallskip}\hline\noalign{\smallskip}
Network & Trail & Grass & Veg. & Sky & Obst. & mIoU & gIoU & FPR & FNR & \text{Params.} & \text{Time} \\
 & \crule[f_trail]{0.2cm}{0.2cm} & \crule[f_grass]{0.2cm}{0.2cm} & \crule[f_vegetation]{0.2cm}{0.2cm} & \crule[f_sky]{0.2cm}{0.2cm} & \crule[f_obstacle]{0.2cm}{0.2cm} & (\%) & (\%) & (\%) & (\%) & ($\million$) & ($\milli\second$) \\
\noalign{\smallskip}\hline\hline\noalign{\smallskip}
FCN-8s & $82.60$ & $85.69$ & $88.78$ & $89.97$ & $40.40$ & $77.49$ & $86.64$ & $6.21$ & $7.15$ & $134.0$ & $101.99$ \\
SegNet & $82.12$ & $84.99$ & $88.64$ & $89.90$ & $27.23$ & $74.58$ & $86.24$ & $6.77$ & $6.98$ & $29.4$ & $79.13$ \\
ParseNet & $85.67$ & $87.33$ & $89.63$ & $89.17$ & $43.08$ & $78.97$ & $87.65$ & $6.34$ & $6.01$ & $20.5$ & $286.54$ \\
FastNet & $85.70$ & $87.53$ & $90.23$ & $90.73$ & $43.76$ & $79.67$ & $88.18$ & $6.09$ & $5.72$ & $21.0$ & $49.31$ \\
DeconvNet & $86.37$ & $87.17$ & $89.63$ & $92.20$ & $34.80$ & $78.04$ & $89.06$ & $6.53$ & $6.78$ & $252.0$ & $168.54$ \\
DeepLab v2 & $88.75$ & $88.87$ & $90.29$ & $91.65$ & $49.55$ & $81.82$ & $89.98$ & $5.94$ & $5.79$ & $43.7$ & $128.90$ \\
AdapNet & $88.99$ & $88.99$ & $91.18$ & $92.89$ & $48.80$ & $82.17$ & $90.67$ & $4.89$ & $5.42$ & $24.4$ & $61.81$ \\
DeepLab v3 & $88.62$ & $88.84$ & $90.55$ & $91.75$ & $51.61$ & $82.28$ & $90.08$ & $5.21$ & $5.82$ & $58.16$ & $82.83$ \\
\hline\noalign{\smallskip}
\textbf{AdapNet++ (ours)} & $\mathbf{89.27}$ & $\mathbf{89.41}$ & $\mathbf{91.43}$ & $\mathbf{93.01}$ & $\mathbf{52.32}$ & $\mathbf{83.09}$ & $\mathbf{90.75}$ & $\mathbf{4.84}$ & $\mathbf{5.37}$ & $\textbf{28.1}$ & $\textbf{72.77}$ \\
\noalign{\smallskip}\hline\noalign{\smallskip}
\end{tabular}
\end{table*}

Finally, we also benchmark on the Freiburg Forest dataset as it contains several modalities and it is the largest dataset to provide labeled training data for unstructured forested environments. We show the results on the Freiburg Forest dataset in \tabref{tab:baselineForest}, where our proposed AdapNet++ outperforms the state-of-the-art by $0.82\%$. Note that this dataset contains large objects such trees and it does not contain thin structures or objects in multiple scales. Therefore, the improvement produced by AdapNet++ is mostly due to the proposed decoder which yields an improved resolution of segmentation along the object boundaries. The actual utility of this dataset is seen in the qualitative multimodal fusion results, where the fusion helps to improve the segmentation in the presence of disturbances such as glare on the optics and snow. Nevertheless, we see the highest improvement of $3.52\%$ in the \textit{obstacle} class, which is the hardest to segment in this dataset as it contains many different types of objects in one category and it has comparatively fewer examples in the dataset

Moreover, we also compare the number of parameters and the inference time with the baseline networks in \tabref{tab:baselineForest}. Our proposed AdapNet++ architecture performs inference in $72.77\milli\second$ on an NVIDIA TITAN X which is substantially faster than the top performing architectures in all the benchmarks. Most of them consume more than twice the amount of time and the number of parameters making them unsuitable for real-world resource constrained applications. Our critical design choices enable AdapNet++ to consume only $10.98\milli\second$ more than our previously proposed AdapNet, while exceeding its performance in each of the benchmarks by a large margin. This shows that AdapNet++ achieves the right performance vs. compactness trade-off which enables it to be employed in not only resource critical applications, but also in applications that demand efficiency and a fast inference time.

\subsection{AdapNet++ Compression}
\label{sec:resultsCompression}

\begin{table}
\footnotesize 
\renewcommand{\arraystretch}{1.4}
\centering
\caption{Comparison of network compression approaches on our AdapNet++ model trained on the Cityscapes dataset and evaluated on the validation set.}
\label{tab:compression}
\begin{tabular}{p{1.3cm}p{0.8cm}p{1.3cm}d{3.2}d{3.2}d{3.2} @{}}
\noalign{\smallskip}\hline\noalign{\smallskip}
Technique & mIoU & Param. & \text{FLOPS} & \multicolumn{2}{c}{\text{ Reduction \% of }} \\
\cline{5-6}
& (\%) & ($\million$) & ($\billion$) & \text{Param.} & \text{FLOPS} \\ 
\noalign{\smallskip}\hline\hline\noalign{\smallskip}
Original & $80.77$ & $30.20$ & $138.47$ & - & - \\
Baseline & $80.67$ & $28.57$ & $136.05$ & $-5.40$ & $-1.75$  \\
\noalign{\smallskip}\hline\noalign{\smallskip}
\multirow{5}{*}{Oracle} & $80.80$ & $28.34$ & $135.64$ & $-6.15$ & $-2.04$ \\
 & $80.56$ & $23.67$ & $125.33$ & $-21.62$ & $-9.49$ \\
 & $80.18$ & $21.66$ & $83.84$ & $-28.28$ & $-39.45$ \\
 & $79.65$ & $19.91$ & $81.72$ & $-34.07$ & $-40.98$ \\
 & $77.95$ & $17.79$ & $79.84$ & $-41.09$ & $-42.34$ \\
 \noalign{\smallskip}\hline\noalign{\smallskip}
 & $\mathbf{80.80}$ & $28.14$ & $135.17$ & \textbf{-6.82} & \textbf{-2.38} \\
\multirow{2}{*}{Oracle with} & $\mathbf{80.58}$ & $23.16$ & $124.14$ & \textbf{-23.31} & \textbf{-10.34} \\
\multirow{2}{*}{skip (Ours)} & $\mathbf{80.21}$ & $21.11$ & $83.01$ & \textbf{-30.10} & \textbf{-40.05} \\
 & $\mathbf{79.68}$ & $19.75$ & $81.53$ & \textbf{-34.60} & \textbf{-41.12} \\
 & $\mathbf{78.05}$ & $17.63$ & $79.48$ & \textbf{-41.62} & \textbf{-42.60} \\
\noalign{\smallskip}\hline\noalign{\smallskip}
\end{tabular}
\end{table}

In this section, we present empirical evaluations of our proposed pruning strategy that is invariant to shortcut connections in \tabref{tab:compression}. We experiment with pruning entire convolutional filters which results in the removal of its corresponding feature map and the related kernels in the following layer. Most existing approaches only prune the first and the second convolution layer of each residual block, or in addition, equally prune the third convolution layer similar to the shortcut connection. However, this equal pruning strategy always leads to a significant drop in the accuracy of the model that is not recoverable~\citep{li2016pruning}. Therefore, recent approaches have resorted to omitting pruning of these connections. Contrarily, our proposed technique is invariant to the presence of identity or projection shortcut connections, thereby making the pruning more effective and flexible. We employ a greedy pruning approach but rather than pruning layer by layer and fine-tuning the model after each step, we perform pruning of entire residual blocks at once and then perform the fine-tuning. As our network has a total of 75 convolutional and deconvolutional layers, pruning and fine-tuning each layer will be extremely cumbersome. Nevertheless, we expect a higher performance employing a fully greedy approach.

\begin{figure}
\centering
\includegraphics[width=\linewidth]{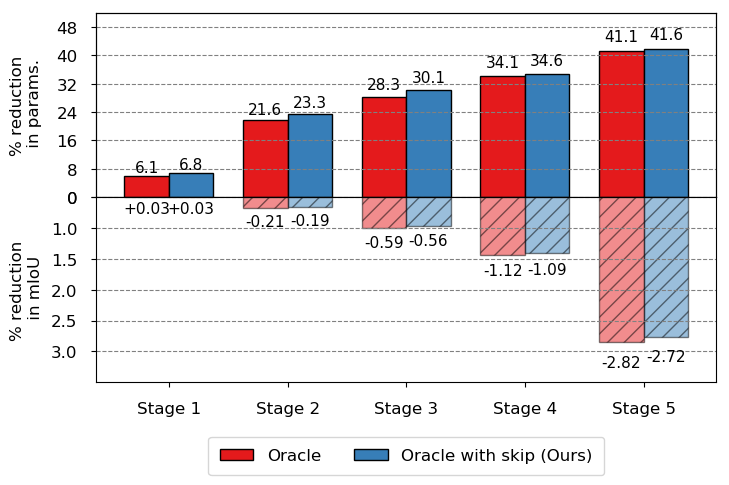}
\caption{Evaluation of network compression approaches shown as the percentage of reduction in the number of parameters with the corresponding decrease in the mIoU score for various baseline approaches versus our proposed technique. The results are shown for the AdapNet++ model trained on the Cityscapes dataset and evaluated on the validation set.}
\label{fig:modelCompression}
\end{figure}

We compare our strategy with a baseline approach~\citep{li2016pruning} that uses the $\ell^1$-norm of the convolutional filters to compute their importance as well as the approach that we build upon that uses the Taylor expansion criteria~\citep{molchanov2016pruning} for the ranking as described in \secref{sec:netCompress}. We denote the approach of \cite{molchanov2016pruning} as Oracle in our results. In the first stage, we start by pruning only the \textit{Res5} block of our model as it contains the most number of filters, therefore, a substantial amount of parameters can be reduced without any loss in accuracy. As shown in \tabref{tab:compression}, our approach enables a reduction of $6.82\%$ of the parameters and $3.3\billion$ FLOPS with a slight increase in the mIoU metric. Similar to our approach the original Oracle approach does not cause a drop in the mIoU metric but achieves a lower reduction in parameters. Whereas, the baseline approach achieves a smaller reduction in the parameters and simultaneously causes a drop in the mIoU score.

Our aim for pruning in the first stage was to compress the model without causing a drop in the segmentation performance, while in the following stages, we aggressively prune the model to achieve the best parameter to performance ratio. Results from this experiment are shown as the percentage in reduction of parameters in comparison to the change in mIoU in \figref{fig:modelCompression}. In the second stage, we prune the convolutional feature maps of \textit{Res2}, \textit{Res3}, \textit{Res4} and \textit{Res5} layers. Using our proposed method, we achieve a reduction of $23.31\%$ of parameters with minor drop of $0.19\%$ in the mIoU score. Whereas, the Oracle approach yields a lower reduction in parameters as well as a larger drop in performance. A similar trend can also be seen for the other pruning stages where our proposed approach yields a higher reduction in parameters and FLOPS with a minor reduction in the mIoU score. This shows that pruning convolutional feature maps with regularity leads to a better compression ratio than selectively pruning layers at different stages of the network. In the third stage, we perform pruning of the deconvolutional feature maps, while in the fourth and fifth stages we further prune all the layers of the network by varying the threshold for the rankings. In the final stage we obtain a reduction of $41.62\%$ of the parameters and $42.60\%$ of FLOPS with a drop of $2.72\%$ in the mIoU score. Considering the compression that can be achieved, this minor drop in the mIoU score is acceptable to enable efficient deployment in resource constrained applications.

\subsection{AdapNet++ Ablation Studies}
\label{sec:ablation}

In order to evaluate the various components of our AdapNet++ architecture, we performed several experiments in different settings. In this section, we study the improvement obtained due to the proposed encoder with the multiscale residual units, a detailed analysis of the proposed eASPP, comparisons with different base encoder network topologies, the improvement that can be obtained by using higher resolution images as input and using multiscale testing. For each of these components, we also study the effect of different parameter configurations. All the ablation studies presented in this section were performed on models trained on the Cityscapes dataset.

\subsubsection{Detailed study on the AdapNet++ Architecture}
\label{sec:dissecting}

\begin{table*}
\footnotesize 
\renewcommand{\arraystretch}{1.4}
\centering
\caption{Effect of the various contributions proposed in the AdapNet++ architecture. The performance is shown for the model trained on the Cityscapes dataset and evaluated on the validation set. $\uparrow_{f}^k$~refers to a deconvolution layer and $c_{f}^k$~refers to a convolution layer with $f$~number of filters, $k \times k$ kernel size and $n$ is the number of classes. PA ResNet-50 refers to the full preactivation ResNet-50 architecture. The weights for the two auxilary losses were set to $\mathcal{L}_1 = 0.5$ and $\mathcal{L}_2 = 0.6$ for this experiment.}
\label{tab:modelEvolution}
\begin{tabular}{p{0.8cm}p{2cm}p{1cm}p{3.6cm}p{1cm}p{1.4cm}p{1.5cm}p{1cm}|p{1cm}}
\noalign{\smallskip}\hline\noalign{\smallskip}
Model & Encoder & MS & Decoder & Skip & ASPP & Aux Loss & He & mIoU \\
 & & Residual & & & dil 3,6,12 & 2x & Init & (\%) \\
\noalign{\smallskip}\hline\hline\noalign{\smallskip}
M1 & ResNet-50 & - & $c^{1}_n\,\uparrow^{16}_n$ & - & - & - & - & $75.22$ \\
M2 & ResNet-50 & \checkmark & $c^{1}_n\,\uparrow^{16}_n$ & - & - & - & - & $76.92$ \\
M3 & ResNet-50 & \checkmark & $c^{1}_n\,\uparrow^{2}_{2n}\,\oplus\,\uparrow^{8}_n$ & 3d & - & - & - & $77.78$ \\
M4 & PA ResNet-50 & \checkmark & $c^{1}_n\,\uparrow^{2}_{2n}\,\oplus\,\uparrow^{8}_n$ & 3d & - & - & - & $78.44$ \\
M5 & PA ResNet-50 & \checkmark & $c^{1}_n\,\uparrow^{2}_{2n}\,\oplus\,\uparrow^{8}_n$ & 3d & \checkmark & - & - & $78.93$ \\
M6 & PA ResNet-50 & \checkmark & $c^{1}_n\,\uparrow^{2}_{2n}\,\oplus\,\uparrow^{2}_{2n}\,\oplus\,\uparrow^{8}_n$ & 3d,2c & \checkmark & - & - & $79.19$ \\
M7 & PA ResNet-50 & \checkmark & $\uparrow^{2}\parallel\,c^3\,c^3\,\uparrow^{2}\parallel\,c^3\,c^3\,c^{1}_n\,\uparrow^{8}_n$ & 3d,2c & \checkmark & - & - & $79.82$ \\
M8 & PA ResNet-50 & \checkmark & $\uparrow^{2}\parallel\,c^3\,c^3\,\uparrow^{2}\parallel\,c^3\,c^3\,c^{1}_n\,\uparrow^{8}_n$ & 3d,2c & \checkmark & \checkmark & - & $80.34$ \\
M9 & PA ResNet-50 & \checkmark & $\uparrow^{2}\parallel\,c^3\,c^3\,\uparrow^{2}\parallel\,c^3\,c^3\,c^{1}_n\,\uparrow^{8}_n$ & 3d,2c & \checkmark & \checkmark & \checkmark & $\mathbf{80.67}$ \\
\noalign{\smallskip}\hline\noalign{\smallskip}
\end{tabular}
\end{table*}

We first study the major contributions made to the encoder as well as the decoder in our proposed AdapNet++ architecture. \tabref{tab:modelEvolution} shows results from this experiment and subsequent improvement due to each of the configurations. The simple base model M1 consisting of the standard ResNet-50 for the encoder and a single deconvolution layer for upsampling achieves a mIoU of $75.22\%$. The model M2 that incorporates our multiscale residual units achieves an improvement of $1.7\%$ without any increase in the memory consumption. Whereas, the multigrid approach from DeepLab~v3~\citep{chen2017rethinking} in the same configuration achieves only $0.38\%$ of improvement in the mIoU score. This shows the novelty in employing our multiscale residual units for efficiently learning multiscale features throughout the network. In the M3 model, we study the effect of incorporating skip connections for refinement. Skip connections that were initially introduced in the FCN architecture are still widely used for improving the resolution of the segmentation by incorporating low or mid-level features from the encoder into the decoder while upsampling. The ResNet-50 architecture contains the most discriminative features in the middle of the network. In our M3 model, we first upsample the encoder output by a factor two, followed by fusing the features from \textit{Res3d} block of the encoder to refinement and subsequently using another deconvolutional layer to upsample back to input resolution. This model achieves a further improvement of $0.86\%$.

In the M4 model, we replace the standard residual units with the full pre-activation residual units which yields an improvement of $0.66\%$. As mentioned in the work by \cite{he2016identity}, the results corroborate that pre-activation residual units yields a lower error than standard residual units due to the ease of training and improved generalization capability. Aggregating multiscale context using ASPP has become standard practice in most classification and segmentation networks. In the M5 model, we add the ASPP module to the end of the encoder segment. This model demonstrates an improved mIoU of $78.93\%$ due to the ability of the ASPP to capture long range context. In the subsequent M6 model, we study if adding another skip refinement connection from the encoder yields a better performance. This was challenging as most combinations along with the \textit{Res3d} skip connection did not demonstrate any improvement. However, adding a skip connection from \textit{Res2c} showed a slight improvement.

In all the models upto this stage, we fused the mid-level encoder features into the decoder using element-wise addition. In order to make the decoder stronger, we experiment with improving the learned decoder representations with additional convolutions after concatenation of the mid-level features. Specifically, the M7 model has three upsampling stages, the first two stages consist of a deconvolution layer that upsamples by a factor of two, followed by concatenation of the mid-level features and two following $3\times3$ convolutions that learn highly discriminative fused features. This model shows an improvement of $0.63\%$ which is primarily due to the improved segmentation along the object boundaries as demonstrated in the qualitative results in \secref{sec:qualitative}. Our M7 model contains a total of 75 convolutional and deconvolutional layers, making the optimization challenging. In order to accelerate the training and to further improve the segmentation along object boundaries, we propose a multiresolution supervision scheme in which we add a weighted auxiliary loss to each of the first two upsampling stages. This model denoted as M8 achieves an improved mIoU of $80.34\%$. In comparison to aforementioned scheme, we also experimented with adding a weighted auxiliary loss at the end of the encoder of the M7 model, however it did not improve the performance, although it accelerated the training. Finally we also experimented with initializing the layers with the He initialization~\citep{he2015delving} scheme (also known as MSRA) in the M9 model which further boosts the mIoU to $80.67\%$. The following section further builds upon the M9 model to yield the topology of our proposed AdapNet++ architecture.

\subsubsection{Detailed study on the eASPP}
\label{sec:studyeASPP}

In this section, we quantitatively and qualitatively evaluate the performance of our proposed eASPP configuration and the new decoder topology. We perform all the experiments in this section using the best performing M9 model described in \secref{sec:dissecting} and report the results on the Cityscapes validation set in \tabref{tab:model1331}. In the first configuration of the M91 model, we employ a single $3\times3$ atrous convolution in the ASPP, similar the configuration proposed in DeepLab~v3~\citep{chen2017rethinking} and use a single $1\times 1$ convolution in the place of the two $3\times 3$ convolutions in the decoder of the M9 model. This model achieves an mIoU score of $80.06\%$ with $41.3\million$ parameters and consumes $115.99\billion$ FLOPS. In order to better fuse the concatenated mid-level features with the decoder and to improve its discriminability, we replace the $1\times 1$ convolution layer with a $3\times 3$ convolution in the M92 model and with two $3\times 3$ convolutions in the M93 model. Both these models demonstrate an increase in performance corroborating that a simple $1\times 1$ convolution is insufficient for object boundary refinement using fusion of mid-level encoder features.

\begin{table}
\footnotesize
\renewcommand{\arraystretch}{1.4}
\centering
\caption{Evaluation of various atrous spatial pyramid pooling configurations. The performance is shown for the model trained on the Cityscapes dataset and evaluated on the validation set. $\uparrow_{f}^k$~refers to a deconvolution layer and $c_{f}^k$~refers to a convolution layer with $f$~number of filters and $k \times k$ kernel size. The weights for the two auxilary losses were set to $\mathcal{L}_1 = 0.5$ and $\mathcal{L}_2 = 0.6$ for this experiment.}
\label{tab:model1331}
\begin{tabular}{p{0.6cm}p{1.65cm}p{1.8cm}|p{0.45cm}p{0.45cm}p{0.9cm}}
\noalign{\smallskip}\hline\noalign{\smallskip}
Model & ASPP Conv. & Decoder Conv. & mIoU & Param & FLOPS \\
 & & & (\%) & ($\million$) & ($\billion$) \\
\noalign{\smallskip}\hline\hline\noalign{\smallskip}
M91 & $[c_{256}^3]$ & $[c_{256}^1]$ & $80.06$ & $41.3$ & $115.99$ \\
M92 & $[c_{256}^3]$ & $[c_{256}^3]$ & $80.27$ & $42.5$ & $142.42$ \\
M93 & $[c_{256}^3]$ & $[c_{256}^3]\times 2$ & $80.67$ & $43.7$ & $169.62$ \\
M94 & $[c_{64}^1\,c_{64}^3\,c_{256}^1]$ & $[c_{256}^3]\times 2$ & $80.42$ & $30.1$ & $138.21$ \\
M95 & $[c_{64}^1\,c_{64}^3\,c_{64}^3\,c_{256}^1]$ & $[c_{256}^3]\times 2$ & $\mathbf{80.77}$ & $\textbf{30.2}$ & $\textbf{138.47}$ \\
\noalign{\smallskip}\hline\noalign{\smallskip}
\end{tabular}
\end{table}

\begin{figure*}
\centering
\footnotesize
\setlength{\tabcolsep}{0.1cm}
\begin{tabular}{P{0.25cm}P{4cm}P{4cm}P{4cm}P{4cm}}
\rot{ASPP} & \raisebox{-0.4\height}{\includegraphics[width=\linewidth]{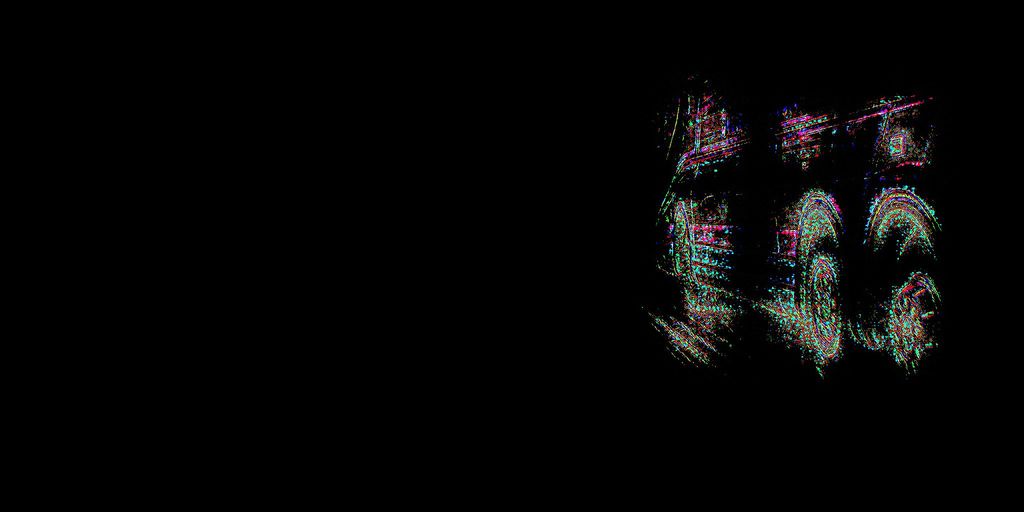}} & 
\raisebox{-0.4\height}{\includegraphics[width=\linewidth]{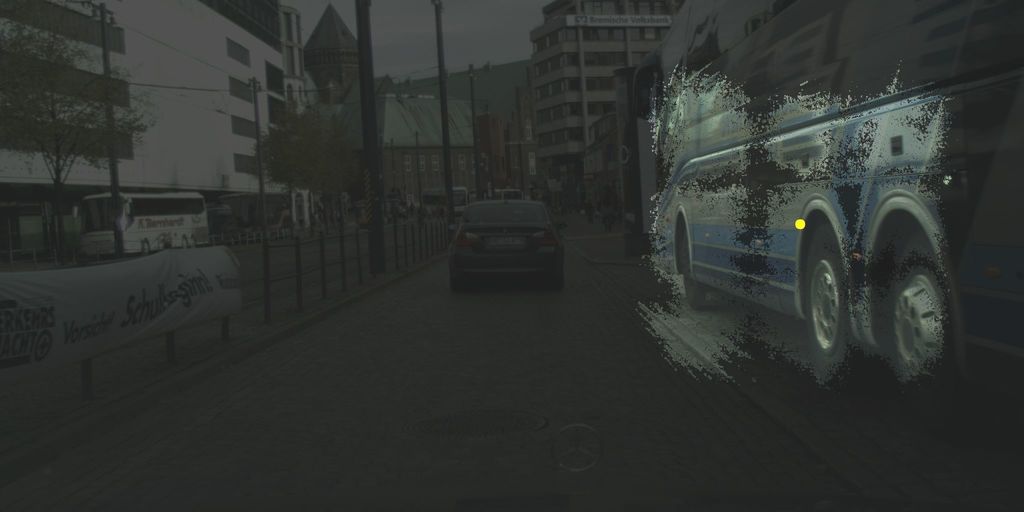}} &
\raisebox{-0.4\height}{\includegraphics[width=\linewidth]{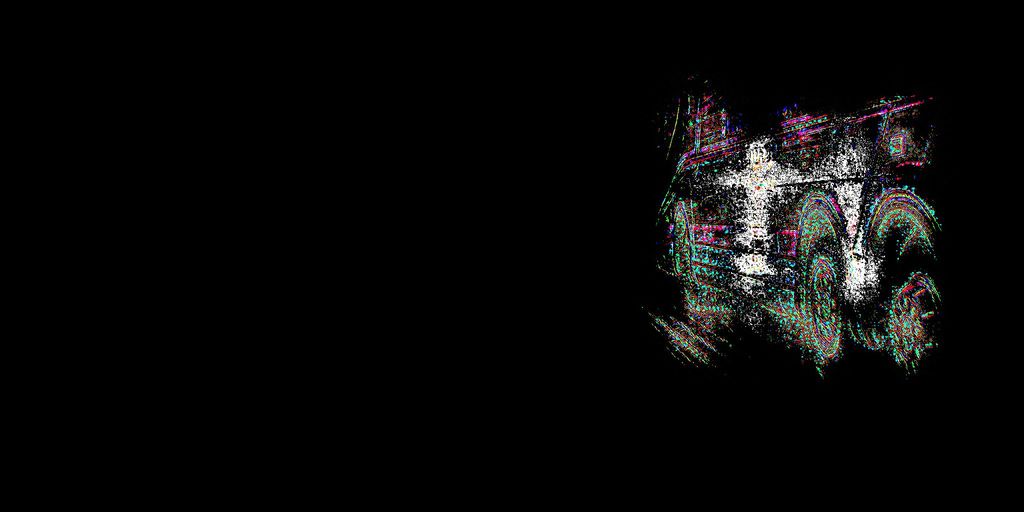}} & 
\raisebox{-0.4\height}{\includegraphics[width=\linewidth]{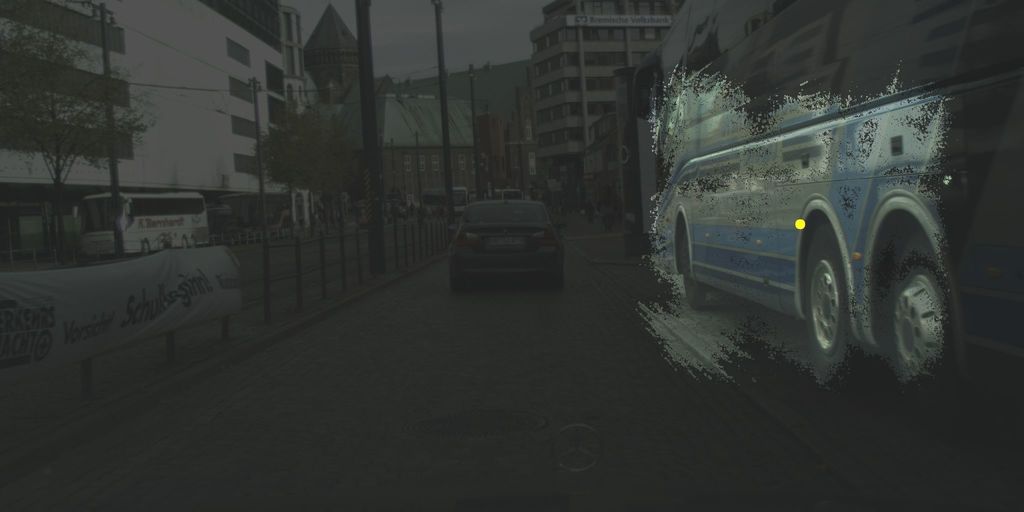}} \\
& \raisebox{-0.4\height}{(a) Sensitive area of dil=12} & \raisebox{-0.4\height}{(b) dil=12 receptive field} & \raisebox{-0.4\height}{(c) Full sensitive area} & \raisebox{-0.4\height}{(d) Full receptive field} \\
\\
\rot{eASPP} & \raisebox{-0.45\height}{\includegraphics[width=\linewidth]{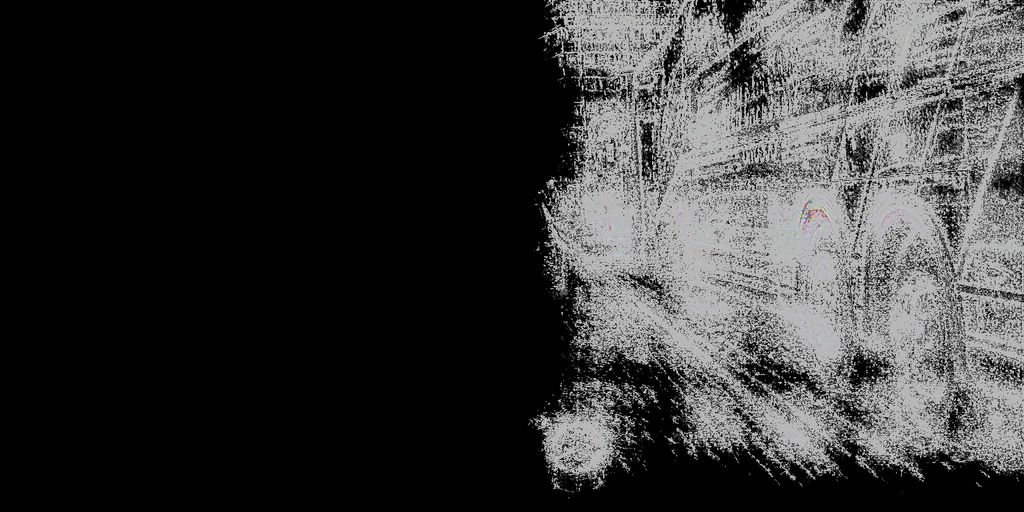}} &
\raisebox{-0.45\height}{\includegraphics[width=\linewidth]{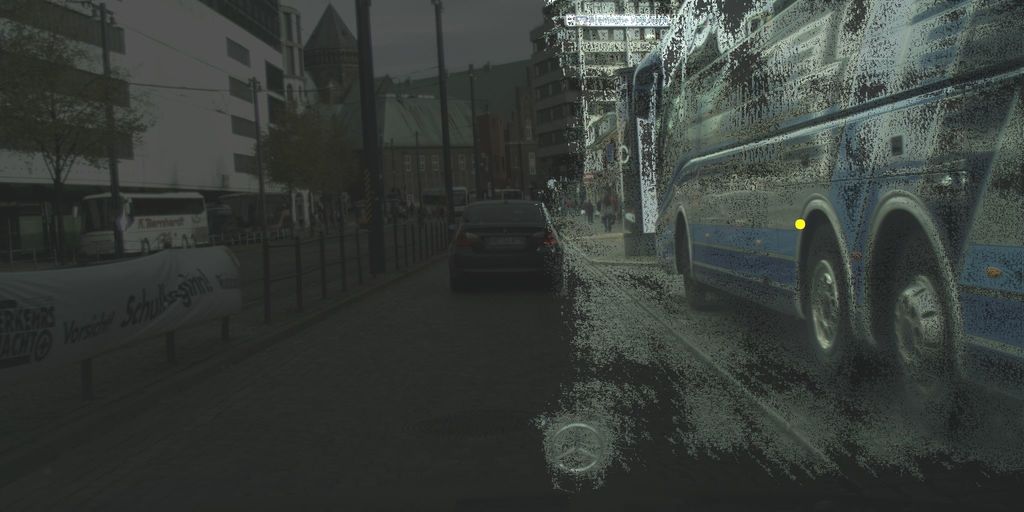}} &
\raisebox{-0.45\height}{\includegraphics[width=\linewidth]{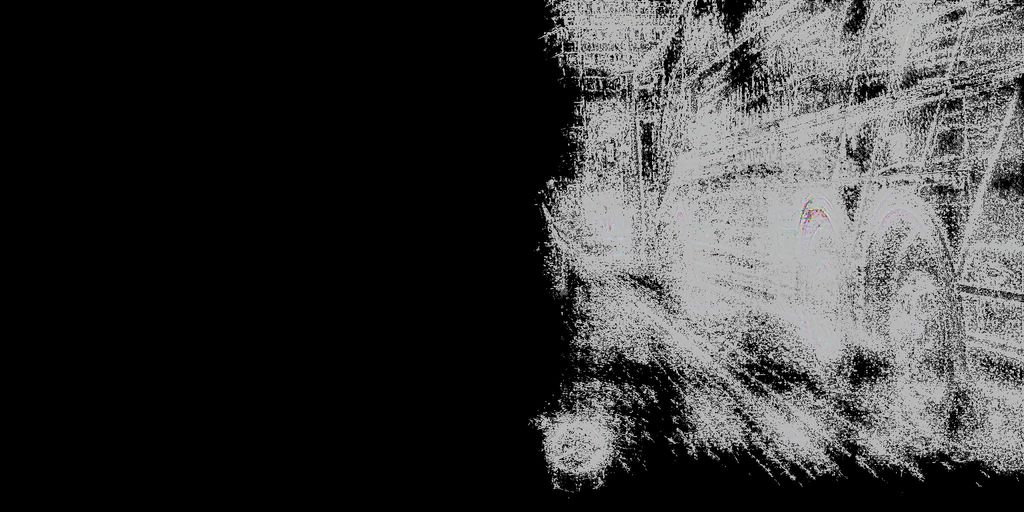}} &
\raisebox{-0.45\height}{\includegraphics[width=\linewidth]{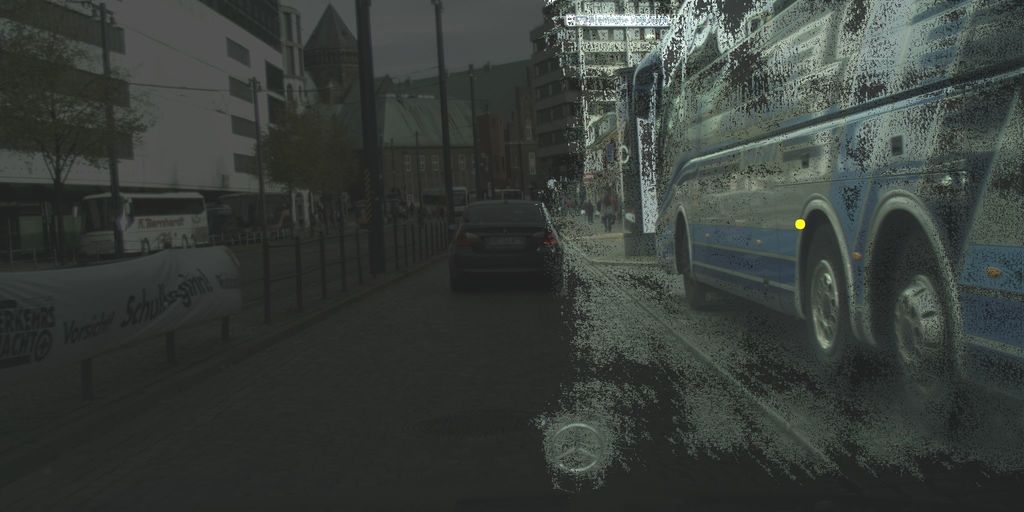}} \\
& \raisebox{-0.45\height}{(e) Sensitive area of dil=12} & \raisebox{-0.45\height}{(f) dil=12 receptive field} & \raisebox{-0.45\height}{(g) Full sensitive area} &  \raisebox{-0.45\height}{(h) Full receptive field} \\
\end{tabular}
\caption{Comparison of the receptive field of ASPP and our proposed eASPP. The receptive field is visualized for the annotated yellow dot. Our proposed eASPP has larger receptive field size and denser pixel sampling in comparison to the ASPP.}
\label{fig:asppReceptive}
\end{figure*}

In an effort to reduce the number of parameters, we employ a bottleneck architecture in the ASPP of the M94 model by replacing the $3\times 3$ atrous convolution with a structure consisting of a $1\times 1$ convolution with half the number of filters, followed by a $3\times 3$ atrous convolution with half the number of filters and another $1\times 1$ convolution with the original amount of filters. This model achieves an mIoU score of $80.42\%$ which accounts to a reduction of $0.25\%$ in comparison to the M93 model, however, it reduces computational requirement by $13.6\million$ parameters and $31.41\billion$ FLOPS which makes the model very efficient. Nevertheless, this drop in performance is not ideal. Therefore, in order to compensate for this drop, we leverage the idea of cascading atrous convolutions that enables an increase in the size of the effective receptive field and the density of the pixel sampling. Specifically, in the M95 model, we add a cascaded $3\times 3$ atrous convolution in place of the single $3\times 3$ atrous convolution in the M94 model. This model achieves a mIoU score of $80.77\%$ which is an increase of $0.35\%$ in the mIoU with only a minor increase of $0.1\million$ parameters in comparison to our M94 model. The originally proposed ASPP module consumes $15.53\million$ parameters and $34.58\billion$ FLOPS, where the cascaded bottleneck structure in the M95 model only consumes $2.04\million$ parameters and $3.62\billion$ FLOPS which is over 10 times more computationally efficient. We denote this cascaded bottleneck structure as eASPP.

\begin{table}
\footnotesize
\centering
\caption{Performance comparison of our proposed eASPP with various other ASPP configurations. The results are reported for the models trained on the Cityscapes dataset and evaluated on the validation set.}
\label{tab:AllAsppComp}
\begin{tabular}{p{4cm}p{0.85cm}p{0.85cm}p{0.9cm}}
\noalign{\smallskip}\hline\noalign{\smallskip}
ASPP Topology & mIoU & Param & FLOPS \\
 & (\%) & ($\million$) & ($\billion$) \\
\noalign{\smallskip}\hline\hline\noalign{\smallskip}
ASPP~v2 & $80.25$ & $18.87$ & $50.96$ \\
ASPP~v3 & $80.67$ & $15.53$ & $34.58$ \\
ASPP~v3 with Separable Conv. & $80.27$ & $3.00$ & $5.56$ \\
DenseASPP & $80.62$ & $4.23$ & $9.74$ \\
eASPP (Ours) & $\mathbf{80.77}$ & $\textbf{2.04}$ & $\textbf{3.62}$ \\
\noalign{\smallskip}\hline\noalign{\smallskip}
\end{tabular}
\end{table}

Furthermore, we present detailed experimental comparisons of our proposed eASPP with other ASPP configurations in \tabref{tab:AllAsppComp}. Specifically, we compare against the initial ASPP configuration proposed in DeepLab v2~\citep{ChenPK0Y16} which we denote as ASPP~v2, the improved ASPP configuration that also incorporates image-level features as proposed in DeepLab~v3~\citep{chen2017rethinking} which we denote as ASPP~v3, the ASPP configuration with separable convolutions and the more recently proposed DenseASPP~\citep{yang2018denseaspp} configuration. In order to have a fair comparison, we use the same AdapNet++ architecture with the different ASPP configurations for this experiment and present the results on the Cityscapes validation set. The four parallel atrous convolution layers in the ASPP~v2 configuration of DeepLab~v2 have the number of feature channels equal to the number of object classes in the dataset, while the ASPP~v3 configuration of DeepLab~v3 has the number of feature channels equal to 256 in the three parallel atrous convolution layers.

The ASPP~v2 model with the convolution feature channels set to the number of object classes achieves a mIoU of $79.22\%$ and increasing the number of convolution feature channels to 256 yields a mIoU of $80.25\%$. By incorporating image-level features using a global pooling layer and removing the fourth parallel atrous convolution in ASPP~v2, the ASPP~v3 model achieves an improved performance of $80.67\%$ with a minor decrease in the parameters and FLOPs. Recently, separable convolutions are being employed in place of the standard convolution layer as an efficient alternative to reduce the model size. Employing atrous separable convolutions in the ASPP~v3 configuration significantly reduces the number of parameters and FLOPs consumed by the model to $3\million$ and $5.56\billion$ respectively. However, this also reduces the mIoU of the model to $80.27\%$ which is comparable to the ASPP~v2 configuration with 256 convolutional filters. The model with the DenseASPP configuration achieves a mIoU of $80.62\%$ which is still lower than the ASPP~v3 configuration but it reduces the number of parameters and FLOPs to $4.23\million$ and $9.74\billion$ respectively. It should be noted that in the work of ~\cite{yang2018denseaspp}, DenseASPP was only compared to ASPP~v2 with the number of convolutional feature channels equal to the number of object classes (mIoU of $79.22\%$). In comparison to the aforementioned ASPP topologies, our proposed eASPP achieves the highest mIoU score of $80.77\%$ with the lowest consumption of parameters and FLOPs. This accounts to a reduction of $86.86\%$ of the number of parameters and $89.53\%$ of FLOPs with a increase in the mIoU compared to the previously best performing ASPP~v3 topology. 

In order to illustrate the phenomenon caused by cascading atrous convolutions in our proposed eASPP, we visualize the empirical receptive field using the approach proposed by \cite{zhou2014object}. First, for each feature vector representing an image patch, we use a $8\times 8$ mean image to occlude the patch at different locations using a sliding window. We then record the change in the activation by measuring the Euclidean distances as a heat map which indicates which regions are sensitive to the feature vector. Although the size of the empirical receptive fields is smaller than theoretical receptive fields, they are better localized and more representative of the information they capture~\citep{zhou2014object}. In \figref{fig:asppReceptive}, we show visualizations of the empirical receptive field size of the convolution layer of the ASPP that has one $3\times 3$ atrous convolution in each branch in comparison to our M95 model that has cascaded $3\times 3$ atrous convolutions. \figref{fig:asppReceptive}(b) and (c) show the receptive field at the annotated yellow dot for the atrous convolution with the largest dilation rate in ASPP and in our eASPP correspondingly. It can be seen that our eASPP has a much larger receptive field that enables capturing large contexts. Moreover, it can be seen that the pixels are sampled much denser in our eASPP in comparison to the ASPP. In \figref{fig:asppReceptive}(d) and (h), we show the aggregated receptive fields of the entire module in which it can be observed that our eASPP has much lesser isolated points of focus and a cleaner sensitive area than the ASPP. We evaluated the generalization of our proposed eASPP by incorporating the module into our AdapNet++ architecture and benchmarking its performance in comparison to DeepLab which incorporates the ASPP. The results presented in \secref{sec:benchm} demonstrate that our eASPP effectively generalizes to a wide range of datasets containing diverse environments.

\subsubsection{Improving Granularity of Segmentation}
\label{sec:skipAblation}

\begin{table}
\footnotesize
\centering
\caption{Effect on varying the number of filters in the skip refinement connections in the M95 model. The performance is shown for the model trained on the Cityscapes dataset and evaluated on the validation set}.
\label{tab:modelSkip}
\begin{tabular}{p{1.8cm}|p{0.8cm}p{0.8cm}p{0.8cm}p{0.8cm}p{0.8cm}p{0.8cm}}
\noalign{\smallskip}\hline\noalign{\smallskip}
Skip Channels & 12 & 24 & 36 & 48 & 60 \\
\noalign{\smallskip}\hline\noalign{\smallskip}
mIoU (\%) & $80.50$ & $\textbf{80.77}$ & $80.67$ & $80.59$ & $80.56$ \\
\noalign{\smallskip}\hline\noalign{\smallskip}
\end{tabular}
\end{table}

In our AdapNet++ architecture, we propose two strategies to improve the segmentation along object boundaries in addition to the new decoder architecture. The first being the two skip refinement stages that fuse mid-level encoder features from \textit{Res3d} and \textit{Res2c} into the decoder for object boundary refinement. However, as the low and mid-level features have a large number of filters (512 in \textit{Res3d} and 256 in \textit{Res2c}) in comparison to the decoder filters that only have 256 feature channels, they will outweigh the high level features and decrease the performance. Therefore, we employ a $1\times 1$ convolution to reduce the number of feature channels in the low and mid-level representations before fusing them into the decoder. In \tabref{tab:modelSkip}, we show results varying the number of feature channels in the $1\times 1$ skip refinement convolutions in the M95 model from \secref{sec:studyeASPP}. We obtain the best results by reducing the number of mid-level encoder feature channels to 24 using the $1\times 1$ convolution layer.

\begin{table*}
\footnotesize
\centering
\caption{Effect on varying the weighting factor of the auxiliary losses in the M95 model. The performance is shown for the model trained on the Cityscapes dataset and evaluated on the validation set.}
\label{tab:auxWeights}
\begin{tabular}{p{1.8cm}|p{0.55cm}p{0.55cm}p{0.55cm}p{0.55cm}p{0.55cm}p{0.55cm}p{0.55cm}p{0.55cm}p{0.55cm}p{0.55cm}
p{0.55cm}p{0.55cm}p{0.55cm}}
\noalign{\smallskip}\hline\noalign{\smallskip}
Aux 1 Weight & 0.3 & 0.4 & 0.4 & 0.5 & 0.5 & 0.5 & 0.6 & 0.6 & 0.6 & 0.7 & 0.7 & 1.0 \\
Aux 2 Weight & 0.2 & 0.2 & 0.3 & 0.4 & 0.6 & 0.7 & 0.4 & 0.5 & 0.7 & 0.5 & 0.6 & 1.0 \\
\noalign{\smallskip}\hline\noalign{\smallskip}
mIoU (\%) & $80.51$ & $80.55$ & $80.68$ & $80.60$ & $80.55$ & $80.49$ & $80.53$ & $\textbf{80.77}$ & $80.69$ & $80.55$ & $80.71$ & $80.37$ \\
\noalign{\smallskip}\hline\noalign{\smallskip}
\end{tabular}
\end{table*}

The second strategy that we employ for improving the segmentation along object boundaries is using our proposed multiresolution supervision scheme. As described in \secref{sec:deepSupervision}, we employ auxiliary loss branches after each of the first two upsampling stages in the decoder to improve the resolution of the segmentation and to accelerate training. Weighing the two auxiliary losses is critical to balance the gradient flow through all the previous layers of the network. We experiment with different loss weightings and report results for the same M95 model in \tabref{tab:auxWeights}. The network achieves the highest performance for auxiliary loss weightings $\lambda_1 = 0.6$ and $\lambda_2 = 0.5$ for $\mathcal{L}_{aux1}$ and $\mathcal{L}_{aux2}$ respectively. 

\begin{figure}
\centering
\includegraphics[width=\linewidth]{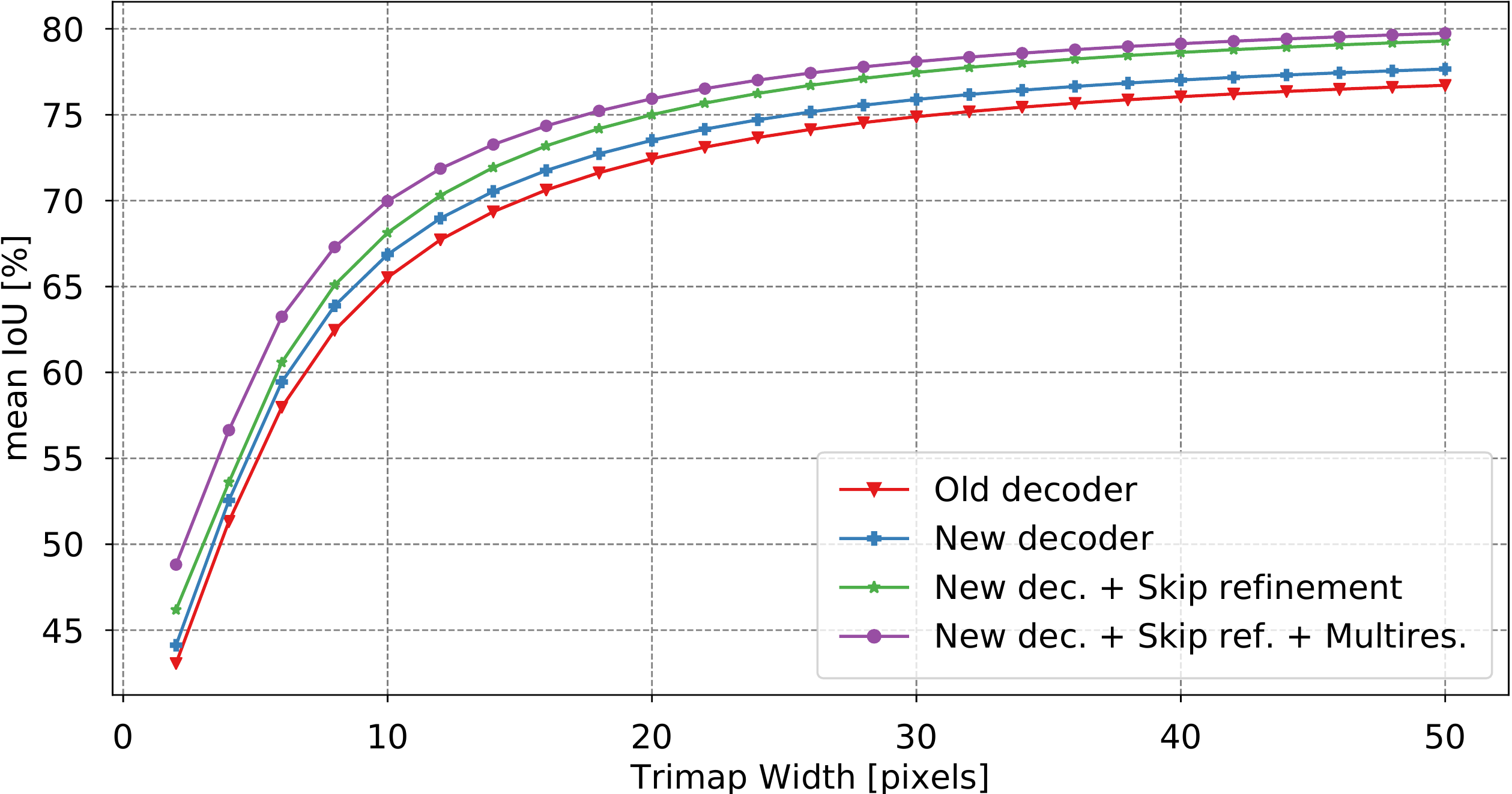}
\caption{Influence of the proposed decoder, skip refinement and multiresolution supervision on the segmentation along objects boundaries using the trimap experiment. The plot shows the mIoU score as a function of the trimap band width along the object boundaries. The results are shown on the Cityscapes dataset and evaluated on the validation set.}
\label{fig:trimap}
\end{figure}

In order to quantify the improvement specifically along the object boundaries, we evaluate the performance of our architecture using the trimap experiment~\citep{kohli2009robust}. The mIoU score for the pixels that are within the trimap band of the void class labels (255) are computed by applying the morphological dilation on the void labels. Results from this experiment shown in \figref{fig:trimap} demonstrates that our new decoder improves the performance along object boundaries compared to the decoder in AdapNet~\citep{valada17icra}, while the M7 model with the new decoder and the skip refinement further improves the performance. Finally, the M8 model consisting of our new decoder with the skip refinement stages and our multiresolution supervision scheme for training significantly improves the segmentation along the boundaries which is more evident when the trimap band is narrow.

\subsubsection{Encoder Topology}

In recent years, several efficient network architectures have been proposed for image classification that are computationally inexpensive and have fast inference times. In order to study the trade-off between accuracy and computational requirement, we performed experiments using five widely employed architectures as the encoder backbone. Specifically, we evaluate the performance using ResNet-50~\citep{He2015}, full pre-activation ResNet-50~\citep{he2016identity}, ResNeXt~\citep{xie2017aggregated}, SEnet~\citep{hu2017squeeze} and Xception~\citep{chollet2016xception} architectures for the encoder topology and augmented them with our proposed modules similar to the M95 model described in \secref{sec:studyeASPP}. 

\begin{table}
\footnotesize
\centering
\caption{Effect of various encoder topologies in the M95 model. The performance is shown for the model trained on the Cityscapes dataset and evaluated on the validation set.}
\label{tab:encoderModel}
\begin{tabular}{@{} p{1.4cm}|p{0.8cm}p{1.3cm}p{0.8cm}p{0.6cm}p{1.2cm} @{}}
\noalign{\smallskip}\hline\noalign{\smallskip}
Encoder & ResNet & PA ResNet & ResNeXt & SEnet & Xception \\
\noalign{\smallskip}\hline\noalign{\smallskip}
mIoU (\%) & $79.32$ & $\mathbf{80.77}$ & $80.30$ & $78.31$ & $78.70$ \\
Param ($\million$) & $30.2$ & $30.2$ & $29.7$ & $32.7$ & $27.5$ \\
FLOPS ($\billion$) & $135.28$ & $138.47$ & $145.81$ & $145.34$ & $137.06$ \\
\noalign{\smallskip}\hline\noalign{\smallskip}
\end{tabular}
\end{table}

Results from this experiment are shown in \tabref{tab:encoderModel}. Note that in the comparisons presented in this section, no model compression has been performed. It can be seen that the full pre-activation ResNet-50 model achieves the highest mIoU score, closely followed by the ResNeXt model. However the ResNeXt model has an additional $7.34\million$ parameters with a slightly lesser number of FLOPS. While, the standard ResNet-50 architecture has $3.19\million$ parameters lesser than the full pre-activation ResNet-50 model, it achieves a lower mIoU score of $79.32\%$. Therefore, we choose the full-preactivation ResNet-50 architecture as the encoder backbone in our proposed AdapNet++ architecture.

\subsubsection{Decoder Topology}

In this section, we compare the performance and computational efficiency of the new strong decoder that we introduce in our proposed AdapNet++ architecture with other existing progressive upsampling decoders. For a fair comparison, we employ the same AdapNet++ encoder in all the models in this experiment and only replace the decoder with the topologies proposed in LRR~\citep{ghiasi2016laplacian} and RefineNet~\citep{lin2017refinenet}. All these decoder topologies that we compare with utilize similar stage-wise upsampling with deconvolution layers and refinement with skip connections from higher resolution encoder feature maps. As a reference, we also compare with a model that employs the AdapNet++ encoder and direct bilinear upsampling for the decoder. Therefore, this reference model does not consume any parameters and FLOPs for the decoder section.

\begin{table}
\footnotesize
\centering
\caption{Performance of the strong decoder that we introduce in our AdapNet++ architecture in comparison with other progressive upsampling decoder topologies. All the models employ the same AdapNet++ encoder with only the decoder replaced with alternative topologies and the results are reported on the Cityscapes validation set.}
\label{tab:decoderAblation}
\begin{tabular}{p{4cm}p{0.85cm}p{0.85cm}p{0.9cm}}
\noalign{\smallskip}\hline\noalign{\smallskip}
Decoder Topology & mIoU & Param & FLOPs \\
 & (\%) & ($\million$) & ($\billion$) \\
\noalign{\smallskip}\hline\hline\noalign{\smallskip}
Bilinear upsampling & $74.83$ & $-$ & $-$ \\
LRR & $79.38$ & $2.01$ & $75.88$ \\
RefineNet & $80.44$ & $14.29$ & $265.33$ \\
AdapNet++ (Ours) & $\mathbf{80.77}$ & $4.59$ & $69.84$ \\
\noalign{\smallskip}\hline\noalign{\smallskip}
\end{tabular}
\end{table}

\tabref{tab:decoderAblation} shows the results from this experiment in which we see that the reference model with direct bilinear upsampling achieves a mIoU score of $74.83\%$. The LRR decoder model outperforms the reference model and the RefineNet decoder model further outperforms the LRR decoder model. However, the RefineNet decoder consumes a significant amount of parameters and FLOPs compared to the LRR decoder. Nevertheless, our proposed decoder in AdapNet++ achieves the highest mIoU score of $80.77\%$, thereby outperforming both the other competing progressive upsampling decoders while consuming the lowest amount of FLOPs and still maintaining a good parameter efficiency.

\subsubsection{Image Resolution and Testing Strategies}
\label{sec:highRes}

We further performed experiments using input images with larger resolutions as well as with left-right flipped inputs and multiscale inputs while testing. In all our benchmarking experiments, we use an input image with a resolution of $768\times 384$ pixels in order enable training of the multimodal fusion model that has two encoder streams on a single GPU. State-of-the-art semantic segmentation architectures use multiple crops from the full resolution of the image as input. For example for the Cityscapes dataset, eight crops of $720\times 720$ pixels from each full resolution image of $2048\times 1024$ pixels are often used. This yields a downsampled output with a larger resolution at the end of the encoder, thereby leading to a lesser loss of information due to downsampling and more boundary delineation. Employing a larger resolution image as input also enables better segmentation of small objects that are at far away distances, especially in urban driving datasets such as Cityscapes. However, the caveat being that it requires multi-GPU training with synchronized batch normalization in order to utilize a large enough mini-batch size, which makes the training more cumbersome. Moreover, using large crops of the full resolution image significantly increases the inference time of the model as the inference time for one image is the sum of the inference time consumed for each of the crops.

\begin{table}
\footnotesize 
\renewcommand{\arraystretch}{1.4}
\centering
\caption{Effect on using a higher resolution input image and employing left-right flip as well as multiscale inputs during testing. The performance is shown for the model trained on the Cityscapes dataset and evaluated on the validation set.}
\label{tab:imageSize}
\begin{tabular}{p{1.5cm}p{0.4cm}p{0.4cm}p{0.7cm}p{0.7cm}p{0.7cm}p{1cm}}
\noalign{\smallskip}\hline\noalign{\smallskip}
Image Size & Flip & MS & mIoU & Acc. & AP & Time \\
(pixels) & & & (\%) & (\%) & (\%) & $(\milli\second)$ \\
\noalign{\smallskip}\hline\hline\noalign{\smallskip}
$768\times384$ & - & - & $80.77$ & $96.04$ & $90.97$ & $72.77$ \\
$768\times384$ & \checkmark & - & $81.35$ & $96.18$ & $90.76$ & $148.93$ \\
$768\times384$ & \checkmark & \checkmark & $82.25$ & $96.36$ & $91.86$ & $1775.96$ \\
\noalign{\smallskip}\hline\noalign{\smallskip}
$896\times448$ & - & - & $81.69$ & $96.06$ & $89.96$ & $88.89$ \\
$896\times448$ & \checkmark & - & $82.28$ & $96.21$ & $90.53$ & $183.57$ \\
$896\times448$ & \checkmark & \checkmark & $83.19$ & $96.48$ & $91.52$ & $2342.31$ \\
\noalign{\smallskip}\hline\noalign{\smallskip}
$1024\times512$ & - & - & $82.47$ & $96.13$ & $90.63$ & $105.94$ \\
$1024\times512$ & \checkmark & - & $83.07$ & $96.28$ & $91.17$ & $219.28$ \\
$1024\times512$ & \checkmark & \checkmark & $84.27$ & $96.66$ & $92.36$ & $3061.11$ \\
\noalign{\smallskip}\hline\noalign{\smallskip}
$2048\times1024$ & - & - & $83.10$ & $96.25$ & $90.87$ & $494.97$ \\
$2048\times1024$ & \checkmark & - & $83.58$ & $96.37$ & $91.14$ & $1022.12$ \\
$2048\times1024$ & \checkmark & \checkmark & $\mathbf{84.54}$ & $\mathbf{96.74}$ & $\mathbf{92.48}$ & $12188.57$ \\
\noalign{\smallskip}\hline\noalign{\smallskip}
\end{tabular}
\end{table}

Nevertheless, we present experimental results with input images of resolution $896\times 448$ pixels, $1024\times 512$ pixels, and eight crops of $720\times 720$ pixels from the full resolution of $2048\times 1024$ pixels, in addition to the resolution of $768\times 384$ pixels that we use. In addition to the varying input image resolutions, we also test with left-right flips and multiscale inputs. However, although this increases the mIoU score it substantially increases the computation complexity and runtime, therefore rendering it not useful in real-world applications. A summary of the results from this experiment are shown in \tabref{tab:imageSize}. It can be seen that with each higher resolution image, the model yields an increased mIoU score and simultaneously consumes a larger inference time. Similarly, left-right flips and multiscale inputs also yield an improvement in the mIoU score. For the input image resolution of $768\times 384$ pixels that we employ in the benchmarking experiments, left-right flips yields an increase of $0.58\%$ in the mIoU, while multiscale inputs in addition, yields a further improvement of $0.9\%$. The corresponding pixel accuracy and and average precision also shows an improvement. The model trained with eights crops of $720\times 720$ pixels from each full resolution image of $2048\times 1024$ pixels demonstrates an improvement of $2.33\%$ in the mIoU score in comparison to the model with a lower resolution image that we use for benchmarking. Furthermore, using left-right flips and multiscale inputs yields an overall improvement of $3.77\%$ in the mIoU and additional improvements in the other metrics in comparison to the benchmarking model. However, it can be seen that the inference time full resolution model is $494.98\milli\second$ and multiscale testing with left-right flips further increases it to $12188.57\milli\second$, while the inference time of the model that uses a image resolution of $768\times 384$ pixels is only $72.77\milli\second$ demonstrating that using full resolution of the image and multiscale testing with left-right flips for real-world robotics applications is impractical.

\subsection{Qualitative Results of Unimodal Segmentation}
\label{sec:unimodalQual}

\begin{figure*}
\centering
\footnotesize
\setlength{\tabcolsep}{0.1cm}
{\renewcommand{\arraystretch}{2}
\begin{tabular}{P{0.4cm}P{3.9cm}P{3.9cm}P{3.9cm}P{3.9cm}}
& Input Image & Baseline Output & AdapNet++ Output & Improvement\textbackslash Error Map \\
\rotatebox{90}{(a) Cityscapes} & \includegraphics[width=\linewidth]{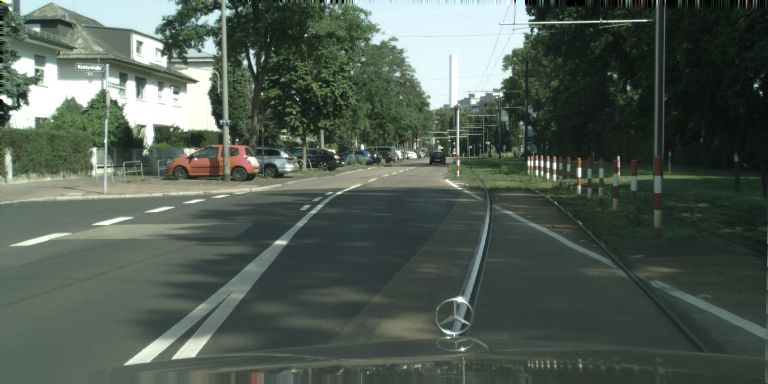} & \includegraphics[width=\linewidth]{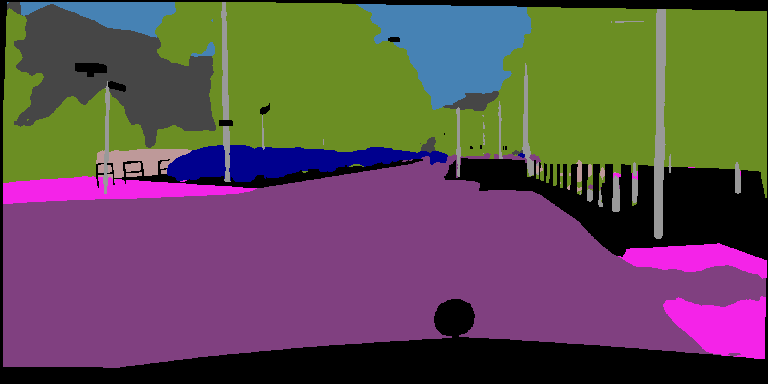} & \includegraphics[width=\linewidth]{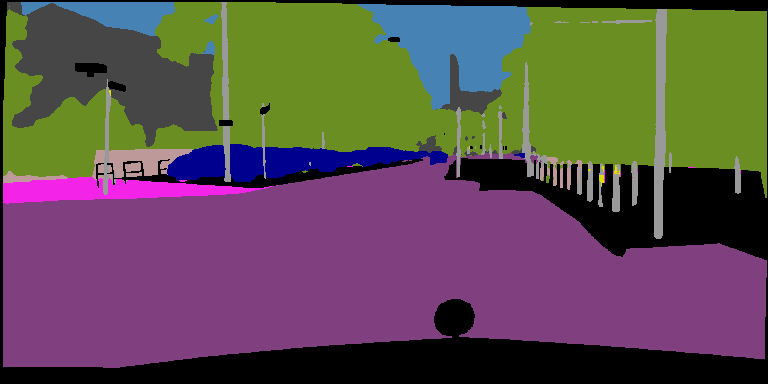} & \fbox{\includegraphics[width=\linewidth]{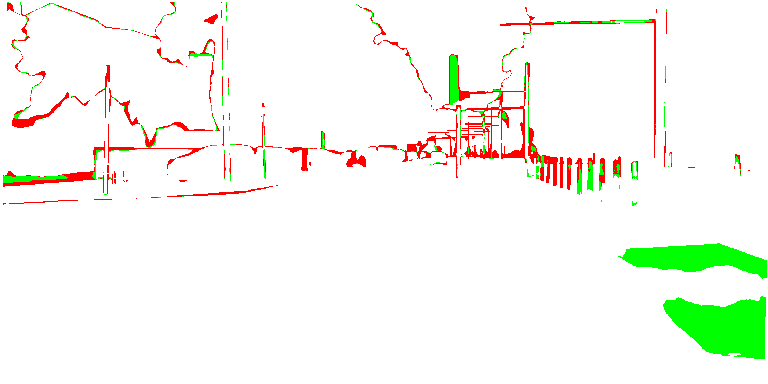}} \\
\rotatebox{90}{(b) Cityscapes} & \includegraphics[width=\linewidth]{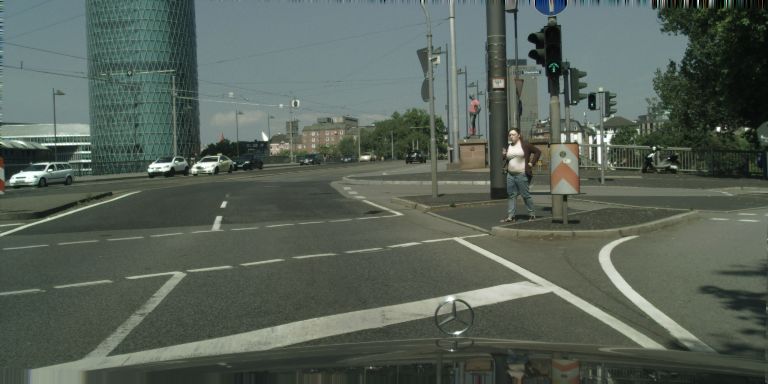} & \includegraphics[width=\linewidth]{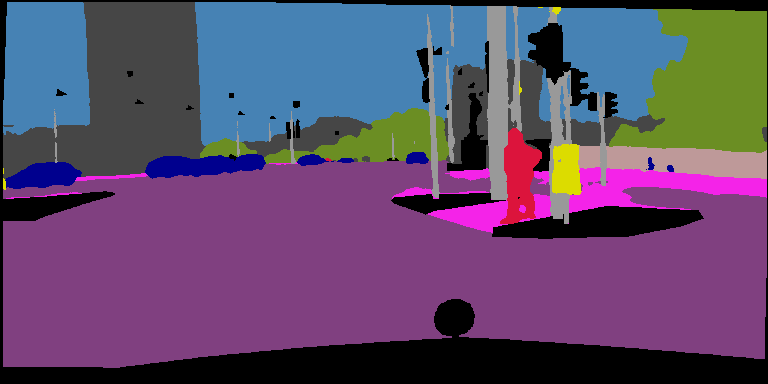} & \includegraphics[width=\linewidth]{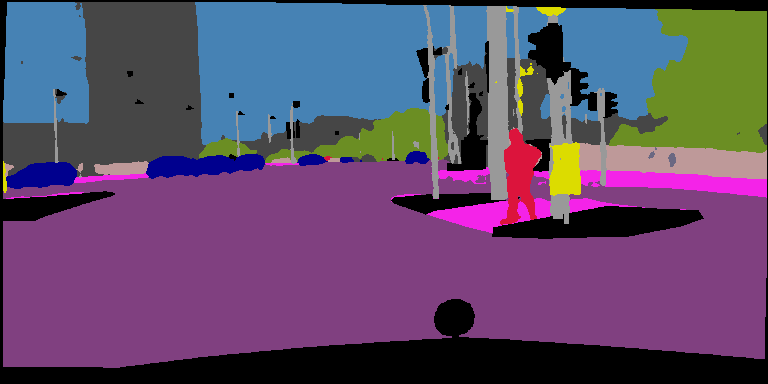} & \fbox{\includegraphics[width=\linewidth]{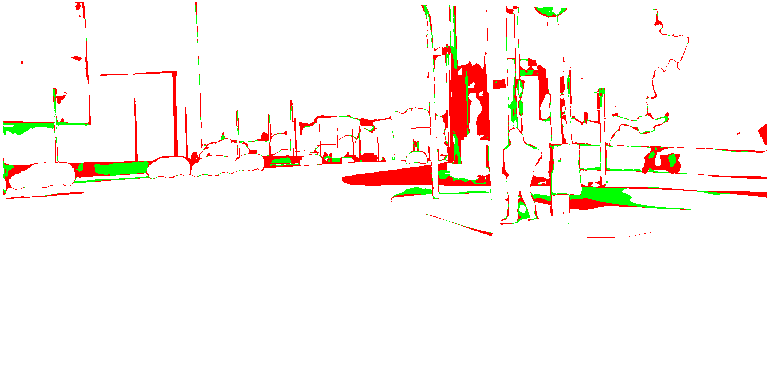}} \\
\rotatebox{90}{(c) Synthia} & \includegraphics[width=\linewidth]{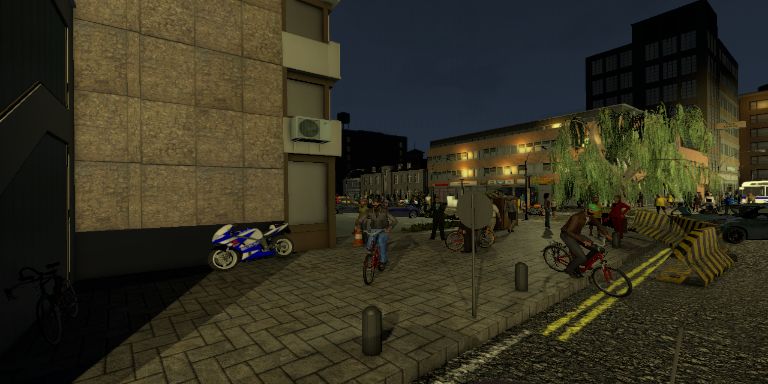} & \includegraphics[width=\linewidth]{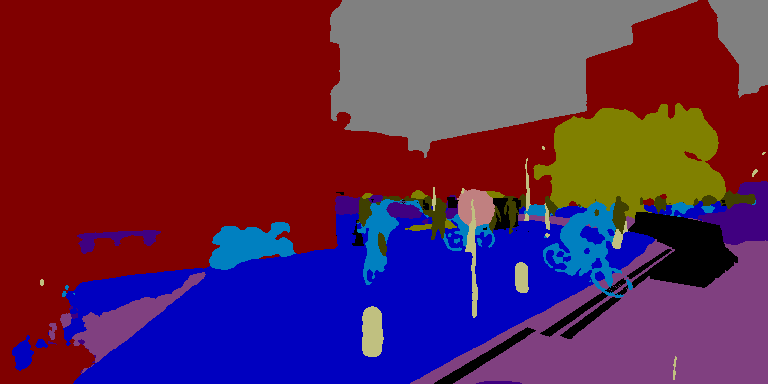} & \includegraphics[width=\linewidth]{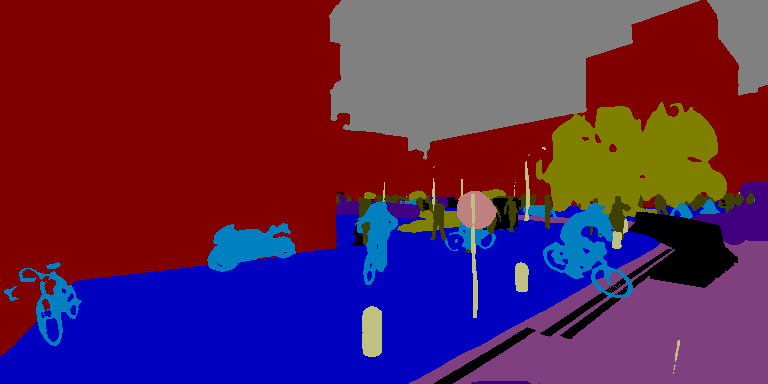} & \fbox{\includegraphics[width=\linewidth]{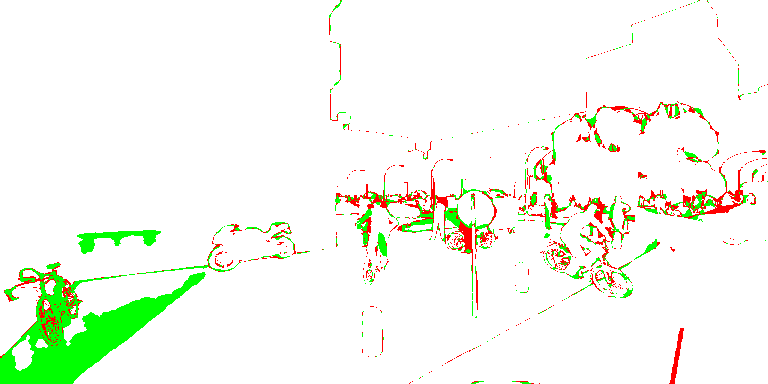}} \\
\rotatebox{90}{(d) Synthia} &
\includegraphics[width=\linewidth]{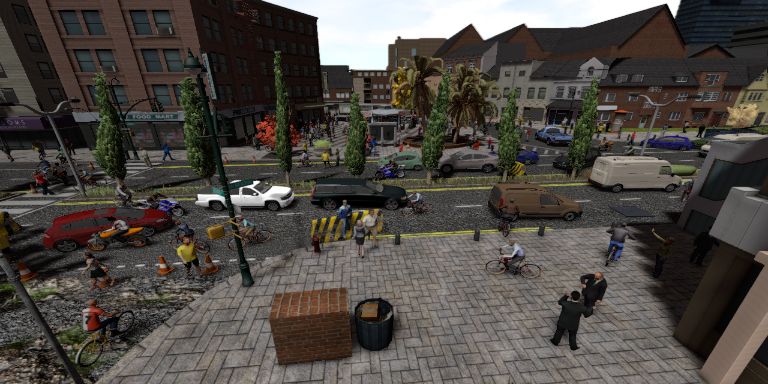} & \includegraphics[width=\linewidth]{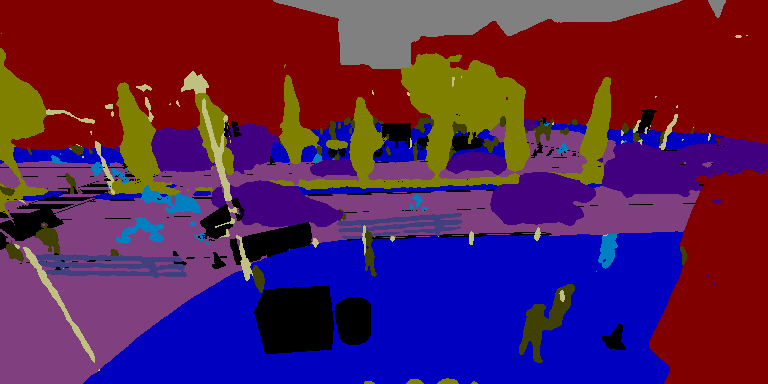} & \includegraphics[width=\linewidth]{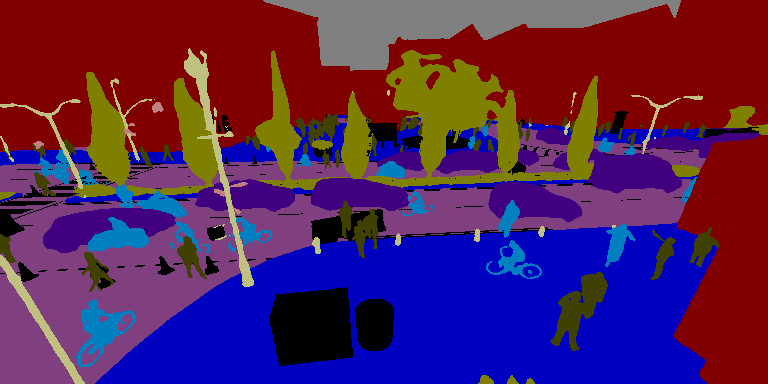} & \fbox{\includegraphics[width=\linewidth]{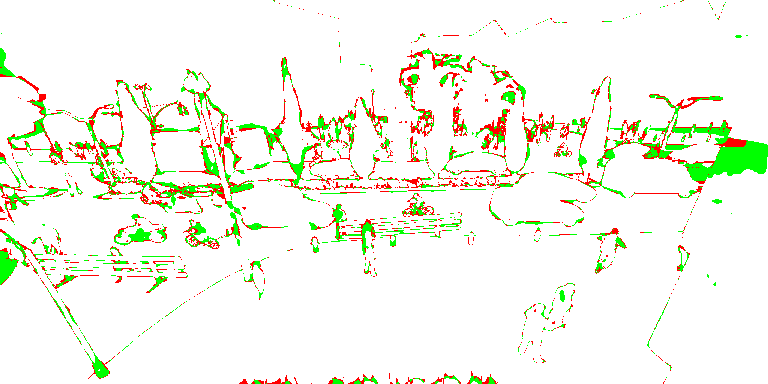}} \\
\rotatebox{90}{(e) SUN RGB-D} & \includegraphics[width=\linewidth]{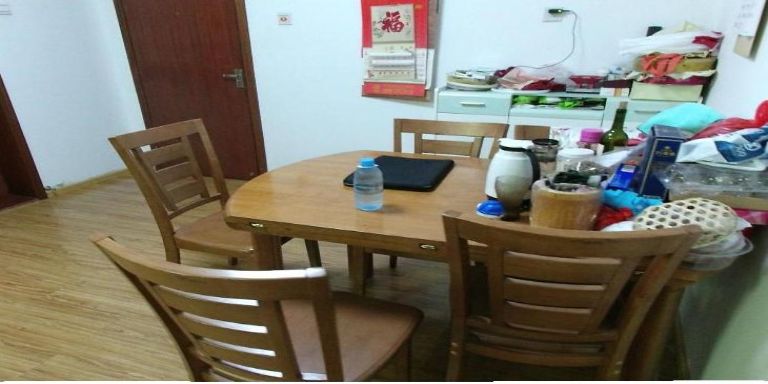} & \includegraphics[width=\linewidth]{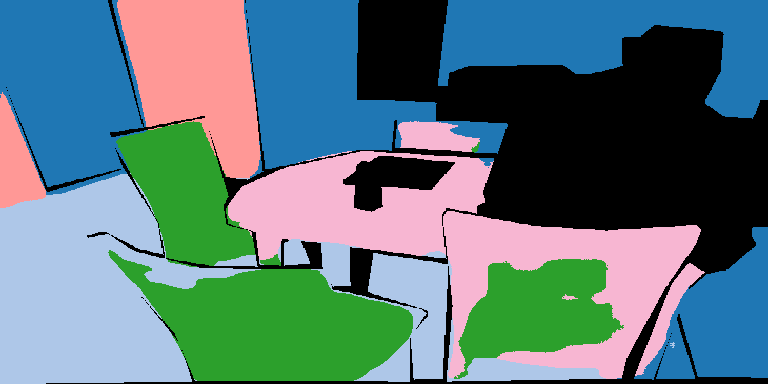} & \includegraphics[width=\linewidth]{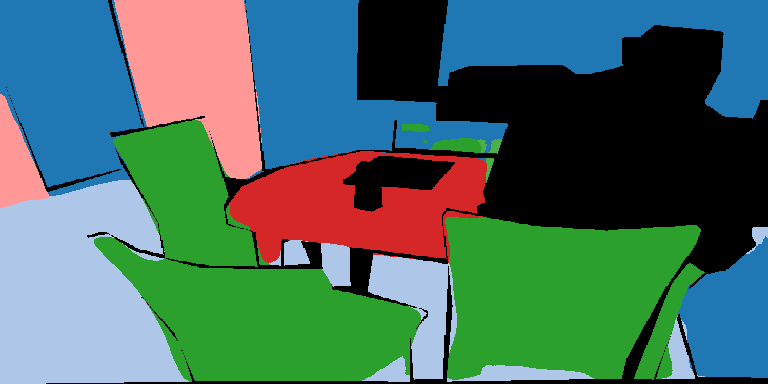} & \fbox{\includegraphics[width=\linewidth]{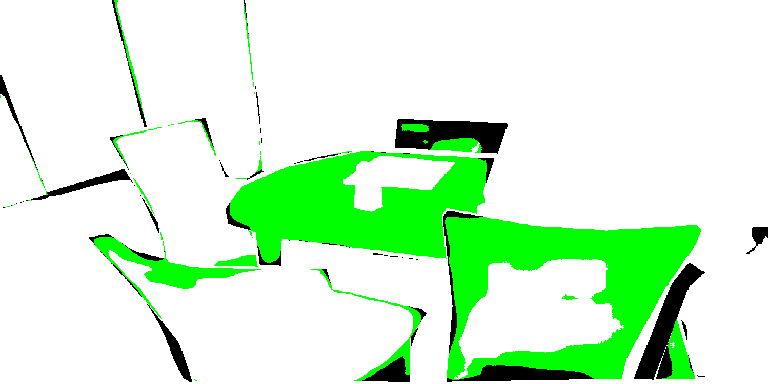}} \\
\rotatebox{90}{(f) SUN RGB-D} & \includegraphics[width=\linewidth]{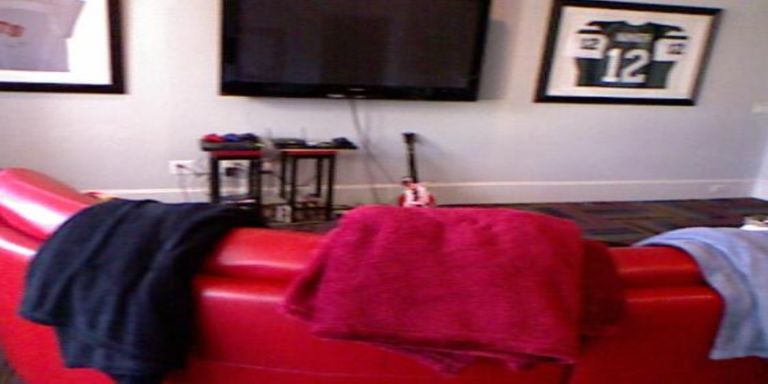} & \includegraphics[width=\linewidth]{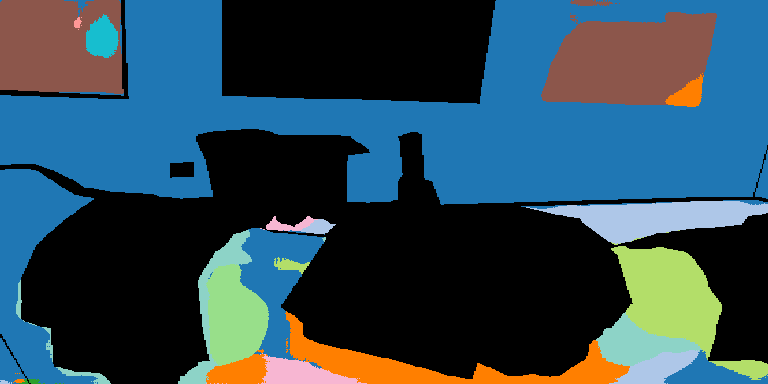} & \includegraphics[width=\linewidth]{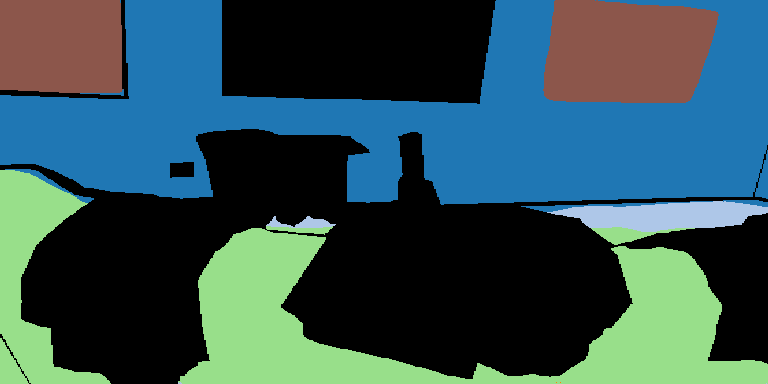} & \fbox{\includegraphics[width=\linewidth]{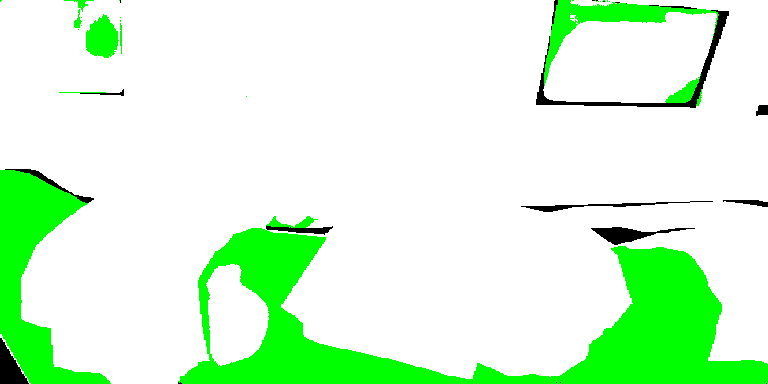}} \\
\rotatebox{90}{(g) SacanNet} & \includegraphics[width=\linewidth]{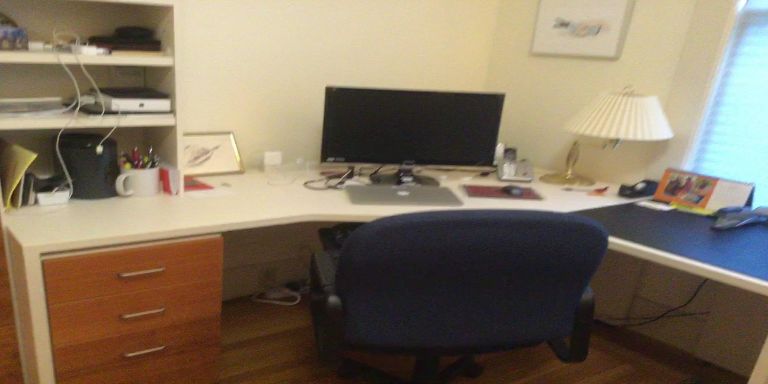} & \includegraphics[width=\linewidth]{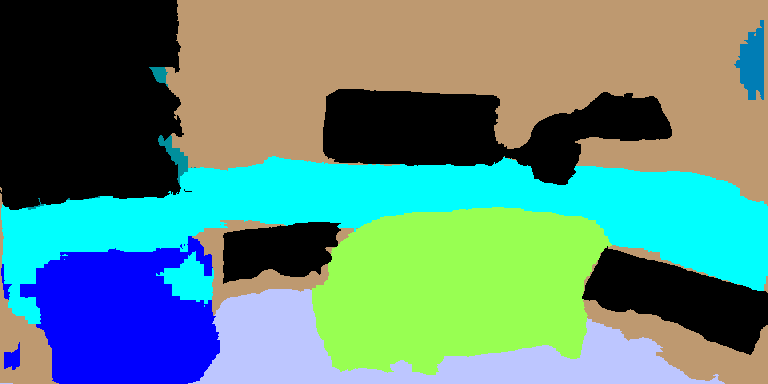} & \includegraphics[width=\linewidth]{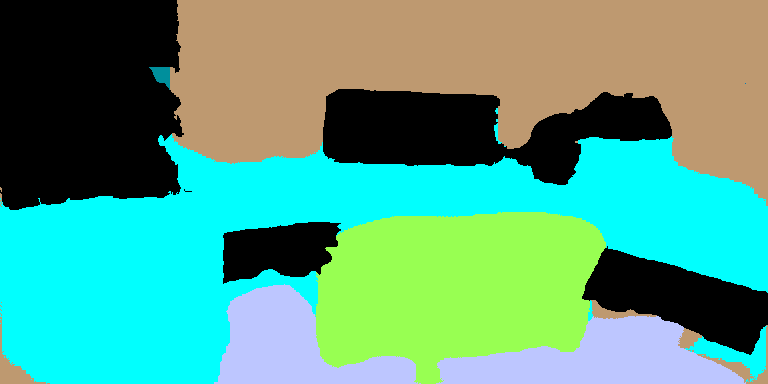} & \fbox{\includegraphics[width=\linewidth]{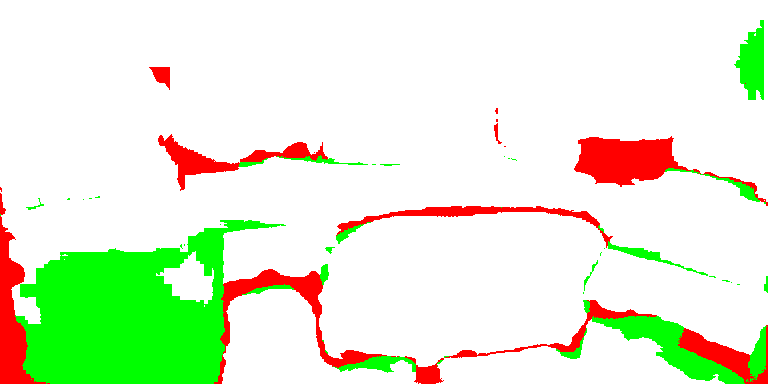}} \\
\rotatebox{90}{(h) SacanNet} & \includegraphics[width=\linewidth]{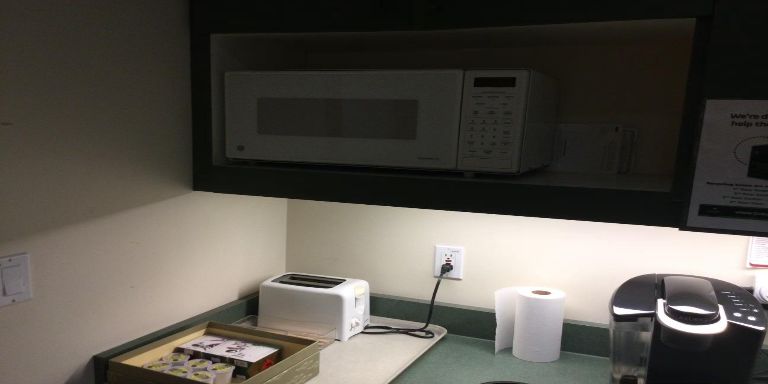} & \includegraphics[width=\linewidth]{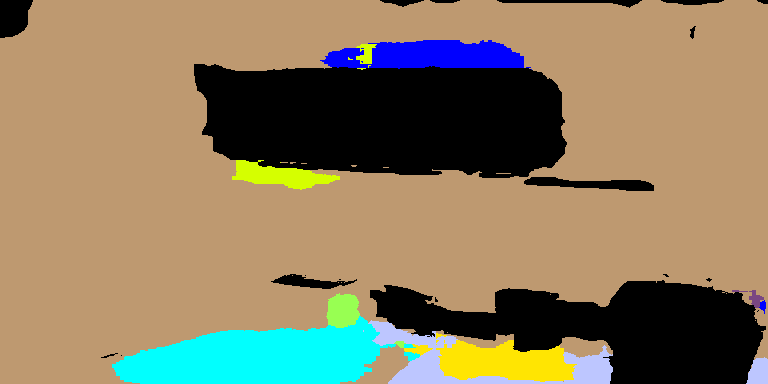} & \includegraphics[width=\linewidth]{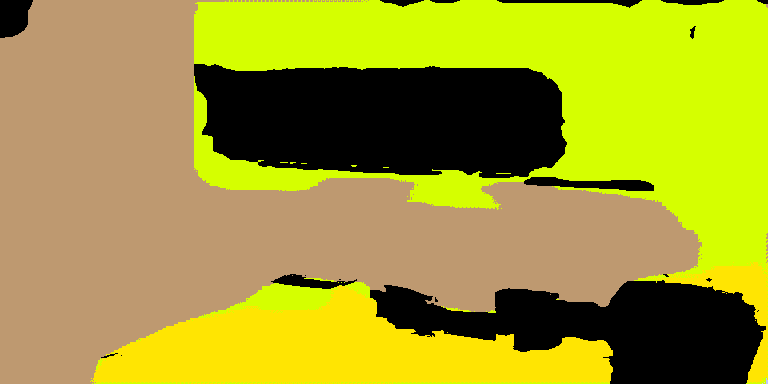} & \fbox{\includegraphics[width=\linewidth]{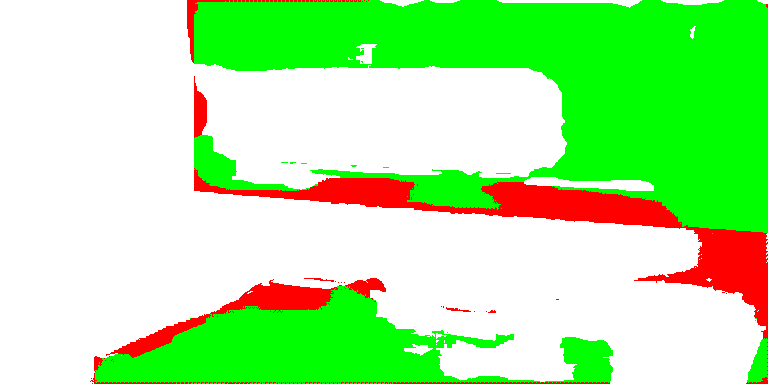}} \\
\rotatebox{90}{(i) Forest} & \includegraphics[width=\linewidth]{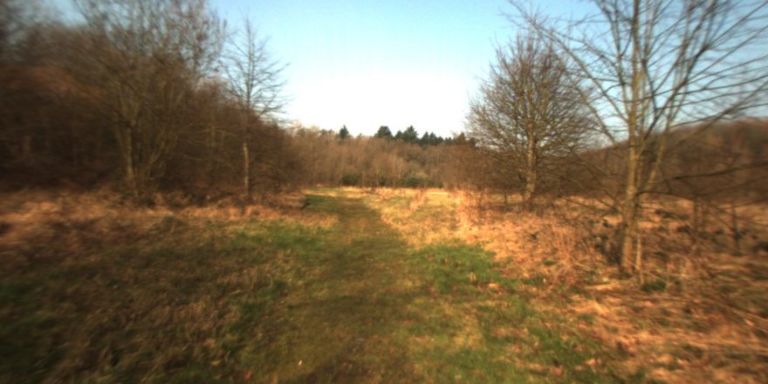} & \includegraphics[width=\linewidth]{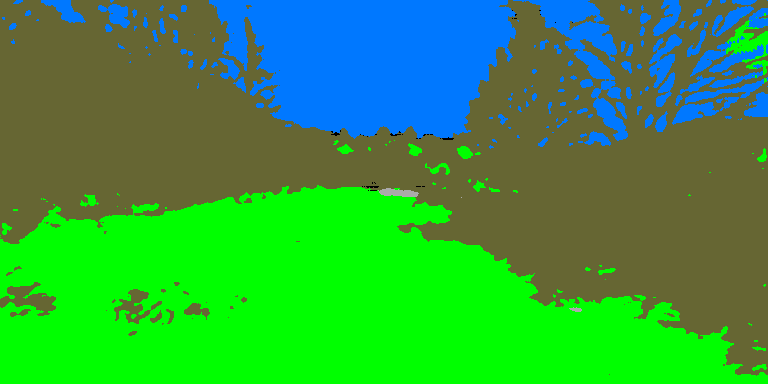} & \includegraphics[width=\linewidth]{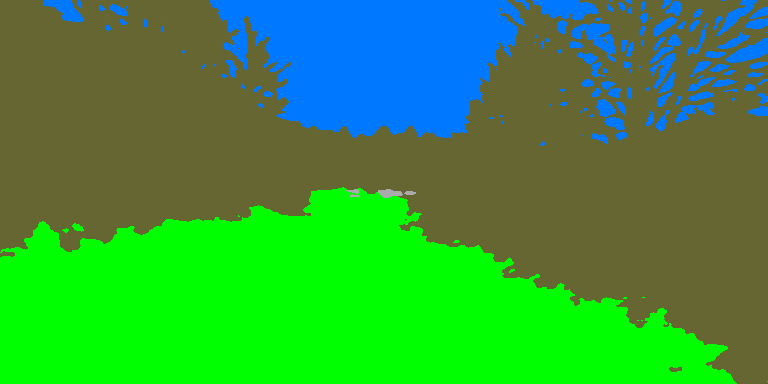} & \fbox{\includegraphics[width=\linewidth]{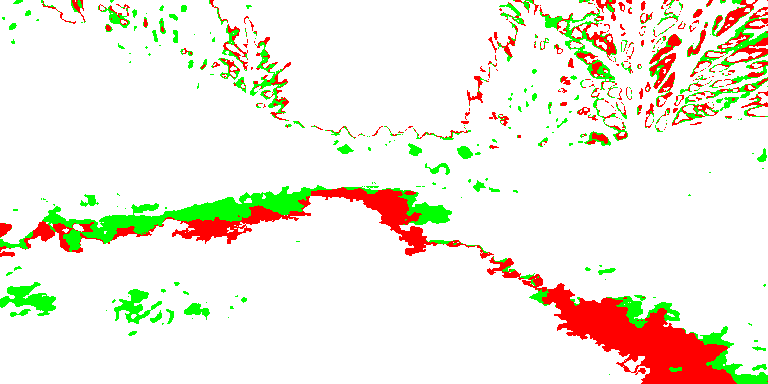}} \\
\rotatebox{90}{(j) Forest} & \includegraphics[width=\linewidth]{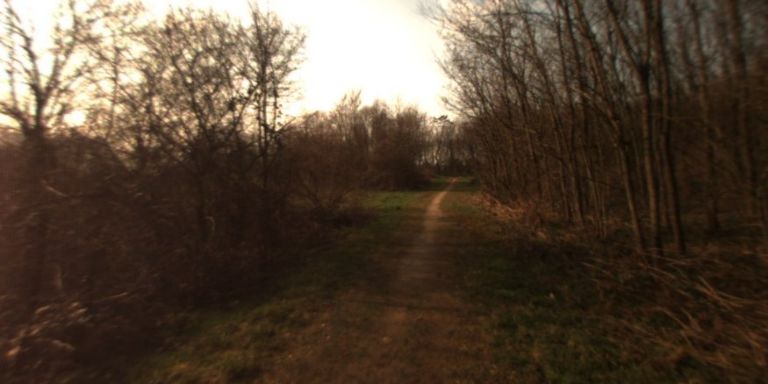} & \includegraphics[width=\linewidth]{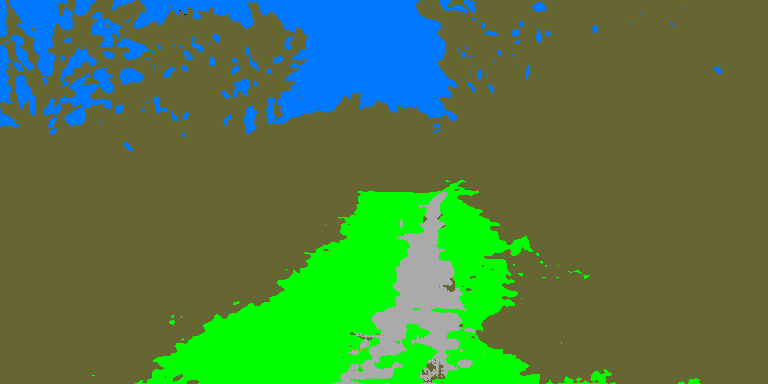} & \includegraphics[width=\linewidth]{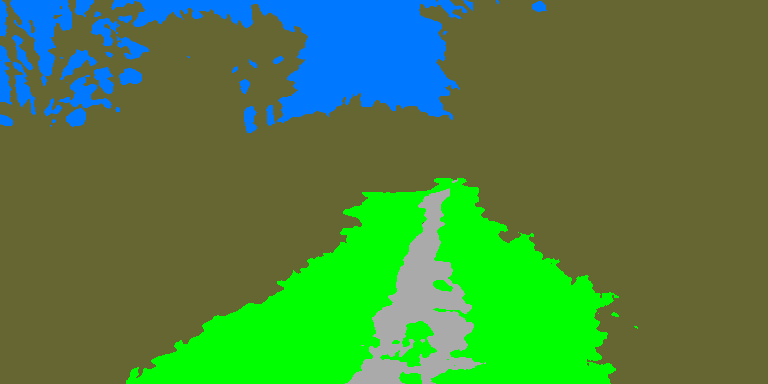} & \fbox{\includegraphics[width=\linewidth]{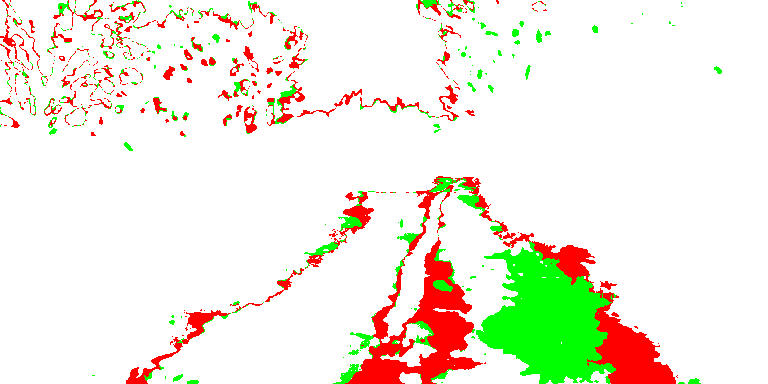}} \\
\end{tabular}}
\caption{Qualitative segmentation results of our unimodal AdapNet++ architecture in comparison to the best performing state-of-the-art model on different datasets. In addition to the segmentation output, we also show the improvement/error map which denotes the misclassified pixels in red and the pixels that are misclassified by the best performing state-of-the-art model but correctly predicted by AdapNet++ in green. The legend for the segmented labels correspond to the colors shown in the benchmarking tables in \secref{sec:benchm}.}
\label{fig:qualitativeUnimodal}
\end{figure*}

In this section, we qualitatively evaluate the semantic segmentation performance of our AdapNet++ architecture in comparison to the best performing state-of-the-art model for each dataset according to the quantitative results presented in \secref{sec:benchm}. We utilize this best performing model as a baseline for the qualitative comparisons presented in this section. \figref{fig:qualitativeUnimodal} shows two examples for each dataset that we benchmark on. The colors for the segmented labels shown correspond to the colors and the object categories mentioned in the benchmarking tables shown in \secref{sec:benchm}. \figref{fig:qualitativeUnimodal}(a) and (b) show examples from the Cityscapes dataset in which the improvement over the baseline output (AdapNet) can be seen in the better differentiation between inconspicuous classes such as \textit{sidewalk} and \textit{road} as well as \textit{pole and sign}. This can be primarily attributed to the eASPP which has a large receptive field and thus captures larger object context which helps to discern the differences between the inconspicuous classes. The improvement due to better boundary segmentation of thin object classes such as \textit{poles} can be seen in the images. 

\figref{fig:qualitativeUnimodal}(c) and (d) show examples from the Synthia dataset, where objects such as \textit{bicycles}, \textit{cars} and \textit{people} are better segmented. The baseline output (AdapNet) shows several missing \textit{cars}, \textit{people} and \textit{bicycles}, whereas the AdapNet++ output accurately captures these objects. Moreover, it can also be seen that the pole-like structures and trees are often discontinuous in the baseline output, while they are more well defined in the AdapNet++ output. In \figref{fig:qualitativeUnimodal}(d), an interesting observation is made where an entire fence is segmented in the baseline output but is absent in the scene. This is due to the fact that the intersection of the sidewalk and the road gives an appearance of a fence which is then misclassified. In the same image, it can also be observed that a small building-like structure on the right is not captured, whereas our AdapNet++ model accurately segments the structure.

\figref{fig:qualitativeUnimodal}(e) and (f) show examples from the indoor SUN RGB-D dataset. Examples from this dataset show significant misclassification due to inconspicuous objects. Often scenes in indoor datasets have large objects that require the network to have very large receptive fields to be able to accurately distinguish between them. \figref{fig:qualitativeUnimodal}(e) shows a scene in which parts of the \textit{chair} and the \textit{table} are incorrectly classified as a \textit{desk} in the output of the baseline model (DeepLab v3). These two classes have very similar structure and appearance which makes distinguishing between them extremely challenging. In \figref{fig:qualitativeUnimodal}(f), we can see parts of the \textit{sofa} incorrectly classified in the baseline model output, whereas the entire object is accurately predicted in the AdapNet++ output. In the baseline output, misclassification can also be seen for the \textit{picture} on the wall which is precisely segmented in the AdapNet++ output.

In \figref{fig:qualitativeUnimodal}(g) and (h), we show examples from the indoor ScanNet dataset. \figref{fig:qualitativeUnimodal}(g) shows misclassification in the output of the baseline model (DeepLab v3) in the boundary where the wall meets the floor and for parts of the \textit{desk} that is misclassified as other furniture. \figref{fig:qualitativeUnimodal}(h) shows a significant improvement in the segmentation of AdapNet++ in comparison to the baseline model. The \textit{cabinet} and \textit{counter} are entirely misclassified as a \textit{desk} and \textit{other furniture} correspondingly in the output of the baseline model, whereas they are accurately predicted by our AdapNet++ mode. 

Finally, \figref{fig:qualitativeUnimodal}(i) and (j) show examples from the unstructured Freiburg Forest dataset where the improvement can largely be seen in discerning the object boundaries of classes such as \textit{grass} and \textit{vegetation}, as well as \textit{trail} and \textit{grass}. By observing these images, we can see that even for us humans it is difficult to estimate the boundaries between these classes. Our AdapNet++ architecture predicts the boundaries comparatively better than the baseline model (DeepLab v3). The improvement in the segmentation can also been seen in the finer segmentation of the \textit{vegetation} and the \textit{trail} path in the AdapNet++ output.

\subsection{Multimodal Fusion Benchmarking}
\label{sec:multimodalFusionEx}

In this section, we present comprehensive results on the performance of our proposed multimodal SSMA fusion architecture in comparison to state-of-the-art multimodal fusion methods, namely, LFC~\citep{valada16iser}, FuseNet~\citep{fusenet2016accv} and CMoDE~\citep{valada17icra}. We employ the same AdapNet++ network backbone for all the fusion models including the competing methods. Therefore, we use the official implementation from the authors as a reference and append the fusion mechanism to our backbone. We also compare with baseline fusion approaches with the AdapNet++ topology as the backbone such as Late Fusion: a $1\times1$ convolution layer appended after individual modality-specific networks and the outputs are merged by adding the feature maps before the $\mathsf{softmax}$, Stacking: channel-wise concatenation of modalities before input to the network, Average: averaging prediction probabilities of individual modality-specific networks followed by $\mathsf{argmax}$, and Maximum: maximum of the prediction probabilities of individual modality-specific networks followed by $\mathsf{argmax}$. Additionally, we also compare against the performance of the unimodal AdapNet++ architecture for each of the modalities in the dataset for reference. We denote our proposed multimodal model as SSMA and the model with left-right flips as well as multiscale testing as SSMA\_msf in our experiments.

\begin{table}
\footnotesize
\centering
\caption{Comparison of multimodal fusion approaches on the Cityscapes validation set (input image dim: $768\times384$). All the fusion models have the same unimodal AdapNet++ network backbone. Results on the test set are shown in \tabref{tab:benchmarkingCityscapes}.}
\label{tab:multimodalCityscapes}
\begin{tabular}{p{1.4cm}p{2.3cm}p{0.8cm}p{0.8cm}p{0.8cm}}
\noalign{\smallskip}\hline\noalign{\smallskip}
Network & Approach & mIoU & Acc. & AP \\
 & & (\%) & (\%) & (\%) \\
\noalign{\smallskip}\hline\hline\noalign{\smallskip}
RGB & Unimodal & $80.80$ & $96.04$ & $90.97$ \\
Depth & Unimodal & $66.36$ & $91.21$ & $80.23$ \\
HHA & Unimodal & $67.66$ & $91.66$ & $81.81$ \\
\noalign{\smallskip}\hline\noalign{\smallskip}
\multirow{9}{*}{RGB-D} & Average & $78.84$ & $95.58$ & $90.49$ \\
& Maximum & $78.81$ & $95.58$ & $90.37$ \\
& Stacking & $80.21$ & $95.96$ & $90.05$ \\
& Late Fusion & $78.75$ & $95.57$ & $90.48$ \\
& LFC & $81.04$ & $96.11$ & $91.10$ \\
& CMoDE & $81.33$ & $96.12$ & $90.29$ \\
& FuseNet & $81.54$ & $96.23$ & $90.24$ \\
& SSMA (Ours) & $82.29$ & $96.36$ & $90.77$ \\
& SSMA\_msf (Ours) & $\mathbf{83.44}$ & $\mathbf{96.59}$ & $\mathbf{92.21}$ \\
\noalign{\smallskip}\hline\noalign{\smallskip}
\multirow{8}{*}{RGB-HHA} & Average & $79.44$ & $95.56$ & $90.27$ \\
& Maximum & $79.40$ & $95.55$ & $90.09$ \\
& Stacking & $80.62$ & $96.01$ & $90.09$ \\
& Late Fusion &  $79.01$ & $95.49$ & $90.25$ \\
& LFC & $81.13$ & $96.14$ & $91.32$ \\
& CMoDE & $81.40$ & $96.12$ & $90.29$ \\
& SSMA (Ours) & $82.64$ & $96.41$ & $90.65$ \\
& SSMA\_msf (Ours) & $\mathbf{83.94}$ & $\mathbf{96.68}$ & $\mathbf{91.99}$ \\
\noalign{\smallskip}\hline\noalign{\smallskip}
\end{tabular}
\end{table}

In \tabref{tab:multimodalCityscapes}, we show the results on the Cityscapes validation set considering visual images (RGB), depth and the HHA encoding of the depth as modalities for the fusion. As hypothesised, the visual RGB images perform the best among the other modalities achieving a mIoU of $80.80\%$. This is especially observed in outdoor scene understanding datasets containing stereo depth images that quickly degrade the information contained, with increasing distance from the camera. Among the baseline fusion approaches, Stacking achieves the highest performance for both RGB-D and RGB-HHA fusion, however, their performance is still lower than the unimodal visual RGB segmentation. This can be attributed to the fact that the baseline approaches are not able to exploit the complementary features from the modalities due to the naive fusion. CMoDE fusion with RGB-HHA achieves the highest performance among state-of-the-art approaches, surpassing the performance of unimodal segmentation. While, our proposed SSMA model for RGB-HHA fusion achieves a mIoU of $83.29\%$ outperforming all the other approaches and setting the new state-of-the-art. The SSMA\_msf model further improves upon the performance of the SMMA model by $1.3\%$. As the Cityscapes dataset does not contain harsh environments, the improvement that can be achieved using fusion is limited to scenes that contain inconspicuous object classes or mismatched relationship. However, the additional robustness that it demonstrates due to multimodal fusion is still notable as shown in the qualitative results in the following sections. Additionally, the benchmarking results on the Cityscapes test set is shown in \tabref{tab:benchmarkingCityscapes}. The results demonstrate that our SSMA fusion architecture with the AdapNet++ network backbone achieves a comparable performance as the top performing DPC and DRN architectures, while outperforming the other networks on the leaderboard.

\begin{table}
\footnotesize
\centering
\caption{Comparison of multimodal fusion approaches on the Synthia validation set (input image dim: $768\times384$). All the fusion models have the same unimodal AdapNet++ network backbone.}
\label{tab:multimodalSynthia}
\begin{tabular}{p{1.4cm}p{2.3cm}p{0.8cm}p{0.8cm}p{0.8cm}}
\noalign{\smallskip}\hline\noalign{\smallskip}
Network & Approach & mIoU & Acc. & AP \\
 & & (\%) & (\%) & (\%) \\
\noalign{\smallskip}\hline\hline\noalign{\smallskip}
RGB & Unimodal & $86.70$ & $97.18$ & $93.17$ \\
Depth & Unimodal & $87.87$ & $97.78$ & $94.23$ \\
\noalign{\smallskip}\hline\noalign{\smallskip}
\multirow{9}{*}{RGB-D} & Average & $89.22$ & $98.03$ & $95.04$ \\
& Maximum & $89.13$ & $98.01$ & $94.97$ \\
& Stacking & $88.95$ & $98.03$ & $94.41$ \\
& Late Fusion & $89.13$ & $98.01$ & $94.66$ \\
& LFC & $89.48$ & $98.09$ & $94.96$ \\
& CMoDE & $89.57$ & $98.13$ & $94.58$ \\
& FuseNet & $89.62$ & $98.34$ & $94.67$ \\
& SSMA (Ours) & $91.25$ & $98.48$ & $95.68$ \\
& SSMA\_msf (Ours) & $\mathbf{92.10}$ & $\mathbf{98.64}$ & $\mathbf{96.37}$ \\
\noalign{\smallskip}\hline\noalign{\smallskip}
\end{tabular}
\end{table}

We benchmark on the Synthia dataset to demonstrate the utility of fusion when both modalities contain rich information. It consists of scenes with adverse perceptual conditions including rain, snow, fog and night, therefore the benefit of multimodal fusion for outdoor environments is most evident on this dataset. As the Synthia dataset does not provide camera calibration parameters, we cannot compute the HHA encoding, therefore we benchmark using visual RGB and depth images. Results from benchmarking on this dataset are shown in \tabref{tab:multimodalSynthia}. Due to the high-resolution depth information, the unimodal depth model achieves a mIoU of $87.87\%$, outperforming segmentation using visual RGB images by $1.17\%$. This demonstrates that accurate segmentation can be obtained using only depth images as input provided that the depth sensor gives accurate long range information. Among the baseline fusion approaches and the state-of-the-art techniques, the CMoDE RGB-D fusion approach achieves the highest mIoU, outperforming the unimodal depth model by $1.7\%$. While our proposed SSMA architecture demonstrates state-of-the-art performance of $91.25\%$ and further improves the mIoU to $92.10\%$ using the SSMA\_msf model. This accounts to a large improvement of $5.4\%$ over the best performing unimodal segmentation model. Other metrics such as the pixel accuracy and average precision also show similar improvement.

\begin{figure}
\centering
\includegraphics[width=\linewidth]{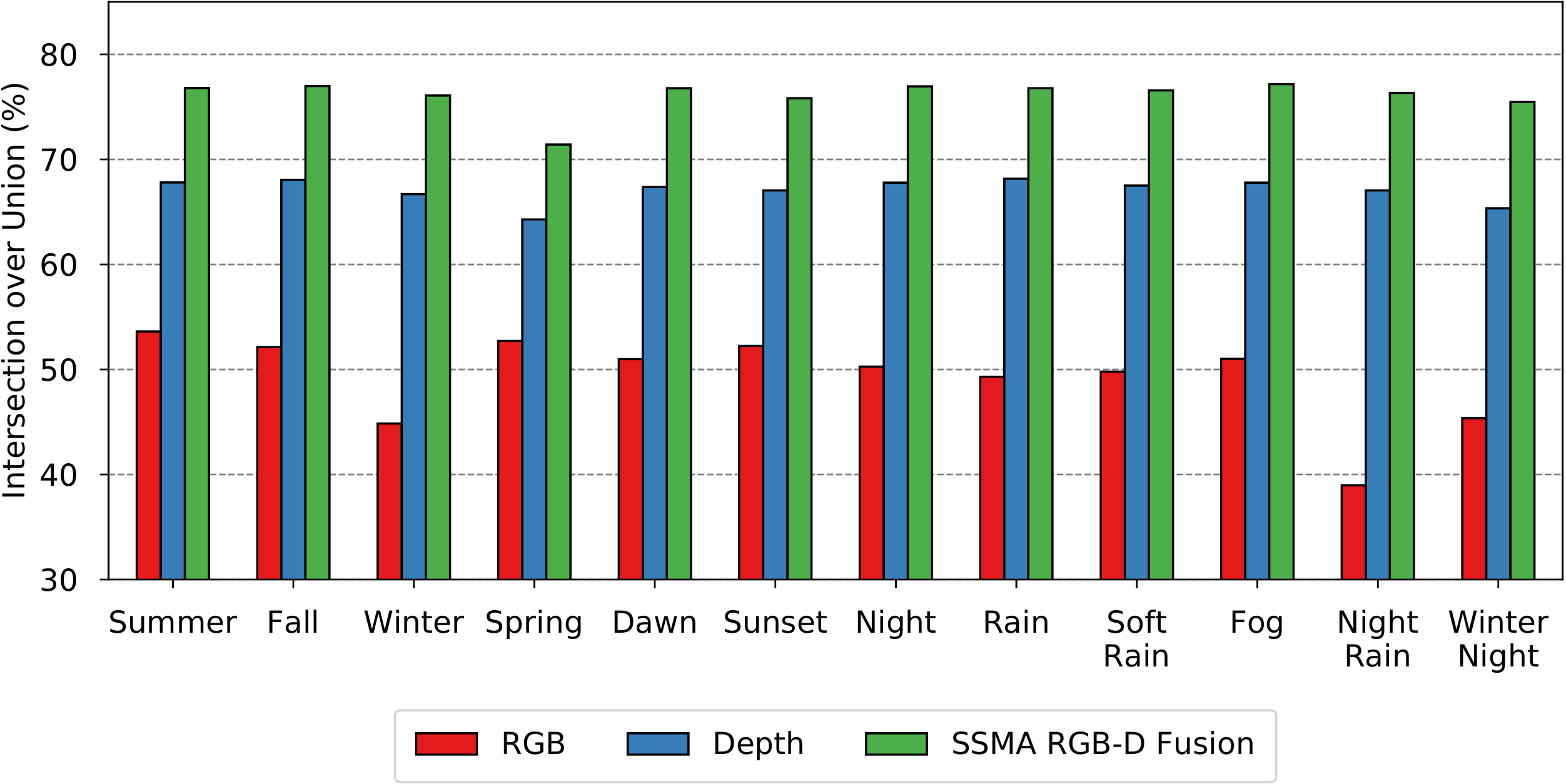}
\caption{Evaluation of our proposed SMMA fusion technique on the Synthia-Sequences dataset containing a variety of seasons and weather conditions. We use the models trained on the Synthia-Rand-Cityscapes dataset and only test on the individual conditions in the Synthia-Sequences dataset to quantify its robustness. Our model that performs RGB-D fusion consistently outperforms the unimodal models which can be more prominently seen qualitatively in \figref{fig:qualitativeFusion}(c,d).}
\label{fig:synthiaSequences}
\end{figure}

One of our main motivations to benchmark on this dataset is to evaluate our SSMA fusion model on a diverse set of scenes with adverse perpetual conditions. For this experiment, we trained our SSMA fusion model on the Synthia-Rand-Cityscapes training set and evaluated the performance on each of the conditions contained in the Synthia-Sequences dataset. The Synthia-Sequences dataset contains individual video sequences in different conditions such as \textit{summer, fall, winter, spring, dawn, sunset, night, rain, soft rain, fog, night rain} and \textit{winter night}. Results from this experiment are shown in \figref{fig:synthiaSequences}. The unimodal visual RGB model achieves an overall mIoU score of $49.27\% \pm 4.04\%$ across the 12 sequences. Whereas, the model trained on the depth maps achieves a mIoU score of $67.07\%\pm 1.12\%$, thereby substantially outperforming the model trained using visual RGB images. 

As this is a synthetic dataset, the depth maps provided are accurate and dense even for structures that are several hundreds of meters away from the camera. Therefore, this enables the unimodal depth model to learn representations that accurately encode the structure of the scene and these structural representations are proven to be invariant to the change in perceptual conditions. It can also be observed that the unimodal depth model performs consistently well in all the conditions with a variance of $1.24\%$, demonstrating the generalization to different conditions. However, the visual RGB model with a variance of $16.30\%$ performs inconsistently across different conditions. Nevertheless, we observe that our RGB-D SSMA fusion model outperforms the unimodal visual RGB model by achieving a mIoU score of $76.51\% \pm 0.49\%$ across the 12 conditions, accounting to an improvement of $27.24\%$. Moreover, the SSMA fusion model has a variance of $0.24\%$, demonstrating better generalization ability across varying adverse perceptual conditions.

\begin{table}
\footnotesize
\centering
\caption{Comparison of multimodal fusion approaches on the SUN RGB-D validation set (input image dim: $768\times384$). All the fusion models have the same unimodal AdapNet++ network backbone.}
\label{tab:multimodalSun}
\begin{tabular}{p{1.4cm}p{2.3cm}p{0.8cm}p{0.8cm}p{0.8cm}}
\noalign{\smallskip}\hline\noalign{\smallskip}
Network & Approach & mIoU & Acc. & AP \\
 & & (\%) & (\%) & (\%) \\
\noalign{\smallskip}\hline\hline\noalign{\smallskip}
RGB & Unimodal & $38.40$ & $76.90$ & $62.78$ \\
Depth & Unimodal & $34.27$ & $73.83$ & $74.39$ \\
HHA & Unimodal & $34.59$ & $74.39$ & $57.18$  \\
\noalign{\smallskip}\hline\noalign{\smallskip}
\multirow{9}{*}{RGB-D} & Average & $40.70$ & $78.58$ & $64.54$ \\
& Maximum & $40.58$ & $78.50$ & $64.04$ \\
& Stacking & $36.48$ & $76.68$ & $57.92$ \\
& Late Fusion & $41.68$ & $79.27$ & $66.63$ \\
& LFC & $41.82$ & $79.36$ & $66.75$ \\
& CMoDE & $41.87$ & $79.84$ & $66.81$ \\
& FuseNet & $41.85$ & $79.56$ & $66.87$ \\
& SSMA (Ours) & $43.90$ & $80.16$ & $66.11$ \\
& SSMA\_msf (Ours) & $\mathbf{44.52}$ & $\mathbf{80.67}$ & $\mathbf{67.92}$ \\
\noalign{\smallskip}\hline\noalign{\smallskip}
\multirow{8}{*}{RGB-HHA} & Average & $41.01$ & $78.54$ & $64.93$ \\
& Maximum & $40.91$ & $78.49$ & $64.78$ \\
& Stacking & $37.49$ & $76.42$ & $57.88$ \\
& Late Fusion & $41.91$ & $79.49$ & $67.31$ \\
& LFC & $42.42$ & $79.55$ & $67.41$ \\
& CMoDE & $42.55$ & $79.94$ & $65.38$ \\
& SSMA (Ours) & $44.43$ & $80.21$ & $64.94$ \\
& SSMA\_msf (Ours) & $\mathbf{45.73}$ & $\mathbf{80.97}$ & $\mathbf{67.82}$ \\
\noalign{\smallskip}\hline\noalign{\smallskip}
\end{tabular}
\end{table}

We also benchmark on the indoor SUN RGB-D dataset which poses a different set of challenges than the outdoor datasets. The improvement from multimodal fusion is more evident here as indoor scenes are often smaller confined spaces with several cluttered objects and the depth modality provides valuable structural information that can be exploited. Results from RGB-D and RGB-HHA multimodal fusion is shown in \tabref{tab:multimodalSun}. Among the individual modalities, segmentation using visual RGB images yields the highest mIoU of $38.40\%$. The model trained on depth images performs $4.13\%$ lower than the visual RGB model. This can be attributed to the fact that the depth images are extremely noisy with numerous missing depth values in the SUN RGB-D dataset. However, our proposed SSMA fusion on RGB-HHA achieves state-of-the-art performance with a mIoU of $44.43\%$, constituting a substantial improvement of $6.03\%$ over the unimodal visual RGB model. Moreover, our SSMA\_msf model further improves upon the mIoU by $1.3\%$. Similar to the performance in other datasets, RGB-HHA achieves a higher mIoU than RGB-D fusion corroborating the fact that CNNs learn more effectively from the HHA encoding but with a small additional preprocessing time. 

\begin{table}
\footnotesize
\centering
\caption{Comparison of multimodal fusion approaches on the ScanNet validation set (input image dim: $768\times384$). All the fusion models have the same unimodal AdapNet++ network backbone. Results on the test set are shown in \tabref{tab:benchmarkingScannet}.}
\label{tab:multimodalScannet}
\begin{tabular}{p{1.4cm}p{2.3cm}p{0.8cm}p{0.8cm}p{0.8cm}}
\noalign{\smallskip}\hline\noalign{\smallskip}
Network & Approach & mIoU & Acc. & AP \\
 & & (\%) & (\%) & (\%) \\
\noalign{\smallskip}\hline\hline\noalign{\smallskip}
RGB & Unimodal & $52.92$ & $77.70$ & $77.28$ \\
Depth & Unimodal & $53.80$ & $80.63$ & $74.46$ \\
HHA & Unimodal & $54.19$ & $80.20$ & $73.90$ \\
\noalign{\smallskip}\hline\noalign{\smallskip}
\multirow{9}{*}{RGB-D} & Average & $58.20$ & $82.31$ & $79.45$ \\
& Maximum & $57.68$ & $82.12$ & $78.35$ \\
& Stacking & $55.89$ & $79.04$ & $77.08$ \\
& Late Fusion & $61.37$ & $82.48$ & $80.15$ \\
& LFC & $62.97$ & $83.70$ & $80.93$ \\
& CMoDE & $64.00$ & $84.94$ & $81.27$ \\
& FuseNet & $63.83$ & $84.24$ & $81.02$ \\
& SSMA (Ours) & $66.29$ & $86.11$ & $81.81$ \\
& SSMA\_msf (Ours) & $\mathbf{67.38}$ & $\mathbf{86.24}$ & $\mathbf{81.93}$ \\
\noalign{\smallskip}\hline\noalign{\smallskip}
\multirow{8}{*}{RGB-HHA} & Average & $58.39$ & $82.36$ & $78.39$ \\
& Maximum & $57.88$ & $82.12$ & $77.34$ \\
& Stacking & $56.48$ & $80.33$ & $77.47$ \\
& Late Fusion & $61.45$ & $82.68$ & $80.12$ \\
& LFC & $63.09$ & $83.67$ & $80.67$ \\
& CMoDE & $64.07$ & $84.84$ & $80.80$ \\
& SSMA (Ours) & $66.34$ & $86.02$ & $81.49$ \\
& SSMA\_msf (Ours) & $\mathbf{67.52}$ & $\mathbf{86.38}$ & $\mathbf{82.21}$ \\
\noalign{\smallskip}\hline\noalign{\smallskip}
\end{tabular}
\end{table}

\tabref{tab:multimodalScannet} shows the results on the ScanNet validation set. ScanNet is the largest indoor RGB-D dataset to date with over 1513 different scenes and $2.5\million$ views. Unlike the SUN RGB-D dataset, ScanNet contains depth maps of better quality with lesser number of missing depth values. The unimodal visual RGB model achieves an mIoU of $52.92\%$ with a pixel accuracy of $77.70\%$, while the unimodal HHA model achieves an mIoU of $54.19\%$ with an accuracy of $80.20\%$. For multimodal fusion, CMoDE using RGB-HAA demonstrates the highest performance among state-of-the-art architectures achieving a mIoU of $64.07\%$. While our proposed SSMA RGB-HHA model outperforms CMoDE by yielding a mIoU of $66.34\%$, which is a significant improvement of $13.42\%$ over the unimodal visual RGB model. Moreover, the SSMA\_msf model further improves the mIoU score to $67.52\%$. To the best of our knowledge, this is the largest improvement due to multimodal fusion obtained thus far. An interesting observation that can be made from the results on SUN RGB-D and ScanNet is that the lowest multimodal fusion performance is obtained using the Stacking fusion approach, reaffirming our hypothesis that fusing more semantically mature features enables the model to exploit complementary properties from the modalities more effectively. We also benchmark on the ScanNet test set and report the results in \tabref{tab:benchmarkingScannet}. Our proposed SSMA fusion architecture with the AdapNet++ network backbone sets the new state-of-the-art on the ScanNet benchmark.

\begin{table}
\footnotesize
\centering
\caption{Comparison of multimodal fusion approaches on the Freiburg Forest validation set (input image dim: $768\times384$). All the fusion models have the same unimodal AdapNet++ network backbone.}
\label{tab:multimodalForest}
\begin{tabular}{p{1.4cm}p{2.3cm}p{0.8cm}p{0.8cm}p{0.8cm}}
\noalign{\smallskip}\hline\noalign{\smallskip}
Network & Approach & mIoU & Acc. & AP \\
 & & (\%) & (\%) & (\%) \\
\noalign{\smallskip}\hline\hline\noalign{\smallskip}
RGB & Unimodal & $83.09$ & $95.15$ & $89.83$ \\
Depth & Unimodal & $73.93$ & $91.42$ & $85.36$ \\
EVI & Unimodal & $80.96$ & $94.20$ & $88.88$ \\
\noalign{\smallskip}\hline\noalign{\smallskip}
\multirow{9}{*}{RGB-D} & Average & $79.51$ & $92.87$ & $90.99$ \\
& Maximum & $81.62$ & $93.93$ & $90.87$ \\
& Stacking & $83.13$ & $95.19$ & $89.95$ \\
& Late Fusion & $82.11$ & $93.95$ & $90.85$ \\
& LFC & $82.53$ & $94.99$ & $90.96$ \\
& CMoDE & $83.21$ & $95.19$ & $90.19$ \\
& FuseNet & $83.10$ & $95.07$ & $90.03$ \\
& SSMA (Ours) & $83.81$ & $95.62$ & $92.78$ \\
& SSMA\_msf (Ours) & $\mathbf{83.99}$ & $\mathbf{95.70}$ & $\mathbf{93.08}$ \\
\noalign{\smallskip}\hline\noalign{\smallskip}
\multirow{8}{*}{RGB-EVI} & Average & $83.00$ & $95.10$ & $90.19$ \\
& Maximum & $83.00$ & $95.10$ & $90.17$ \\
& Stacking & $83.18$ & $95.21$ & $90.11$\\
& Late Fusion & $82.80$ & $95.01$ & $90.07$ \\
& LFC & $83.00$ & $95.13$ & $90.28$ \\
& CMoDE & $83.31$ & $95.22$ & $90.19$ \\
& SSMA (Ours) & $83.90$ & $95.56$ & $92.28$ \\
& SSMA\_msf (Ours) & $\mathbf{84.18}$ & $\mathbf{95.64}$ & $\mathbf{92.60}$ \\
\noalign{\smallskip}\hline\noalign{\smallskip}
\end{tabular}
\end{table}

Finally, we present benchmarking results on the Freiburg Forest dataset that contains three inherently different modalities including visual RGB images, Depth data and EVI. EVI or Enhanced Vegetation Index was designed to enhance the vegetation signal in high biomass regions and it is computed from the information contained in three bands, namely, Near-InfraRed, Red and Blue channels~\citep{running1999modis}. As this dataset contains scenes in unstructured forested environments, EVI provides valuable information to discern between inconspicuous classes such as \textit{vegetation} and \textit{grass}. \tabref{tab:multimodalForest} shows the results on this dataset for multimodal fusion of RGB-D and RGB-EVI. For unimodal segmentation, the RGB model yields the highest performance, closely followed by the model trained on EVI. While for multimodal segmentation, our SSMA model trained on RGB-EVI yields the highest mIoU of $83.90\%$ and our SMMA\_msf model further improves upon the performance and achieves a mIoU of $84.18\%$. Both these models outperform existing multimodal fusion methods and set the new state-of-the-art.

\subsection{Multimodal Fusion Discussion}

To summarize, the models trained on visual RGB images perform the best in general in comparison to unimodal segmentation with other modalities. However, when the depth data is less noisy and the environment is a confined indoor space, the model trained on depth or HHA encoded depth outperforms visual RGB models. Among the multimodal fusion baselines, late fusion and Stacking, each perform well in different environments. Stacking performs better in outdoor environments, while late fusion performs better indoors. This can be attributed to the fact that the late fusion method fuses semantically mature representations. Therefore, in indoor environments, modalities such as depth maps from stereo cameras are less noisy than in outdoors and as the environment is confined, all the objects in the scene are well represented with dense depth values. This enables the late fusion architecture to leverage semantically rich representations for fusion. However in outdoor environments, depth values are very noisy and no information is present for objects at far away distances. Therefore, the semantic representations from the depth stream are considerably less informative for certain parts of the scene which does not allow the Late Fusion network to fully exploit complementary features and hence it does not provide significant gains. On the other hand, in indoor or synthetic scenes where the depth modality is dense and rich with information, late fusion generally outperforms the stacking approach.

Among the current state-of-the-art methods, CMoDE outperforms all other approaches in most of the diverse environments. To recapitulate, CMoDE employs a class-wise probabilistic late fusion technique that adaptively weighs the modalities based on the scene condition. However, our proposed SSMA fusion techniques outperforms CMoDE in all the datasets and sets the new state-of-the-art in multimodal semantic segmentation. This demonstrates that fusion of modalities is an inherently complex problem that depends on several factors such as the object class of interest, the spatial location of the object and the environmental scene context. Our proposed SSMA fusion approach dynamically adapts the fusion of semantically mature representations based on the aforementioned factors, thereby enabling our model to effectively exploit complementary properties from the modalities. Moreover, as the dynamicity is learned in a self-supervised fashion, it efficiently generalizes to different diverse environments, perceptual conditions and the types of modalities employed for fusion. Another advantage of this dynamic adaptation property of our multimodal SSMA fusion mechanism is the intrinsic tolerance to sensor failure. In case one of the modalities becomes unavailable, the SSMA module can be trained to switch the output of the unavailable modality-specific encoder off by generating gating probabilities as zeros. This enables the multimodal model to still yield a valid segmentation output using only the modality that is available and the performance is comparable to that of the unimodal model with the remaining modality.

\subsection{Generalization of SSMA fusion to Other Tasks}
\label{sec:multimodalGeneralization}

In order to demonstrate the generalization of our proposed SSMA module for multimodal fusion in other tasks, we report results for the scene type classification task on the ScanNet benchmark. The goal of the scene type classification task is to classify scans of indoor scenes into 13 distinct categories, namely, \textit{apartment, bathroom, bedroom/hotel, bookstore/library, conference room, copy/mail room, hallway, kitchen, laundry room, living room/lounge, misc, office, storage/basement/garage}. The benchmark ranks the methods according to recall and the intersection-over-union metric (IoU). For our approach, we employ the top-down 2D projection of the textured scans as one modality and the jet-colorized depth map of the top-down 2D projection of the scans as another modality. We utilize the SE-ResNetXt-101~\citep{hu2018squeeze} architecture for the unimodal model and for the multimodal network backbone. Our multimodal architecture has a late fusion topology with two individual modality-specific SE-ResNetXt-101 streams that are fused after block 5 using our SSMA module. The output of the SSMA module is fed to a fully connected layer that has number of output units equal to the number of scene classes in the dataset. We evaluate the performance of our multimodal SSMA fusion model against the individual modality-specific networks, as well as the multimodal fusion baselines such as Average, Maximum, Stacking and Late Fusion, as described in \secref{sec:multimodalFusionEx}.

\begin{table}
\footnotesize
\centering
\caption{Performance of multimodal SSMA fusion for the scene type classification task on the ScanNet benchmark. Results are reported on the validation set.}
\label{tab:multimodalScanNetSceneType}
\begin{tabular}{p{1.2cm}p{2.3cm}p{1cm}p{1cm}}
\noalign{\smallskip}\hline\noalign{\smallskip}
Network & Approach & mIoU & mRecall \\
 & & (\%) & (\%) \\
\noalign{\smallskip}\hline\hline\noalign{\smallskip}
RGB & Unimodal & $33.16$ & $43.37$ \\
Depth & Unimodal & $35.07$ & $44.82$ \\
\noalign{\smallskip}\hline\noalign{\smallskip}
\multirow{5}{*}{RGB-D} & Average & $34.92$ & $43.99$ \\
& Maximum & $34.75$ & $43.79$ \\
& Stacking & $32.47$ & $41.89$ \\
& Late Fusion & $35.17$ & $44.48$ \\
& SSMA (Ours) & $\mathbf{37.45}$ & $\mathbf{54.28}$ \\
\noalign{\smallskip}\hline\noalign{\smallskip}
\end{tabular}
\end{table}

Results from this experiment on the ScanNet validation set are shown in \tabref{tab:multimodalScanNetSceneType}. It can be seen that the unimodal depth model outperforms the RGB model in both the mean IoU (mIoU) score and the mean recall (mRecall). Among the multimodal fusion baselines, only the Late Fusion network outperforms the unimodal depth model by a small margin in the mIoU score but it achieves a lower mean recall. However, our multimodal SSMA fusion model achieves the state-of-the-art performance with a mIoU score of $37.45\%$ and a mean recall of $54.28\%$. This accounts for an improvement of $2.28\%$ in the mIoU score and $9.8\%$ in the mean recall over the Late Fusion model, and a larger improvement over the performance of the unimodal depth model. Since the only difference in the Late Fusion architecture and the SSMA architecture is how the multimodal fusion is carried out, the improvement achieved by the SSMA model can be solely attributed to the dynamic fusion mechanism of our SSMA module. We also benchmarked on the ScanNet test set for scene classification, in which our multimodal SSMA model achieves a mIoU score of $35.5\%$ and a mean recall of $49.8\%$, thereby setting the state-of-the-art for scene type classification on this benchmark.

\subsection{Multimodal Fusion Ablation Studies}
\label{sec:multimodalFusionAblation}

In this section, we study the influence of various contributions that we make for multimodal fusion. Specifically, we evaluate the performance by comparing the fusion at different intermediate network stages. We then evaluate the utility of our proposed channel attention scheme for better correlation of mid-level encoder and high-level decoder features. Subsequently, we experiment with different SSMA bottleneck downsampling rates and qualitatively analyze the convolution activation maps of our fusion model at various intermediate network stages to study the effect of multimodal fusion on the learned network representations.

\subsubsection{Experiments with Real-Time Backbone Networks}
\label{sec:realTimeExperiments}

In the interest of real-time performance, we additionally trained multimodal models in our proposed SSMA fusion framework with different real-time-intended backbone networks. Specifically, we performed experiments using two different backbone networks: ERFnet~\citep{romera2018erfnet} and MobileNet~v2~\citep{sandler2018mobilenetv2}. For the ERFnet fusion model, we replace the two modality-specific encoders in our multimodal fusion configuration with the ERFnet encoder and we replace our decoder with the ERFnet decoder. While for the fusion model with the MobileNet~v2 backbone, we employ the MobileNet~v2 topology for the two modality-specific encoders in our multimodal fusion configuration and we append our eASPP as well as the decoder from our AdapNet++ architecture. Note that ERFnet is a semantic segmentation architecture with both an encoder and decoder, while MobileNet~v2 is only a classification architecture and therefore it only has an encoder topology.

\begin{table}
\footnotesize
\centering
\caption{Performance of multimodal SSMA fusion technique with different real-time backbone networks. Results are shown on the Cityscapes validation set (input image dim: $768\times384$).}
\label{tab:multimodalCityscapesRT}
\begin{tabular}{p{2cm}p{2.5cm}p{1cm}p{0.8cm}}
\noalign{\smallskip}\hline\noalign{\smallskip}
Network & Modalities & mIoU & Time \\
Backbone & & (\%) & ($\milli\second$) \\
\noalign{\smallskip}\hline\hline\noalign{\smallskip}
\multirow{3}{*}{ERFnet} & RGB & $62.71$ & $43.87$ \\
& HHA & $51.84$ & $43.87$ \\
& RGB-HHA SSMA & $64.60$ & $66.06$ \\
\noalign{\smallskip}\hline\noalign{\smallskip}
\multirow{3}{*}{MobileNet v2} & RGB & $74.78$ & $47.43$ \\
& HHA & $61.89$ & $47.43$ \\
& RGB-HHA SSMA & $77.17$ & $73.62$ \\
\noalign{\smallskip}\hline\noalign{\smallskip}
\multirow{3}{*}{AdapNet++} & RGB & $80.80$ & $72.77$ \\
& HHA & $67.66$ & $72.77$ \\
& RGB-HHA SSMA & $82.64$ & $99.96$ \\
\noalign{\smallskip}\hline\noalign{\smallskip}
\end{tabular}
\end{table}

Results from this experiment for RGB-HHA fusion along with the unimodal RGB and unimodal HHA performance are shown in \tabref{tab:multimodalCityscapesRT}. The ERFnet model in our multimodal SSMA fusion configuration achieves a mIoU score of $64.60\%$ with an inference time of $66.06\milli\second$, thereby outperforming both the unimodal RGB ERFnet model and the unimodal HHA ERFnet model. The MobileNet~v2 model in our multimodal SSMA fusion configuration further outperforms the ERFnet fusion model by $12.57\%$ in the mIoU score with an inference time of $73.62\milli\second$. In comparison to these network backbones, our AdapNet++ fusion model achieves a mIoU score of $82.84\%$ with an inference time of $99.96\milli\second$. Each of these multimodal fusion models outperform their unimodal counterparts and have an inference time in the range of $66\milli\second - 99 \milli\second$. This demonstrates the modularity of our fusion framework that enables the selection of an appropriate network backbone according to the desired frame rate.

\subsubsection{Detailed Study on the Fusion Architecture}
\label{sec:fusionAblationDetails}

\begin{table}
\footnotesize 
\renewcommand{\arraystretch}{1.4}
\centering
\caption{Effect of the various contributions proposed for multimodal fusion using AdapNet++ and the SMMA architecture. The performance is shown for RGB-HHA fusion on the Cityscapes dataset and evaluated on the validation set.}
\label{tab:ssmaFusionEvolution}
\begin{tabular}{p{0.8cm}p{0.5cm}p{0.5cm}p{1.2cm}p{0.7cm}p{0.7cm}p{0.7cm}}
\noalign{\smallskip}\hline\noalign{\smallskip}
Model & \multicolumn{3}{c}{SSMA Fusion} & mIoU & Acc. & AP \\
\cline{2-4}
& ASPP & Skip & Ch. Agg. & (\%) & (\%) & (\%) \\
\noalign{\smallskip}\hline\hline\noalign{\smallskip}
F0 & - & - & - & $80.77$ & $96.04$ & $90.97$ \\
F1 & \checkmark & - & - & $81.55$ & $96.19$ & $91.15$ \\
F2 & \checkmark & \checkmark & - & $81.75$ & $96.25$ & $91.09$ \\
F3 & \checkmark & \checkmark & \checkmark & $\mathbf{82.64}$ & $\mathbf{96.41}$ & $\mathbf{90.65}$ \\
\noalign{\smallskip}\hline\noalign{\smallskip}
\end{tabular}
\end{table}

In our proposed multimodal fusion architecture, we employ a combination of mid-level fusion and late-fusion. Results from fusion at each of these stages is shown in \tabref{tab:ssmaFusionEvolution}. First, we employ the main SSMA fusion module at the end of the two modality-specific encoders, after the eASPPs and we denote this model as F1. The F1 model achieves a mIoU of $81.55\%$, which constitutes to an improvement of $0.78\%$ over the unimodal F0 model. We then employ an SSMA module at each skip refinement stage to fuse the mid-level skip features from each modality-specific stream. The fused skip features are then integrated into the decoder for refinement of high-level decoder features. The F2 model that performs multimodal fusion at both stages, yields a mIoU of $81.75\%$, which is not significant compared to the improvement that we achieve in fusion of the mid-level features into the decoder in our unimodal AdapNet++ architecture. As described in \secref{sec:approachFusion}, we hypothesise that this is due to the fact that the mid-level representations learned by a network do not align across different modality-specific streams. Therefore, we employ our channel attention mechanism to better correlate these features using the spatially aggregated statistics of the high-level decoder features. The model that incorporates this proposed attention mechanism achieves an improved mIoU of $82.64\%$, which is an improvement of $1.09\%$ in comparison to $0.2\%$ without the channel attention mechanism. Note that the increase in quantitative performance due to multimodal fusion is more apparent in the indoor or synthetic datasets as shown in \secref{sec:multimodalFusionEx}. The experiment shown in \tabref{tab:ssmaFusionEvolution} shows a correspondingly larger increase for the contributions presented in this table. However, as we present the unimodal ablation studies on the Cityscapes dataset, we continue to show the multimodal ablation studies on the same dataset.

\begin{table}
\footnotesize 
\renewcommand{\arraystretch}{1.4}
\centering
\caption{Effect of varying the SSMA bottleneck downsampling rate $\eta$ on the RGB-HHA fusion performance. Results are shown for the model trained on the Cityscapes dataset and evaluated on the validation set.}
\label{tab:ssmaRate}
\begin{tabular}{p{1.9cm}p{1.5cm}p{0.9cm}p{0.9cm}p{0.9cm}}
\noalign{\smallskip}\hline\noalign{\smallskip}
Encoder & Skip & mIoU & Acc. & AP \\
SSMA & SSMA & (\%) & (\%) & (\%) \\
\noalign{\smallskip}\hline\hline\noalign{\smallskip}
$\eta_{enc} = 2$ & $\eta_{skip} = 2$ & $82.15$ & $96.32$ & $91.17$ \\
$\eta_{enc} = 4$ & $\eta_{skip} = 4$ & $82.11$ & $96.27$ & $91.34$ \\
$\eta_{enc} = 8$ & $\eta_{skip} = 4$ & $82.21$ & $96.32$ & $91.61$ \\
$\eta_{enc} = 16$ & $\eta_{skip} = 4$ & $82.25$ & $96.31$ & $91.12$ \\ 
$\eta_{enc} = 16$ & $\eta_{skip} = 6$ & $\mathbf{82.64}$ & $\mathbf{96.41}$ & $\mathbf{90.65}$ \\ 
\noalign{\smallskip}\hline\noalign{\smallskip}
\end{tabular}
\end{table}

The proposed SSMA module has a bottleneck structure in which the middle convolution layer downsamples the number of feature channels according to a rate $\eta$ as described in \secref{sec:ssmaBlock}. As we perform fusion both at the mid-level and at the end of the encoder section, we have to estimate the downsampling rates individually for each of the SSMA blocks. We start by using values from a geometric sequence for the main encoder SSMA downsampling rate $\eta_{enc}$ and correspondingly vary the values for the skip SSMA downsampling rates $\eta_{skip}$. Results from this experiment shown in \tabref{tab:ssmaRate} demonstrates that the best performance is obtained for $\eta_{enc} = 16$ and $\eta_{skip} = 6$ which also increases the parameter efficiency compared to lower downsampling rates.

\subsubsection{Fusion Stage and Reliance on Modalities}
\label{sec:fusionAblationLocation}

In this section, we study the SSMA fusion configuration in terms of learning dependent or independent probability weightings that are used to recalibrate the modality-specific feature maps dynamically. The SSMA configuration that we depict in \figref{fig:ssmaUnit} learns independent probability weighting where the activations at a specific location in the feature maps from modality A can be independently enhanced or suppressed regardless of whether the activations at the corresponding location in the feature maps from modality B are going to be enhanced or suppressed. An alternative dependent configuration can be employed by replacing the sigmoid with a $\mathsf{softmax}$, where the $\mathsf{softmax}$ takes the activation at a specific location from a feature map from modality A and the activation in the corresponding location from the feature map from modality B, and outputs dependent probabilities that are used to weigh the modality-specific activations. This dependent configuration acts as punishing the modality that makes the mistake while rewarding the other. While the independent configuration also considers if a modality is making a mistake, it does not necessarily punish one and reward the other, it has the ability to punish both or reward both in addition. We study the performance of the multimodal fusion in these two dependent and independent configurations in this section. Additionally, we study the effect of learning the fusion in a fully supervised manner by employing an explicit loss function at the output of each SSMA module after the fusion. We also study the overall configuration on where the SSMA module is to be placed, at the end of the encoder stage where the features are highly discriminative or at the end of the decoder stage where the features are more high-level and semantically mature.

\begin{table}
\footnotesize 
\centering
\renewcommand{\arraystretch}{1.2}
\caption{Comparison of multimodal SSMA fusion at different network stages and by employing a dynamic dependent probability weighting or independent probability weighting configuration. Results are shown for RGB-HHA fusion on the Cityscapes validation set with (supervised) and without (self-supervised) an explicit loss for the SSMA module.}
\label{tab:ssmaLoc}
\begin{tabular}{p{0.8cm}p{0.8cm}p{0.7cm}p{0.7cm}|p{0.7cm}p{0.7cm}|p{0.6cm}}
\noalign{\smallskip}\hline\noalign{\smallskip}
Model & SSMA & \multicolumn{2}{c|}{Encoder Stage} & \multicolumn{2}{c|}{Decoder Stage} & IoU \\
\cline{3-6}
& Loss & Dep. & Indep. & Dep. & Indep. & (\%) \\
\noalign{\smallskip}\hline\hline\noalign{\smallskip}
S1 & - & \checkmark & - & - & - & $82.39$ \\
S2 & - & - & \checkmark & - & - & $\mathbf{82.64}$ \\
S3 & - & - & - & \checkmark & - & $80.96$ \\
S4 & - & - & - & - & \checkmark & $81.14$ \\
S5 & \checkmark & \checkmark & - & - & - & $82.32$ \\
S6 & \checkmark & - & \checkmark & - & - & $82.52$ \\
S7 & \checkmark & - & - & \checkmark & - & $80.86$ \\
S8 & \checkmark & - & - & - & \checkmark & $81.06$ \\
\noalign{\smallskip}\hline\noalign{\smallskip}
\end{tabular}
\end{table}

\tabref{tab:ssmaLoc} shows the results from this experiment where we present the multimodal fusion performance for the model in the dependent SSMA and independent SSMA configuration, as well as when the SSMA module is placed at the encoder stages as in our standard configuration or at the decoder stage, and with and without an explicit supervision for the fusion. The results are presented for RGB-HHA fusion on the Cityscapes validation set. It can be seen that the models S1 to S4 without the explicit supervision outperform the corresponding models S5 to S8 with the explicit supervision demonstrating that learning the fusion in self-supervised manner is more beneficial. Comparing the performance of models that employ the SSMA fusion at the encoder stages S1, S2, S5, S6, with the corresponding models at employ the SSMA fusion at the decoder stage S3, S4, S7, S8, we see that the encoder fusion models substantially outperform the decoder fusion models. Finally, comparing the models with the dependent SSMA configuration S1, S3, S5, S7 with the corresponding models the the independent SSMA configuration, we observe that independent configuration always outperforms the dependent configuration. In summary, these results demonstrate that our multimodal SSMA fusion scheme that learns independent probability weightings in a self-supervised manner, when employed at the encoder stages outperforms the other configurations.

\subsubsection{Influence of Multimodal Fusion on Activation Maps}

\begin{figure*}
\centering
\footnotesize
\setlength{\tabcolsep}{0.1cm}
{\renewcommand{\arraystretch}{2}
\begin{tabular}{P{0.3cm}P{3.2cm}P{3.2cm}P{3.2cm}P{3.2cm}P{3.2cm}}
& Person & Pole & Table & Bathtub & Trail \\
\rotatebox{90}{Input Mod 1} & \includegraphics[width=\linewidth]{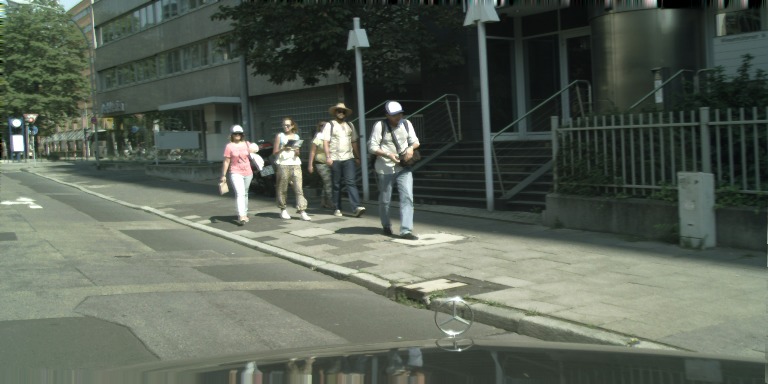} & \includegraphics[width=\linewidth]{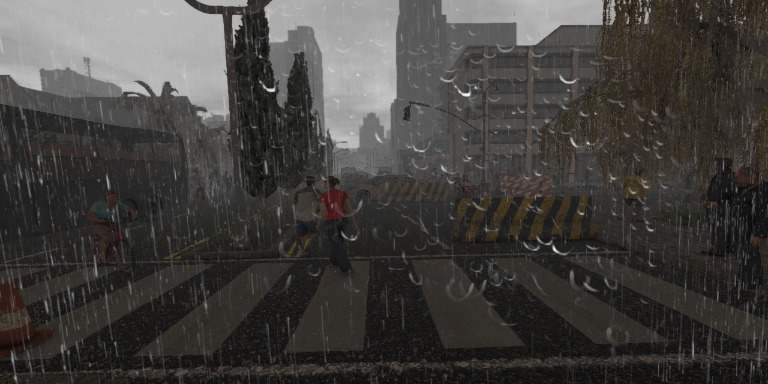} & \includegraphics[width=\linewidth]{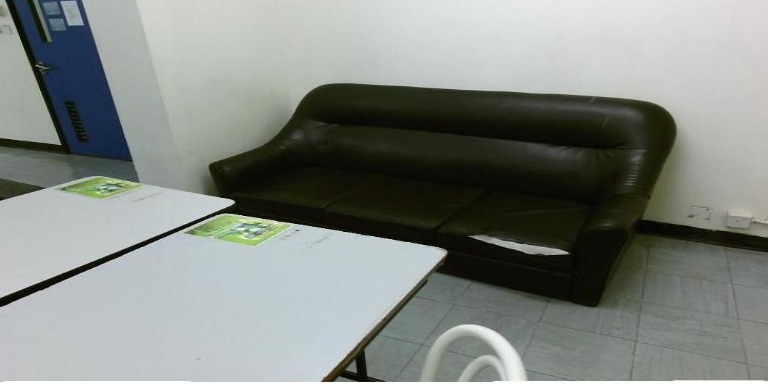} & \includegraphics[width=\linewidth]{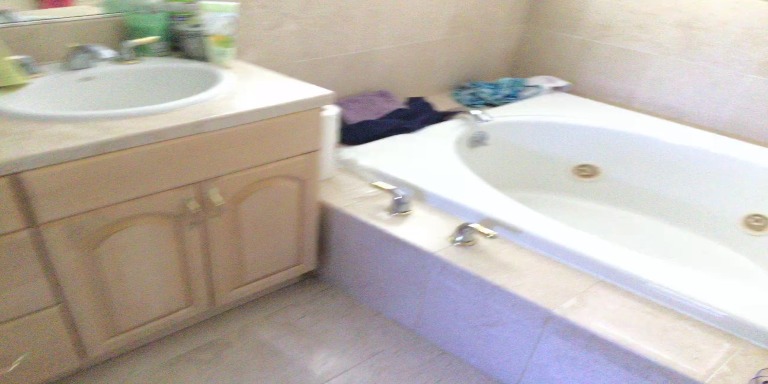} & \includegraphics[width=\linewidth]{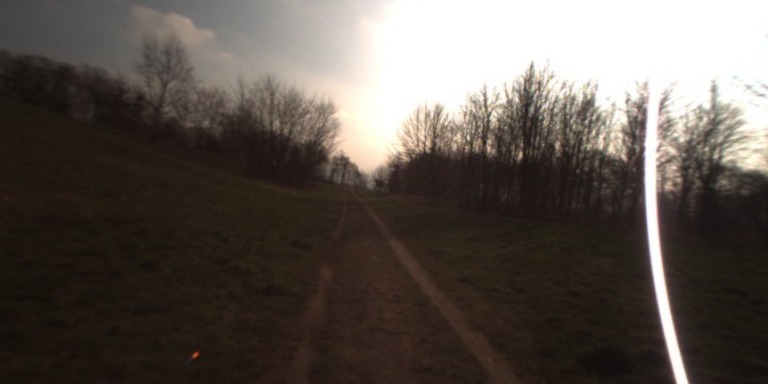} \\
\rotatebox{90}{Input Mod 2} & \includegraphics[width=\linewidth]{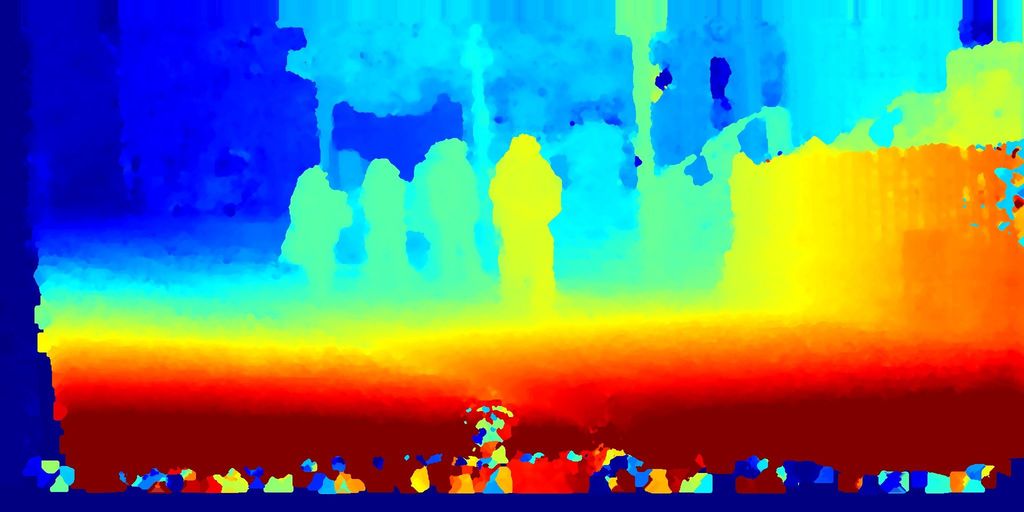} & \includegraphics[width=\linewidth]{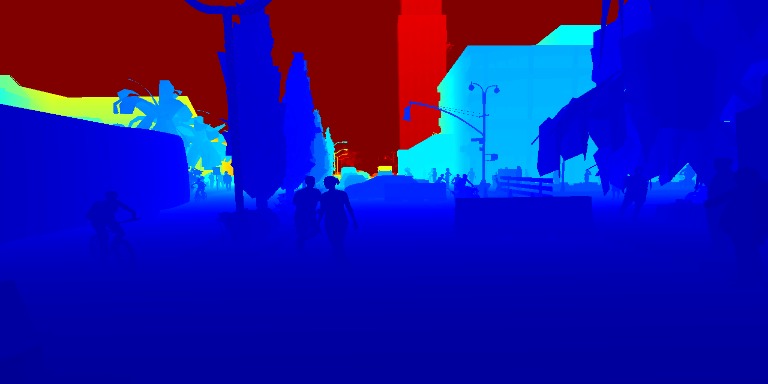} & \includegraphics[width=\linewidth]{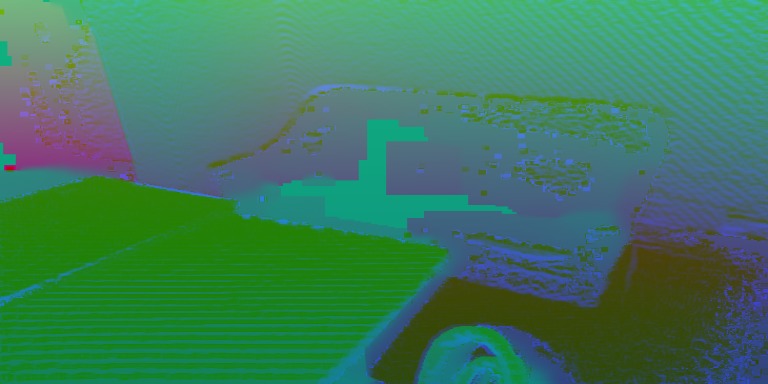} & \includegraphics[width=\linewidth]{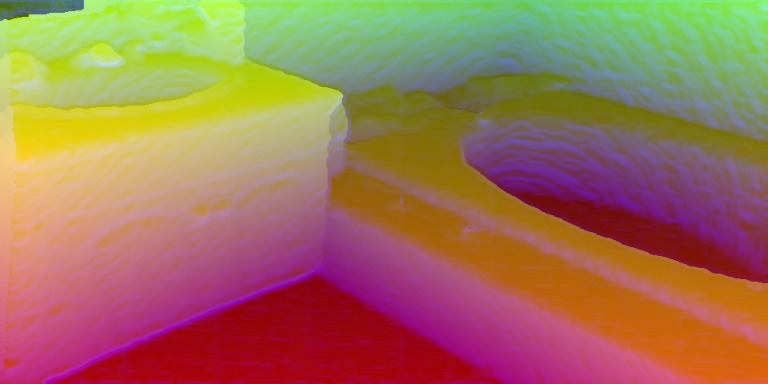} & \includegraphics[width=\linewidth]{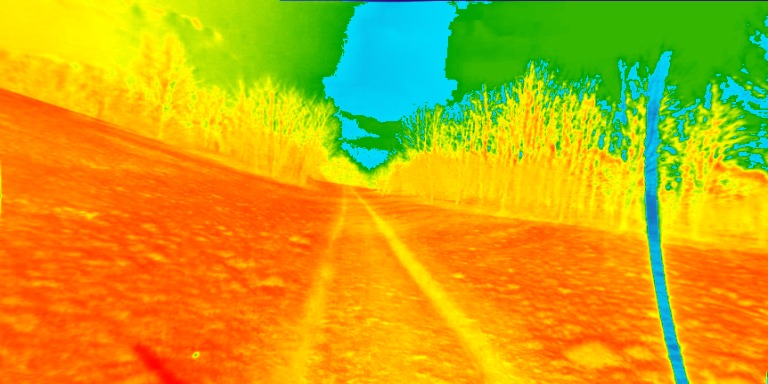} \\
\rotatebox{90}{Predicted Out} & \includegraphics[width=\linewidth]{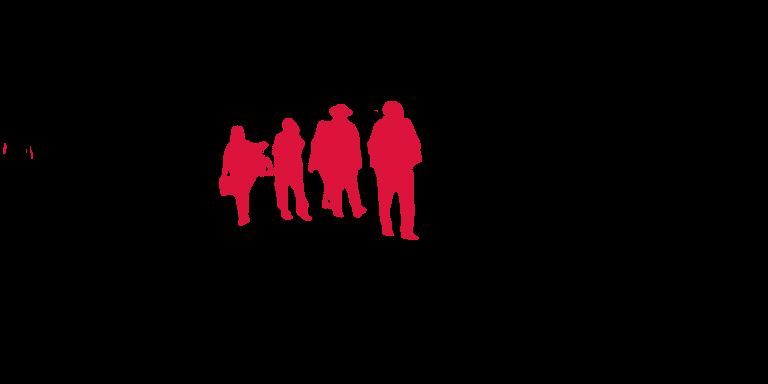} & \includegraphics[width=\linewidth]{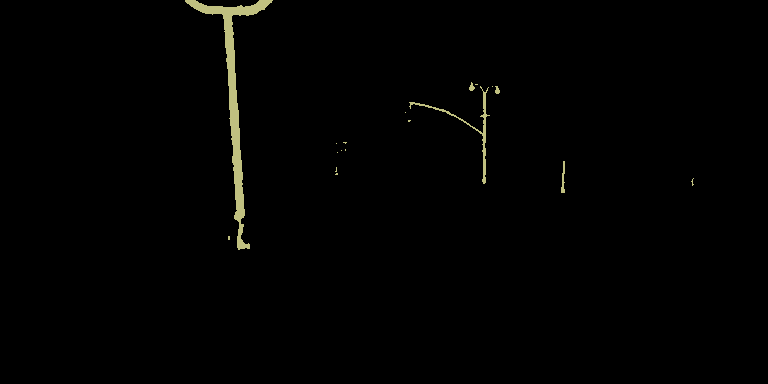} & \includegraphics[width=\linewidth]{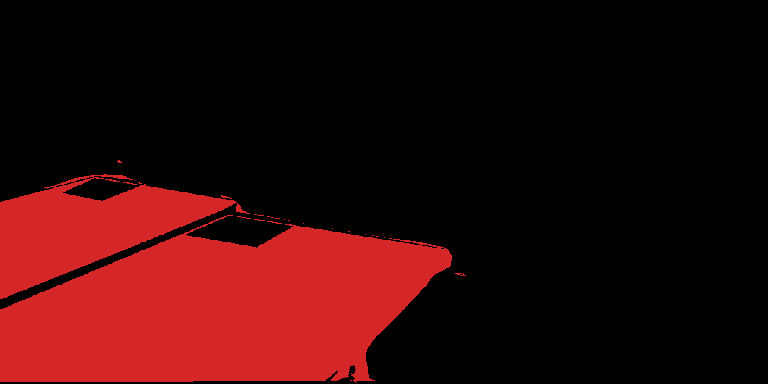} & \includegraphics[width=\linewidth]{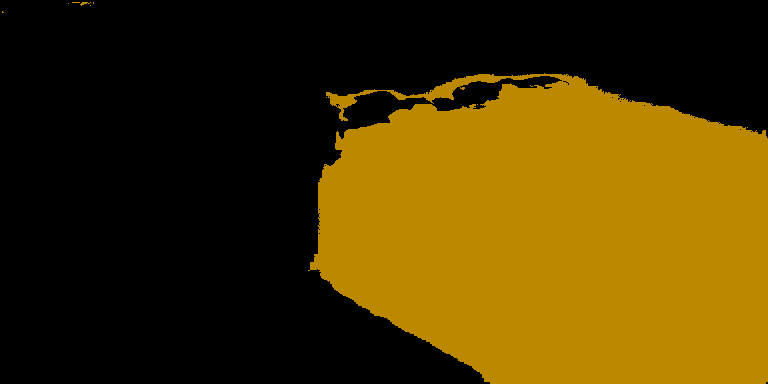} & \includegraphics[width=\linewidth]{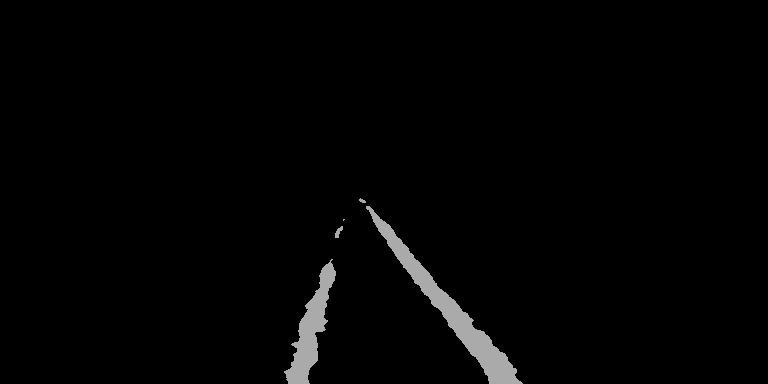} \\
\rotatebox{90}{Input $X^a$} & \includegraphics[width=\linewidth]{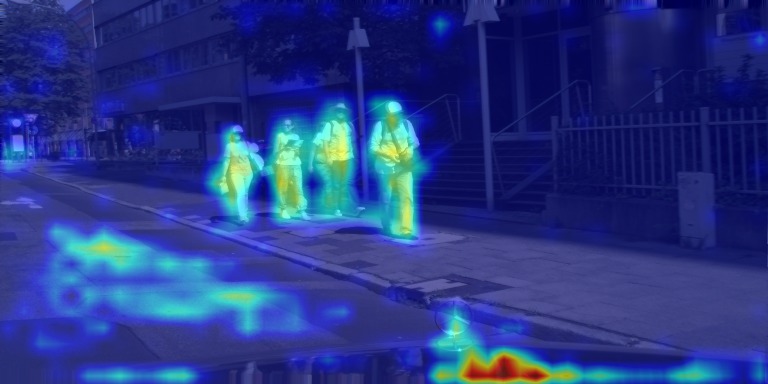} & \includegraphics[width=\linewidth]{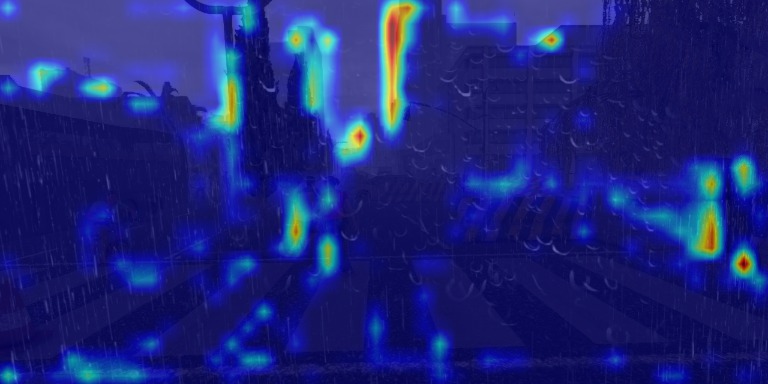} & \includegraphics[width=\linewidth]{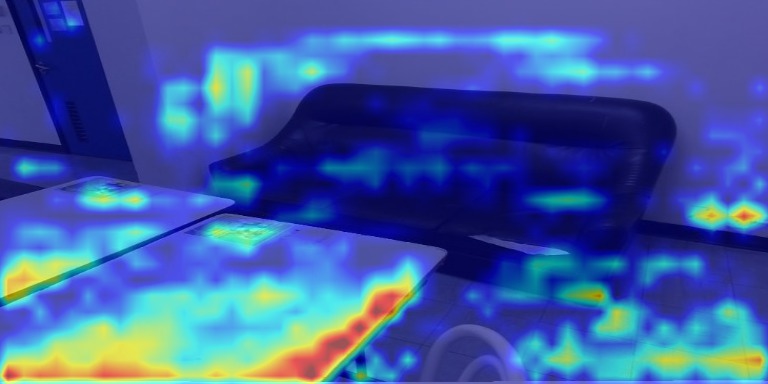} & \includegraphics[width=\linewidth]{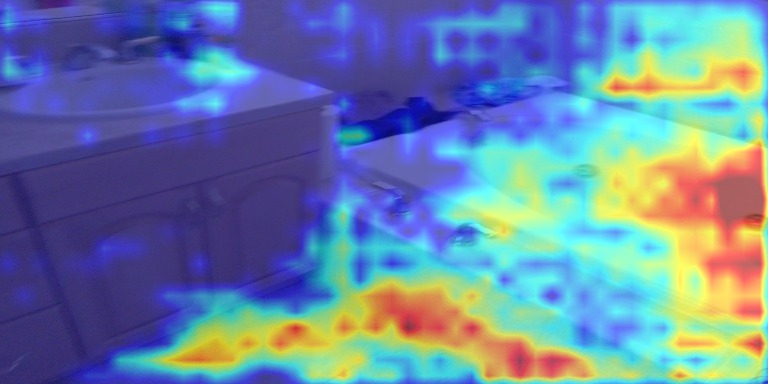} & \includegraphics[width=\linewidth]{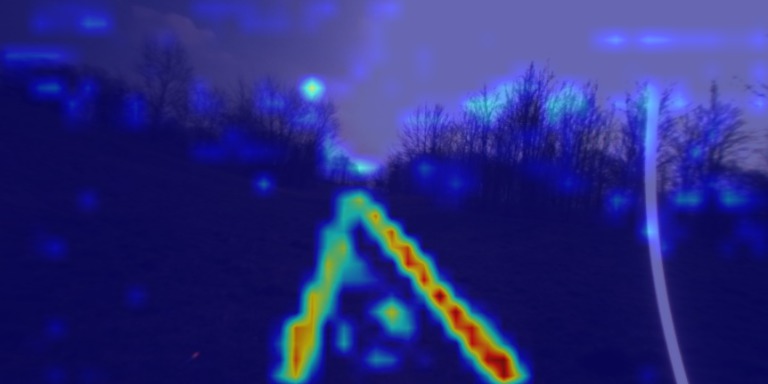} \\
\rotatebox{90}{Input $X^b$} & \includegraphics[width=\linewidth]{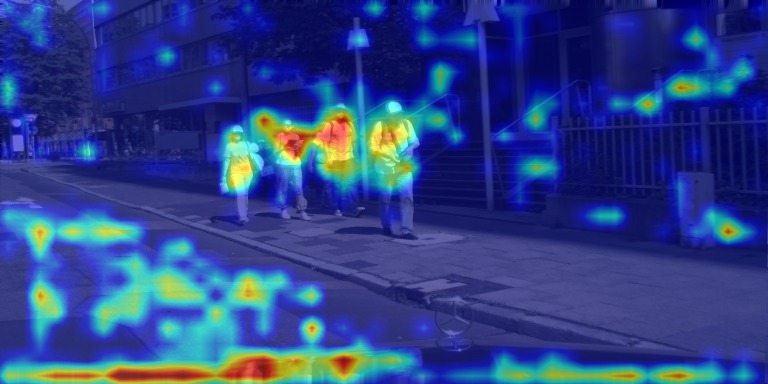} & \includegraphics[width=\linewidth]{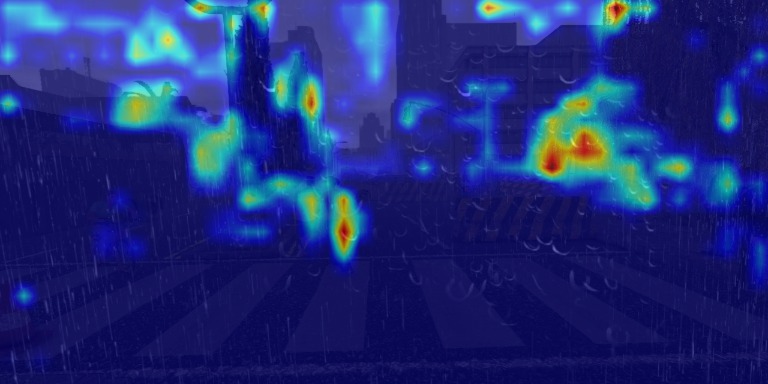} & \includegraphics[width=\linewidth]{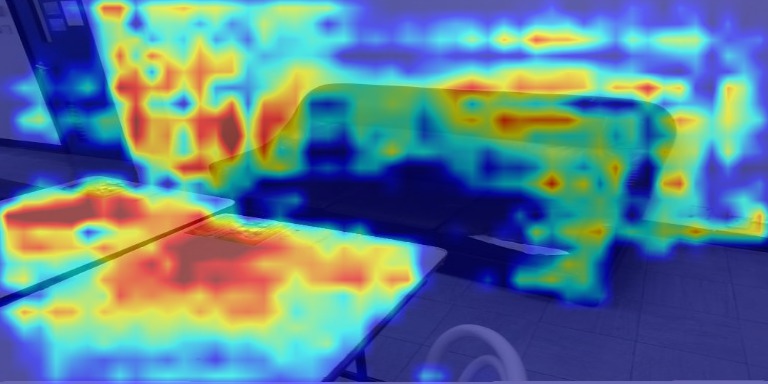} & \includegraphics[width=\linewidth]{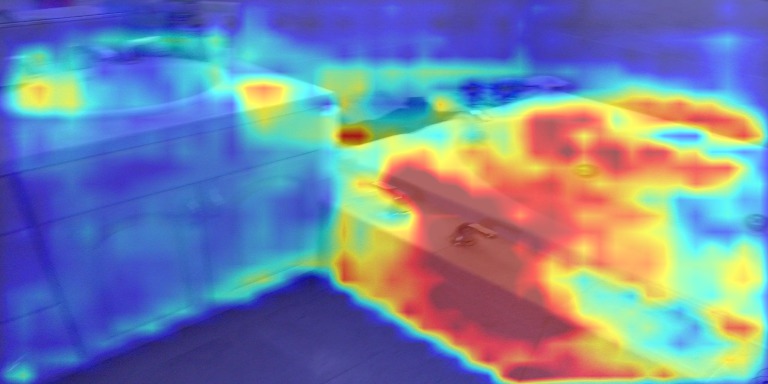} & \includegraphics[width=\linewidth]{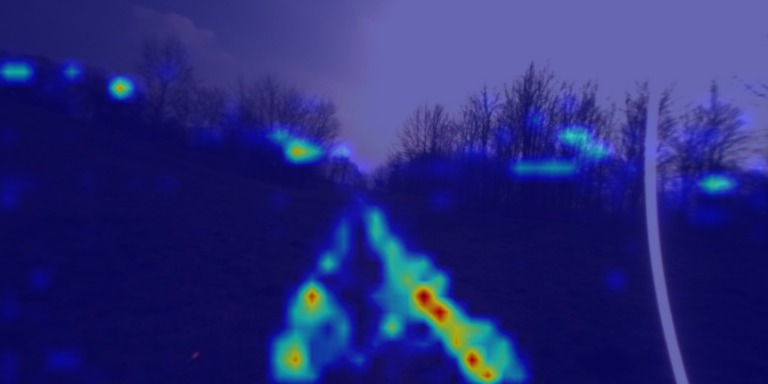} \\
\rotatebox{90}{Recalib. $\hat{X}^a$} & \includegraphics[width=\linewidth]{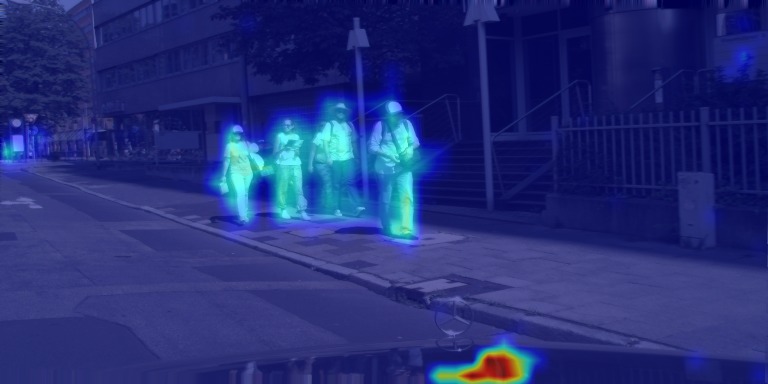} & \includegraphics[width=\linewidth]{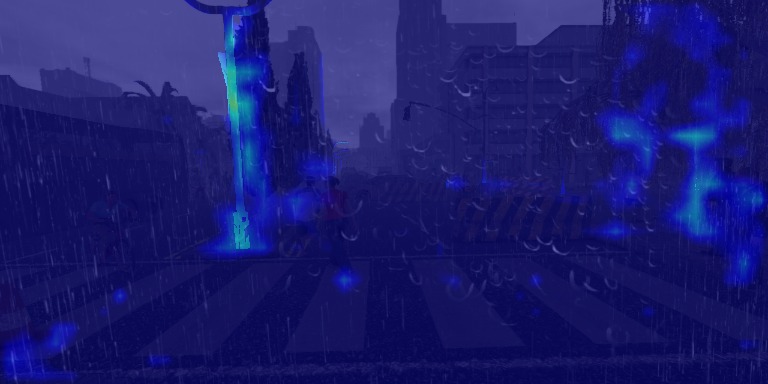} & \includegraphics[width=\linewidth]{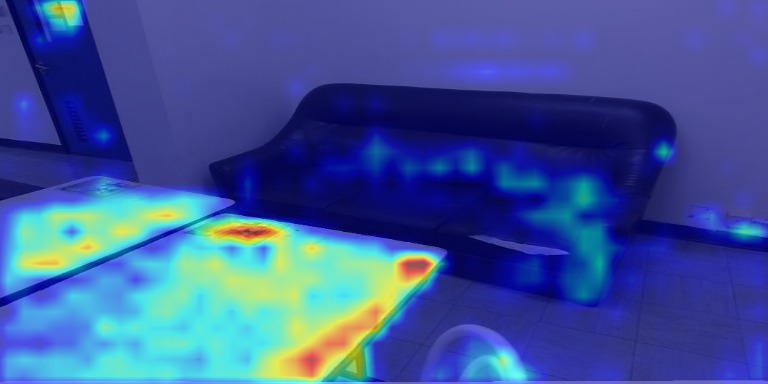} & \includegraphics[width=\linewidth]{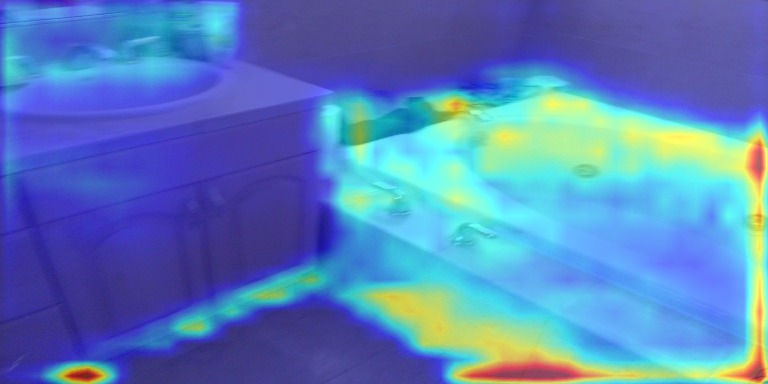} & \includegraphics[width=\linewidth]{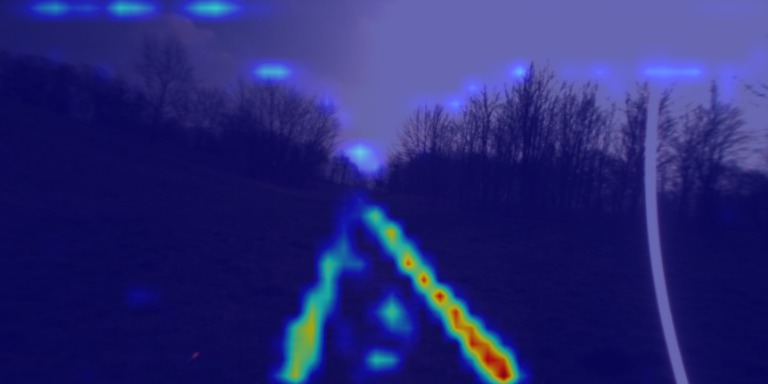} \\
\rotatebox{90}{Recalib. $\hat{X}^b$} & \includegraphics[width=\linewidth]{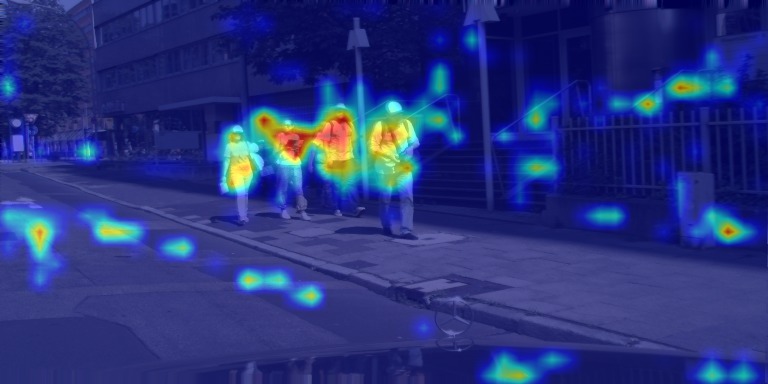} & \includegraphics[width=\linewidth]{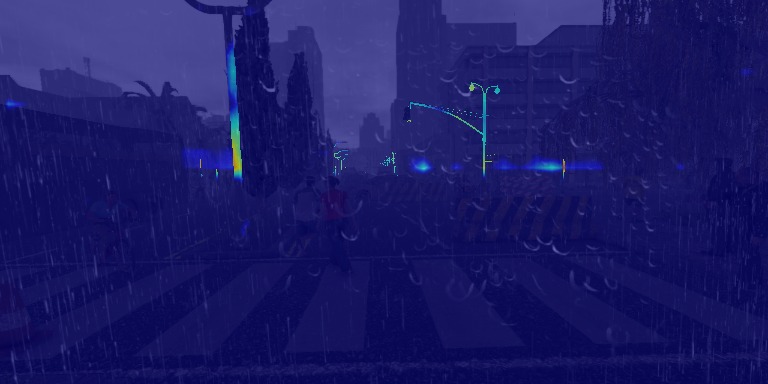} & \includegraphics[width=\linewidth]{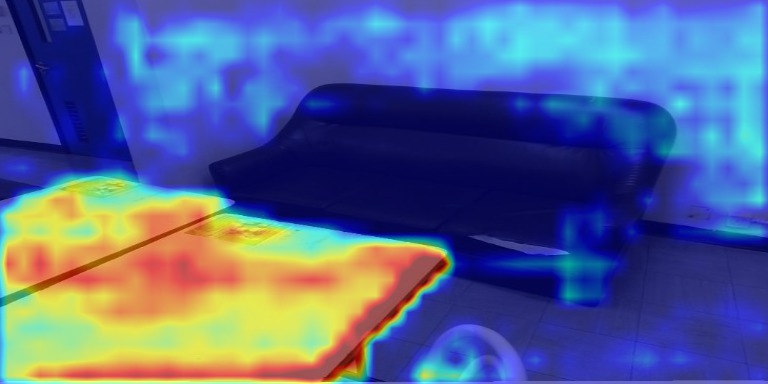} & \includegraphics[width=\linewidth]{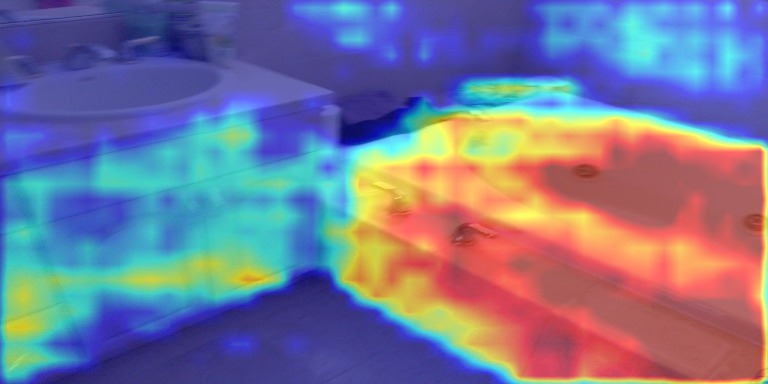} & \includegraphics[width=\linewidth]{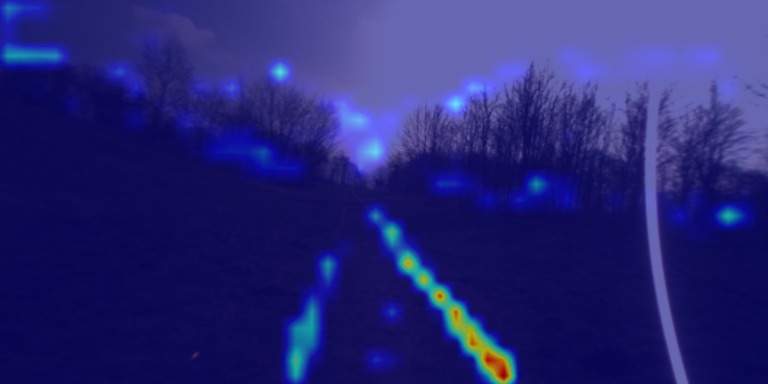} \\
\rotatebox{90}{Fused $f$} & \includegraphics[width=\linewidth]{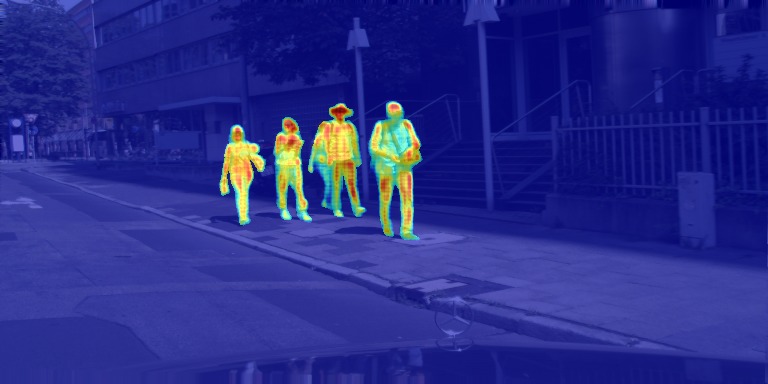} & \includegraphics[width=\linewidth]{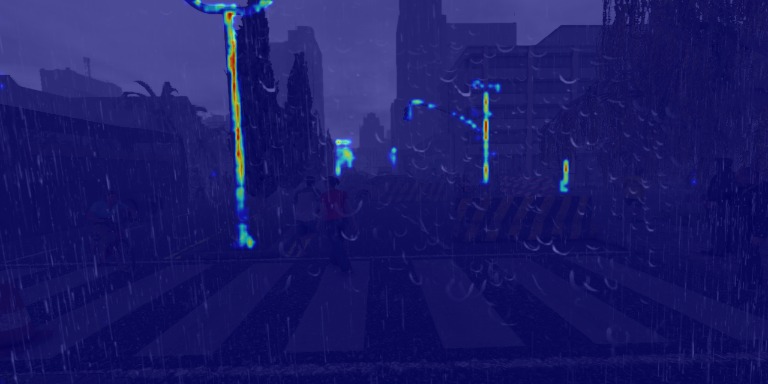} & \includegraphics[width=\linewidth]{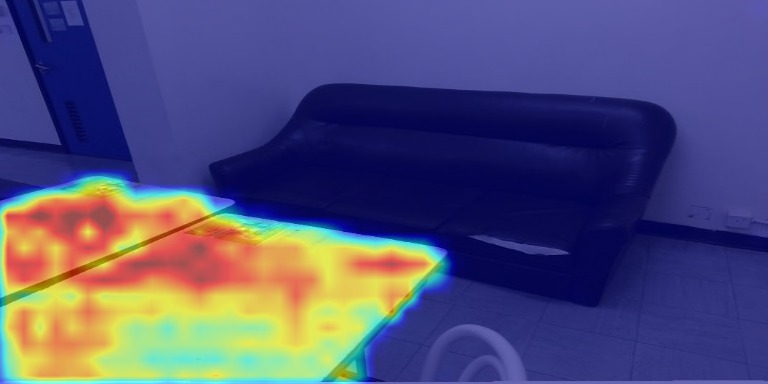} & \includegraphics[width=\linewidth]{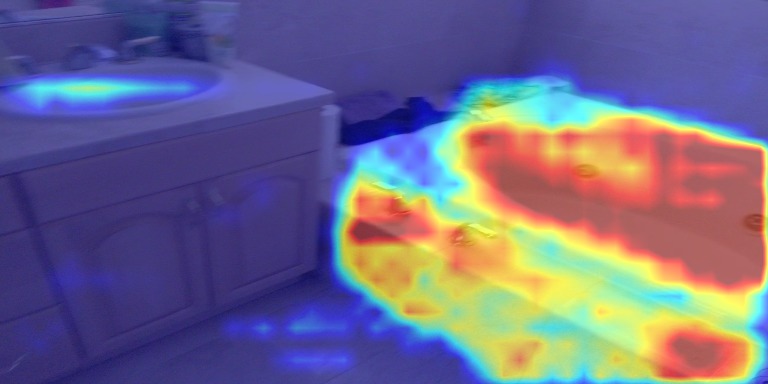} & \includegraphics[width=\linewidth]{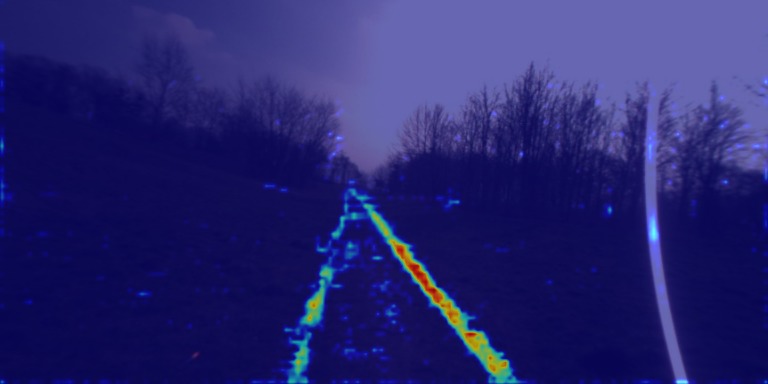} \\
& Cityscapes & Synthia & SUN RGB-D & ScanNet & Freiburg Forest \\ 
\end{tabular}}
\caption{Visualization of the activation maps with respect to a particular class at various stages of the network before and after multimodal fusion. $X^a$ and $X^b$ are at the outputs of the modality-specific encoder which is input to the SSMA fusion module, $\hat{X}^a$ and $\hat{X}^b$ are the at the feature maps after recalibration inside the SSMA block, and $f$ is after the fusion of both modalities inside the SSMA block. Both the input modalities and the corresponding segmentation output for the particular object category is also shown.}
\label{fig:asppActivations}
\end{figure*}

In an effort to present visual explanations for the improvement in performance due to multimodal fusion, we study the activation maps at various intermediate network stages before and after the multimodal fusion using the GradCam++ technique~\citep{chattopadhyay2017grad}. The approach introduces pixel-wise weighting of the gradients of the output with respect to a particular spatial location in the convolutional feature map to generate a score. The score provides a measure of the importance of each location in feature map towards the overall prediction of the network. We apply a colormap over the obtained scores to generate a heat map as shown in \figref{fig:asppActivations}. We visualize the activation maps at five different stages of the network. Firstly, at the output of each modality-specific encoder $X^a$ and $X^b$ which is the input to the SSMA fusion block. Secondly, after recalibrating the individual modality-specific feature maps inside the SSMA block $\hat{X}^a$ and $\hat{X}^b$, and finally after the fusion with the $3\times 3$ convolution inside the SSMA block $f$. \figref{fig:asppActivations} illustrates one example for each dataset that we benchmark on with the activation maps, the input modalities and the corresponding segmentation output for the particular object category. 

For the Cityscapes dataset, we show the activation maps for the \textit{person} category. It can be seen that the activation map $X^a$ from the visual RGB stream is well defined for the \textit{person} class but it does not show high activations centered on the objects, whereas the activation map from the depth stream $X^b$ is more noisy but high activations are shown on the objects. For the locations in the input depth map that show noisy depth data, the activation map correspondingly shows prominent activations in these regions. After the recalibration of the feature maps, both $\hat{X}^a$ and $\hat{X}^b$  are less noisy and maintaining the structure with high activations. Furthermore, the activation map of the fused convolution $f$ shows very well defined high activations that almost correspond to the segmented output.

The second column in \figref{fig:asppActivations} shows the activations for the \textit{pole} class in the Synthia dataset. As the scene was captured during rainfall, the objects in the visual RGB image are indistinguishable. However, the depth map still maintains some structure of the scene. Studying both the modality-specific activation maps at the input to the SSMA module shows substantial amount of noisy activations spread over the scene. Therefore, the unimodal visual RGB model only achieves an IoU of $74.94\%$ for the pole class. Whereas, after the recalibration of the feature maps, the activation maps show significantly reduced noise. It can be seen the recalibrated activation map $\hat{X}^b$  of the depth stream shows more defined high activations on the pole, whereas $\hat{X}^a$ of the visual RGB stream shows less amount of activations indicating that the network suppresses the noisy RGB activations in order to better leverage the well defined features from the depth stream. Activations of the final fused convolution layer show higher activations on the pole than either of the recalibrated activation maps demonstrating the utility of multimodal fusion. This enables the fusion model to achieve an improvement of $8.4\%$ for the pole class.

The third column of \figref{fig:asppActivations} shows the activation maps for the \textit{table} class in the SUN RGB-D dataset. Interestingly, both the modality-specific activation maps at the input show high activations at different locations indicating the complementary nature of the features in this particular scene. However, the activation map $X^b$ from the HHA stream also shows high activations on the couch in the background which would cause misclassifications. After the recalibration of the HHA feature maps, the activation map $\hat{X}^b$ no longer has high activations on the couch but it retains the high activations on the table. While, the recalibrated activation map $\hat{X}^a$ of the visual RGB stream shows significantly lesser noisy activations. The activation map of the fused convolution $f$ shows well defined high activations on the table, more than the modality-specific input activation maps. The enables the SSMA fusion model to achieve an improvement of $4.32\%$ in the IoU for the table category.

For the ScanNet dataset, we show the activation maps for the \textit{bathtub} category in the fourth column in \figref{fig:asppActivations}. It can be seen that the modality specific activation maps at the input of the SSMA module shows high activations at complementary locations, corroborating the utility of exploiting features from both modalities. Moreover, the activation map $X^b$ from the HHA stream shows significantly higher activations on the object of interest than the RGB stream. This also aligns with the quantitative results, where the unimodal HHA model outperforms the model trained on visual RGB images. After the recalibration of the feature maps inside the SSMA block, the activation maps show considerably lesser noise while maintaining the high activations at complementary locations.  The activation map of the fused convolution $f$ shows only high activations on the bathtub and resembles the actual structure of the segmented output.

The last column of \figref{fig:asppActivations} shows the activation maps for the \textit{trail} category in the Freiburg Forest dataset. Here we show the fusion with visual RGB and EVI. The EVI modality does not provide substantial complementary information for the trail class in comparison to the RGB images. This is also evident in the visualization of the activations at the input of the SSMA module. The activation maps of the EVI modality after the recalibration show significantly lesser noise but also lesser amount of high activation regions than the recalibrated activation maps of the visual RGB stream. Nevertheless, the activation map after the fusion $f$ shows more defined structure of the trail than either of the modality-specific activation maps of the input to the SSMA module.

\subsection{Qualitative Results of Multimodal Segmentation}
\label{sec:qualitative}

\begin{figure*}
\centering
\footnotesize
\setlength{\tabcolsep}{0.05cm}
{\renewcommand{\arraystretch}{1}
\begin{tabular}{P{0.25cm}P{3.3cm}P{3.3cm}P{3.3cm}P{3.3cm}P{3.3cm}}
\arrayrulecolor{red}
& Input Modality1 & Input Modality2 & Unimodal Output & Multimodal Output & Improvement\textbackslash Error Map \\
\rotatebox{90}{(a) Cityscapes} & \includegraphics[width=\linewidth]{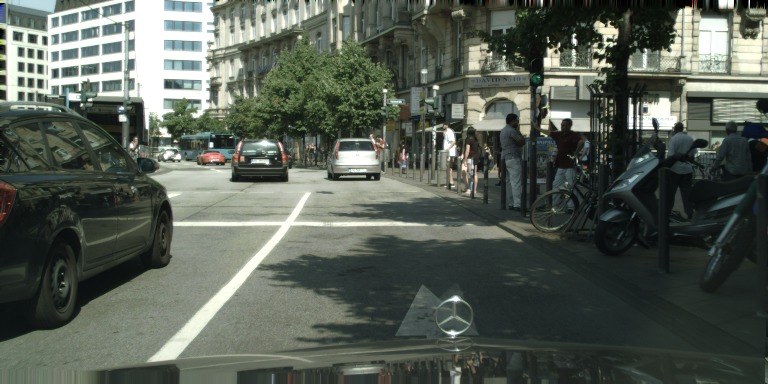} & \includegraphics[width=\linewidth]{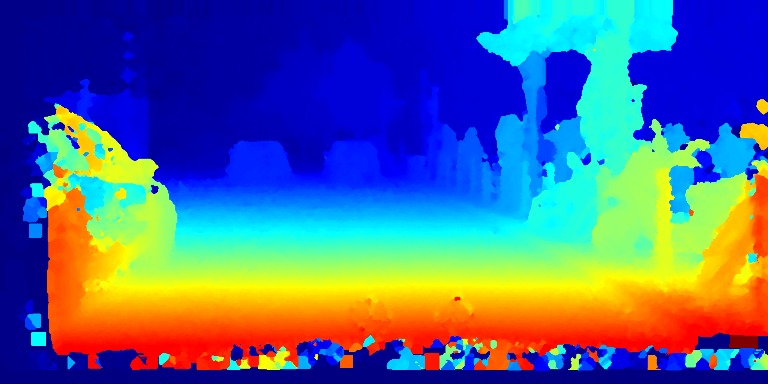} & \includegraphics[width=\linewidth]{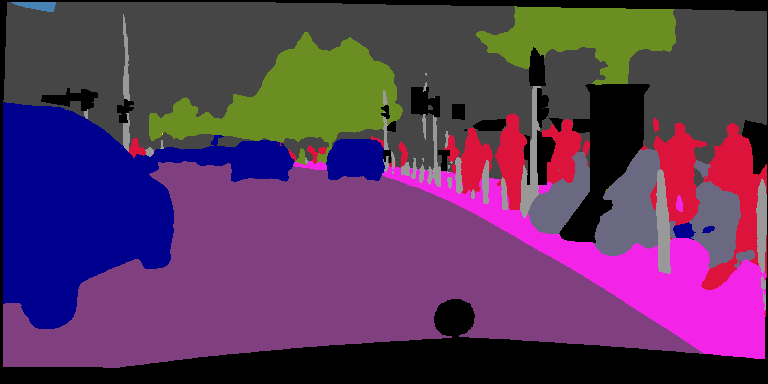} & \includegraphics[width=\linewidth]{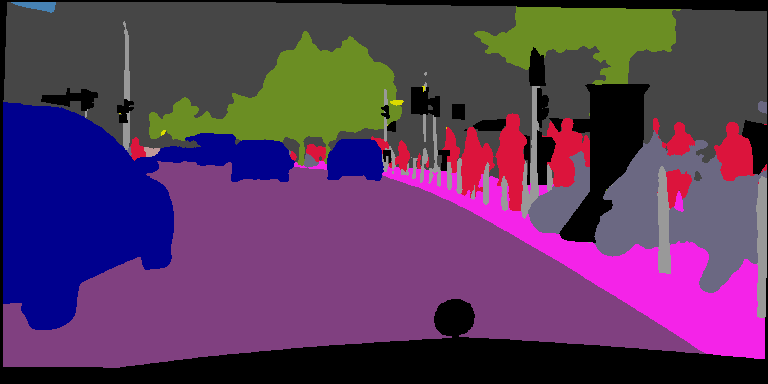} & \fbox{\includegraphics[width=\linewidth]{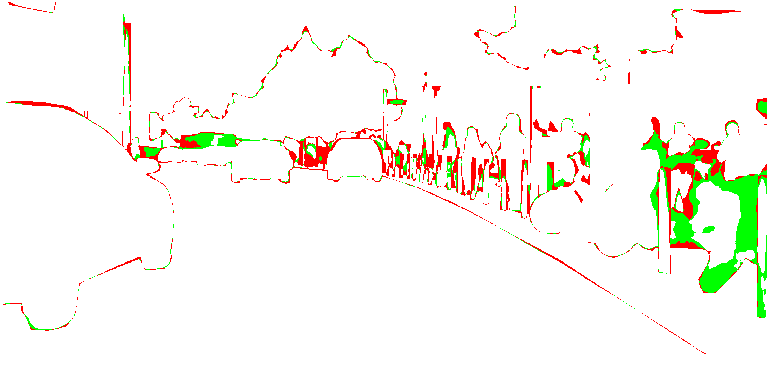}} \\
\rotatebox{90}{(b) Cityscapes} & \includegraphics[width=\linewidth]{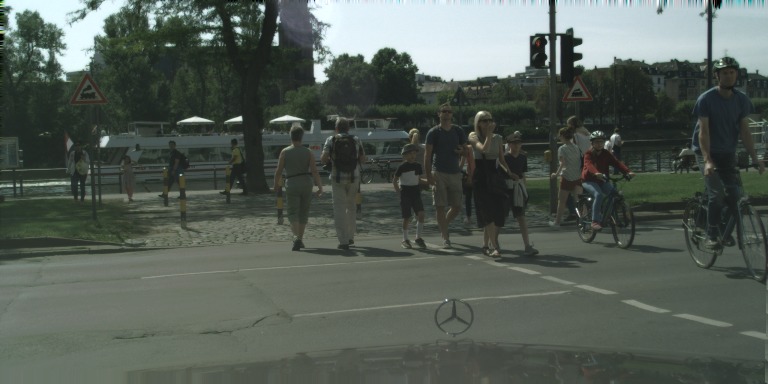} & \includegraphics[width=\linewidth]{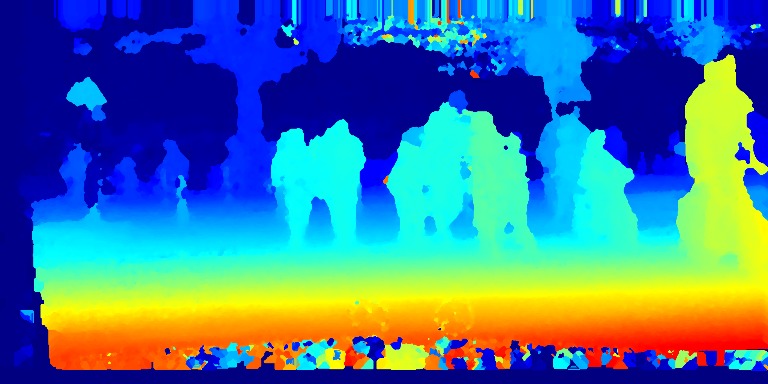} & \includegraphics[width=\linewidth]{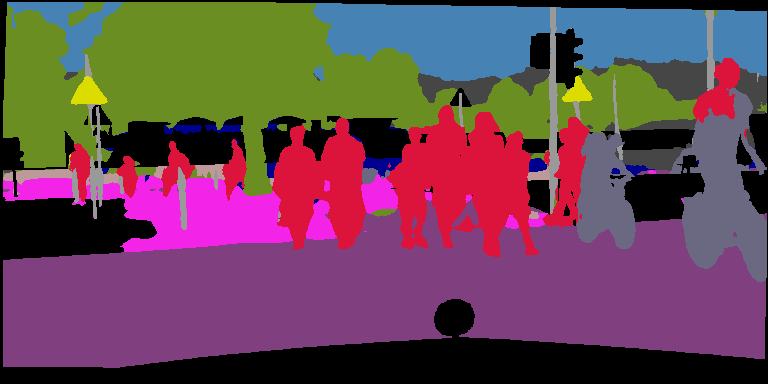} & \includegraphics[width=\linewidth]{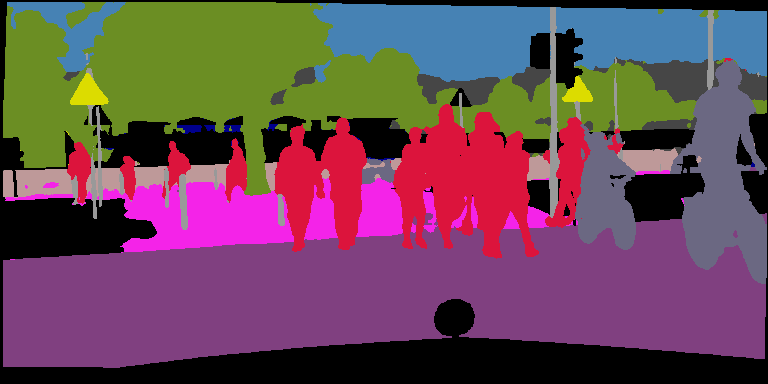} & \fbox{\includegraphics[width=\linewidth]{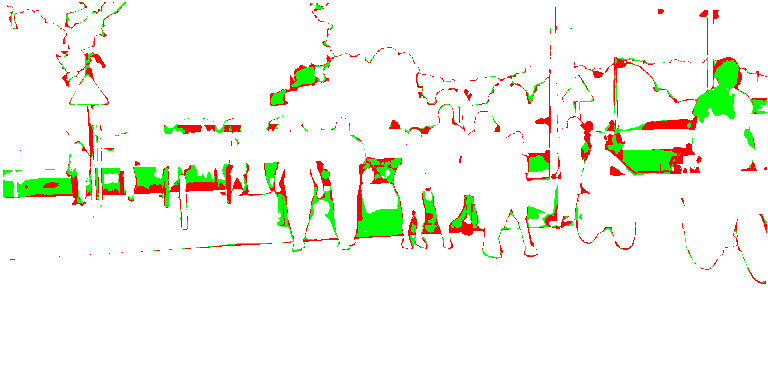}} \\
\rotatebox{90}{(c) Synthia} & \includegraphics[width=\linewidth]{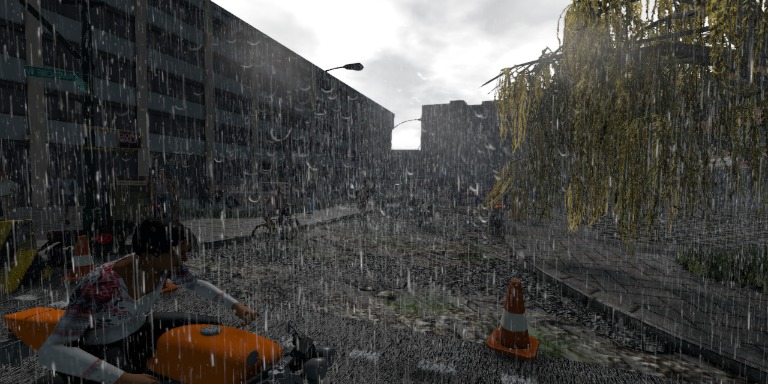} & \includegraphics[width=\linewidth]{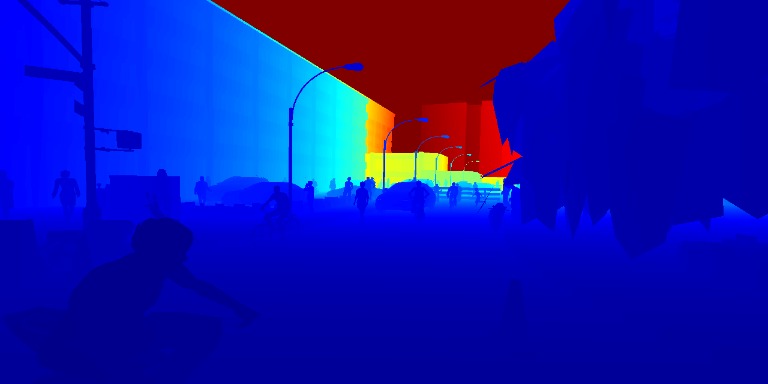} & \includegraphics[width=\linewidth]{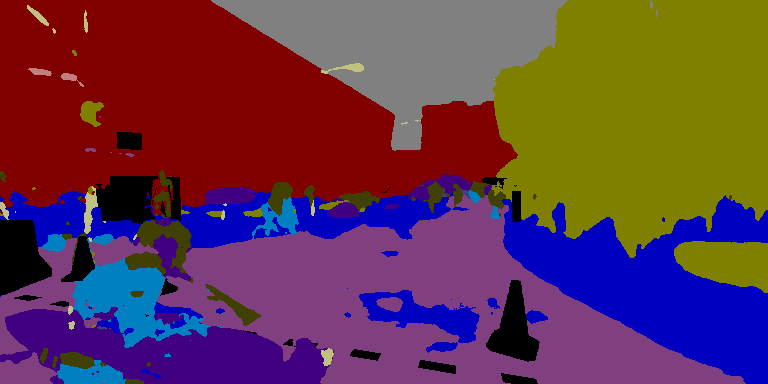} & \includegraphics[width=\linewidth]{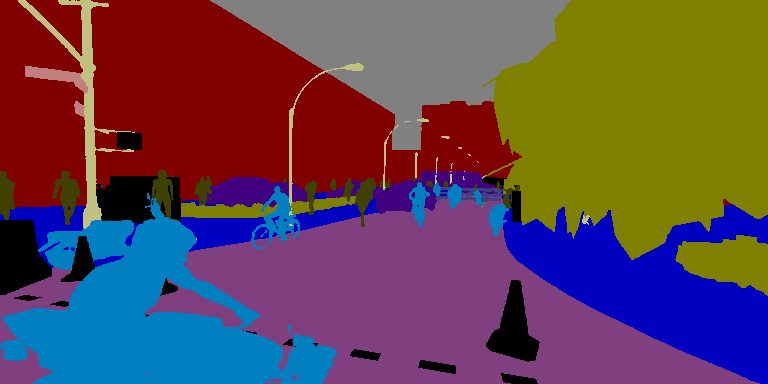} & \fbox{\includegraphics[width=\linewidth]{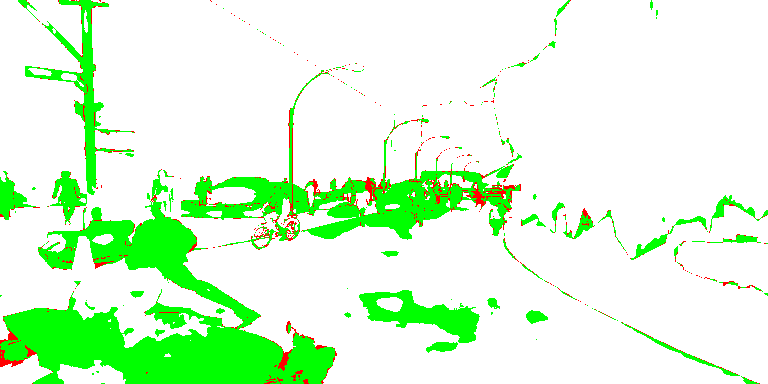}} \\
\rotatebox{90}{(d) Synthia} & \includegraphics[width=\linewidth]{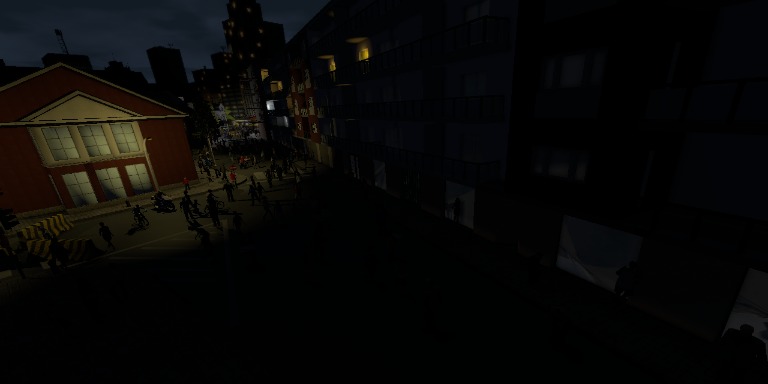} & \includegraphics[width=\linewidth]{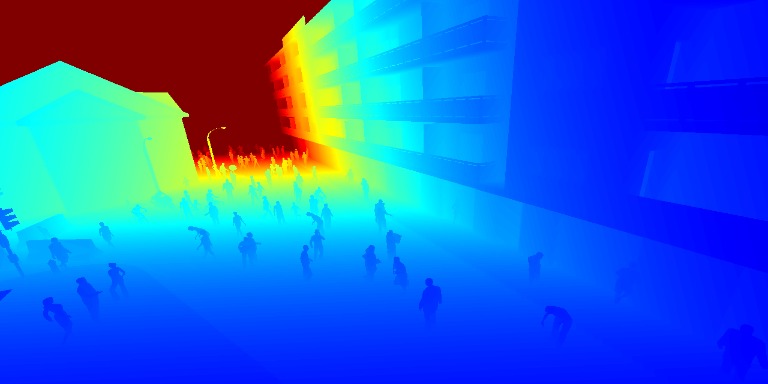} & \includegraphics[width=\linewidth]{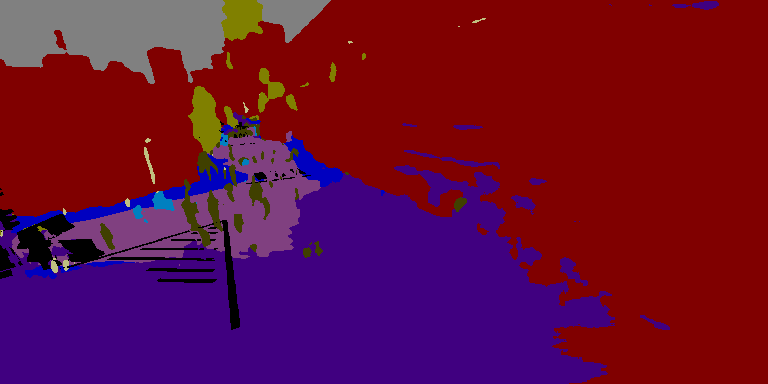} & \includegraphics[width=\linewidth]{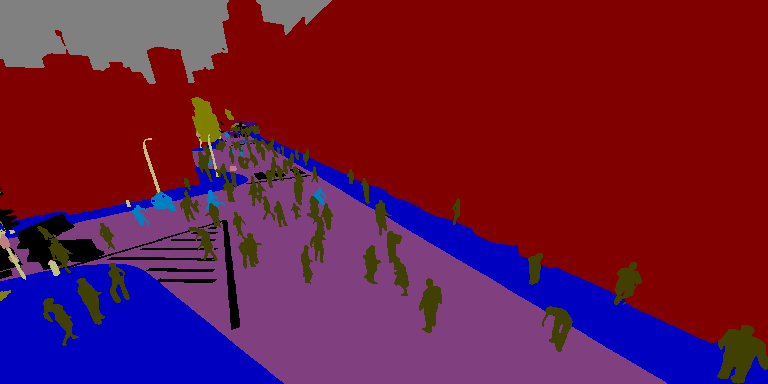} & \fbox{\includegraphics[width=\linewidth]{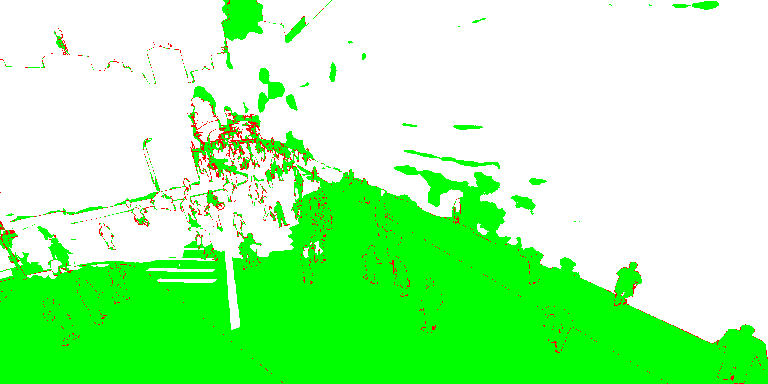}} \\
\rotatebox{90}{(e) SUN} & \includegraphics[width=\linewidth]{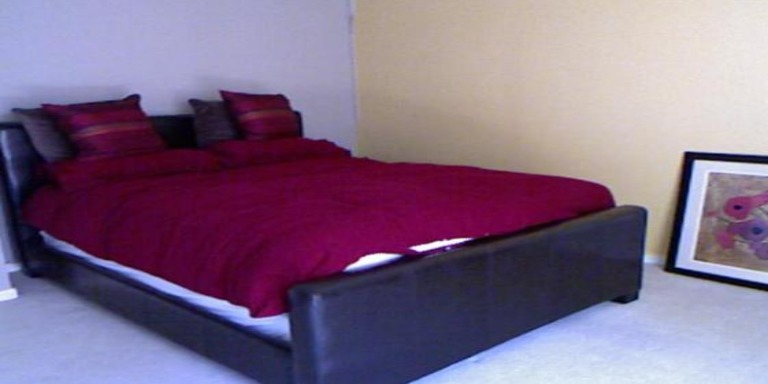} & \includegraphics[width=\linewidth]{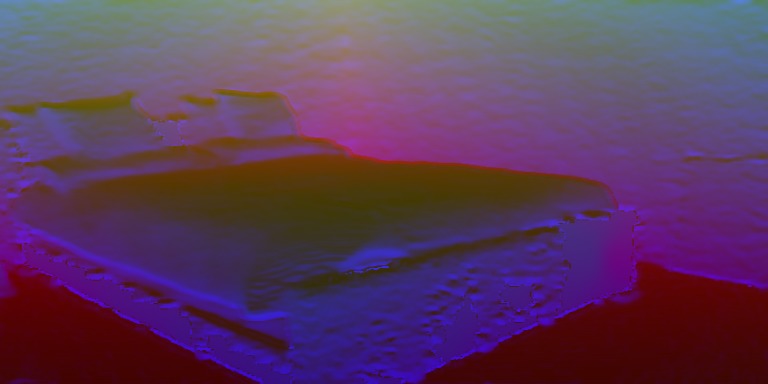} & \includegraphics[width=\linewidth]{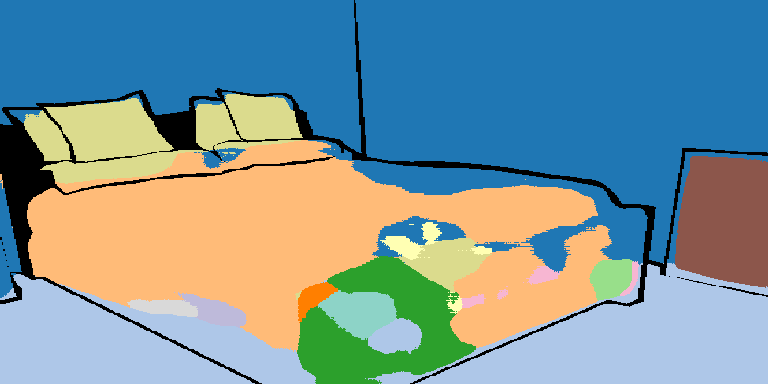} & \includegraphics[width=\linewidth]{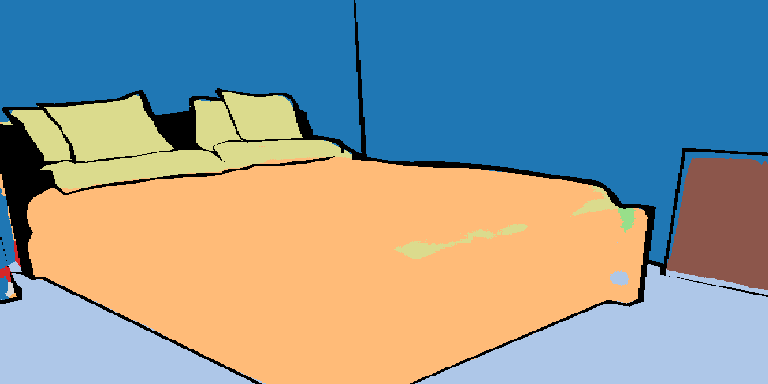} & \fbox{\includegraphics[width=\linewidth]{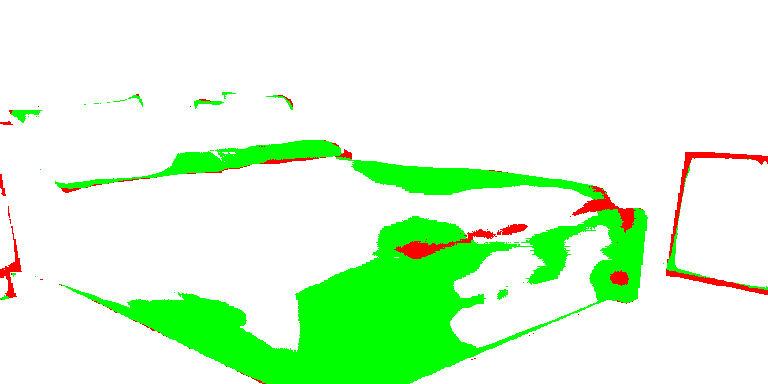}} \\
\rotatebox{90}{(f) SUN} & \includegraphics[width=\linewidth]{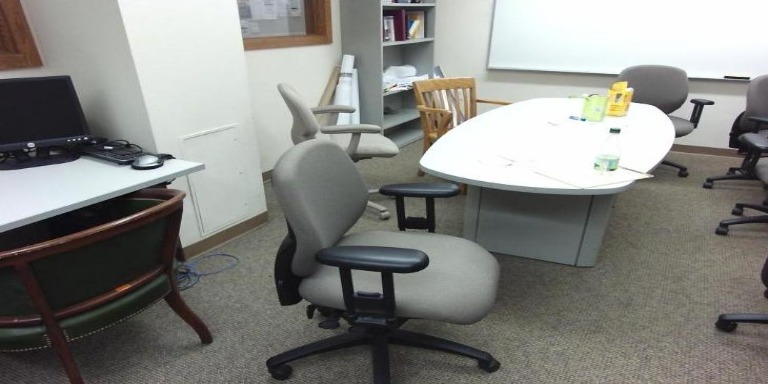} & \includegraphics[width=\linewidth]{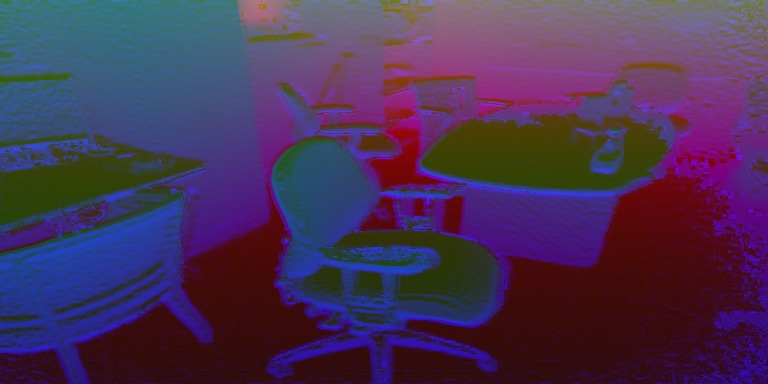} & \includegraphics[width=\linewidth]{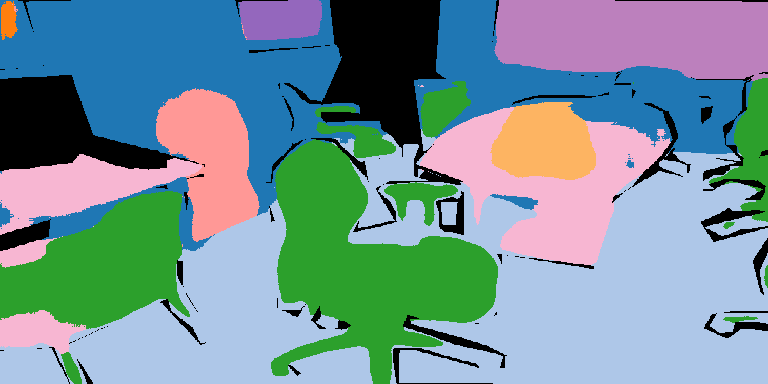} & \includegraphics[width=\linewidth]{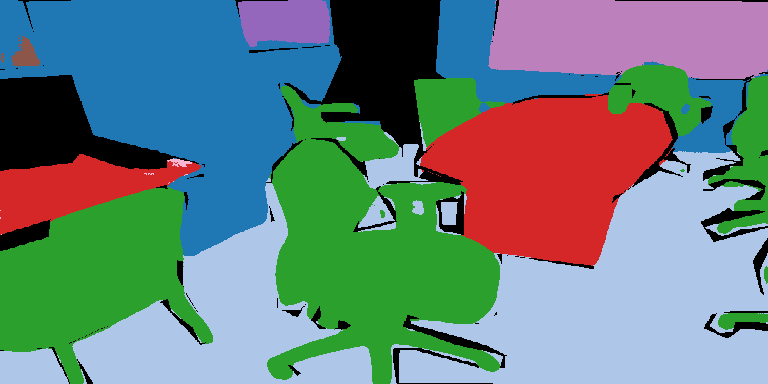} & \fbox{\includegraphics[width=\linewidth]{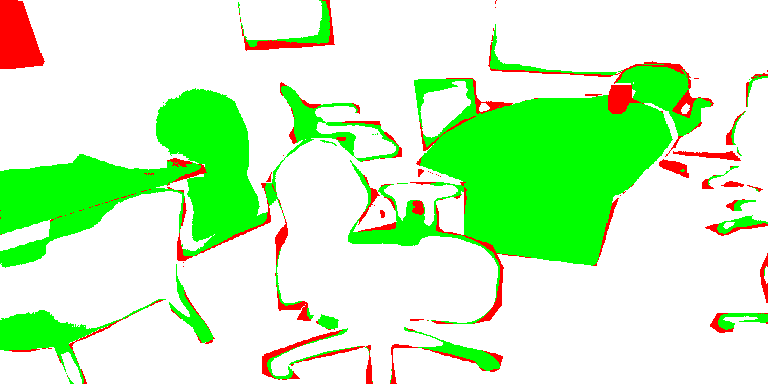}} \\
\rotatebox{90}{(g) SacanNet} & \includegraphics[width=\linewidth]{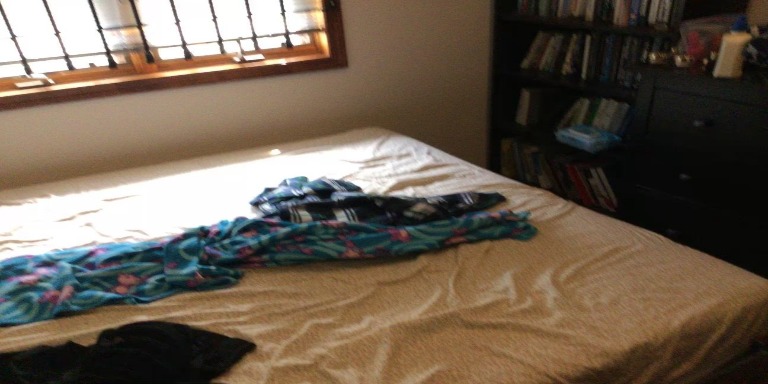} & \includegraphics[width=\linewidth]{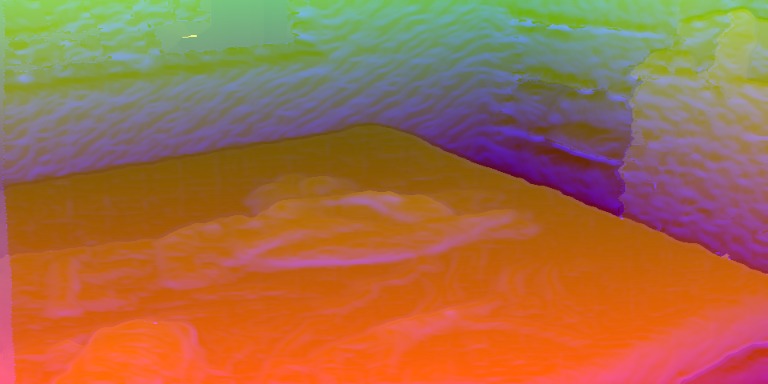} & \includegraphics[width=\linewidth]{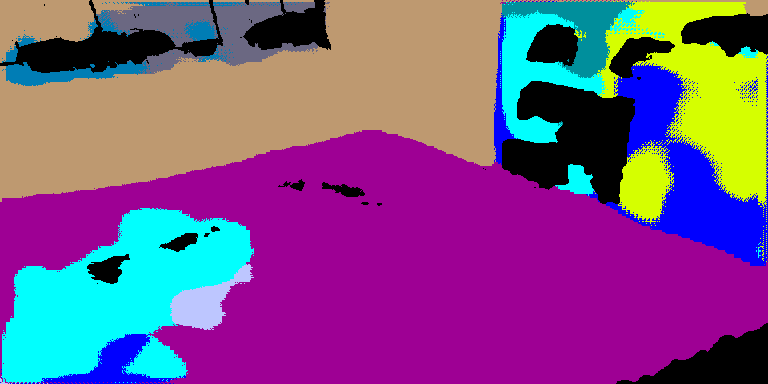} & \includegraphics[width=\linewidth]{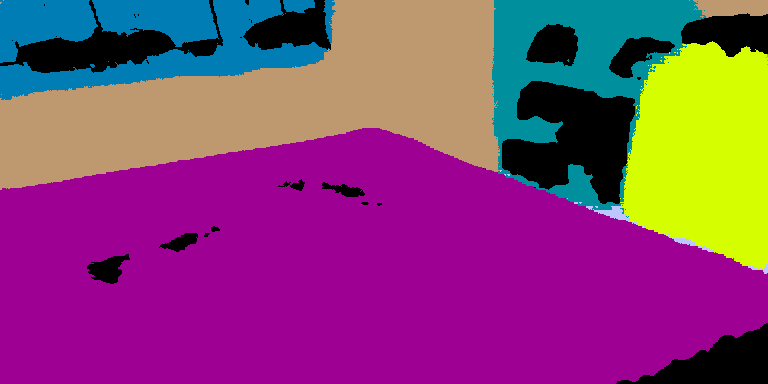} & \fbox{\includegraphics[width=\linewidth]{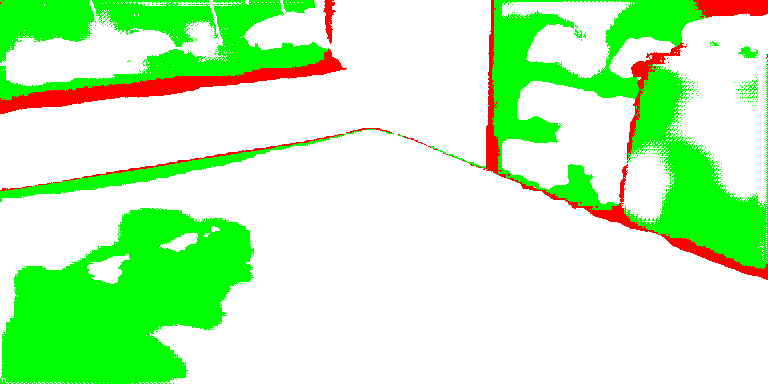}} \\
\rotatebox{90}{(h) SacanNet} & \includegraphics[width=\linewidth]{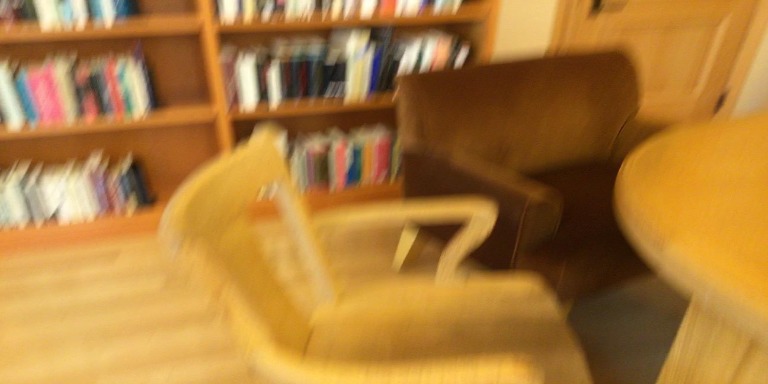} & \includegraphics[width=\linewidth]{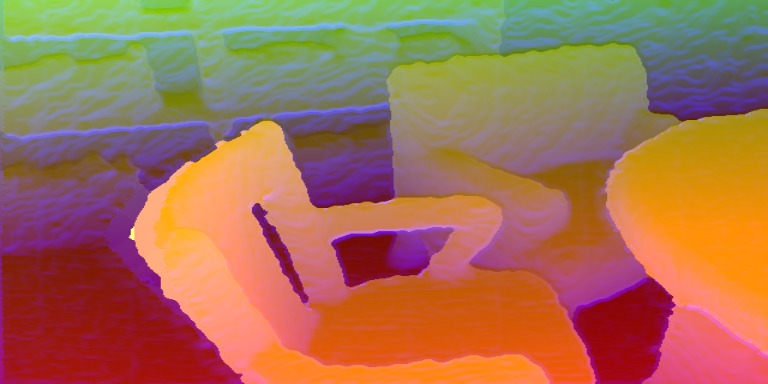} & \includegraphics[width=\linewidth]{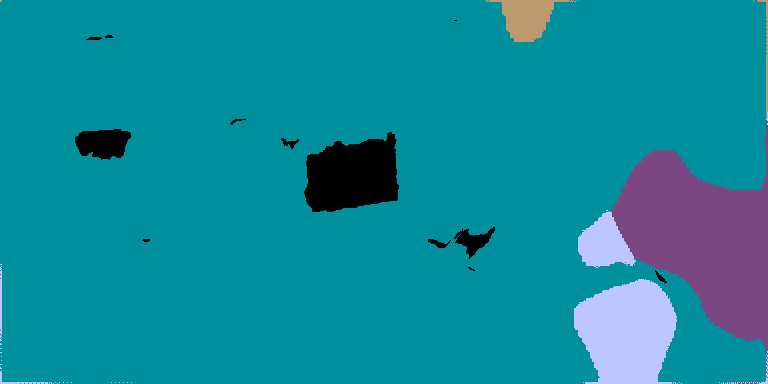} & \includegraphics[width=\linewidth]{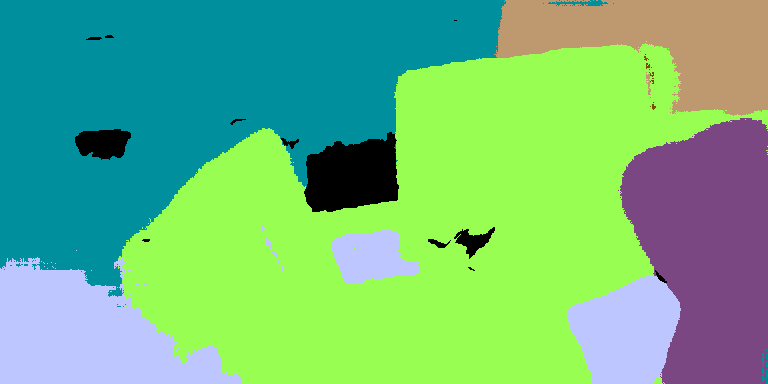} & \fbox{\includegraphics[width=\linewidth]{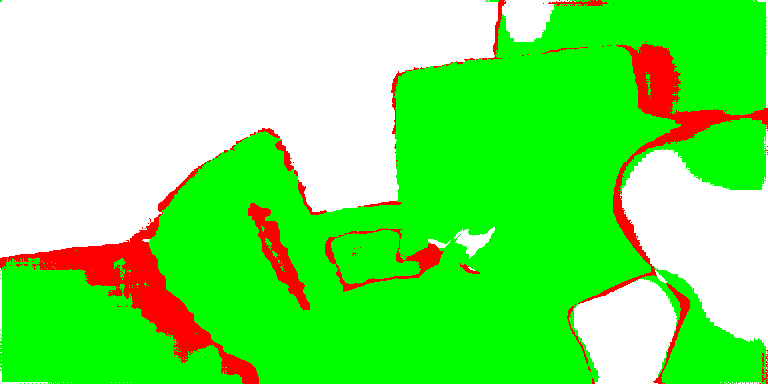}} \\
\rotatebox{90}{(i) Forest} & \includegraphics[width=\linewidth]{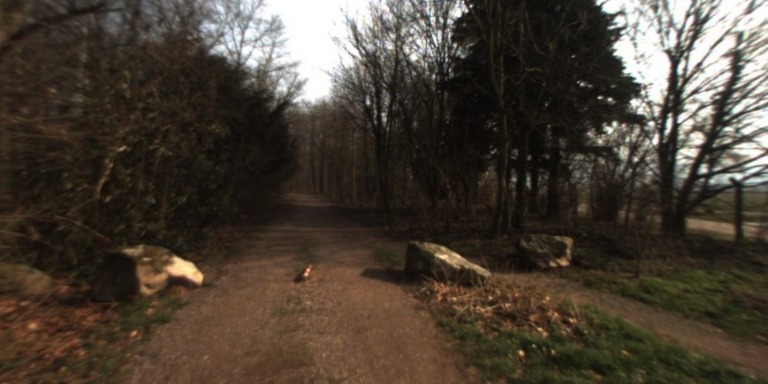} & \includegraphics[width=\linewidth]{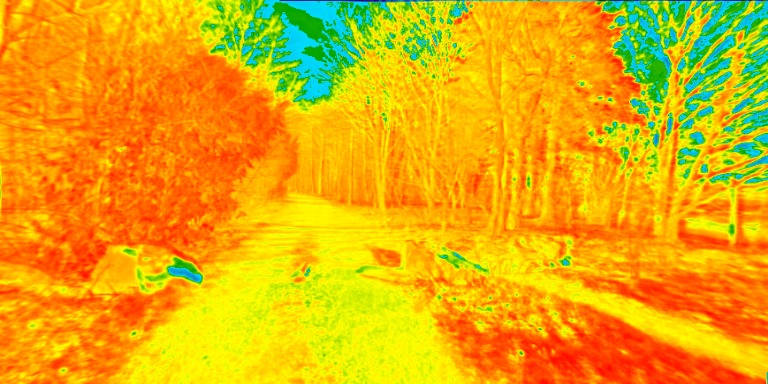} & \includegraphics[width=\linewidth]{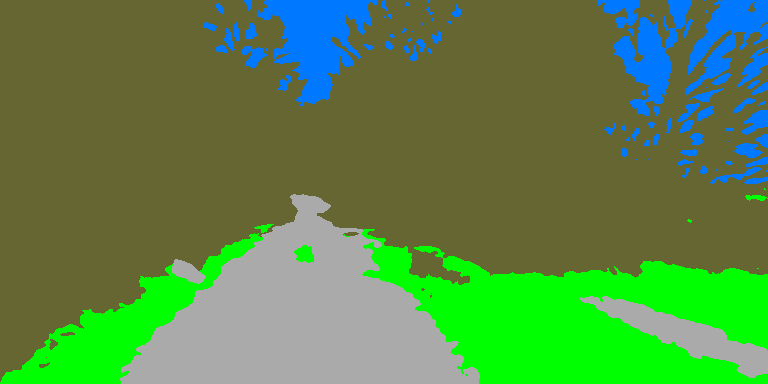} & \includegraphics[width=\linewidth]{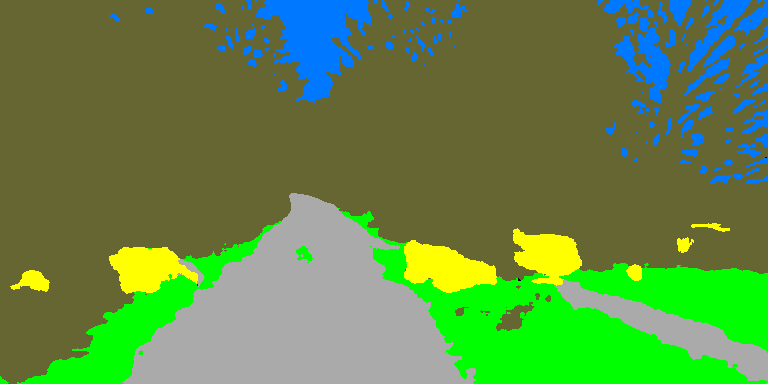} & \fbox{\includegraphics[width=\linewidth]{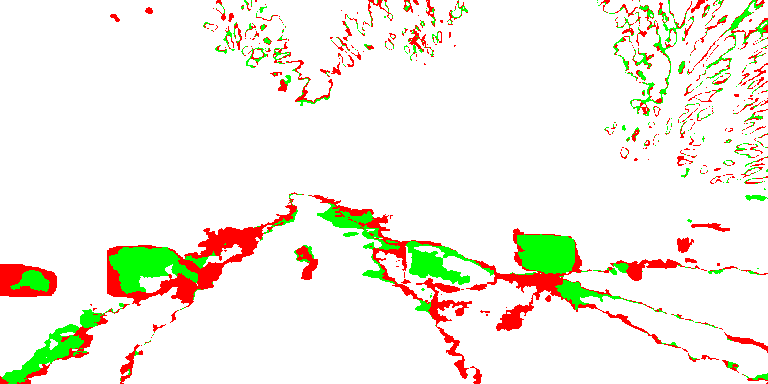}} \\
\rotatebox{90}{(j) Forest} & \includegraphics[width=\linewidth]{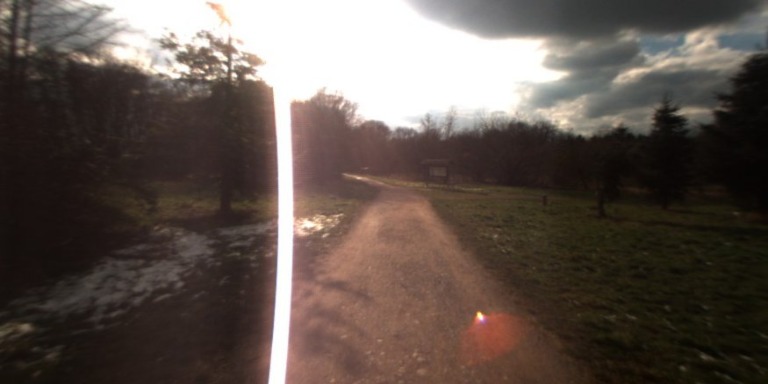} & \includegraphics[width=\linewidth]{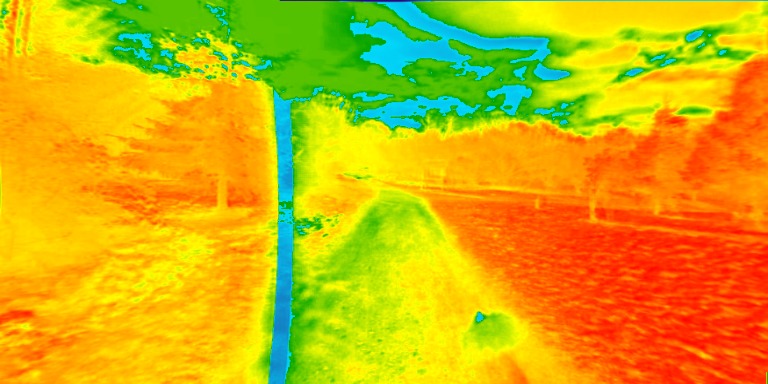} & \includegraphics[width=\linewidth]{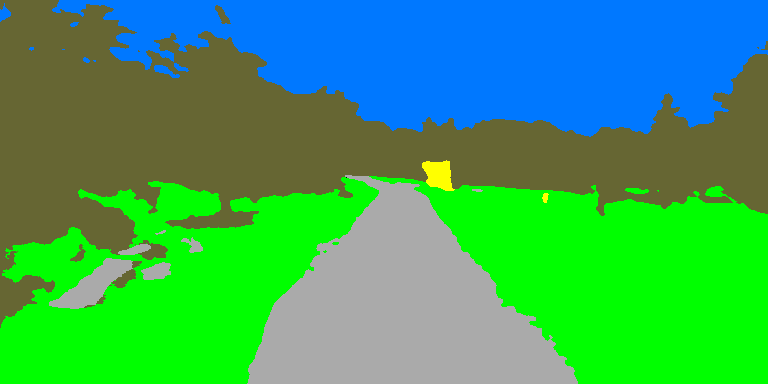} & \includegraphics[width=\linewidth]{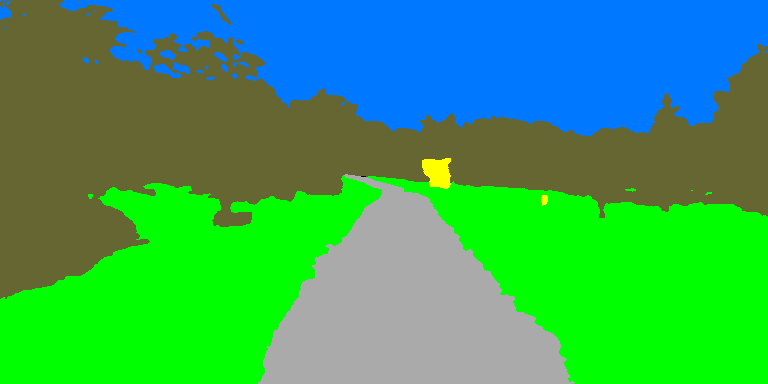} & \fbox{\includegraphics[width=\linewidth]{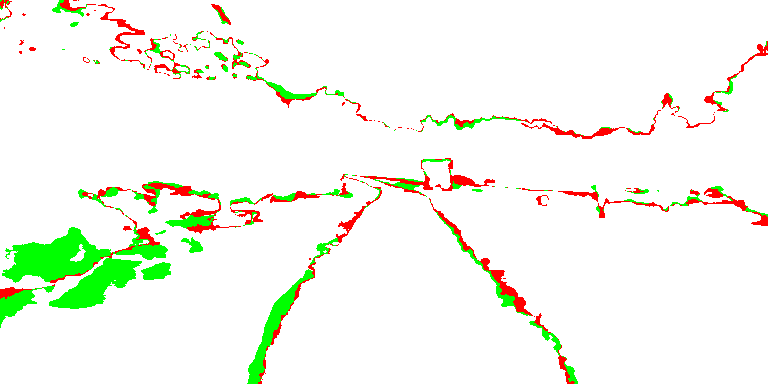}} \\
\midrule[0.05cm]
\rotatebox{90}{(k) Cityscapes} & \includegraphics[width=\linewidth]{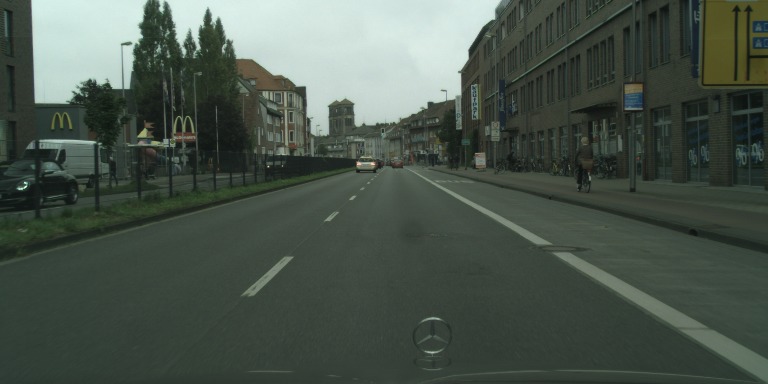} & \includegraphics[width=\linewidth]{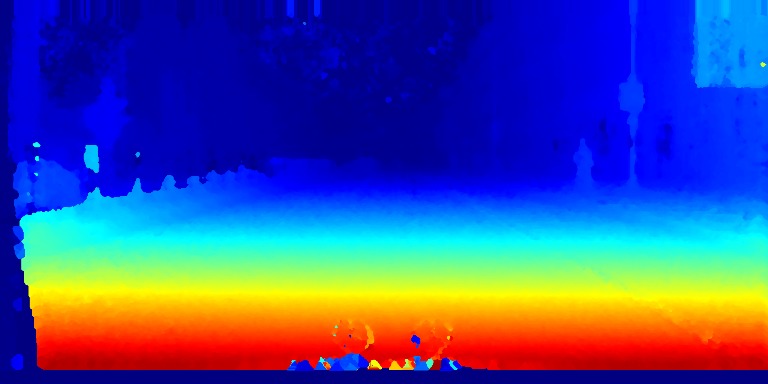} & \includegraphics[width=\linewidth]{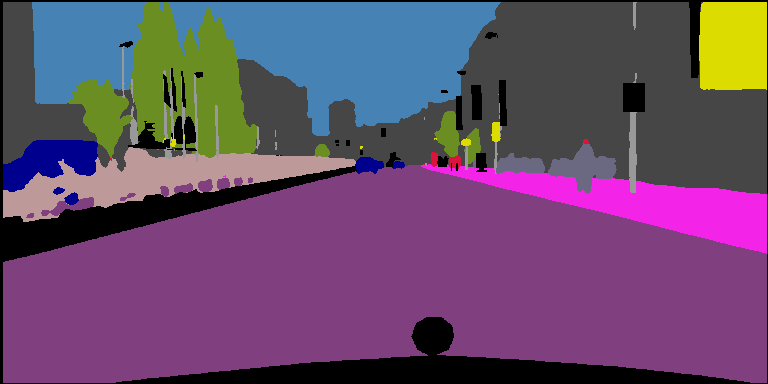} & \includegraphics[width=\linewidth]{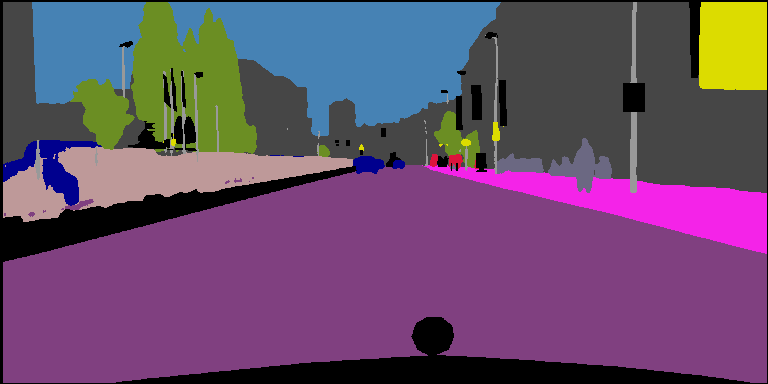} & \fbox{\includegraphics[width=\linewidth]{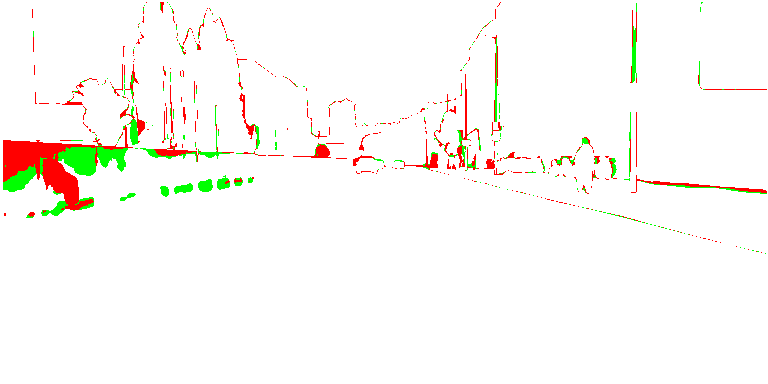}} \\
\rotatebox{90}{(l) SUN} & \includegraphics[width=\linewidth]{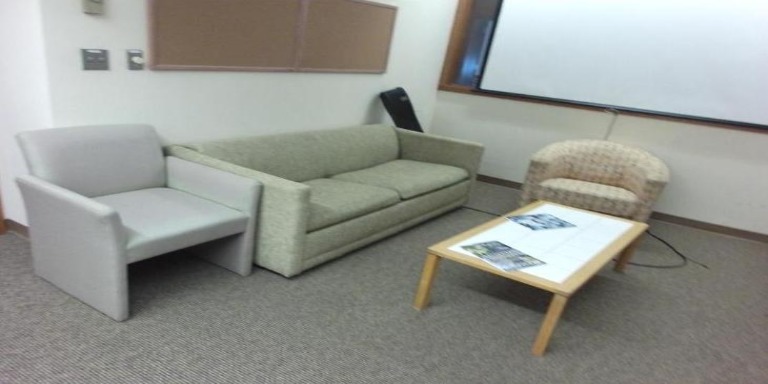} & \includegraphics[width=\linewidth]{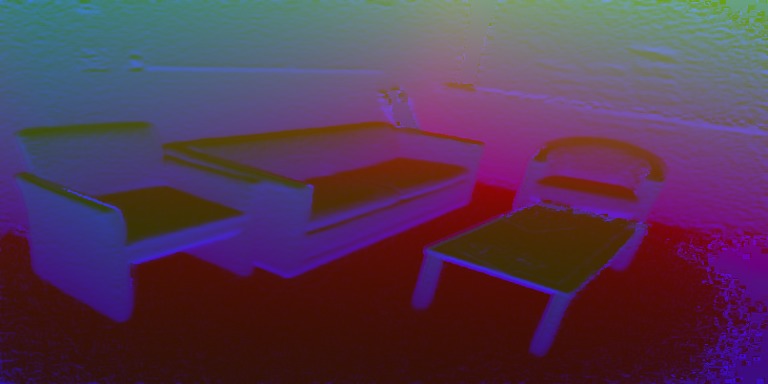} & \includegraphics[width=\linewidth]{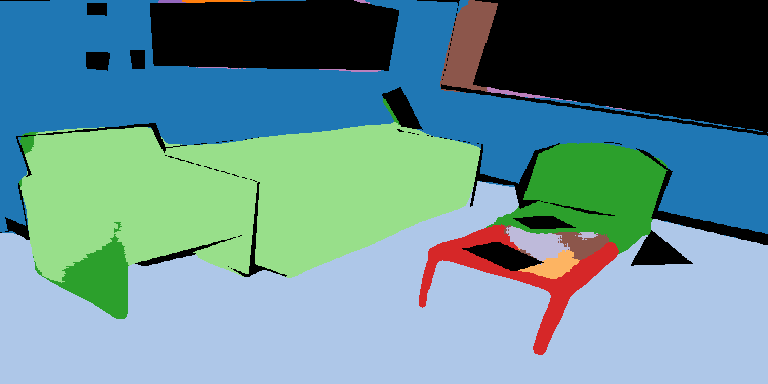} & \includegraphics[width=\linewidth]{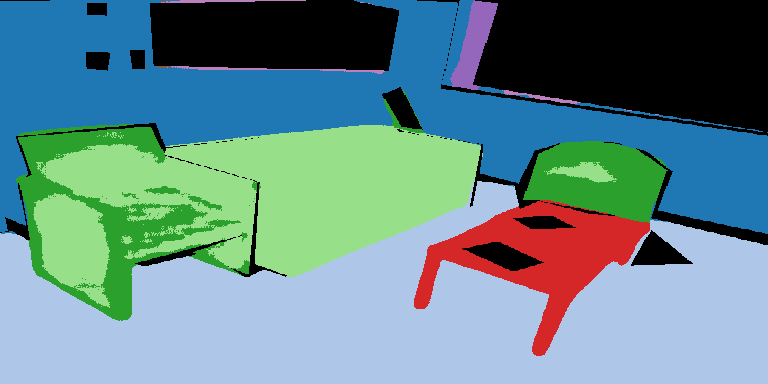} & \fbox{\includegraphics[width=\linewidth]{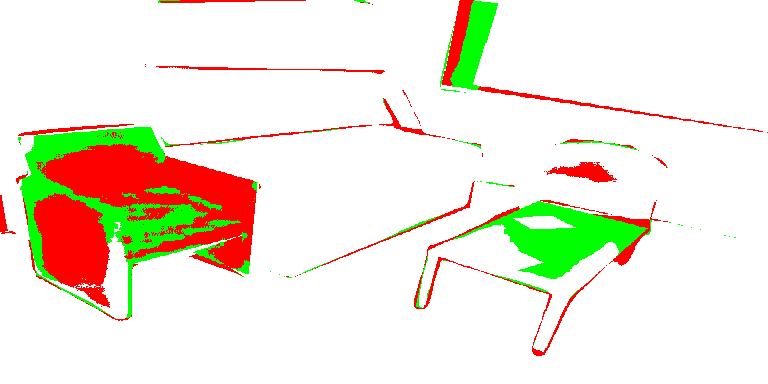}} \\
\end{tabular}}
\caption{Qualitative multimodal fusion results in comparison to the output of the unimodal visual RGB model on each of the five datasets that we benchmark on. The last two rows show failure modes. In addition to the segmentation output, we also show the improvement/error map which denotes the misclassified pixels in red and the pixels that are misclassified by the best performing state-of-the-art model but correctly predicted by AdapNet++ model in green. The legend for the segmented labels correspond to the colors shown in the benchmarking tables in \secref{sec:benchm}.}
\label{fig:qualitativeFusion}
\end{figure*}

\figref{fig:qualitativeFusion} illustrates the visualized comparisons of multimodal semantic segmentation for each of the five benchmark datasets. We compare with the output of unimodal AdapNet++ architecture and show the improvement\textbackslash error map which denotes the improvement over the unimodal AdapNet++ output in green and the misclassifications in red. \figref{fig:qualitativeFusion}(a) and (b) show interesting examples from the Cityscapes dataset. In both these examples, we can see a significant improvement in the segmentation of \textit{cyclists}. As cyclists constitute to a person riding a bike, often models assign a part of the pixels on the person riding a bike as the person class, instead of the \textit{cyclist} class. 

Another common scenario is when there is a person standing a few meters behind a parked bike, the model misclassifies the \textit{person} as a \textit{cyclist} but since he is not on the bike, the right classification would be the person category. In these examples, we can see that by leveraging the features from the depth modality our network makes accurate predictions in these situations. In \figref{fig:qualitativeFusion}(a), we can also see that parts of the car at several meters away is not completely segmented in the unimodal output but it is accurately captured in the multimodal output. Furthermore, in the unimodal output of \figref{fig:qualitativeFusion}(b), we see parts of the \textit{sidewalk} behind the people is misclassified as \textit{road} and parts of the \textit{fence} that is several meters away is misclassified as a \textit{sidewalk}. As the distinction between these object categories can clearly be seen in the depth images, our fusion model accurately identifies these boundaries.

\figref{fig:qualitativeFusion}(c) and (d) show examples on the Synthia dataset. Here we show the first scene during rainfall and the second scene during night-time. In the unimodal output of first scene, we can see significant misclassifications in all the object categories, except \textit{building} and \textit{vegetation} that are substantially large in size. Whereas, the multimodal fusion model is able to leverage the depth features to identify the objects in the scene. In \figref{fig:qualitativeFusion}(d), even for us humans it is impossible to see the \textit{people} on the road due to the darkness in the scene. As predicted, the unimodal visual RGB model misclassifies the entire road with people as a \textit{car}, which could lead to disastrous situations if it occurred in the real-world. The multimodal model is able to accurately predict the scene with almost no error in the predictions.

In \figref{fig:qualitativeFusion}(e) and (f), we show examples on the indoor SUN RGB-D dataset. Due to the large number of object categories in this dataset, several inconspicuous classes exist. Leveraging structural properties of objects from the HHA encoded depth can enable better discrimination between them. \figref{fig:qualitativeFusion}(e) shows a scene where the unimodal model misclassifies parts of the wooden \textit{bed} as a \textit{chair} and parts of the \textit{pillow} as the \textit{bed}. We can see that the multimodal output significantly improves upon the unimodal counterpart. \figref{fig:qualitativeFusion}(f) shows a complex indoor scene with substantial clutter. The unimodal model misclassifies the \textit{table} as a \textit{desk} and a hatch in the wall is misclassified as a \textit{door}. Moreover, partly occluded \textit{chairs} are not entirely segmented in the unimodal output. The HHA encoded depth shows well defined structure of these objects, which enables the fusion approach to precisely segment the scene. Note that the \textit{window} in the top left corner of \figref{fig:qualitativeFusion}(f) is mislabeled as a \textit{desk} in the groundtruth.

\figref{fig:qualitativeFusion}(g) and (h) show examples of indoor scenes from the ScanNet dataset. In the unimodal output of \figref{fig:qualitativeFusion}(g), overexposure of the image near the windows causes misclassification of parts of the \textit{window} as a \textit{picture} and crumpled bedsheets as well as the \textit{bookshelf} is missclassified as a \textit{desk}. While the multimodal segmentation output does not demonstrate these errors. \figref{fig:qualitativeFusion}(h) shows an image with motion-blur due to camera motion. The motion blur causes a significant percentage of the image to be misclassified as the largest object in the scene and this case as a \textit{bookshelf}. Analyzing the HHA encoded depth map, we can see that it does not contain overexposed sections or motion-blur, rather it strongly emphasizes the structure of the objects in the scene. By leveraging features from the HHA encoded depth stream, our multimodal model is able to accurately predict the object classes in the presence of these perceptual disturbances.

In \figref{fig:qualitativeFusion}(i) and (j), we show results on the unstructured Freiburg Forest dataset. \figref{fig:qualitativeFusion}(i) shows an oversaturated image due to sunlight which causes boulders on the \textit{grass} to be completely absent in the unimodal segmentation output. Oversaturation causes boulders to appear with a similar texture as the \textit{trail} or \textit{vegetation} class. However, the RGB-EVI multimodal model is able to leverage the complementary EVI features to segment these structures. \figref{fig:qualitativeFusion}(j) shows an example scene with glare on the camera optics and snow on the ground. In the unimodal semantic segmentation output, the presence of these disturbances often causes localized misclassifications in the areas where they are present. Whereas, the multimodal semantic segmentation model compensates for these disturbances exploiting the complementary modalities.

The final two rows in \figref{fig:qualitativeFusion} show interesting failure modes where the multimodal fusion model demonstrates incorrect predictions. In \figref{fig:qualitativeFusion}(k), we show an example from the Cityscapes dataset which contains an extremely thin \textit{fence} connected by wires along the median of the road. The thin wires are not captured by the depth modality and it is visually infeasible to detect it from the RGB image. Moreover due its thin structure, the vehicles on the opposite lane are clearly visible. This causes both the unimodal and multimodal models to partially segment the vehicles behind the fence which causes incorrect predictions according to the groundtruth. However, we can see that the multimodal model still captures more of the \textit{fence} structure than the unimodal model. In \figref{fig:qualitativeFusion}(l), we show an example from the SUN RGB-D dataset in which misclassifications are produced due to inconspicuous classes. The scene contains two object categories that have very similar appearance in some scenes, namely, \textit{chair} and \textit{sofa}. The \textit{chair} class is denoted in dark green, while the \textit{sofa} class is denoted in light green. As we see in this scene, a single-person \textit{sofa} is considered to be a \textit{chair} according to the groundtruth, whereas only the longer \textit{sofa} in the middle is considered to be in the \textit{sofa} class. In this scene, the single person \textit{sofa} is adjacent to the longer \textit{sofa} which causes the network to predict the pixels on both of them as the \textit{sofa} class. 

\subsection{Visualizations Across Seasons and Weather Conditions}
\label{sec:seasonsQual}

\begin{figure*}
\centering
\footnotesize
\setlength{\tabcolsep}{0.05cm}
{\renewcommand{\arraystretch}{1}
\begin{tabular}{P{0.25cm}P{3.3cm}P{3.3cm}P{3.3cm}P{3.3cm}P{3.3cm}}
& Input Modality1 & Input Modality2 & Unimodal Output & Multimodal Output & Improvement\textbackslash Error Map \\
\rotatebox{90}{(a) Summer} & \includegraphics[width=\linewidth]{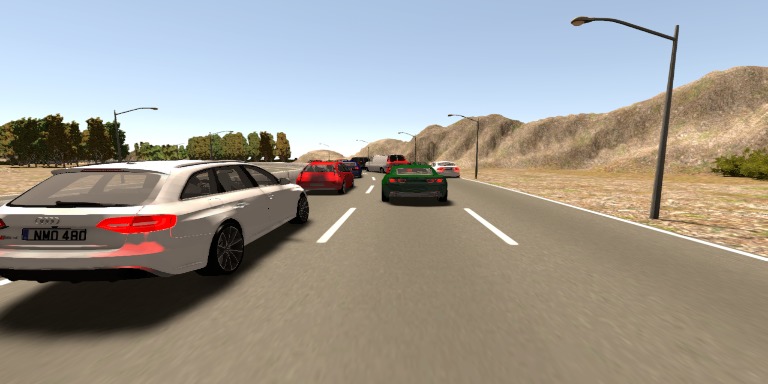} & \includegraphics[width=\linewidth]{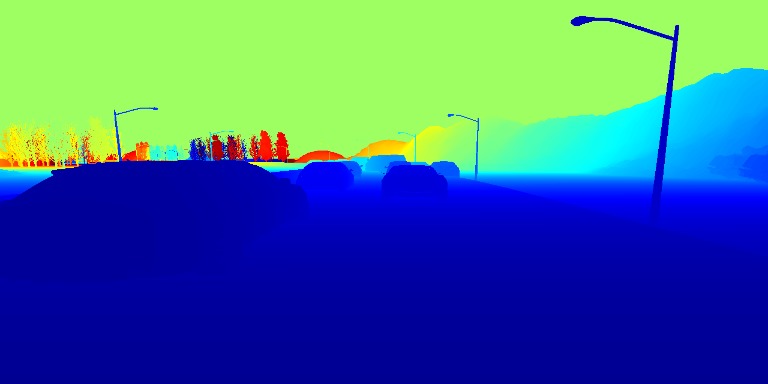} & \includegraphics[width=\linewidth]{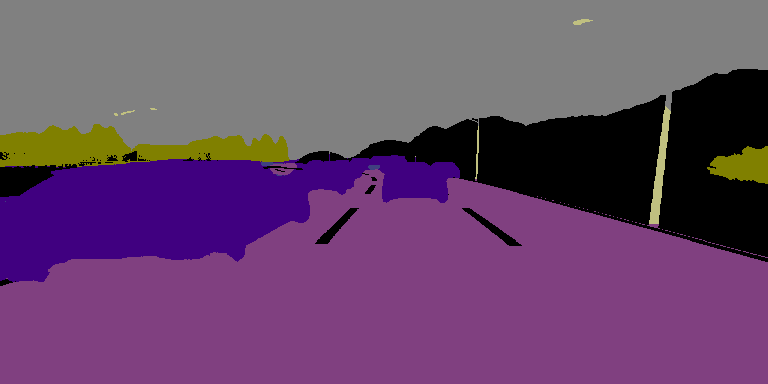} & \includegraphics[width=\linewidth]{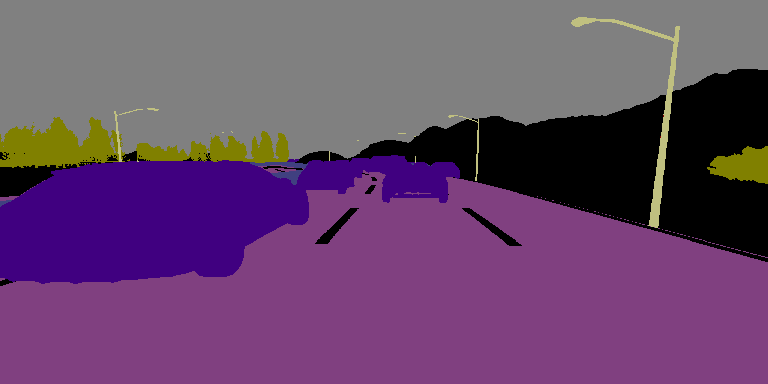} & \fbox{\includegraphics[width=\linewidth]{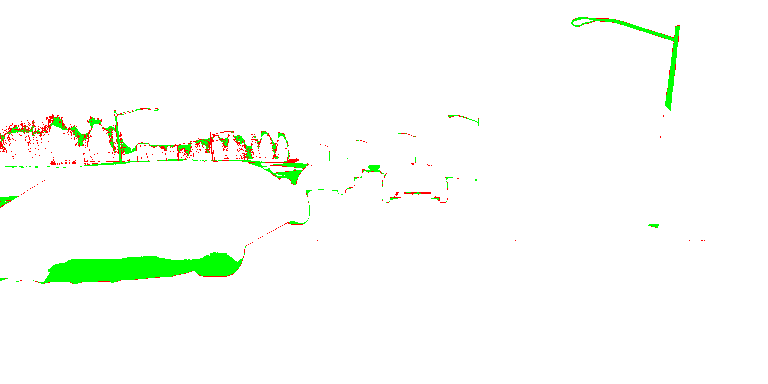}} \\
\rotatebox{90}{(b) Fall} & \includegraphics[width=\linewidth]{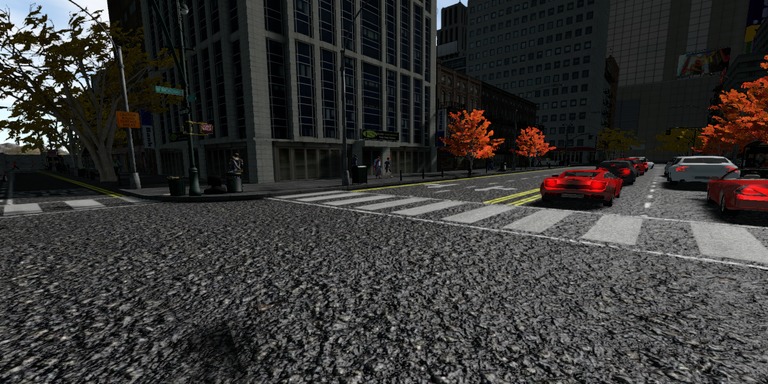} & \includegraphics[width=\linewidth]{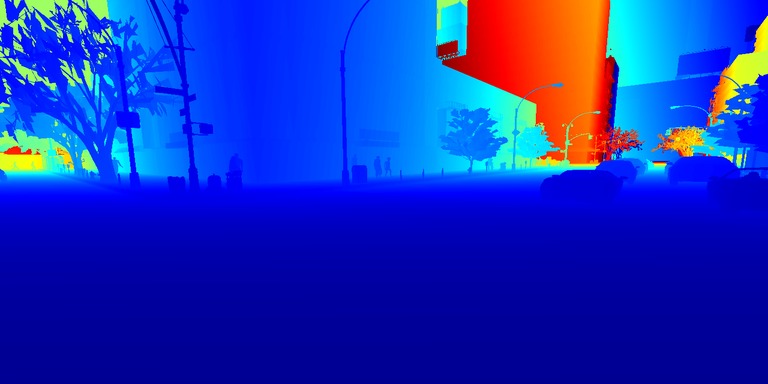} & \includegraphics[width=\linewidth]{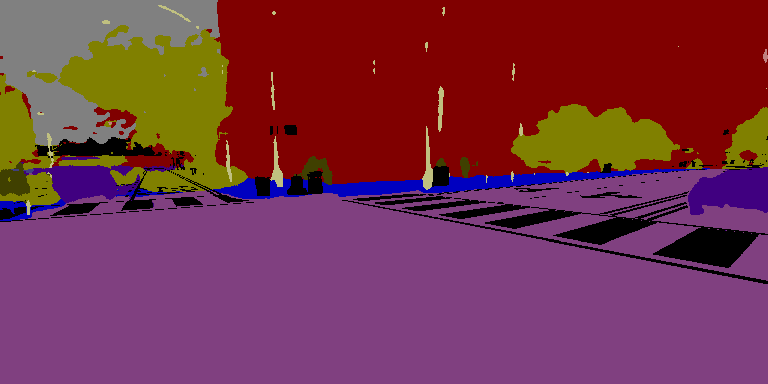} & \includegraphics[width=\linewidth]{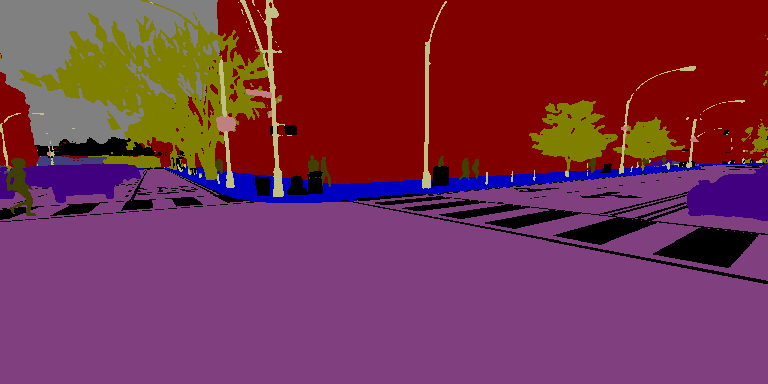} & \fbox{\includegraphics[width=\linewidth]{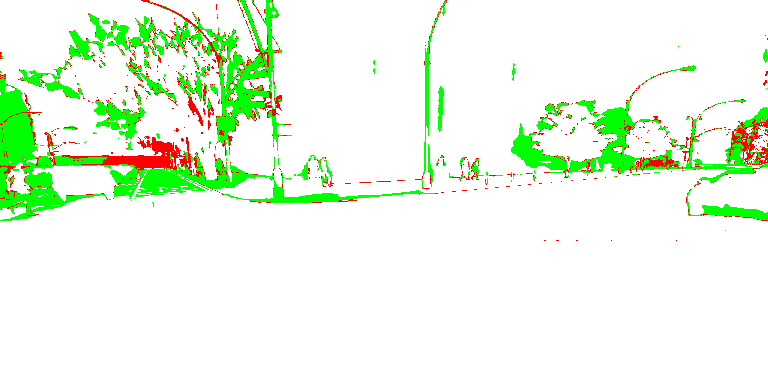}} \\
\rotatebox{90}{(c) Winter} & \includegraphics[width=\linewidth]{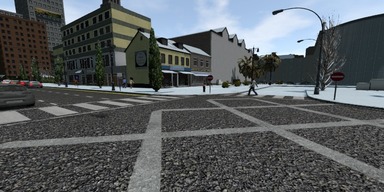} & \includegraphics[width=\linewidth]{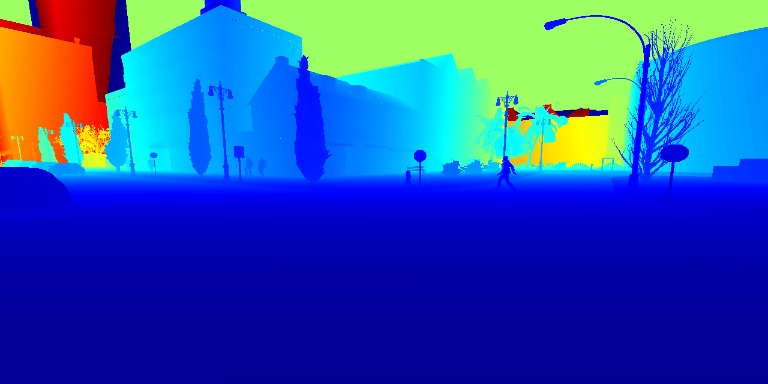} & \includegraphics[width=\linewidth]{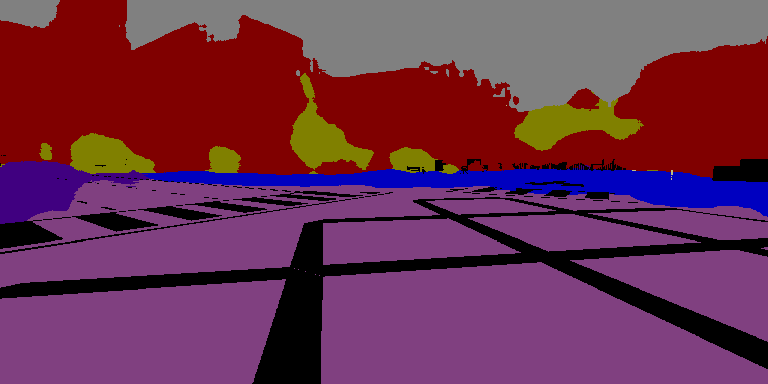} & \includegraphics[width=\linewidth]{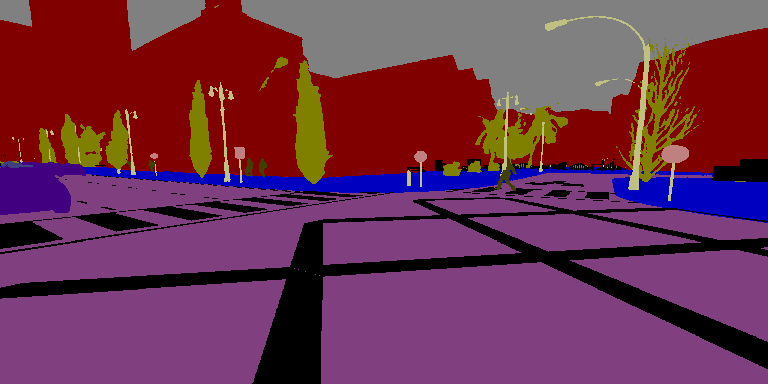} & \fbox{\includegraphics[width=\linewidth]{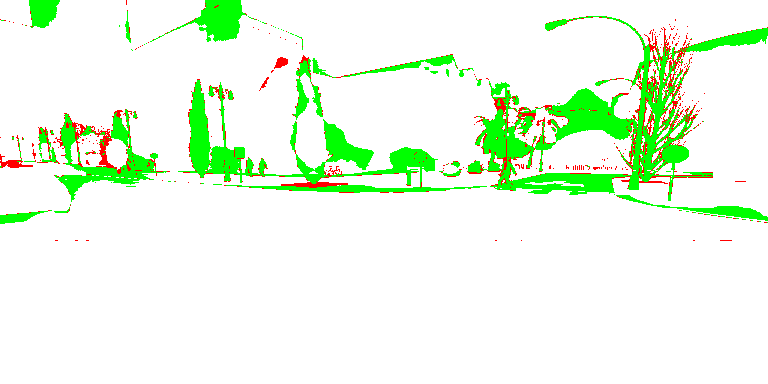}} \\
\rotatebox{90}{(d) Spring} & \includegraphics[width=\linewidth]{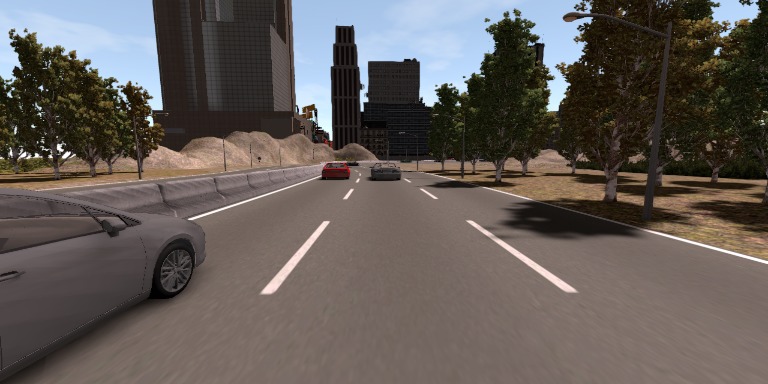} & \includegraphics[width=\linewidth]{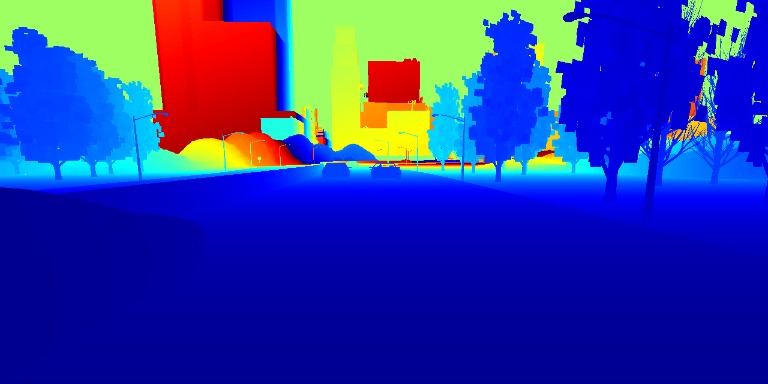} & \includegraphics[width=\linewidth]{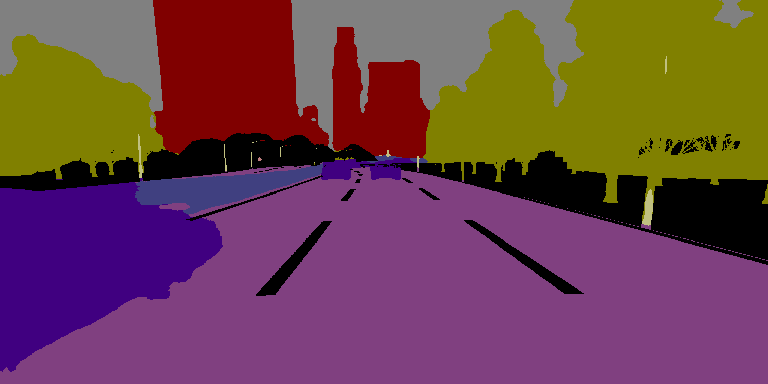} & \includegraphics[width=\linewidth]{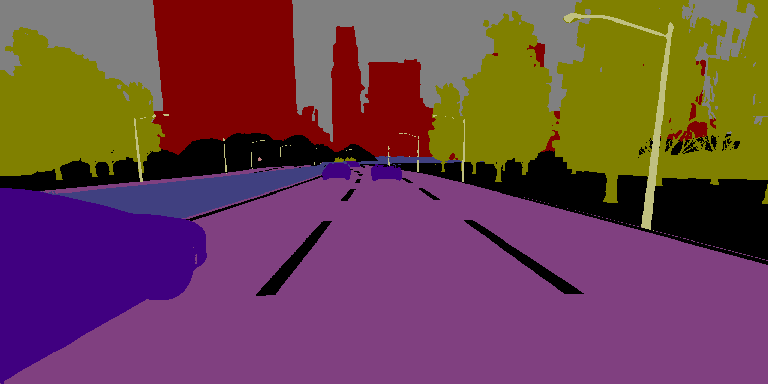} & \fbox{\includegraphics[width=\linewidth]{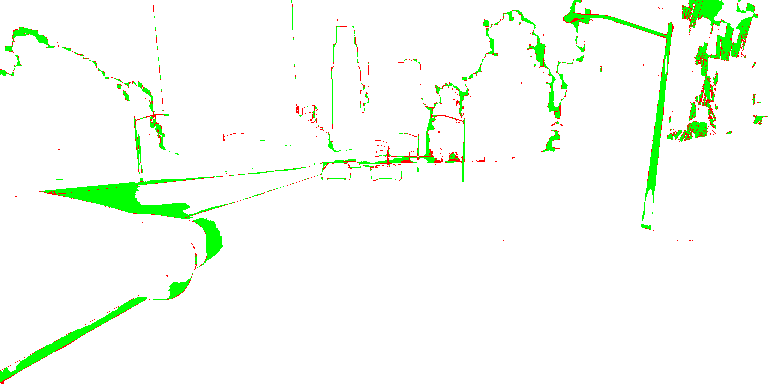}} \\
\rotatebox{90}{(e) Dawn} & \includegraphics[width=\linewidth]{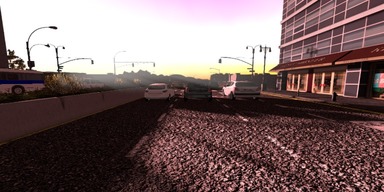} & \includegraphics[width=\linewidth]{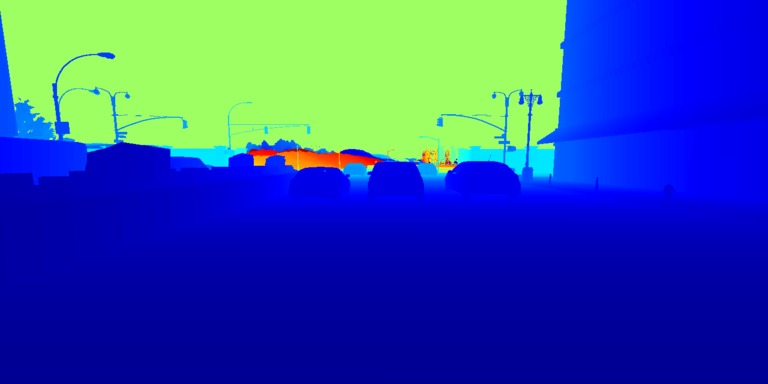} & \includegraphics[width=\linewidth]{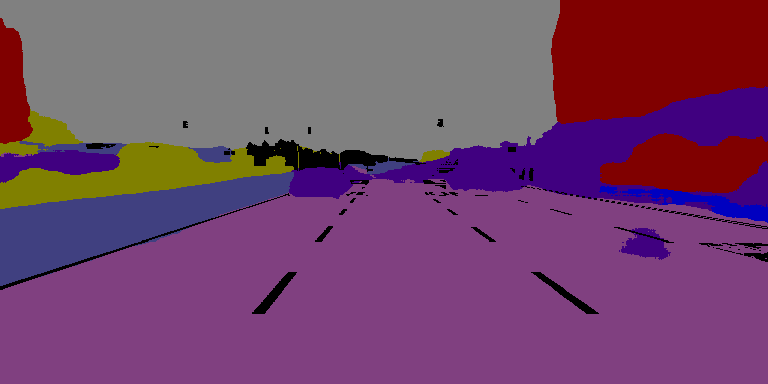} & \includegraphics[width=\linewidth]{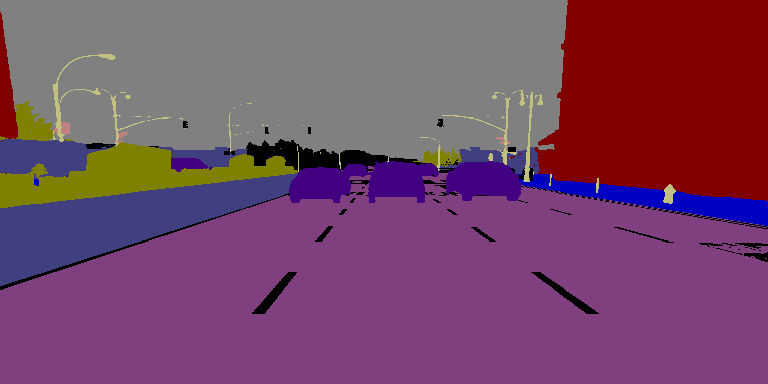} & \fbox{\includegraphics[width=\linewidth]{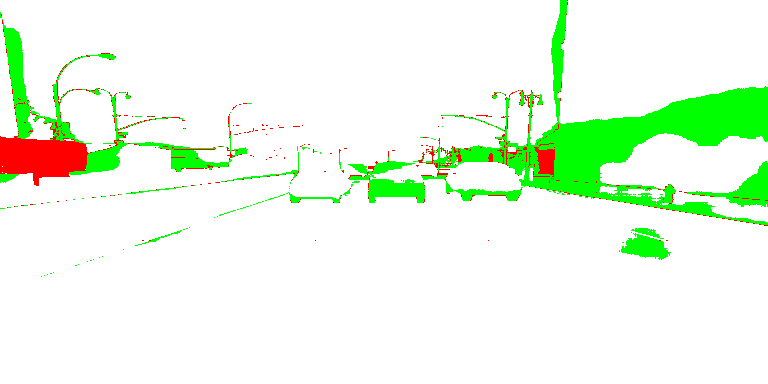}} \\
\rotatebox{90}{(f) Sunset} & \includegraphics[width=\linewidth]{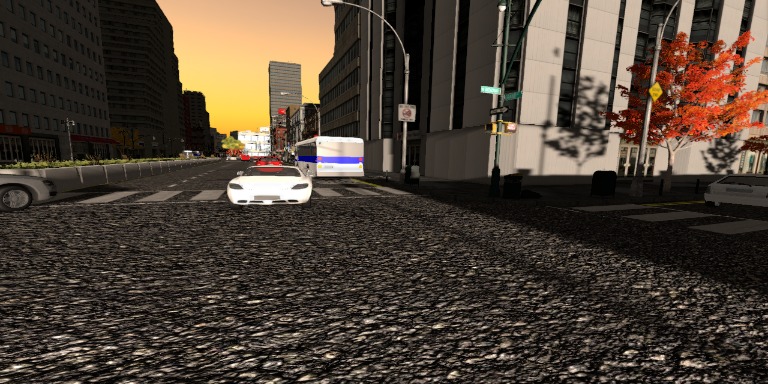} & \includegraphics[width=\linewidth]{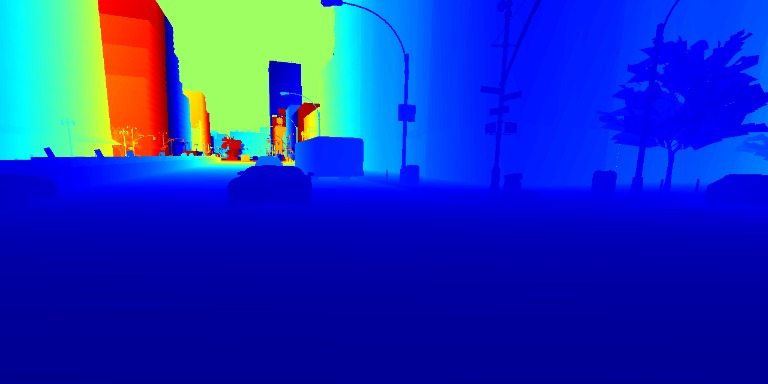} & \includegraphics[width=\linewidth]{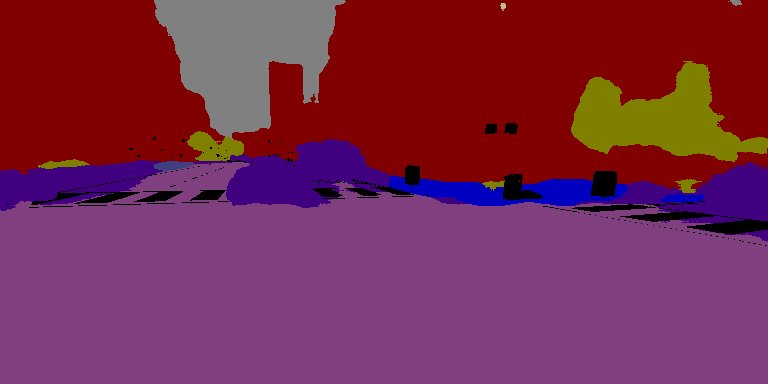} & \includegraphics[width=\linewidth]{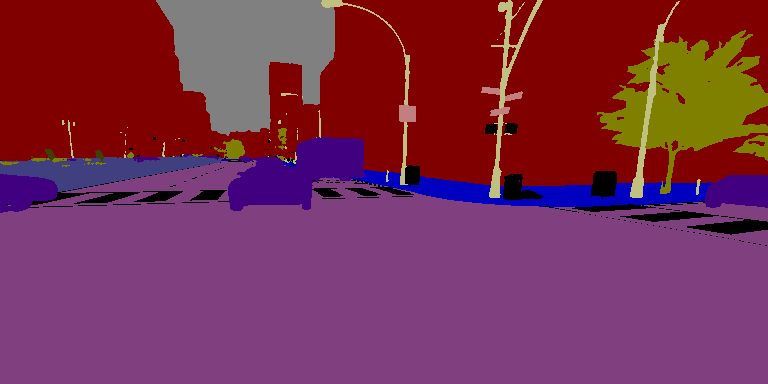} & \fbox{\includegraphics[width=\linewidth]{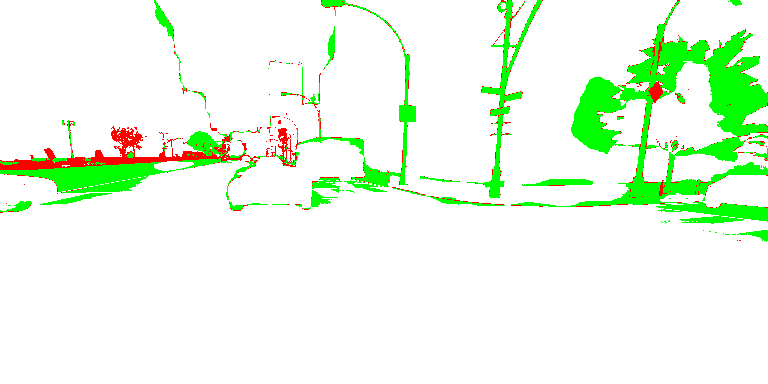}} \\
\rotatebox{90}{(g) Night} & \includegraphics[width=\linewidth]{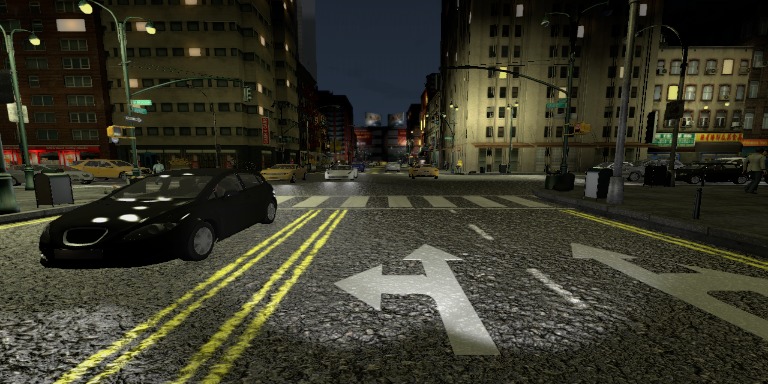} & \includegraphics[width=\linewidth]{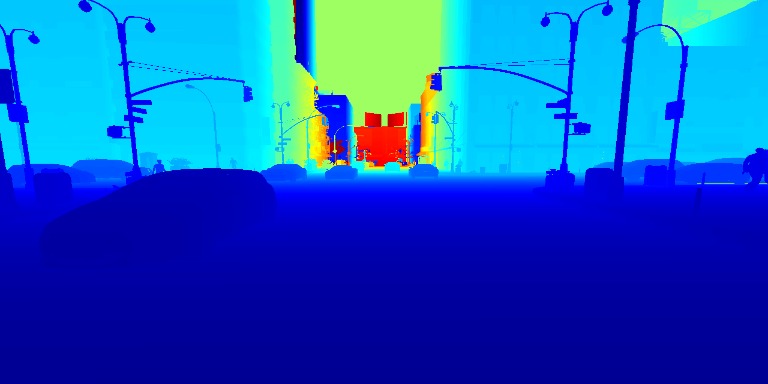} & \includegraphics[width=\linewidth]{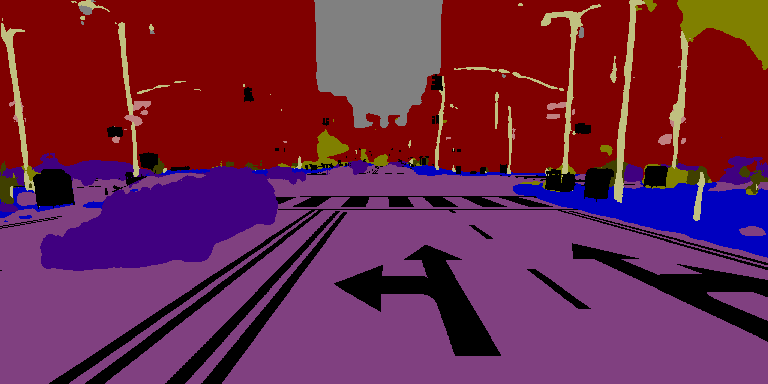} & \includegraphics[width=\linewidth]{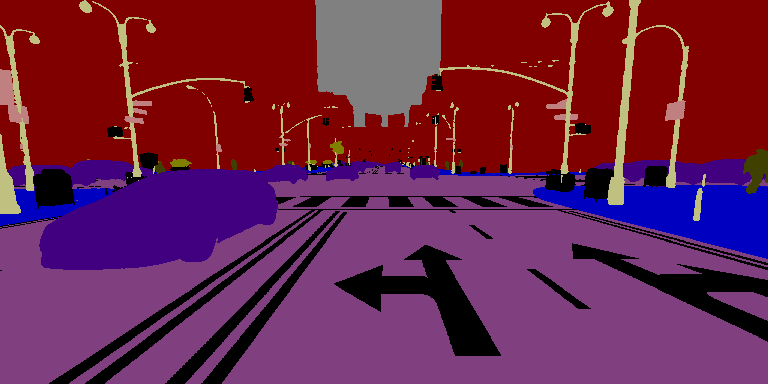} & \fbox{\includegraphics[width=\linewidth]{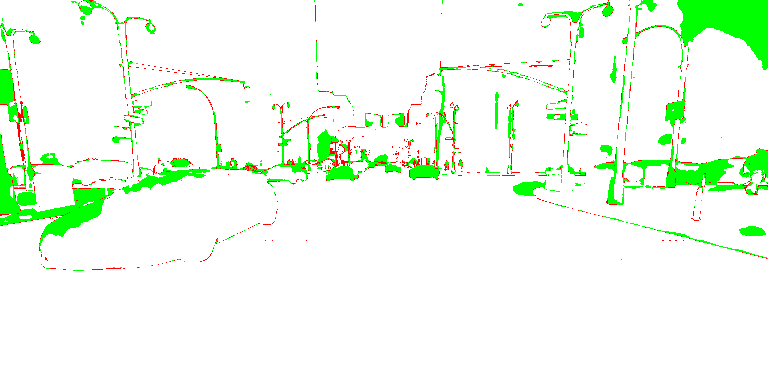}} \\
\rotatebox{90}{(h) Rain} & \includegraphics[width=\linewidth]{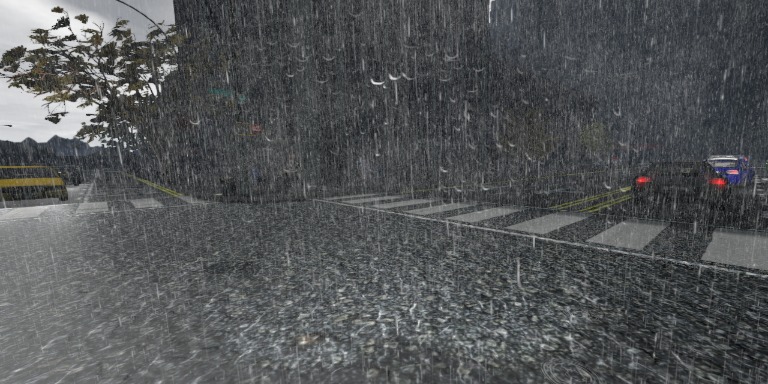} & \includegraphics[width=\linewidth]{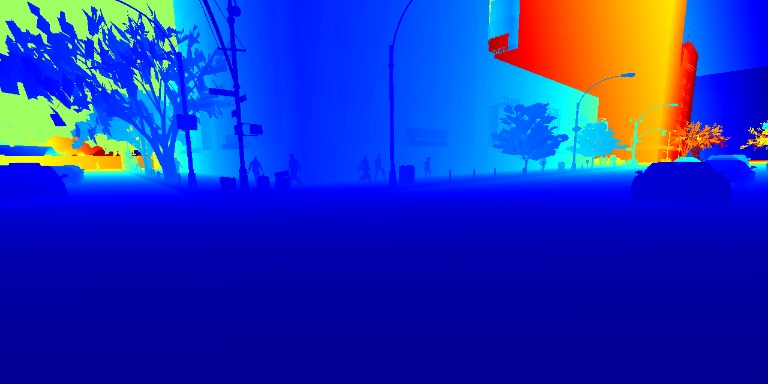} & \includegraphics[width=\linewidth]{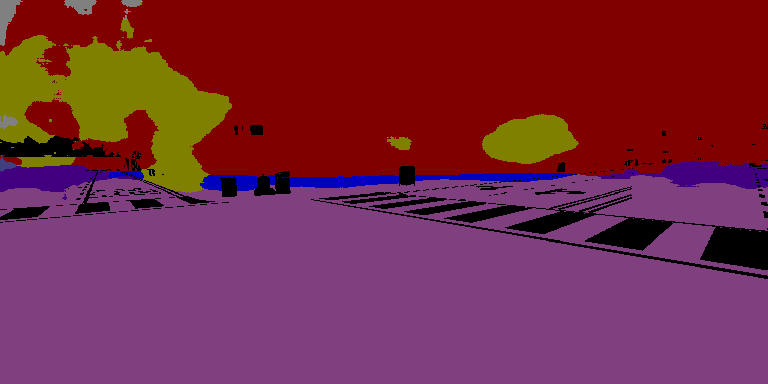} & \includegraphics[width=\linewidth]{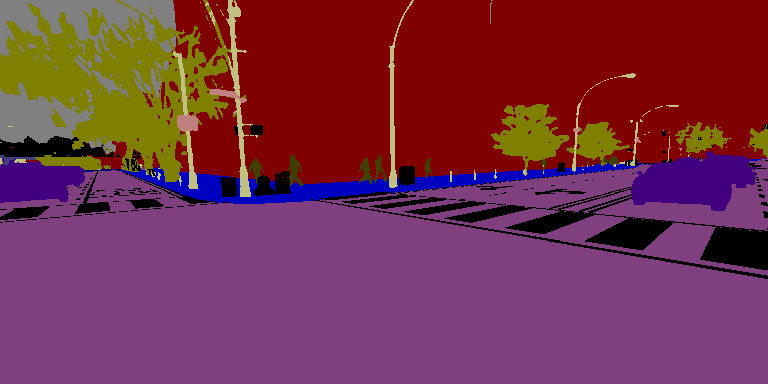} & \fbox{\includegraphics[width=\linewidth]{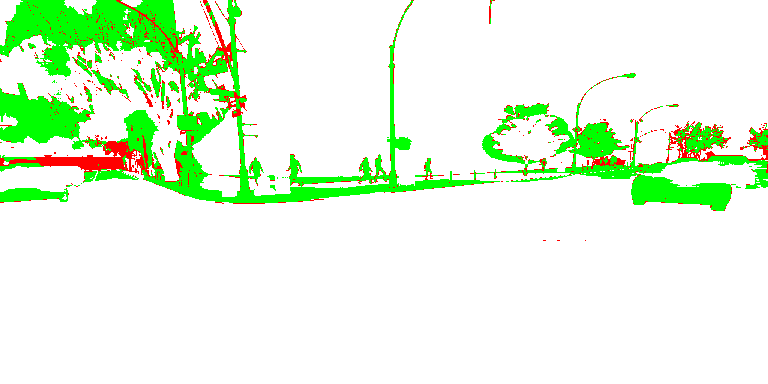}} \\
\rotatebox{90}{(i) Soft Rain} & \includegraphics[width=\linewidth]{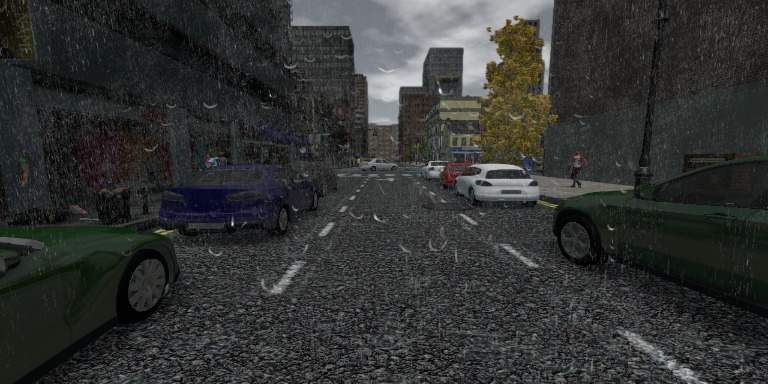} & \includegraphics[width=\linewidth]{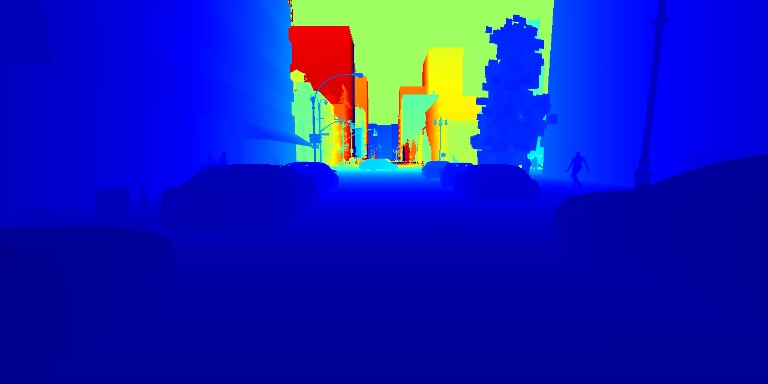} & \includegraphics[width=\linewidth]{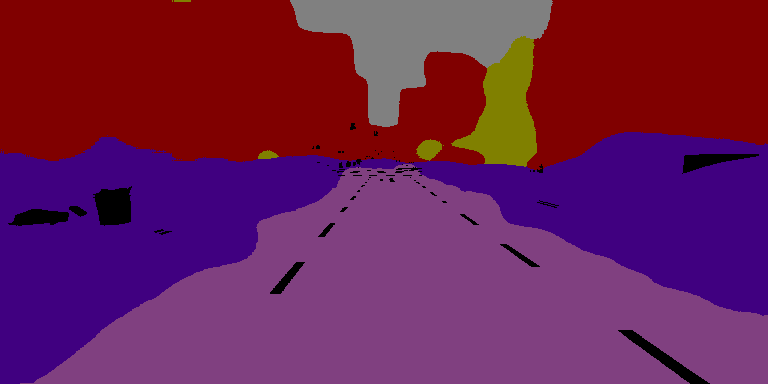} & \includegraphics[width=\linewidth]{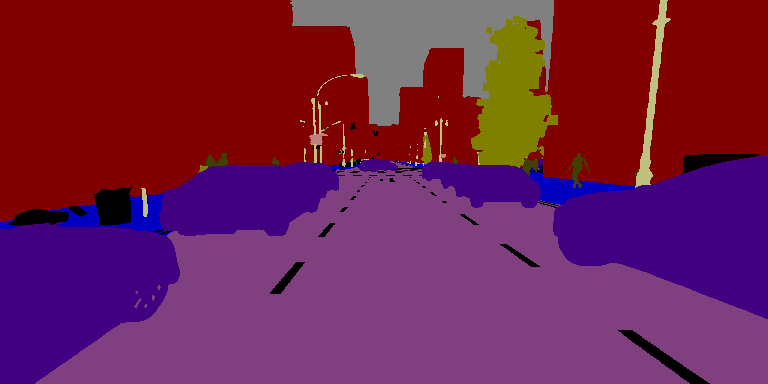} & \fbox{\includegraphics[width=\linewidth]{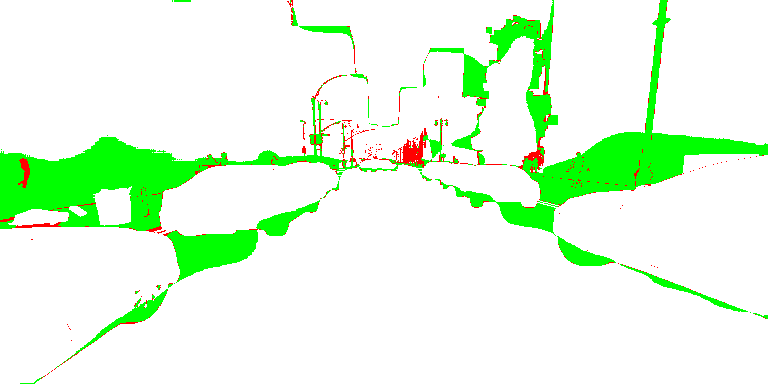}} \\
\rotatebox{90}{(j) Fog} & \includegraphics[width=\linewidth]{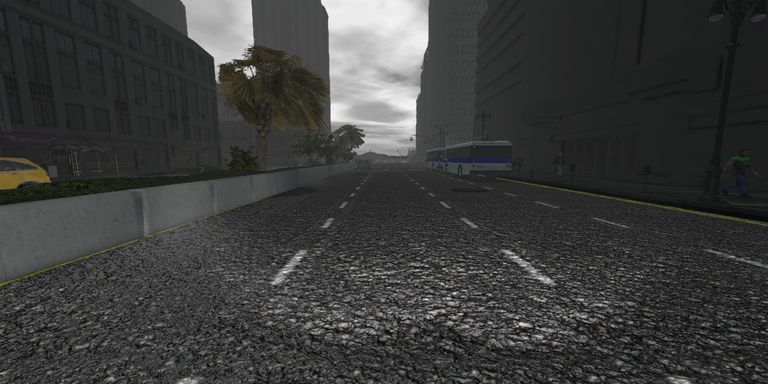} & \includegraphics[width=\linewidth]{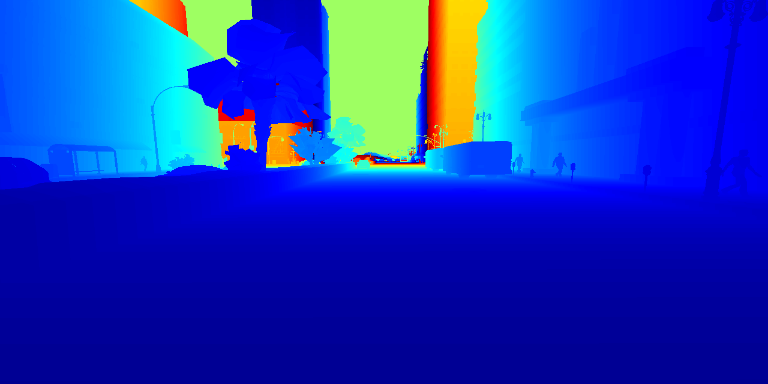} & \includegraphics[width=\linewidth]{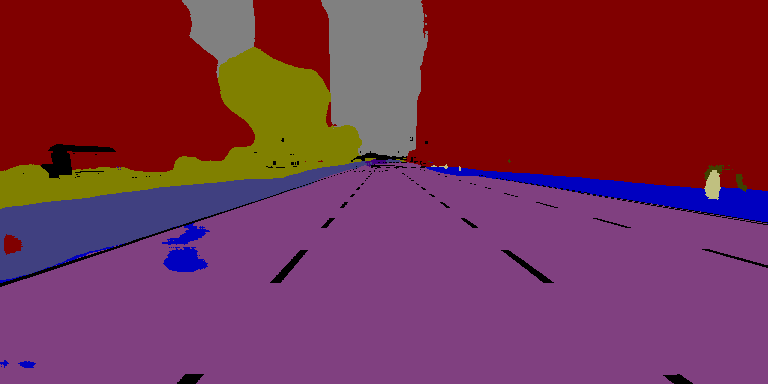} & \includegraphics[width=\linewidth]{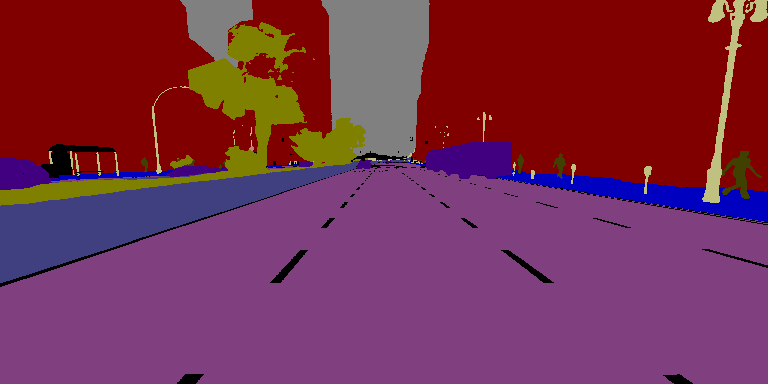} & \fbox{\includegraphics[width=\linewidth]{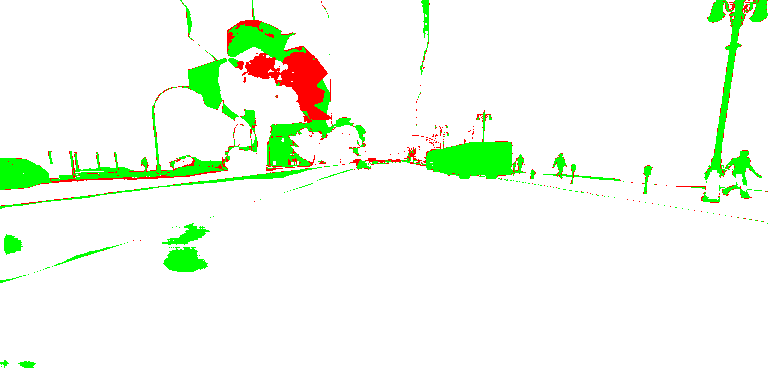}} \\
\rotatebox{90}{(k) Ngt Rain} & \includegraphics[width=\linewidth]{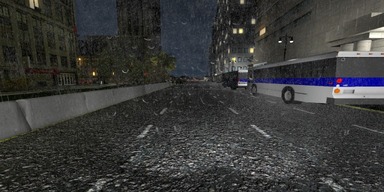} & \includegraphics[width=\linewidth]{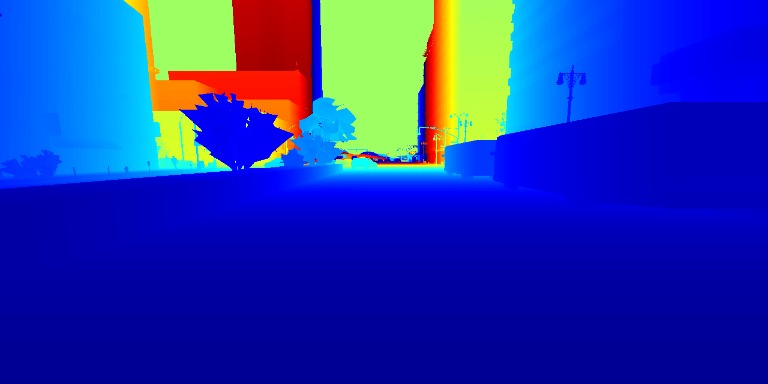} & \includegraphics[width=\linewidth]{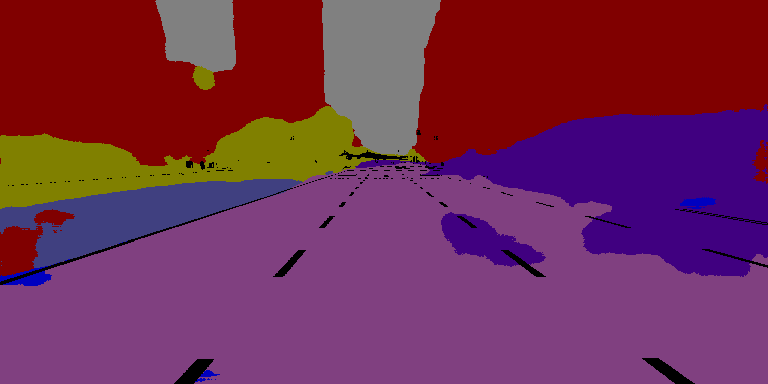} & \includegraphics[width=\linewidth]{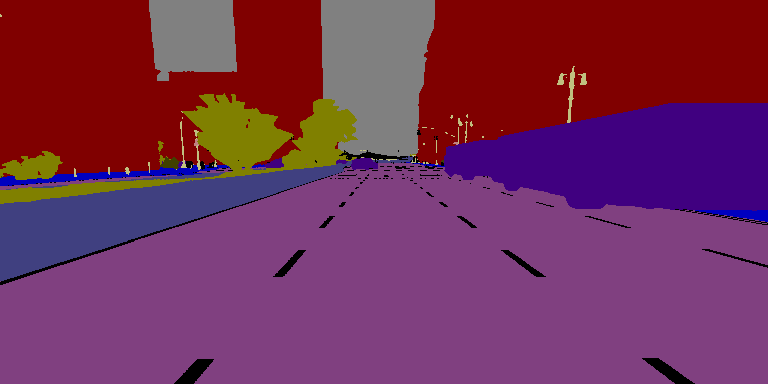} & \fbox{\includegraphics[width=\linewidth]{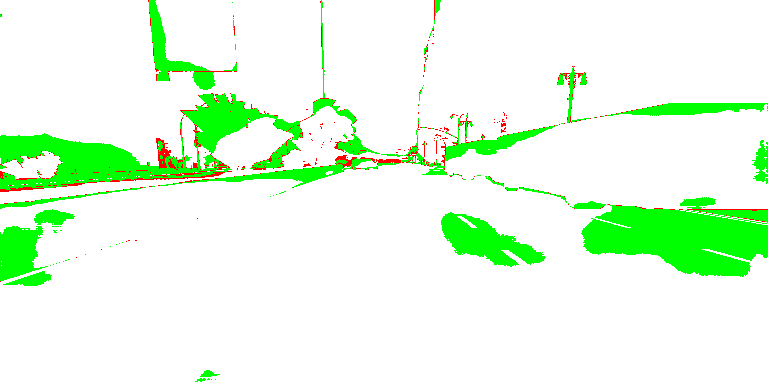}} \\
\rotatebox{90}{(l) Wntr Ngt} & \includegraphics[width=\linewidth]{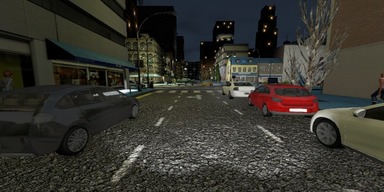} & \includegraphics[width=\linewidth]{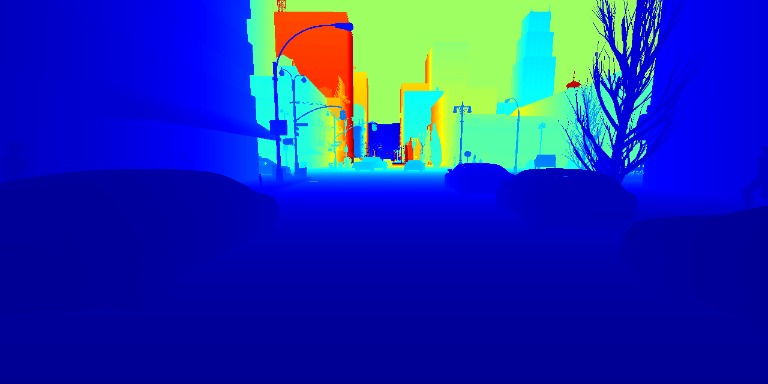} & \includegraphics[width=\linewidth]{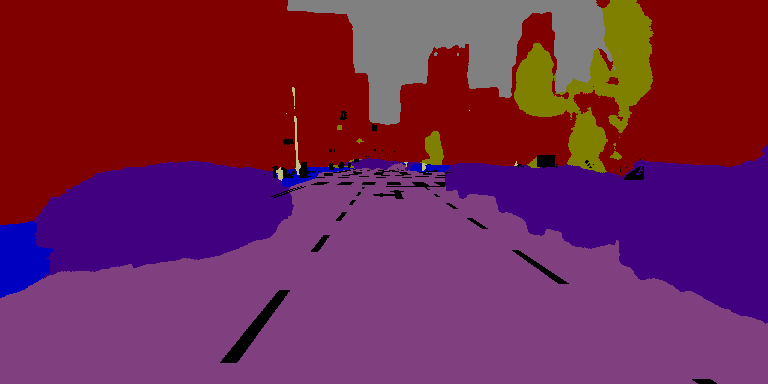} & \includegraphics[width=\linewidth]{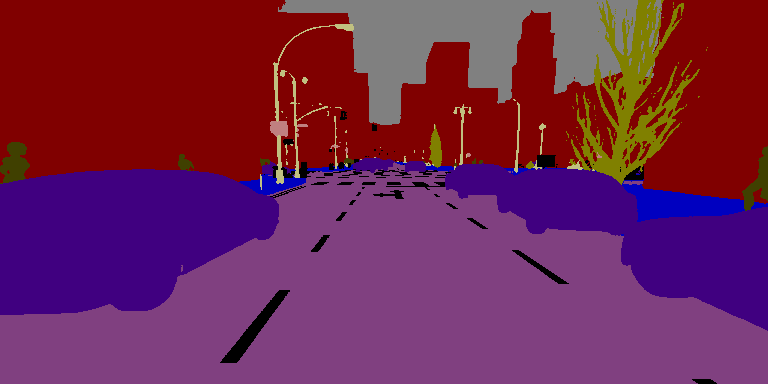} & \fbox{\includegraphics[width=\linewidth]{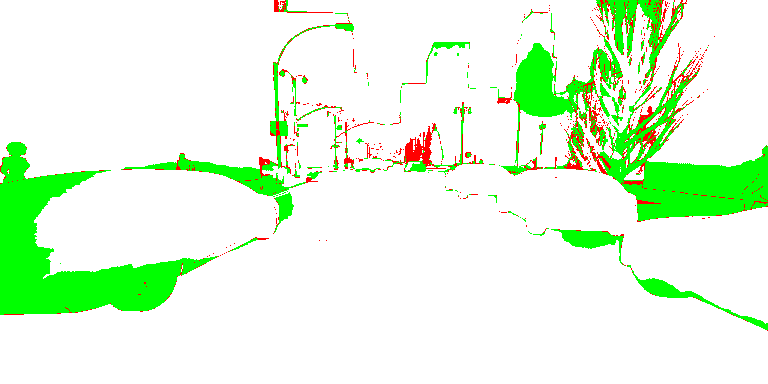}} \\
\end{tabular}}
\caption{Qualitative multimodal semantic segmentation results in comparison the model trained on visual RGB images on the Synthia-Seasons dataset. In addition to the segmentation output, we also show the improvement/error map which denotes the misclassified pixels in red and the pixels that are misclassified by the best performing state-of-the-art model but correctly predicted by AdapNet++ in green. The legend for the segmented labels correspond to the colors shown in the benchmarking tables in \secref{sec:benchm}.}
\label{fig:qualitativeSeasons}
\end{figure*}

In this section, we present qualitative results on the Synthia-Sequences dataset that contains video sequences of 12 different seasons and weather conditions. We visualize the segmentation output of the multimodal and unimodal models for which the qualitative results are shown in \tabref{fig:synthiaSequences}. For this experiment, the models were trained on the Synthia-Rand-Cityscapes dataset and only evaluated on the Synthia-Sequences dataset. The Synthia-Sequences dataset contains a diverse set of conditions such as \textit{summer, fall, winter, spring, dawn, sunset, night, rain, soft rain, fog, night rain} and \textit{winter night}. We show qualitative evaluations on each of these conditions by comparing the multimodal segmentation performance with the output obtained from the unimodal visual RGB model in \figref{fig:qualitativeSeasons}. This aim of this experiment is twofold: to study the robustness of the model to adverse perceptual conditions such as rain, snow, fog and nightime; Secondly, to evaluate the generalization of the model to unseen scenarios. 

From the examples shown in \figref{fig:qualitativeSeasons}, we can see the diverse nature of the scenes containing environments such as highway driving, inner-city with skyscrapers and small sized cities. The visual RGB images in all of the scenes show the changing weather conditions that cause vegetation to change color in \figref{fig:qualitativeSeasons}(b), snow on the ground and leaf-less trees in \figref{fig:qualitativeSeasons}(c), glaring light due to sunrise in \figref{fig:qualitativeSeasons}(c), orange hue due to sunset in \figref{fig:qualitativeSeasons}(f), dark scene with isolated lights in \figref{fig:qualitativeSeasons}(g), noisy visibility due to rain in \figref{fig:qualitativeSeasons}(h) and blurred visibility due to fog in \figref{fig:qualitativeSeasons}(j). Even for humans it is extremely hard to identify objects in some of these environments. The third column shows the output obtained from the unimodal visual RGB model. The output shows significant misclassifications in scenes that contain rain, fog, snow or nighttime, whereas the multimodal RGB-D model precisely segments the scene by leveraging the more stable depth features. The improvement map in green shown in the last column of \figref{fig:qualitativeSeasons}, demonstrates substantial improvement over unimodal segmentation and minimal error for multimodal segmentation.  The error is noticeable only along the boundaries of objects that are far away, which can be remedied using a higher resolution input image. \figref{fig:qualitativeSeasons}(e) and (j) show partial failure cases. In the first example in \figref{fig:qualitativeSeasons}(e), the occluded bus on the left is misclassified as a \textit{fence} due to its location beyond the sidewalk, where often fences appear in the same configuration. While, in \figref{fig:qualitativeSeasons}(j), a segment of the \textit{vegetation} several meters away is misclassified as a part of the \textit{building} behind. However, overall the multimodal network is able to generalize to unseen environments and visibility conditions demonstrating the efficacy of our approach.

%% file: sections/conclusion.tex
\section{Conclusion}
\label{sec:conclusion}

In this paper, we proposed an architecture for multimodal semantic segmentation that incorporates our self-supervised model adaptation blocks which dynamically adapt the fusion of features from modality-specific streams at various intermediate network stages in order to optimally exploit complementary features. Our fusion mechanism is simultaneously sensitive to critical factors that influence the fusion, including the object category, its spatial location and the environmental scene context in order to fuse only the relevant complementary information. We also introduced a channel attention mechanism for better correlating the fused mid-level modality-specific encoder features with the high-level decoder features for object boundary refinement. Moreover, as the fusion mechanism is self-supervised, we demonstrated that it effectively generalizes to the fusion of different modalities, beyond the commonly employed RGB-D data and across different environments ranging from urban driving scenarios to indoor scenes and unstructured forested environments.\looseness=-1

In addition, we presented a computationally efficient unimodal semantic segmentation architecture that consists of an encoder with our multiscale residual units and an efficient atrous spatial pyramid module, complemented by a strong decoder with skip refinement stages. Our proposed multiscale residual units outperform the commonly employed multigrid method and our proposed efficient atrous spatial pyramid pooling achieves a $10\times$ reduction in the number of parameters with a simultaneous increase in performance compared to the standard atrous spatial pyramid pooling. Additionally, we proposed a holistic network-wide pruning approach to further compress our model to enable efficient deployment. We presented exhaustive theoretical analysis, visualizations, quantitative and qualitative results on Cityscapes, Synthia, SUN RGB-D, ScanNet and Freiburg Forest datasets. The results demonstrate that our unimodal AdapNet++ architecture achieves state-of-the-art performance on Synthia, ScanNet and Freiburg Forest benchmarks while demonstrating comparable performance on Cityscapes and SUN RGB-D benchmarks with a significantly lesser number of parameters and a substantially faster inference time in comparison to other state-of-the-art models. More importantly, our multimodal semantic segmentation architecture sets the new state-of-the-art on all the aforementioned benchmarks, while demonstrating exceptional robustness in adverse perceptual conditions.